%% file: main.tex
\documentclass[11pt]{article}
\usepackage[letterpaper, left=1in, right=1in, top=1in,bottom=1in]{geometry}
\usepackage{amsfonts}       
\usepackage{nicefrac}       
\usepackage{bbm}
\usepackage[ruled,noend]{algorithm2e}
\usepackage{bm}
\usepackage{tikz}
\usepackage{graphicx}
\usepackage{pgf}
\usepackage{mathtools}
\usepackage{amsmath,amsthm,amssymb}

\usepackage{auxiliary}

\usepackage{url}      

\usepackage[pagebackref=true]{hyperref} \PassOptionsToPackage{pagebackref=true}{hyperref}   

\usepackage{cleveref}

\usepackage{color}              
\usepackage[suppress]{color-edits}
\definecolor{orange}{rgb}{0.9,0.5,0.0}
\addauthor{ww}{orange}
\newtheorem{example}{Example}

\title{Efficient 
decentralized multi-agent
learning\\
in asymmetric bipartite queueing systems}

\author{
Daniel Freund\thanks{Massachusetts Institute of Technology, \texttt{dfreund@mit.edu}}
\and Thodoris Lykouris\thanks{Massachusetts Institute of Technology, \texttt{lykouris@mit.edu}} \and  Wentao Weng\thanks{Massachusetts Institute of Technology, \texttt{wweng@mit.edu}}}

\date{First version: June 2022\\
Current version: August 2023\footnote{A preliminary version of this work was accepted for presentation at the Conference on Learning Theory (COLT) 2022. Compared to the first version of the paper, the current version expands upon the related work and adds intuition on the technical content.}}

\begin{document}

\maketitle

\begin{abstract}\input{abstract}

\end{abstract}

\addtocounter{page}{-1}
\thispagestyle{empty}

\newpage

\section{Introduction}\label{sec:intro}
\input{intro}


\section{Preliminaries}\label{sec:prelims}
\input{prelims}

\section{Decentralization with known service rates}\label{sec:algorithm-nolearning}
\input{algo-nolearning}

\section{Learning service rates via forced exploration}\label{sec:algorithm-learning}
\input{algo-learning}

\section{Adaptive exploration via optimistic service-rate estimates}\label{sec:algorithm-ucb}
\input{algo-ucb}

\section{Extension to dynamic queues}\label{sec:dynamic}
\input{algo-dynamic}

\section{Numerical results}\label{sec:simulation}
\input{algo-simulation}

\section{Conclusions}\label{sec:conclusions}
\input{conclusions}

\section*{Acknowledgments}

The authors thank the anonymous review team at COLT 2022 and at Operations Research for their valuable feedback as well as participants of the Simons Semester on Data-Driven Decision Processes and the Dagstuhl Seminar on Scheduling for insightful discussions.

\bibliographystyle{alpha}
\bibliography{references}

\appendix

\newpage

\section{Motivating applications, constraints, and assumptions}\label{app:example}\input{app_example}

\section{Notions of stability}\label{app:stability}\input{app_stability}

\section{Comparison to static matching}\label{app:comparison_to_static}\input{comparison_to_static}

\section{Omitted proofs from Section~\ref{sec:algorithm-nolearning}}\label{app:section_no_learning}
\input{app_section_no_learning}

\section{Omitted proofs from Section~\ref{sec:algorithm-learning}}\label{app:section_dam_fe}
\input{app_dam_fe}

\section{Omitted proofs from Section~\ref{sec:algorithm-ucb}} \label{app:section-ucb}
\input{app_dam_ucb}

\section{Omitted proofs from Section~\ref{sec:dynamic}} \label{app:section-dynamic}
\input{app_dam_dynamic}

\section{Omitted experiment details  and figures from Section~\ref{sec:simulation}}\label{app:simulation}
\input{app_simulation}

\section{Notation table}\label{app:notation}
\input{app_notation}

\end{document}

%% file: abstract.tex
We study decentralized multi-agent learning in bipartite queueing systems, a standard model for service systems. In particular, $N$ agents request service from $K$ servers in a fully decentralized way, i.e, by running the same algorithm without communication. Previous decentralized algorithms are restricted to symmetric systems, have performance that is degrading exponentially in the number of servers, require communication through shared randomness and unique agent identities, and are computationally demanding. In contrast, we provide a simple learning algorithm that, when run decentrally by each agent, leads the queueing system to have efficient performance in general asymmetric bipartite queueing systems while also having additional robustness properties. Along the way, we provide the first provably efficient UCB-based algorithm for the centralized case of the problem.

%% file: intro.tex
Motivated by packet routing in computer networks and resource allocation in cognitive radio, bipartite queueing systems have risen as a canonical setting to capture carryover effects in sequential learning \cite{KrishnasamySenJohariShakkottai21,gaitonde2023price,sentenac2021decentralized}. In this setting, there are $N$ agents and $K$ servers. Each agent $i$ receives jobs with a fixed arrival rate $\lambda_i$ and selects a server $j$ to route their job. The server selects (at most) one of the requesting agents $i$ and successfully serves her job with service rate~$\mu_{i,j}$. Any non-served job returns to its respective agent and is stored in a queue in front of her. 

Although this kind of queueing system has long been a standard approach to model service systems (e.g., surveys such as \cite{chen2020survey} and \cite{Srikant_Ying_2014}), a learning lens has only recently been introduced to this context. In particular, Krishnasamy, Sen, Johari, and Shakkottai \cite{KrishnasamySenJohariShakkottai16,KrishnasamySenJohariShakkottai21} introduced this line of work by studying a centralized view of the problem where a learner is allowed to jointly control all agents (there, agents correspond to different classes of jobs). That said, many queueing systems exhibit a decentralized nature, in which agents do not have the ability to communicate (see Appendix~\ref{app:example} for motivating applications). Very recently, two works initiated the study of decentralized multi-agent learning in queueing systems for the symmetric case where the service rates are only affected by the server $j$, i.e., $\mu_{i,j}=\mu_j$ for all agents $i$. Gaitonde and Tardos~\cite{gaitonde2023price} studied the quality of outcomes when agents are strategic and use no-regret learning algorithms to maximize their individual welfare and showed that the system only stabilizes if it has twice as much capacity as a central controller requires.
Closer to our work, Sentenac, Boursier, and Perchet~\cite{sentenac2021decentralized} provided a collaborative scheme that, when followed by all agents, provides bounded average-time queue lengths, i.e., it stabilizes the system for any positive traffic slackness without online communication or knowledge of the service rates.

Despite providing the first decentralized learning algorithm for bipartite queueing systems, \cite{sentenac2021decentralized}  has some important shortcomings. First, the algorithm does not scale well to systems with even a moderate number of servers $K$. In particular, the queue length guarantee is exponential in~$K^{2}$; moreover, each queue needs to solve a computationally expensive mathematical program in every time slot.~\footnote{As we discuss in Section \ref{sec:simulation}, running the algorithm becomes challenging even for $K=10$ servers.} Second, although not communicating during the online component, the algorithm requires significant initial coordination to hardwire subsequent communication. In particular, agents need to have distinct identifiers and the randomness in their algorithms is shared. Finally, the algorithm requires the system to be symmetric. We note that, even in the much simpler centralized setting, the only asymmetric result in the literature  \cite{KrishnasamySenJohariShakkottai21} requires the strong structural assumption that each agent has a unique and well-separated optimal server and that no two agents have the same optimal server (see discussion in Appendix~\ref{app:comparison_to_static}).

In this work, we design the first decentralized learning algorithm for online queueing systems that achieves the following desiderata. It is \emph{sample-efficient} in the sense that its queue-length guarantee is polynomial in the system parameters $K$ and $N$. It is \emph{computationally efficient} as its running time is linear in $K$ and independent of $N$. It is \emph{fully decentralized} in the sense that all agents use exactly the same simple algorithm without needing to have a unique identifier or shared randomness. Finally, it works for any \emph{asymmetric} bipartite queueing system without the strong structural assumptions on the optimal assignment of agents to servers of \cite{KrishnasamySenJohariShakkottai21}.

\subsection{Our contributions}
Our first contribution lies in determining an appropriate selection rule on the server side (Section~\ref{sec:prelims}). Selection rules in prior works on decentralized learning in bipartite queueing systems include random or the oldest job among the requests a server receives \cite{gaitonde2023price,sentenac2021decentralized}. We propose a variation of these rules allowing higher flexibility in the system. In particular, each agent accompanies its request with a bid and the server selects the job with the highest bid. The oldest-job rule arises as the special case where the bid is equal to the age of the job. Our bidding selection rule allows for adaptive policies that account for the current queue lengths rather than learning a global static schedule. Decentrally computing a static schedule requires all agents to solve the same optimization problem and to estimate all system parameters (e.g., \cite{sentenac2021decentralized} learn arrival rates by operating queues in a last-come-first-serve manner). This is highly sensitive to the (decentrally) learned parameters and leads to the exponential dependence on $K^2$ in \cite{sentenac2021decentralized}.

\vspace{0.1in}
\noindent{\textbf{Decentralized Auction Mechanism.}} One popular centralized adaptive policy for queueing systems is the $\textsc{MaxWeight}$ policy, which was developed for efficient scheduling in networking \cite{Tassiulas_Ephremides_1992,Mandelbaum_Stolyar_2004,Stolyar_2004}. At any time~$t$, the \textsc{MaxWeight} policy finds an allocation that maximizes a weighted bipartite matching $\boldsymbol{\sigma}$ where each edge links an agent $i$ and a server $j$ with weight equal to the corresponding service rate $\mu_{i,j}$ times the queue length of the agent $Q_i(t)$. There are a few challenges in adapting this idea to a decentralized learning setting. First, each agent $i$ only knows the queue length $Q_i(t)$ of her own queue and does not have information about the queue lengths of other agents. Second, these queue lengths fluctuate over time due to the variability in the arrival processes and, as a result, aiming to communicate the information on queue lengths through hardwiring is not amenable in this context. Finally, the service rates $\mu_{i,j}$ are also not known and need to be learned in an online manner. 

To devise a decentralized approximate version for $\textsc{MaxWeight}$, we start from the easier case where agents know the service rates and hence each agent $i$ can compute the weights $\{w_{i,j}\}_{j=1}^K$ associated to herself (Section~\ref{sec:algorithm-nolearning}). Note that even with that knowledge, the different agents do not know each others' queue lengths, and consequently cannot just compute the max-weight matching.
To address this informational bottleneck,  we adapt the $\textsc{Auction Mechanism}$, an approximate max-weight matching algorithm which was independently discovered as ``auction algorithm'' towards massively parallel machines \cite{bertsekas1988auction} and as ``approximate auction mechanism'' towards welfare maximization in multi-item auctions \cite{demange1986multi}. In that setting, there are $N$ players (agents) and $K$ items (servers) and each player $i$ has a valuation (weight) $w_{i,j}$ for item $j$. The $\textsc{Auction Mechanism}$ runs parallel ascending-price auctions; each item has a price $p_j$ and players who are not assigned to an item make increasing bids for item~$j^{\star}$ maximizing their payoff $w_{i,j^{\star}}-p_j$ at the current price as long as this payoff is positive. Once no player has reason to submit a new bid, the \textsc{Auction Mechanism} has converged to an approximate max-weight matching. 

The algorithmic crux in our approach lies in designing a decentralized version of the above algorithm, which we term $\textsc{Decentralized Auction Mechanism}$ or $\textsc{DAM}$, as a shorthand; to do so we  have to tackle three challenges. First, the bidding selection rule of the servers must be aligned with an ascending auction setting (the highest bidder is the one who is selected). However, agents do not see whether they are selected by the server when they make a request but rather only observe whether they get served (which includes randomness due to the service rates). Second, each agent's queue length is dynamically updated over time and therefore the corresponding weights $w_{i,j}$ are also changing. This makes it tricky to run a mechanism that is designed to find a matching for static weights. Finally, the $\textsc{Auction Mechanism}$ requires that all players see the same prices in order to ensure that the item is allocated to the highest bidder while our setting does not allow shared centralized information.

$\textsc{DAM}$ deals with these aforementioned challenges by operating in epochs of fixed length $\epochlength$ and using the queue length at the start of the epoch, $t_0$, in order to determine the weights of each server, i.e., $w_{i,j}=\mu_{i,j}Q_i(t_0)$ for the whole epoch. To avoid centralized prices, each agent operates with its own prices $p_{i,j}$, which she updates throughout the epoch. To counter the randomness arising from the service rates, agents only update their bids if they did not receive service for at least $\checkperiod$ time slots (coming from concentration bounds). To ensure that the epoch makes progress in decreasing the queue lengths, we maintain a Lyapunov function (the sum of sqaures of queue lengths). The epoch length $\epochlength$ strikes a balance between a) being long enough so that the expected drift in the Lyapunov function until agents converge to a matching is outweighed by the negative drift  after convergence and b) being short enough so that queue lengths at $t_0$ are representative of the queue lengths within the epoch.

\vspace{0.1in}
\noindent{\textbf{Forced Exploration.}} To incorporate learning in the algorithm (Section~\ref{sec:algorithm-learning}), each agent also maintains empirical estimates for the service rates and acts based on optimistic estimates. A challenge that arises in this case is
bias in the estimates due to interference:
when an agent is not served by the server she requested, she does not know whether this occurred due to the randomness on the service rates or because the server selected another agent. To account for this, we add \emph{forced exploration} to our mechanism: at the start of an epoch, each agent decides randomly whether she will explore or exploit in the epoch. If she explores, she commits to a random server and bids a random number  higher than any bid an \emph{exploiting} agent can make. This random number is the same over the entire  epoch which guarantees that, within an epoch, at most one exploring agent is getting served by any particular server. As a result, the estimates she obtains from this server are indeed unbiased. We note that this forced exploration idea was used in the  queueing learning paper of \cite{KrishnasamySenJohariShakkottai21} though the centralized nature of their setting avoids  the kind of potential bias we encounter.

\vspace{0.1in}
\noindent{\textbf{Adaptive Exploration.}} A natural question that arises is whether we can instead  use adaptive exploration similar to the standard Upper Confidence Bound (UCB) algorithm for multi-armed bandits \cite{auer2002finite}. In Section~\ref{sec:algorithm-ucb}, we provide a UCB-based algorithm for our setting; on a technical level, this result extends the reach of techniques based on optimism at the face of uncertainty. Typically, in such algorithms, any time that we select a suboptimal action (in our case, server), we receive some error \emph{at the current round} and 
refine the corresponding confidence interval. The analysis of such multi-armed bandit algorithms is indifferent to when these suboptimal selections were made. On the other hand, in a queueing setting, it does not suffice to consider \emph{how many} times a suboptimal action is selected but we also need to account for the system state \emph{when} the action was selected. In particular, an agent with a longer queue length making an error causes a larger increase in the Lyapunov function. We resolve this problem by charging the cost of each error to an interval rather than one specific time slot. Within this interval each time slot absorbs a cost equal to a tiny fraction of queue lengths. We show that the negative drift under \textsc{MaxWeight} outweighs the positive component of the drift caused by such errors. To our knowledge, this is the first UCB-based scheduling algorithm even beyond our bipartite setting (see Section~\ref{ssec:related_work}).

In addition, our UCB-based algorithm also extends to a model in which queues arrive and depart dynamically (Section \ref{sec:dynamic}). Though our forced exploration algorithm may fail to stabilize the system under such a dynamic model, our adaptive exploration algorithm can easily be adapted to it. We show that this adaptation stabilizes the system as long as in every time slot there is traffic slackness and an upper bound on the number of queues in the system. Our analysis involves a careful separation of the drift impact of each queue, and reveals that the impact is only based on the time a queue stays in the system. We then establish an upper bound on these drifts by connecting the gross life time of queues and the total number of queues.

Finally, we also conduct a numerical comparison (Section~\ref{sec:simulation}) of our algorithm to ones that have previously been considered in this setting: \textsc{ADEQUA} and \textsc{EXP3.P.1}. There we display both the faster convergence of our algorithms and their greater robustness with respect to (i) asymmetric service rates, (ii) time-varying arrival rates, (iii) and dynamic arrivals/departures of queues.

\subsection{Related work}\label{ssec:related_work}
\input{related_work_new.tex}

%% file: related_work_new.tex

\noindent\textbf{Decentralized queueing models.} The study of decentralized queueing models has long been motivated by
wireless networks. This is often done in a 
one-hop network: given a set of links with queues of jobs and a collection of link pairs that interfere with each other, in each time slot a set of links can be served if no two links within the set interfere with each other~\cite{Tassiulas_Ephremides_1992}. 
Algorithms aim to be practical, i.e., decentralized with low communication and computational requirements, and efficient, i.e., maximum throughput (stabilizes the system whenever possible) and queue lengths that scale polynomially with the system size and the inverse of its traffic slackness  (Definition~\ref{def:comp-slack}).

It is a daunting task to find an algorithm that is both practical and efficient in general one-hop networks. The seminal work by Tassiulas and Ephremides~\cite{Tassiulas_Ephremides_1992} shows that the \textsc{MaxWeight} algorithm, which weighs links by their queue length and selects a maximum weighted  independent set, is efficient. However, it is impractical to run \textsc{MaxWeight} for wireless networks as it is a centralized algorithm that requires solving a computationally difficult problem in each time slot. To overcome these hurdles, some papers propose greedy algorithms as approximations and impose assumptions on the link interference structure~\cite{ChenLCD06, LinS05}. One common assumption is the node-exclusive interference (NEI)  model, where there are nodes over a graph and each link connects two nodes; two links interfere if they share a common node~\cite{ChenLCD06}. Under NEI, \textsc{MaxWeight} selects a maximum-weight matching in each time slot and it is known that a greedy algorithm for maximum-weight matching achieves at least half of the optimal weight. Motivated by this, a number of papers investigates decentralized greedy algorithms with low complexity that achieve at least half of the throughput under NEI~\cite{ChenLCD06,LinS05}, and studies the throughput of its variants in other interference models~\cite{dimakis_walrand_2006,JooLS09,BirandCRSZZ12}. However, despite being easy to decentralize and incurring low complexity, greedy algorithms are usually not throughput optimal.

Another approach towards maximum throughput with reduced computational requirements was initiated by~\cite{tassiulas1998linear}. Their central idea is to randomly generate a new schedule and mix it with previous ones to obtain a slightly better schedule (in terms of weight) in each slot. Based on the generate-then-mix idea, some papers design decentralized algorithms that are of low complexity and nearly throughput optimal in NEI using message passing between nodes~\cite{ModianoSZ06,GuptaLS09,BuiSS09}. Although this stream of work ensures maximum throughput, the algorithms do not usually have strong efficiency guarantees: queue lengths may be exponentially large due to the difficulty of sampling an approximately optimal schedule  \cite{ShahTT11}.

A more recent stream of papers, motivated by Carrier Sense Multiple Access (CSMA)~\cite{KleinrockT75}, designs decentralized throughput-optimal algorithms under a general interference model where nodes cannot pass messages but can sense whether a neighbor requests service. These algorithms maintain an independent clock for each link with different parameters~\cite{YunYSE12}. A link waits for its clock to expire before its service and pauses its clock if it senses the service of a conflict link.~\cite{JiangWalrand10} shows that a system operating in this manner is effectively a reversible continuous time Markov chain and provides a gradient descent algorithm for each link to update their parameter. Their proof of maximum throughput relies on a time-scale separation assumption, later relaxed in~\cite{JiangSSW10}.~\cite{RajagopalanSS09, shah2012randomized,GhaderiS10} apply similar ideas to sample maximum-weight independent sets using Metropolis-Hastings. \cite{NiTS12,ShahST11,JiangW11} remove the requirement of knowledge sharing among links, and extend to discrete-time models and imperfect sensing ability. That said, algorithms based on CSMA usually lead to exponential queue lengths~\cite{BoumanBL11,LotfinezhadM11}. One exception is the CSMA policy in \cite{LotfinezhadM11} which is efficient when the interference graph is a Lattice or Torus and there are sufficiently many nodes. Their result requires two difficult-to-establish regularity assumptions of CSMA policies (Section VIII.D in~\cite{abs-1009-5944}).

Our work contributes to the broad literature of wireless networks by identifying a wide class of interference models (asymmetric bipartite queueing systems) for which we can design a practical and efficient scheduling algorithm without requiring additional assumptions. This is in contrast to the general interference model where it is impossible to design (even  centralized) efficient low-complexity algorithms, under a computational hardness assumption~\cite{ShahTT11}. Our work thus contributes to the broad literature of wireless networks by identifying a wide class of interference models (asymmetric bipartite queueing systems) for which we can design a practical and efficient scheduling algorithm  without requiring additional assumptions. As alluded to before, the bipartite queueing system is a standard model for service systems, with applications in communication networks \cite{DBLP:journals/tit/TassiulasE93}, call centers \cite{DBLP:journals/ior/GurvichW10} and healthcare \cite{armony2015patient}; see \cite{chen2020survey} for a detailed discussion. Restricted to this special case of NEI, $\textsc{MaxWeight}$ is known to be delay-optimal in heavy-traffic~\cite{Mandelbaum_Stolyar_2004,Stolyar_2004,shah2012switched,maguluri2016heavy}. Indeed, $\textsc{MaxWeight}$ just needs to identify a maximum-weight bipartite matching, which facilitates algorithms such as the Auction Mechanism~\cite{bertsekas1988auction,demange1986multi} or max-product belief propagation~\cite{BayatiSS05}. In fact, \cite{bayati2007iterative} applied the auction mechanism to improve the computational performance of $\textsc{MaxWeight}$. The Auction Mechanism was also studied in distributed computing~\cite{zavlanos2008distributed,NaparstekL11,kalathil2014decentralized} but the resulting algorithms use intense communication between agents.

\vspace{0.1in}
\noindent\textbf{Learning in queues.}
Even ignoring decentralization, bandit learning has only recently been incorporated in queueing systems. In the centralized setting, $\textsc{MaxWeight}$ needs to know the service rates although they can be adaptively selected by an adversary~\cite{liang2018minimizing}; learning of service rates was introduced in a single-queue context \cite{walton2014two} and was only recently studied for more involved systems, see survey in~\cite{walton2021learning}. Apart from \cite{KrishnasamySenJohariShakkottai16,KrishnasamySenJohariShakkottai21} with which we more heavily contrast, a few other works try to efficiently address the exploration-exploitation trade-off between learning and scheduling in wireless networks~\cite{KrishnasamyAJS18,krishnasamy2018augmenting}, load balancing~\cite{choudhury2021job}, best-channel identification~\cite{stahlbuhk2021learning} and queues with abandonment~\cite{zhong2022learning}. However, all of those algorithms either use non-adaptive forced exploration to learn parameters or consider settings in which exploration is unnecessary~\cite{KrishnasamyAJS18}. 
Adaptive exploration is used in \cite{HsuXLB22} to learn utility functions of customers entering a queueing system. They assume known service probabilities, and aim to learn optimal utility while maintaining queue stability, whereas we directly learn to schedule. 
Closer to our setting, \cite{StahlbuhkSM19} provides a decentralized learning algorithm in the NEI model based on UCB; their algorithm is based on greedy maximal matchings and is thus not throughput optimal. Subsequently to our work, \cite{YangSrikantYing22} 
consider a bipartite queueing system with non-stationary service probabilities and augment \textsc{MaxWeight} with discounting UCB to ensure maximum throughput. They establish asymptotic queue length bounds and the proof heavily relies on the decoupling between the queueing and learning process due to the discounting of UCB. Even in a centralized queueing setting, prior to our work, it was 
open whether simple adaptive exploration algorithm such as UCB can provide favorable queue lengths while achieving throughput optimality.

\vspace{0.1in}
\noindent\textbf{Multi-player multi-armed bandit.} If one disregards the queueing aspect of our problem, the problem becomes equivalent to the one of Multi-player Multi-Armed Bandits that has seen emerging interest in the past years. In this setting, $N$ agents need to decide which arm to select among $K$ arms. The selection rule that is used in this case is typically that, if more than one arm is selected, then no one is served. This setting has been studied both with symmetric~\cite{avner2014concurrent,rosenski2016multi} and asymmetric rewards~\cite{avner2014concurrent,rosenski2016multi,mehrabian2020practical,bistritz2021game,liu2021bandit}. Although the initial papers allowed the agents to observe collisions (which makes learning easier), subsequent works provided algorithms that do not require agents to observe collisions which is more similar to our work~\cite{bubeck2021cooperative,boursier2019sic,lugosi2021multiplayer}. Comparing these two lines of works, the selection rule we consider gives us more flexibility in the algorithm design. On the other hand, in our setting, the cost associated with an error depends on when the error occurred due to its effect on the queues, which adds a technical complexity that is not present in multi-player multi-armed bandits. It is an interesting open question to understand whether decentralized multi-agent learning in bipartite queueing systems can be achieved with the stricter collision selection rule considered in those works. 
In terms of objectives, in the bandit setting one aims to minimize \emph{regret}, defined as the deficit of total reward compared with an optimal benchmark. However, in a queueing network setting, it is hard to compare with an optimal benchmark: even with known parameters, computing the centralized optimal policy is intractable \cite{DBLP:journals/mor/PapadimitriouT99}. In the bandit setting with known distributions, the ex-ante optimal policy just selects the arm with the highest mean.
Thus, queueing results aim for \emph{stability}, i.e., having an upper bound on queue lengths that does not scale with time (a lower upper bound is better).

\vspace{0.1in}
\noindent\textbf{Multi-agent reinforcement learning.} Our setting can also be captured in the general framework of multi-agent reinforcement learning (MARL) where each agent receives reward and transitions to a new system state based on the actions of all agents. Depending on the reward, agents can be competitive (zero-sum game), cooperative (shared reward function) or in a mixed game. Existing work focuses on designing decentralized algorithms so that agents converge fast to certain equilibrium observing only their own actions and rewards \cite{DBLP:journals/tsmc/BusoniuBS08, jin2021v,gao2021finite,sayin2021decentralized}. Our model can be viewed as a cooperative setting where the global reward for agents is the sum of queue lengths. However, in our setting, agents only partially observe the system state and the global reward through their own queue lengths. To the best of our knowledge, such a limit of local observation in MARL is only recently studied by \cite{qu2022scalable} where the authors design an algorithm  that provably converges to a near-optimal local policy. Unlike our setting, their result assumes a finite number of states and discounted reward. In addition, their focus is on the convergence speed to an optimal policy assuming sufficient exploration; while our focus is on efficient exploration strategies in terms of the cost in queue lengths.

%% file: prelims.tex
\noindent\textbf{Multi-agent queueing system.}
Consider the asymmetric discrete-time queueing system where there is a set of queues $\mathcal{N}$, also referred to as \emph{agents}, and a set of servers $\mathcal{K}$ of cardinality $N$ and~$K$ respectively. For ease of notation, queues and servers are indexed in~$[N]=\{1,\ldots,N\}$ and~$[K]=\{1,\ldots,K\}$. Each queue $i\in\mathcal{N}$ is associated with an arrival rate  $\lambda_i\in[0,1]$ and service rates $\mu_{i,j}\in[0,1]$ for each server $j\in \mathcal{K}$. The service rates of a server may differ based on the queue that requests service which makes our queueing system asymmetric.

At each time slot $t =1,2,\ldots$, each queue $i\in\mathcal{N}$ receives a new job with probability $\lambda_i$ and either selects to request service from a server $J(i,t) \in \mathcal{K}$ or chooses not to request service which we denote
by~$J(i,t)=\perp$. If the request is successful, the job on the head of the queue is served; otherwise no job is served and we proceed to the next round. On the server side, each server $j\in \mathcal{K}$  receives requests from a set of queues $R(j,t) = \{i \colon J(i,t) = j\}$ and either selects  to serve a queue $I(j,t)\in R(j,t)$ or no queue at all, $I(j,t)=\perp$, based on a tie-breaking rule (see below). Requests from non-selected queues $i\in R(j,t)\backslash \{I(j,t)\}$ fail. If $I(j,t) \neq \perp$, the request from $I(j,t)$ is successful with probability~$\mu_{I(j,t), j}$. It will be useful to define $\mu_{\perp,j}=1$ and $\mu_{i,\perp}=0$ for all $i\in\mathcal{N}$ and $j\in\mathcal{K}$.

\vspace{0.1in}
\noindent\textbf{Server selection rule.}
We now expand on how  server $j$ selects which, if any, job to serve  among the set of requests $R(j,t)$ it receives. The literature generally considered three tie-breaking rules. Previous works on non-queueing multi-player multi-armed bandits assume that, if there is a collision~($|R(j,t)|>1$), then no job is served \cite{avner2014concurrent, rosenski2016multi, boursier2019sic}. On the queueing side, the simplest rule involves selecting a random request from $R(j,t)$ \cite{gaitonde2023price}. The most successful rules for this setting posits that the server selects the job in $R(j,t)$ that has been created the earliest \cite{gaitonde2023price,sentenac2021decentralized}. Our approach requires a variation of this rule that allows for somewhat greater flexibility. Specifically, we allow each queue to send a bid to the server together with its request. The server then selects the queue with the highest bid.\footnote{Ties are assumed to be broken arbitrarily; our algorithms will ensure that, almost surely, ties do not exist.}
The oldest-job rule arises in our generalization if all queues sent the age of the job as their bid. 

To formally define the queue dynamics, let $Q_i(t)$ be the number of jobs in queue $i$ at the beginning of time slot $t$. Define $A_i(t), S_i(t) \in \{0,1\}$ where $A_i(t) = 1$ if there is a new job arrival, and $S_i(t) = 1$ if the request from queue $i$ is successful. Then queue $i$ evolves as 
\begin{equation}\label{eq:queuedynamic}
    Q_i(t+1) = \big(Q_i(t) + A_i(t) - S_i(t)\big)^+,
\end{equation}
where $x^+ = \max(x,0).$ All queues are initially empty, and thus $\forall i, Q_i(1) = 0$. We note that even when a queue is empty, it is still allowed to send a ``null'' request to a server. To ease exposition, we also assume that the queue observes whether this null request was successfully completed.

\vspace{0.1in}
\noindent\textbf{Objective.}
Our goal is to design an algorithm $\textsc{Alg}$ to guide the queues' selection of which server to send their request. This algorithm needs to be \emph{fully decentralized}, i.e., operate without knowledge of the arrival and service rates, or the number and label of queues in the system. All queues follow $\textsc{Alg}$ and cannot communicate further about their service status, their queue lengths, etc.

On a high level, the performance of this algorithm is evaluated based on the average time it takes a job to be served. More precisely, our objective measures
\begin{equation*}\label{eq:average-queue-length}
    \textsc{Obj}(T)=\frac{1}{T}\expect{\sum_{t=1}^T \sum_{i=1}^N \lambda_i Q_i(t)}.
\end{equation*}
To understand this quantity, we can
rewrite it as $\|\boldsymbol{\lambda}\|_1\sum_{i\in\mathcal{N}} \frac{\lambda_i}{\|\boldsymbol{\lambda}\|_1} Q_i(t)$, where $\|\boldsymbol{\lambda}\|_1=\sum_i|\lambda_i|$ denotes the 1-norm of vector $\boldsymbol{\lambda}$, i.e., the gross arrival rates of the system. This is multiplied with the queue length 
$Q_i(t)$. From the perspective of a job arriving at time slot~$t$
, $\expect{\sum_{i\in\mathcal{N}} \frac{\lambda_i}{\|\boldsymbol{\lambda}\|_1} Q_i(t)}$ is the expected number of jobs ahead of it. We summarize the notation used in Appendix~\ref{app:notation}.

Our goal is to design a fully decentralized algorithm $\textsc{Alg}$ such that, if all queues follow $\textsc{Alg}$ this objective will be upper bounded by a term $C$. We call such an algorithm \emph{efficient} if its resulting upper bound $C$ depends polynomially on $K$ and $N$ (ideally we wish the dependence on $N$ to be polylogarithmic so that we can afford a really large number of agents), and does not depend on $T$.

\paragraph{Stability of the queueing system.}
Our objective ensures  \emph{strong stability} of the underlying system, which asks for $\lim_{T \to \infty} \frac{1}{T}\expect{\sum_{t=1}^T \sum_{i\in \set{N}} Q_i(t)} < \infty$ (we discuss other notions of stability in Appendix~\ref{app:stability-static}).  Of course, we cannot hope to obtain an efficient fully decentralized algorithm unless the underlying centralized queueing system makes such a guarantee achievable. We thus require that the arrival and service rates admit the necessary and sufficient conditions for the (centralized) system to be \emph{stable} (Theorem 3.2 in \cite{Tassiulas_Ephremides_1992}).

To formally state these conditions, consider a system evolving over $T$ time slots with a centralized controller. At each queue $i\in \mathcal{N}$ the number of jobs arriving during this time horizon is roughly $T\lambda_i$ --- to maintain a bounded queue length, each queue needs this many jobs to be serviced. Let $\phi_{i,j}$ denote the fraction of time slots in which the centralized controller  sends 
jobs from queue $i$ to server $j\in \mathcal{K}$. Since in each slot a server only chooses one job,
the centralized controller never sends more than one job per time slot to a server. Hence, these fractions form the set
\begin{equation*}
\Phi:=\Big\{\boldsymbol{\phi}:\sum_{j'=1}^K \phi_{i,j'} \leq 1;~\sum_{i'=1}^N \phi_{i',j} \leq 1;~\phi_{i,j} \geq 0\Big\}.
\end{equation*}
Each job sent from queue $i$ to server $j$ is serviced with probability equal to $\mu_{i,j}$; thus, the expected number of jobs from queue $i$ that are serviced in this centralized setting is equal to~$T\mu_{i,j}\phi_{i,j}$. Summed over all servers $j$, the expected number of jobs from queue $i$ that are served is then ~$T\sum_j\mu_{i,j}\phi_{i,j}$. For our system to be stable,
$T\lambda_i\leq T\sum_j \mu_{i,j} \phi_{i,j}$ 
, i.e., the number of jobs that arrive to queue $i$ should be no more than the number of jobs from queue $i$ that are completed. To formalize this intuition we define
$\mu_i(\boldsymbol{\phi})=\sum_{j\in\mathcal{K}}\mu_{i,j}\phi_{i,j}$; as the \emph{effective service rate of queue~$i$ under a flow~$\boldsymbol{\boldsymbol{\phi}}$} and denote the vector of all queues' effective service rates with flow $\boldsymbol{\phi}$ by $\boldsymbol{\mu}(\boldsymbol{\phi})$. Then the set of all vectors of queue-processing rates that are realizable by the existing asymmetric service rates $\boldsymbol{\mu}$ is:

\begin{equation*}
    \mathcal{M}:=\Big\{\boldsymbol{\mu}(\boldsymbol{\phi}) : \boldsymbol{\phi}\in\Phi\Big\}.
\end{equation*}
If the arrival rates are not in $\mathcal{M}$, i.e., $\boldsymbol{\lambda}\not\in\mathcal{M}$, it means that no matter what the central controller's policy is, there exists at least one queue whose arrival rate $\lambda_i$ is larger than the queue's effective queue-processing service rate. This means that its queue length will build up over time. Indeed, even if $\lambda_i=\sum_{i,j}\phi_{i,j}\mu_{i,j}$, stochastic fluctuations would still cause the expected queue length at $i$ to grow large. 
The traffic slackness $\varepsilon$ measures how far away from that scenario we are and how easy the goal to serve all jobs in a centralized manner is.

\begin{definition}\label{as:stability}
A  multi-agent queueing system has \emph{traffic slackness} $\varepsilon$ if  $(1+\varepsilon)\boldsymbol{\lambda}\in \mathcal{M}$.
\end{definition}
 A system is stable if it admits traffic slackness $\varepsilon>0$; the smaller this quantity is the more difficult the setting. We assume that the fully decentralized algorithm knows a traffic slackness~$\varepsilon\in (0,1]$ for the system (if a system has traffic slackness~$\varepsilon>\varepsilon'$, then it also has traffic slackness $\varepsilon'$). We also assume that there is a universal lower bound $\delta$ on all non-zero service rates, i.e., for all $i\in\mathcal{N}$ and $j\in\mathcal{K}$, $\mu_{i,j}\in\{0\}\cup[\mulower,1]$ and this lower bound is known to the algorithm.

\vspace{0.1in}
\noindent\textbf{Connection to symmetric service rates.} When the service rate $\mu_{i,j}$ is symmetric, i.e., $\mu_{i,j}=\mu_j$ for all $i\in \mathcal{N}$ and $j\in\mathcal{K}$, the stability condition becomes $\sum_{i=1}^{n} \lambda_i < \sum_{j=1}^{\min(n,K)} \mu_j~~ \forall n \in [N]$,
where arrival and service rates are indexed in decreasing order, i.e., $\lambda_1 \geq \ldots \geq \lambda_N, \mu_1 \geq \ldots \mu_K.$  To illustrate the traffic slackness in that case it is also
useful to define
$\Delta = \min_{n \in [N]} \sum_{j=1}^{\min(n,K)}\mu_j - \sum_{i=1}^{n} \lambda_i$ as in \cite{sentenac2021decentralized}. A traffic slackness $\varepsilon$ then implies $
    (1+\varepsilon) \sum_{i=1}^n \lambda_i \leq \sum_{j=1}^{\min(n,K)} \mu_j~~ \forall n \in [N]$.
If a symmetric multi-agent queueing system has gap $\Delta = \min_{n \in [N]} \sum_{j=1}^{\min(n,K)}\mu_j - \sum_{i=1}^{n} \lambda_i$, it follows that $\min_{n \in [N]}\sum_{j=1}^{\min(n,K)}\mu_j - (1+\frac{\Delta}{K})\sum_{i=1}^{n} \lambda_i \geq 0$, implying traffic slackness $\trafficslack$ for all $\trafficslack\leq \Delta/K$.

\begin{remark}
It may be useful to compare the traffic slackness $\varepsilon$ with the gap quantity in multi-armed bandits \cite{auer2002finite}.
In bandits, a small gap makes the estimation problem more difficult, i.e., it is difficult to identify the best arm, but does not necessarily lead to large regret (e.g., when the gap is $1/T$). In contrast, in service systems a small traffic slackness causes instability even with full knowledge of all parameters. In particular, it is known that the average queue length of any algorithm is at least $\nicefrac{1}{\varepsilon}$. Letting $\varepsilon$ be a decreasing function of $T$ necessarily leads to an unstable system. Hence, the usual assumption in queueing is that the system has infinite horizon and $\varepsilon$ is a fixed parameter independent of time, see \cite{KrishnasamySenJohariShakkottai21,gaitonde2023price}.
 \end{remark}

 \vspace{0.1in}
 \noindent\textbf{Motivating applications, assumptions and constraints.} We now briefly comment on the main constraints that our model posits. We require \emph{decentralization} (queues possess only local information), \emph{asymmetry} on the service rates, \emph{unique requests} (at most one request per round for each queue), \emph{learning} (initially unknown service/arrival probabilities), and \emph{binary feedback} (queues cannot distinguish between rejected and accepted-but-failed requests). For our results to go through, we also make the following assumptions: \emph{synchronization} (existence of a common clock), \emph{lower bounds} of $\delta$ on non-zero service probabilities and $\varepsilon$ on the traffic slackness, \emph{collision handling} based on the highest-bid selection rule, and \emph{collaboration} (absence of selfish behavior). Our two main motivating applications in considering this model include cognitive radio and online service platforms; in particular, decentralization and learning are prominent in both applications. We discuss which constraints and assumptions are satisfied in either application in Appendix~\ref{app:example} and point to open directions aiming to remove the assumptions we make in Section~\ref{sec:conclusions}.

%% file: algo-nolearning.tex
A centralized way to achieve stability is by maximizing $\sum_{i \in \set{N}} Q_i(t) \mu_{i,\sigma(i)}$ where $Q_i(t)$ is the queue length of agent $i$ and $\sigma(i)$ is its assigned server \cite{Tassiulas_Ephremides_1992,Srikant_Ying_2014}; we refer to this policy as \textsc{MaxWeight}. However, in a decentralized setting, agents do not know their relative queue length to abide by this optimal assignment. As a result, our goal is to design a fully decentralized algorithm for the agents that induces a matching which is \emph{approximately max-weight most of the times} and \emph{show that this suffices to induce low queue lengths}. In this section, we focus on the simpler task of deriving such a fully decentralized algorithm when the service rates are known in advance; in Sections~\ref{sec:algorithm-learning} and~\ref{sec:algorithm-ucb}, we extend this to the case where the agents need to also learn them in an online fashion.
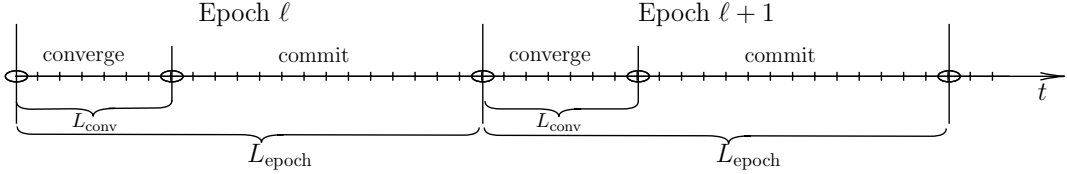
\begin{figure}
    \centering
    \scalebox{0.7}{\input{figure/algofig.tex}}
    \caption{High-level protocol of the $\textsc{Decentralized Auction Mechanism}$}
    \label{fig:algofig}
\end{figure}
Our algorithm (Algorithm~\ref{algo:DAM-master}), which we term \emph{Decentralized Auction Mechanism}, works in epochs of fixed length~$\epochlength$. Each epoch~$\ell$ consists of two parts: an initial (shorter) part of length~$\convlength$ where queues converge to the desired approximate maximum weight matching in a decentralized fashion and a subsequent part where they commit to selecting the matched server until the end of the epoch. Fig.~\ref{fig:algofig} provides a pictorial representation. To formally define the algorithm, recall that~$\delta$ is the uniform lower bound on non-zero service probabilities and we let $\xi =  \frac{\trafficslack^2}{3200K^2(\log N + K)}$. We instantiate the epoch length $\epochlength$, the exploration length $\convlength$ as well as the length of a checking period $\checkperiod$ as follows:
\begin{equation}\label{eq:parameter-setting}
\begin{aligned}
\checkperiod &= 
\left\lceil\max\left(3, \left(\frac{2}{\ln(1-\mulower)}\right)^2, \frac{2\ln \xi}{\ln(1-\mulower)}\right)\right\rceil,\\
\convlength &=
\left \lceil \frac{99K\checkperiod}{\trafficslack}(\log N + K) \right\rceil \text{, and }
\epochlength 
=\left \lceil (\frac{32}{\trafficslack} + 1) \convlength \right\rceil.
\end{aligned}
\end{equation}

 \input{DAM-master}

 \subsection{Algorithmic crux: Approximate max-weight matching within an epoch}
In the initial time slots of an epoch $t=t_0,\ldots, t_0 + \convlength -1$, the queues run a decentralized version of the Auction Mechanism \cite{bertsekas1988auction, demange1986multi}. This ensures, with high probability, that when this initial phase terminates, the queues have converged to a matching, $\sigma$, of queues to servers, which approximately maximizes $\sum_{i=1}^N w_{i,\sigma(i)}$ for weights $w_{i,j}=Q_i(t_0)\mu_{i,\sigma(i)}$ that reflect the queue lengths at the beginning of the epoch. Note that at time slot $t>t_0$, the queue lengths are no longer equal to~$Q_i(t_0)$; thus, the matching identified is not necessarily an approximate max-weight matching. However, as long as each epoch consists of sufficiently few time slots, we can bound the gap to the max-weight matching throughout the epoch, and leverage this bound to prove stability.
 
Let $f_{i,j}$ denote the indicator of whether queue $i$ is matched to server $j$ in a matching. The goal of maximizing the aforementioned sum of weights can thereby be written as in the \eqref{eq:primal} of the following mathematical program. Solving this program requires knowledge of all the weights and therefore of the queue lengths $Q_i(t_0)$ for all queues $i\in\mathcal{N}$, information that is not available to any individual queue. 
However, the dual of the program allows the queues to compete for each server based on the respective queue-server weights. In particular, by viewing the dual variables $\pi_i$ and $p_j$ as the payoff
 of queue $i$ and the price of server~$j$ respectively, we obtain the following economic interpretation:
 queue $i$ needs to pay $p_j$ to obtain value ~$w_{i,j}$ by using server $j$. Therefore, its payoff~$\pi_i$ is at least $w_{i,j}-p_j$, as stated in \eqref{eq:dual}. With this interpretation in hand, we adopt the Auction Mechanism to attain a feasible dual solution and approximate the optimal primal.
 
 \begin{minipage}{2.75in}
\begin{equation}
\label{eq:primal}
\tag{\text{Primal}}
\begin{aligned}
\text{max} & \sum_{i=1}^N\sum_{j=1}^K w_{i,j}f_{i,j}\\
\text{s.t.} & \sum_{j \in \mathcal{K}} f_{i,j} \leq 1, \forall i \in \mathcal{N}\\
& \sum_{i \in [N]} f_{i,j} \leq 1, \forall j \in \mathcal{K} \\
& f_{i,j} \geq 0, \forall i,j \in \mathcal{N} \times \mathcal{K}
\end{aligned}
\end{equation}
\end{minipage}
\begin{minipage}{3.5in}
\begin{equation}
\label{eq:dual}
\tag{\text{Dual}}
\begin{aligned}
\text{min} & \sum_{i=1}^N \pi_i + \sum_{j=1}^K p_j\\
\text{s.t.} & \pi_i + p_j \geq w_{i,j}, \forall (i,j)\in[N]\times[K]\\
& \pi_i,p_j \geq 0.
\end{aligned}
\end{equation}
\end{minipage}
The high-level centralized strategy in the Auction Mechanism is to keep centralized prices $p_j$ (initialized to $0$). At any time, all servers with a price larger than $p_j>0$ are matched to some queue which bids $p_j$. The algorithm aims to improve the matching by trying to assign an unmatched queue~$i$ to the server providing the highest payoff at the current price, i.e., $j^{\star}=\arg\max_{j\in\mathcal{K}}(w_{i,j}-p_j)$. If that payoff is non-negative, the queue bids $p_j+\beta$ for a small increment~$\beta>0$. The process ends when all unmatched queues have negative payoff from any server at the current price; at this point, the resulting dual solution is at most $K\beta$ larger than the primal. 

There are two challenges in extending this approach to a decentralized queueing setting. First, the above algorithm requires unmatched queues to seek service from some server at an increased price. However, in a decentralized queueing setting, the queue does not know if it is unmatched as it does not observe the reason why it was not served (not getting served could be  due to either the randomness in service probabilities or collisions with other queues). To address this issue (lines \ref{algoline:no-update-price}-\ref{algoline:no-update-price-step} of Algorithm~\ref{algo:DAM-queue}), the queue's algorithm consistently selects the same server until it is not served for a checking period of $\checkperiod$ time slots, defined in Eq.~\ref{eq:parameter-setting}
. Since the probability of an unsuccessful request is always at most $1-\delta$ for any queue-server pair $(i,j)$ with $\mu_{i,j}>0$, concentration bounds imply that, at this point, the queue is confident (with high probability)
that it is no longer matched to the server. The second challenge is that, due to the lack of communication, different queues no longer maintain the same prices for the servers. To deal with this, at any time slot $t>t_0$, each queue operates with its own prices $\{p_{i,j}(t)\}_{j\in\mathcal{K}}$ which are increased by an increment $\beta_{i,j}=\frac{1}{16}\trafficslack w_{i,j}$ (line \ref{algoline:price-update} in Algorithm~\ref{algo:DAM-queue}). 
The final algorithm is formalized in Algorithm~\ref{algo:DAM-queue}. To modularly apply the algorithm and analysis to the setting of Section~\ref{sec:algorithm-learning}, we use an estimate $\tilde{\mu}_{i,j}$ instead of the actual service rates $\mu_{i,j}$. For the purposes of this section, $\tilde{\mu}_{i,j}=\mu_{i,j}$.

\input{DAM-converge}

Finally, after the queue converges to its matched server and bid, it proceeds by committing to selecting them until the end of the epoch as illustrated in Algorithm~\ref{algo:DAM-commit}. To avoid tie-breaking among requests with the same bid, the price update size is set as $\frac{1}{16}\trafficslack(1-\eta_i)$, where $\eta_i$ is a sufficiently small amount $\eta_i$ drawn uniformly at random in $(0,10^{-9})$. This makes such ties disappear almost surely and is helpful in ensuring that, for any given bid vector, each server is providing service to at most one queue. Such perturbation does not affect our analysis as it only inflates the convergence speed by $\frac{1}{1-10^{-9}}$, which is negligible for the proof of Lemma~\ref{lem:algo-converge-main}.

 \input{DAM-commit}

\subsection{Main result for DAM.K and proof sketch}
The main result of this section is a bound on the time-averaged weighted queue lengths when all queues follow $\textsc{DAM.K}$ (Algorithm \ref{algo:DAM-master}), and is stated as follows.

\begin{theorem} \label{thm:queue-nolearning}
If all queues follow $\textsc{DAM.K}$, then for any $T > 0$, it holds that:
\begin{equation*}\label{eq:thmnolearning}
\expect{\frac{1}{T}\sum_{t=1}^T \sum_{i\in\mathcal{N}}
\lambda_i Q_i(t)} = O\left(\frac{K^2}{\trafficslack^3}\left(\log N + K\right)\checkperiod\right).
\end{equation*}
\end{theorem}
The queue length bound in Theorem~\ref{thm:queue-nolearning} only has a logarithmic dependence on the number of agents. To see how this is possible, note that the queue lengths are weighted by arrival probabilities of agents and the sum of arrival probabilities is upper bounded by $K$ due to traffic slackness. In the case of $\textsc{MaxWeight}$, one can show the weighted sum of queue lengths is upper bounded by $\frac{K}{\varepsilon}$. In our decentralized case the epoch length causes an additional dependence on $N$ as the queues follow $\textsc{DAM.converge}$ to find an approximately optimal matching; this requires a horizon whose length depends on $N$ (see Section \ref{sec:property}). Since  queues send messages to servers in parallel in each time slot, the convergence to the matching only takes $O(\log N)$ time slots.

The proof of the theorem relies on a drift analysis based on the following Lyapunov function, the sum of squares of queue lengths:~$V(\mathbf{Q}(t)) = \sum_{i=1}^N Q^2_i(t).$
The drift at time slot $t$ captures the change in this function, i.e., $\mathbf{D}_t=V\big(\mathbf{Q}(t+1)\big)-V\big(\mathbf{Q}(t)\big)$ and is a random variable taking randomness over the lengths of the queues, arrivals, and services at time $t$. Similarly, the drift in an epoch $\ell=1,2,\ldots$ is $\mathbf{D}_{t_0,\ldots, t_0+\epochlength}=\sum_{\tau=t_0}^{t_0+\epochlength - 1}\mathbf{D}_{\tau}$. Our proof relies on showing that whenever the queue lengths are sufficiently large (as measured by $\|\mathbf{Q}(t_0)\|_1$)at the beginning of an epoch, the expected drift in that epoch is negative; this ensures a bound on the left hand side of Theorem~\ref{thm:queue-nolearning}.

Our algorithm consistently selects the same server for a checking period of length $\checkperiod$ as defined in Eq.~\ref{eq:parameter-setting}. This allows the queue to determine if it is matched since in that case, with high probability, at least one request will be successful. Formally, we define the \emph{good checking event} $\goodevent_\ell$ for an epoch $\ell$ with time slots $[t_0,t_0 + \epochlength - 1]$ by
\[
\goodevent_\ell = \left\{\forall t \in [t_0 + \checkperiod - 1,t_0 + \convlength - 1],j \in \mathcal{K}, \quad \exists t' \in [t - \checkperiod + 1,t], \quad S_{I(j,t'),j}(t') = 1 \right\}.
\]
The following lemma lower bounds its probability; the proof follows standard concentration arguments and is provided in Appendix~\ref{app:good_event}).

\begin{lemma}\label{lem:checkperiod-good}
For epoch $\ell\geq 1$, the good  checking event $\goodevent_{\ell}$ holds with probability at least  $1 - \frac{1}{32}\trafficslack.$ 
\end{lemma}
\noindent The first key result (proof in Section~\ref{sec:property}) shows that, if all queues follow $\textsc{DAM.converge}$ (Algorithm \ref{algo:DAM-queue}) in an epoch then they converge to an approximate max-weight matching.

\begin{lemma}\label{lem:algo-converge-main}
Assume that all queues follow $\textsc{DAM.converge}$ in an epoch $\ell$ with time slots $[t_0,t_0+\convlength-1]$ 
and let $\set{S}$ be the set of feasible solutions in (\ref{eq:primal}). Under the good checking event $\goodevent_\ell$, it holds that $\sigma$ is a matching where for $i \neq i'$, either $\sigma(i)=\perp$ or $\sigma(i) \neq \sigma(i')$ and 
\begin{equation*}
\sum_{i=1}^N w_{i,\sigma(i)} \geq (1 - \frac{1}{16}\trafficslack) \max_{\phi \in \set{S}} \sum_{i=1}^N \sum_{j=1}^K\phi_{i,j}w_{i,j}.
\end{equation*}
\end{lemma}
\noindent Lemma \ref{lem:algo-converge-main} shows that, during the $\textsc{commit}$ phase of the epoch, queues operate based on an approximate max-weight matching. Since that phase is larger than the $\textsc{converge}$ phase, this enables us to bound the expected drift of the epoch in the next lemma (proof in Section~\ref{sec:queue-analysis}).

\begin{lemma}\label{lem:epoch-drift-bound}
If queues follow $\textsc{DAM.converge}$ in time slots $\{t_0,t_0+\convlength-1\}$ and $\textsc{DAM.commit}$ in time slots $\{t_0+\convlength,\ldots, t_0 + \epochlength-1\}$, then the expected drift in the corresponding epoch is: 
\begin{equation*}
\expect{V(\bold{Q}(t_0 + \epochlength)) - V(\bold{Q}(t_0))) \mid \bold{Q}(t_0)} \leq 5780 \frac{K}{\trafficslack^2}\convlength^2 - 2\convlength\sum_{i=1}^N \lambda_i Q_i(t_0).
\end{equation*}
\end{lemma}

\begin{proof}[Proof sketch of Theorem~\ref{thm:queue-nolearning}.]
Summing Lemma~\ref{lem:epoch-drift-bound} over all epochs and by linearity of
expectations:
\begin{align*}&\mspace{25mu}\sum_{\tau=0}^{\ell} \expect{V(\bold{Q}((\tau+1)\epochlength + 1)) - V(\bold{Q}(\tau \epochlength + 1))} \\
&\leq \frac{5780(\ell+1)K\convlength^2}{\trafficslack^2} - 2\convlength\expect{\sum_{i=1}^N \lambda_i \sum_{\tau=0}^{\ell} Q_i(\tau \epochlength + 1)}.
\end{align*}
Since all queues are initially empty, the  left-hand side in the above inequality is non-negative and:
\[
\expect{\sum_{i=1}^N \lambda_i \sum_{\tau=0}^{\ell} Q_i(\tau \epochlength + 1)} \leq \frac{2890(\ell+1)K\convlength}{\trafficslack^2}.
\]
The final guarantee comes from arguing that the queue lengths do not change significantly within an epoch, rearranging, and expanding $\convlength$ and $\epochlength$. The full proof is in Appendix~\ref{app:proof_thm_queue_nolearning}.
\end{proof}

\subsection{Convergence to an approximate max-weight matching (Lemma~\ref{lem:algo-converge-main})} \label{sec:property}

We now investigate how Algorithm \ref{algo:DAM-queue} leads to efficient decentralized scheduling. To do so we show that, with high probability,  
queues and servers converge to a
matching after a bounded number of time slots (Lemma~\ref{lem:algo-converge}), which is approximately max-weight (Lemma~\ref{lem:algo-efficient}). We note that results established in this section hold for general $w_{i,j}$ without assuming $w_{i,j} = \mu_{i,j} Q_i(t_0)$; this will be essential in seamlessly applying them for our analysis in Sections~\ref{sec:algorithm-learning} and \ref{sec:algorithm-ucb}. 

\vspace{0.1in}
\noindent\textbf{Convergence in each epoch.}
Ideally, after a certain number of time slots, queues and servers form a matching that remains unchanged for the rest of the epoch. To formalize that, we say that our decentralized queueing system \emph{converges} in time slot $t$ if every server $j\in\mathcal{K}$ receives a request from at most one queue, i.e., $|R_j(t)| \leq 1$; this induces a matching between queues and servers where $I(j,t)$ is matched to $j$. The following lemma (proof in Appendix~\ref{app:not_unmatched}) shows that, once the system converges to a particular matching, this remains unaltered for the remainder of the epoch.

\begin{lemma}\label{lem:def-converge}
Condition on the good checking event $\goodevent_{\ell}$ and assume that at
time slot $t$,
$|R_j(t)| \leq 1$ for all $j \in \mathcal{K}$. Then $R_j(t') = R_j(t), \forall j \in \mathcal{K}, t' > t.$
\end{lemma}

\noindent Lemma~\ref{lem:def-converge} implies that it is sufficient to study the earliest time slot in which the decentralized queueing system converges. Our convergence analysis is based on a potential function argument. The intuition is that, if a queue is unselected and has positive payoff for at least one server, it updates its price within $\checkperiod+1$ time slots. This is formalized below (proof in Appendix~\ref{app:unmatched_queues_increase_prices}).

\begin{lemma}\label{lem:interval-update}
Fix a time slot $t$ and a server $j$. If $|R_j(t)| > 1$, then all but one queue in $R_j(t)$ will update one of their prices at least once in $[t + 1,t + \checkperiod + 1].$
\end{lemma}

\noindent Armed with these two lemmas, we provide a bound on the number of time slots the decentralized queueing system takes to converge (proof in Appendix~\ref{app:algo_converge}). These time slots are treated as \emph{lost} in our drift analysis but their contribution is dominated by the drift after the system converges.

\begin{lemma}\label{lem:algo-converge}
Condition on the good checking event $\goodevent_{\ell}$. Then a decentralized queueing system where all queues follow \textsc{DAM.converge}, converges in at most $\frac{99K\checkperiod}{\trafficslack}\left(\log N + K\right)$ time slots.
\end{lemma}

\begin{proof}[Proof sketch.] Consider a time slot $t \geq t_0$. Let $\set{A}(t)$ denote the set of queues that have at least one positive-payoff server at time slot $t$. To show the convergence of the queueing system, we define the following potential function: $\Psi(t)=\sum_{i\in\mathcal{A}(t)}C_i(t)$ where $C_i(t)=\sum_{j\in\mathcal{K}}(1+\lceil\frac{w_{i,j}-p_{i,j}}{\beta_{i,j}}\rceil)$ denotes how many times a price of queue $i$ can be updated if all updates are incrementing a price by $\beta_{i,j}=\frac{1}{16}\varepsilon w_{i,j}$. Note that $C_i(t)=\frac{w_{i,j}-p_{i,j}}{\beta_{i,j}}\leq \frac{17K}{\varepsilon}$ 
which implies that $\Psi(t)\leq \frac{17K|\set{A}(t)|}{\varepsilon}$.

By Lemma~\ref{lem:interval-update}, for each interval $[t + 1,t + \checkperiod + 1]$ and each server $j$, at least $|R_j(t)| - 1$ queues update their price. As a result, for any such interval, the potential function $\Psi(t)$ decreases by at least $|\set{A}(t)|-K$. This implies that, for any time $t$, the potential function will stop updating by time $t + \Psi(t)\cdot (\checkperiod+1)$. At this point, the system has converged: every server $j$ is receiving at most $1$ request, i.e., for any $i_1,i_2 \in \set{A}(t), i_1 \neq i_2,$ it holds that $J(i_1,t) \neq J(i_2,t)$. This holds as, if a server received request by more than one queue, one of them would increase their price further decreasing the potential function. This implies that the system converges after $O(\frac{NK}{\epsilon})$. 

To show the final guarantee, we exploit the fact that the decrease in the potential happens at a rate of $|\set{A}(t)| - K$; hence when $|\set{A}(t)| \gg K$, the potential decreases much faster than the above bound implies. By considering geometrically decreasing sizes of $|\set{A}(t)|$, we can replace the linear dependence of $N$ by $\log(N)$ for the case when $|\mathcal{A}(t)|>2K$. Finally, for the last steps, our bounds are linear in the size of $|\mathcal{A}(t)|$ which causes the dependence on $K^2$ in the guarantee. 
\end{proof}

\noindent\textbf{Approximate maximum weight.}
To show that the converged matching is approximately max-weight, we adapt ideas from the centralized Auction Mechanism \cite{bertsekas1988auction} to incorporate the individual prices and allow for a multiplicative rather than an additive guarantee. In particular, for a fixed~$\alpha>0$, a matching~$\sigma$, a price vector~$\hat{\bold{p}}\in\mathbb{R}^K$, and a payoff vector~$\hat{\boldsymbol{\pi}}\in\mathbb{R}^N$, we define $\alpha-$complementary slackness as follows.
\begin{definition}\label{def:comp-slack}
A tuple $(\sigma,\hat{\bold{p}},\hat{\boldsymbol{\pi}})$ satisfies~$\alpha$-complementary slackness if it holds that:
\begin{itemize}
\item $\hat{p}_j=0$ for any unmatched server $j\in\mathcal{K}$, and $\hat{p}_j \geq 0$ for any matched server $j \in \mathcal{K}$,
\item $\hat{\pi}_i=\max\left(\max_{j \in \mathcal{K}} w_{i,j} - \hat{p}_j,0\right)$ for every queue $i\in\mathcal{N}$,
\item $\hat{\pi}_i=0$ if~$i$ is not matched; otherwise, $w_{i,\sigma(i)}>0$ and $\hat{\pi}_i + \hat{p}_{\sigma(i)} \leq (1+\alpha)w_{i,\sigma(i)}$. 
\end{itemize}
\end{definition}
The following lemma connects this notion to an $\alpha$-approximate max-weight matching. The proof follows similar arguments to \cite{bertsekas1988auction} and is provided in Appendix~\ref{app:complementary_slackness_to_approx_matching} for completeness.

\begin{lemma}\label{lem:lowslack-goodweight}
If for a matching $\sigma$, there exist price $\hat{\bold{p}}$ and payoff $\hat{\boldsymbol{\pi}}$ such that the tuple  $(\sigma, \hat{\bold{p}},\hat{\boldsymbol{\pi}})$ satisfies $\alpha$-complementary slackness, then 
$\sum_{i=1}^N w_{i,\sigma(i)} \geq (1-\alpha)\max_{\phi \in \Phi} \sum_{i=1}^N\sum_{j=1}^K w_{i,j}\phi_{i,j}$.
\end{lemma}
To use the above lemma, we show below that the matching $\sigma$ enjoys complementary slackness for an appropriate choice of prices $\hat{\bold{p}}$ and payoffs $\hat{\boldsymbol{\pi}}$. 

\begin{lemma}\label{lem:algo-efficient}
Condition on event $\goodevent_{\ell}$. If the queueing system with queues using $\textsc{DAM.converge}$ 
converges to a matching $\sigma$, then there exist a payoff vector $\hat{\boldsymbol{\pi}}$ and a price vector~$\hat{\bold{p}}$, such that the tuple $(\sigma,\hat{\bold{p}}, \hat{\boldsymbol{\pi}})$ satisfies
$\frac{1}{16}\trafficslack-$complementary slackness. 
\end{lemma}

\begin{proof}[Proof sketch.]
Based on the prices $p_{i,j}(t)$ that each queue maintains during $\textsc{DAM.converge}$, we can define a corresponding payoff $\pi_i(t) = \left(\max_{j \in \mathcal{K}} w_{i,j} - p_{i,j}(t)\right)^+$. Note that, if all queues operate based on the same prices, the complementary slackness conditions are satisfied with respect to this pair. However, unlike the centralized setting, each queue retains its own price and therefore we need price-payoff analogues that are common across all queues. To do so, we define:
\begin{equation}
\label{eq:def-hat-price}
\hat{p}_j(t) = \left\{
\begin{aligned}
&\max_{i \in R_j(t)} p_{i,j}(t),~\text{if }R_j(t) \not = \emptyset \\
&0,~\text{otherwise.}
\end{aligned}
\right.
\end{equation}
Equivalently, we define $\hat{\pi}_i = \max(0, \max_{j \in \mathcal{K}} w_{i,j} - \hat{p}_j(t))$. The proof follows from showing that the matching $\sigma$ that queues converge to (Lemma~\ref{lem:algo-converge}), combined with payoffs $\hat{\boldsymbol{\pi}}$ and prices $\hat{\bold{p}}$, satisfy the approximate complementary slackness. The full proof is provided in Appendix~\ref{app:satisfies_slackness}.
\end{proof}

\begin{proof}[Proof of \Cref{lem:algo-converge-main}]
Condition on $\goodevent_{\ell}$. \Cref{lem:algo-converge} shows that queues converge to a matching $\sigma$ for time slots $[t_0 + \convlength, t_0 + \epochlength - 1].$ Moreover, there exists a vector of payoffs $\boldsymbol{\hat{\pi}}$ and a price vector~$\bold{\hat{p}}$ such that~$(\sigma, \boldsymbol{\hat{\pi}}, \bold{\hat{p}})$ satisfies $\frac{1}{16}\trafficslack-$complementary slackness by \Cref{lem:algo-efficient}. Therefore, Lemma~\ref{lem:lowslack-goodweight} implies that $\sum_{i=1}^N w_{i,\sigma(i)} \geq (1-\frac{\trafficslack}{16})\max_{\phi \in \set{S}}\sum_{i=1}^N \sum_{j=1}^K w_{i,j}\phi_{i,j}$, which completes the proof.
\end{proof}

\subsection{Bounding the expected drift during an epoch (Lemma~\ref{lem:epoch-drift-bound})}\label{sec:queue-analysis}
Having established that the queueing system converges relatively fast to an approximately max-weight matching in most of the epochs, we now show that this implies a negative drift for our Lyapunov function. Unlike the previous part, the analysis in this section relies on having weights induced by the accurate service rates, i.e., $w_{i,j}=\mu_{i,j}Q_i(t_0)$ for an epoch starting at $t_0$.

To bound the total drift in the Lyapunov function, we split its contribution in three terms: a) the increase in the Lyapunov function $V$ in the $\textsc{converge}$ period of the epoch, b) the increase in the Lyapunov function when the good checking event $\goodevent$ fails and therefore we are not guaranteed to converge to an approximate max-weight matching, and c) the expected decrease in the Lyapunov function if the $\textsc{commit}$ period operates with an approximate max-weight matching. Formally:

\begin{equation}\label{eq:drift-decomposition}
\begin{aligned}
\expect{V(\bold{Q}(t_0 + \epochlength)) - V(\bold{Q}(t_0)) \mid \bold{Q}(t_0)} 
&= \underbrace{\expect{V(\bold{Q}(t_0 + \convlength)) - V(\bold{Q}(t_0)) \mid \bold{Q}(t_0)}}_{\text{drift until queues have converged to a matching}} \\
&\hspace{-1in}+ \underbrace{\expect{V(\bold{Q}(t_0 + \epochlength)) - V(\bold{Q}(t_0 + \convlength)) \mid \bold{Q}(t_0), \goodevent_{\ell}^c}\Pr\{\goodevent_{\ell}^c\}}_{\text{drift when the good event does not hold}}\\
&\hspace{-1in}+ \underbrace{\expect{V(\bold{Q}(t_0 + \epochlength)) - V(\bold{Q}(t_0 + \convlength)) \mid \bold{Q}(t_0), \goodevent_{\ell}}\Pr\{\goodevent_{\ell}\}}_{\text{drift after queues have converged to a good matching}} .
\end{aligned}
\end{equation}
The proof follows by showing that, when queue lengths are not too small, the expected decrease by the third term dominates the expected increase caused by the other two. The reason why the third term leads to an expected decrease in the Lyapunov function stems from Lemma~\ref{lem:algo-converge-main}, combined with the stability assumption (Definition~\ref{as:stability}). The expected increase from the first and second terms is bounded by the relative relationship between $\convlength$ (length of $\textsc{converge}$) to $\epochlength$ (length of epoch) and the probability of the good checking event $\goodevent$ not holding respectively. 
Thiis is formalized in the following two lemmas. Lemma~\ref{lem:epoch-drift-converge} bounds the first term and Lemma~\ref{lem:epoch-drift-matching} bounds the sum of the two latter terms; their proofs are provided in Appendix~\ref{app:epoch-drift-converge} and Appendix~\ref{app:epoch-drift-matching} respectively.

\begin{lemma}\label{lem:epoch-drift-converge}
It holds that 
\begin{equation*}
\expect{V(\bold{Q}(t_0+\convlength)) - V(\bold{Q}(t_0)) \mid \bold{Q}(t_0)} \leq 2\convlength\sum_{i=1}^N \lambda_i Q_i(t_0) + K\convlength^2.
\end{equation*}
\end{lemma}

\begin{lemma}\label{lem:epoch-drift-matching}
It holds that
\begin{align*}
&\mspace{32mu}\expect{V(\bold{Q}(t_0+\epochlength)) - V(\bold{Q}(t_0 + \convlength)) \mid \bold{Q}(t_0)} \\
&\leq (\epochlength - \convlength)\left(5K(\epochlength-1+\convlength)-\frac{1}{8}\trafficslack\sum_{i=1}^N \lambda_i Q_i(t_0)\right).
\end{align*}
\end{lemma}

\begin{proof}[Proof of \Cref{lem:epoch-drift-bound}]
By \Cref{lem:epoch-drift-converge} and \Cref{lem:epoch-drift-matching}, it holds
\begin{align*}
&\mspace{25mu}\expect{V(\bold{Q}(t_0 + \epochlength)) - V(\bold{Q}(t_0))) \mid \bold{Q}(t_0)} \\
&=\expect{V(\bold{Q}(t_0 + \convlength)) - V(\bold{Q}(t_0))) \mid \bold{Q}(t_0)} + \expect{V(\bold{Q}(t_0 + \epochlength)) - V(\bold{Q}(t_0 + \convlength))) \mid \bold{Q}(t_0)} \\
&\leq 2\convlength\sum_{i=1}^N \lambda_i Q_i(t_0) + K\convlength^2 + (\epochlength-\convlength)\left(5K(\epochlength-1+\convlength) - \frac{1}{8}\trafficslack \sum_{i=1}^N \lambda_i Q_i(t_0)\right) \\
&\leq 5K(\epochlength+1)^2 + \sum_{i=1}^N \lambda_i Q_i(t_0) \left(2\convlength - \frac{\epochlength-\convlength}{8}\trafficslack\right).
\end{align*}
Since $\epochlength = \left \lceil (\frac{32}{\trafficslack} + 1)\convlength\right \rceil$, it holds that $\epochlength + 1 \leq \frac{34}{\trafficslack}\convlength.$ In addition, $\frac{\epochlength-\convlength}{8}\trafficslack \geq 4\convlength.$ Plugging this into the above, we obtain the desired result.
\end{proof}

%% file: figure/algofig.tex
\tikzset{every picture/.style={line width=0.75pt}} 

\begin{tikzpicture}[x=1.2pt,y=0.6pt,yscale=-1,xscale=1]

\draw    (99.5,167.24) -- (570.02,167.24) ;
\draw [shift={(572.02,167.24)}, rotate = 180] [color={rgb, 255:red, 0; green, 0; blue, 0 }  ][line width=0.75]    (10.93,-3.29) .. controls (6.95,-1.4) and (3.31,-0.3) .. (0,0) .. controls (3.31,0.3) and (6.95,1.4) .. (10.93,3.29)   ;
\draw    (99.5,167.24) -- (547.02,167.24) (109.5,163.24) -- (109.5,171.24)(119.5,163.24) -- (119.5,171.24)(129.5,163.24) -- (129.5,171.24)(139.5,163.24) -- (139.5,171.24)(149.5,163.24) -- (149.5,171.24)(159.5,163.24) -- (159.5,171.24)(169.5,163.24) -- (169.5,171.24)(179.5,163.24) -- (179.5,171.24)(189.5,163.24) -- (189.5,171.24)(199.5,163.24) -- (199.5,171.24)(209.5,163.24) -- (209.5,171.24)(219.5,163.24) -- (219.5,171.24)(229.5,163.24) -- (229.5,171.24)(239.5,163.24) -- (239.5,171.24)(249.5,163.24) -- (249.5,171.24)(259.5,163.24) -- (259.5,171.24)(269.5,163.24) -- (269.5,171.24)(279.5,163.24) -- (279.5,171.24)(289.5,163.24) -- (289.5,171.24)(299.5,163.24) -- (299.5,171.24)(309.5,163.24) -- (309.5,171.24)(319.5,163.24) -- (319.5,171.24)(329.5,163.24) -- (329.5,171.24)(339.5,163.24) -- (339.5,171.24)(349.5,163.24) -- (349.5,171.24)(359.5,163.24) -- (359.5,171.24)(369.5,163.24) -- (369.5,171.24)(379.5,163.24) -- (379.5,171.24)(389.5,163.24) -- (389.5,171.24)(399.5,163.24) -- (399.5,171.24)(409.5,163.24) -- (409.5,171.24)(419.5,163.24) -- (419.5,171.24)(429.5,163.24) -- (429.5,171.24)(439.5,163.24) -- (439.5,171.24)(449.5,163.24) -- (449.5,171.24)(459.5,163.24) -- (459.5,171.24)(469.5,163.24) -- (469.5,171.24)(479.5,163.24) -- (479.5,171.24)(489.5,163.24) -- (489.5,171.24)(499.5,163.24) -- (499.5,171.24)(509.5,163.24) -- (509.5,171.24)(519.5,163.24) -- (519.5,171.24)(529.5,163.24) -- (529.5,171.24)(539.5,163.24) -- (539.5,171.24) ;
\draw    (100,120) -- (100,210) ;
\draw    (310,120) -- (310,210) ;
\draw    (520,120) -- (520,210) ;
\draw   (310.68,213.41) .. controls (310.67,218.08) and (312.99,220.42) .. (317.66,220.44) -- (410.71,220.83) .. controls (417.38,220.86) and (420.7,223.2) .. (420.68,227.87) .. controls (420.7,223.2) and (424.04,220.88) .. (430.71,220.91)(427.71,220.9) -- (511.65,221.25) .. controls (516.32,221.27) and (518.66,218.95) .. (518.68,214.28) ;
\draw    (170,140) -- (170,190) ;
\draw    (380,140) -- (380,190) ;
\draw   (100,190) .. controls (100.09,194.67) and (102.47,196.95) .. (107.14,196.86) -- (124.46,196.52) .. controls (131.13,196.39) and (134.51,198.65) .. (134.6,203.32) .. controls (134.51,198.65) and (137.79,196.25) .. (144.46,196.12)(141.46,196.18) -- (163.15,195.76) .. controls (167.82,195.67) and (170.1,193.29) .. (170.01,188.62) ;
\draw   (311.01,189.62) .. controls (311.01,194.29) and (313.34,196.62) .. (318.01,196.62) -- (334.23,196.62) .. controls (340.9,196.62) and (344.23,198.95) .. (344.23,203.62) .. controls (344.23,198.95) and (347.56,196.62) .. (354.23,196.62)(351.23,196.62) -- (373.01,196.62) .. controls (377.68,196.62) and (380.01,194.29) .. (380.01,189.62) ;
\draw   (100.35,212.41) .. controls (100.33,217.08) and (102.65,219.42) .. (107.32,219.44) -- (200.37,219.83) .. controls (207.04,219.86) and (210.36,222.2) .. (210.34,226.87) .. controls (210.36,222.2) and (213.7,219.88) .. (220.37,219.91)(217.37,219.9) -- (301.31,220.25) .. controls (305.98,220.27) and (308.32,217.95) .. (308.34,213.28) ;

\draw (559,172) node [anchor=north west][inner sep=0.75pt]    {\Large $t$};
\draw (414,226) node [anchor=north west][inner sep=0.75pt]    {\Large $\epochlength$};
\draw (124,199) node [anchor=north west][inner sep=0.75pt]    {\textbf{$\convlength$}};
\draw (333,199) node [anchor=north west][inner sep=0.75pt]    {\textbf{$\convlength$}};
\draw (111,143) node [anchor=north west][inner sep=0.75pt]   [align=left] {\large converge };
\draw (321,143) node [anchor=north west][inner sep=0.75pt]   [align=left] {\large converge };
\draw (217,139) node [anchor=north west][inner sep=0.75pt]   [align=left] {\large commit};
\draw (427,139) node [anchor=north west][inner sep=0.75pt]   [align=left] {\large commit};
\draw (203,226) node [anchor=north west][inner sep=0.75pt]    {\Large $\epochlength$};
\draw (181,99) node [anchor=north west][inner sep=0.75pt]   [align=left] {\Large Epoch $\displaystyle \ell $};
\draw (379,99) node [anchor=north west][inner sep=0.75pt]   [align=left] {\Large Epoch $\displaystyle \ell +1$};

\draw   (170, 167.24) circle [x radius= 5, y radius= 5]   ;
\draw   (170, 167.24) circle [x radius= 5, y radius= 5]   ;
\draw   (310, 167.24) circle [x radius= 5, y radius= 5]   ;
\draw   (310, 167.24) circle [x radius= 5, y radius= 5]   ;
\draw   (380, 167.24) circle [x radius= 5, y radius= 5]   ;
\draw   (380, 167.24) circle [x radius= 5, y radius= 5]   ;
\draw   (520, 167.24) circle [x radius= 5, y radius= 5]   ;
\draw   (520, 167.24) circle [x radius= 5, y radius= 5]   ;
\draw   (100, 167.24) circle [x radius= 5, y radius= 5]   ;
\draw   (100, 167.24) circle [x radius= 5, y radius= 5]   ;
\end{tikzpicture}

%% file: DAM-master.tex
\begin{algorithm}[H]
\LinesNumbered
\DontPrintSemicolon
\caption{Decentralized Auction Mechanism for Known service rates ($\textsc{DAM.K}$)
\label{algo:DAM-master}}
\SetKwInOut{Input}{input}\SetKwInOut{Output}{output}
\Input{Traffic slackness $\trafficslack$; Lower bound of nonzero service rates $\mulower$; Service rates $\{\mu_{i,j}\}_{j\in\mathcal{K}}$}
Initialize check period $\checkperiod$, converging length $\convlength$, and epoch length $\epochlength$ as in \eqref{eq:parameter-setting}\;
\tcc{initialize a random perturbation in $(0,10^{-9})$ for tie-breaking}
$\eta_i \gets \text{a uniform random number in }(0,10^{-9})$\;
\For{$\ell = 1\ldots$}{
$t_0\gets (\ell-1)\epochlength+1$\;
    \tcc{queues converge to matching $\sigma$ and bids $\bold{p}$ in  
     $\convlength$ time slots}
    $\sigma(i),p_{i,\sigma(i)}\gets \textsc{DAM.converge}(t_0,\convlength, \checkperiod,\trafficslack,\{\mu_{i,j}\}_{j \in \set{K}},\eta_i)$\;
    \tcc{queues submit jobs to converged server until epoch's end}
    call $\textsc{DAM.commit}(t_0 + \convlength, t_0 + \epochlength - 1, \sigma(i), p_{i,\sigma(i)})$
}
\end{algorithm}

%% file: DAM-converge.tex
\begin{algorithm}[H]
\LinesNumbered
\DontPrintSemicolon
\caption{$\textsc{DAM.converge}$}
\label{algo:DAM-queue}
\SetKwInOut{Input}{input}\SetKwInOut{Output}{output}
\Input{epoch start $t_0$; Converge length $\convlength$; Check period $\checkperiod$; Queue length $Q_i(t_0)$;\\ Traffic slackness $\trafficslack$; Estimated service rates $\{\tilde{\mu}_{i,j}\}_{j\in\mathcal{K}}$; Price perturbation $\eta_i$}
\tcc{Initialize weights $w_{i,j}$, price $p_{i,j}$, and the event log $\tau_i$.}
$w_{i,j} \gets \tilde{\mu}_{i,j}Q_i(t_0)$; \label{algoline:set-weight}
$p_{i,j}(t_0) \gets 0$ for all $j \in \mathcal{K}$;
 $\tau_i(t_0-1) \gets t_0 -1$ \;
\For{$t = t_0$ \KwTo $t_0 + \convlength - 1$}{
        \If{$t > t_0$ {\bf and} $t - \tau_i(t - 1) \leq \checkperiod$} { \label{algoline:no-update-price}
            \tcc{Within checking period. No update on price.}
            $J(i,t) \gets J(i,t-1)$  and $p_{i,j}(t) \gets p_{i,j}(t-1), \forall j \in \mathcal{K}$ \label{algoline:no-update-price-step}
        }
        \Else {
            \tcc{Request is not selected. Update price.}
            $j^{\star}
            \gets \arg \max_j w_{i,j} - p_{i,j}(t - 1)$\;
            \If {$w_{i,j^{\star}
            } - p_{i,j^{\star}
            }(t - 1) > 0$} {  \label{algoline:when-to-request}
                $p_{i,j^{\star}
                }(t) \gets p_{i,j^{\star}
                }(t - 1) + \frac{1}{16}\trafficslack(1-\eta_i) w_{i,j^{\star}
                }$ and
                $J(i,t) \gets j^{\star}
                $   \label{algoline:price-update}
            }
            \Else {
                \tcc{No gain on choosing any server. Stop requesting.}
                $J(i,t) \gets \perp$  and   $p_{i,J(i,t)}\gets 0 $
            }
        }
        Send request to server $J(i,t)$ with price $p_{i,J(i,t)}(t)$ \;
        $\tau_i(t) \gets \tau_i(t - 1)$ \;
        \tcc{Update $\tau_i(t)$ when either price changes or a request is successful}
        \If {Price updates for some server $j$ {\bf or} Request to $J(i,t)$ is successful}{
            $\tau_i(t) \gets t$
        }
}
\tcc{set the committed server and price}
$\sigma(i) \gets J(i,t_0 + \convlength - 1)$ and
$p_{i,\sigma(i)} \gets p_{i,J(i, t_0 + \convlength - 1)}(t_0 + \convlength - 1)$ \;
\Return{$\sigma(i), p_{i,\sigma(i)}$}
\end{algorithm}

%% file: DAM-commit.tex
\begin{algorithm}[H]
\LinesNumbered
\DontPrintSemicolon
\caption{$\textsc{DAM.commit}$}
\label{algo:DAM-commit}
\SetKwInOut{Input}{input}\SetKwInOut{Output}{output}
\Input{Start $t_s$; End $t_e$; Committed server $j$; Committed price $p$}
\For{$t = t_s$ \KwTo $t_e - 1$}{
    
    \If {$j \neq \perp$}{
        Send request to server $j$ with price $p$
    }
}
\end{algorithm}

%% file: algo-learning.tex
In this section, we remove the assumption that service rates are known at 
the beginning of the algorithm. Instead, each queue does its own exploration to learn these parameters.
Our approach follows an explore-exploit paradigm similar to the $\textsc{Q-UCB}$ approach of Krishnasamy et al.~\cite{KrishnasamySenJohariShakkottai21} who provide a learning algorithm for the centralized and symmetric version based on \emph{forced exploration}. In the next section we replace this forced exploration with adaptive exploration.

\subsection{Algorithmic crux: Coordinating exploration despite queue interference}

The idea of forced exploration in a centralized algorithm such as \textsc{Q-UCB} is that some time slots are devoted to exploring queue-server matchings to estimate their efficacy. To ensure that the exploration rounds are not causing significant overhead in the performance of the algorithm, the exploration probability is decaying over time. This idea of decaying exploration probabilities seamlessly extends to our decentralized setting, with the slight modification that time slots need to now be replaced by epochs. In particular, our algorithm $\textsc{DAM.FE}$ (Algorithm~\ref{algo:DAM-online-master}) decides if the queue will explore ($E_\ell=1$) or exploit ($E_{\ell}=0$); the exploration probability is equal to $\min(1,\frac{K}{\ell^\expratio})$ for some exploration parameter~$\expratio\in(0,1)$. At the beginning of an exploration epoch, the queue samples uniformly a server (line~\ref{algoline:uniform_server} in Algorithm~\ref{algo:DAM-online-master}) and consistently selects it throughout the epoch.

However, in the decentralized setting, queues can no longer coordinate their exploration and therefore may explore  servers that are also requested by another queue -- this interference could bias the estimates computed through this exploration. 
To circumvent this roadblock, we introduce two important levers in our forced exploration procedure.
First, in exploration epochs, queues make bids that are higher than the ones of any queue which is a slightly boosted version of $t_0 + \epochlength + 1$ (line~\ref{algoline:high_bid} in Algorithm~\ref{algo:DAM-online-master}), while a queue that is not exploring has bid upper bounded by its queue length in time slot $t_0$ (line \ref{algoline:set-weight} in Algorithm \ref{algo:DAM-queue} and line \ref{algoline:set-ucb-rate} in Algorithm \ref{algo:DAM-online-master}), which is at most $t_0$. Therefore, the bid from an exploring queue always dominate that from an exploiting queue.

\input{DAM-online-master}

Second, to ensure a consistent tie-breaking across exploring queues, we boost the bid by a random number $\eta_i$ that is fixed for each queue (line~\ref{algoline:tiebreaking} in Algorithm~\ref{algo:DAM-online-master}). Since  exploration bids are constant within an epoch and larger than exploitation bids, if there exists at least one exploring queue, the server consistently selects the same exploring queue. Hence, to ensure that our estimates are unbiased, we need to only update the estimates for that particular queue. To do so, our estimation procedure $\textsc{DAM.update}$ (Algorithm~\ref{algo:DAM-online-estimation}) only makes updates after the first success of a queue: after this point, we know that that queue is the selected queue and therefore the remaining samples are unbiased. The latter claim is formalized in the following lemma (proof in Appendix~~\ref{app:dam_update}).

\begin{lemma}\label{lem:update-unbiased}
Let $t_0$ be the start of an epoch and let $\mathcal{Y}_{i,j}(t_0)$ by the set of samples between queue $i\in \mathcal{N}$ and server $j\in \mathcal{K}$ collected by Algorithm \ref{algo:DAM-online-estimation} up to time slot $t_0$. Then all random variables in $\mathcal{Y}_{i,j}(t_0)$ are almost surely independent and Bernoulli distributed with mean $\mu_{i,j}.$
\end{lemma}

Finally, in exploitation epochs, the queues use optimistic estimates for the service rates to converge to a matching (line~\ref{algoline:set-ucb-rate} in Algorithm~\ref{algo:DAM-online-master}). To handle service rates that are $0$, if a server has never served a queue, it maintains an estimate of $0$.

\input{DAM-online-estimation}

\subsection{Main result for DAM.FE and proof sketch}\label{sec:dam-fe-sketch}
To provide the queue-length guarantee of $\textsc{DAM.FE}$, we define 
\begin{align}C_1 &= \max\left(80, \frac{1}{K^2\checkperiod \mulower^2}\right), C_2 = \max\left(80\ln(2NK\epochlength), \frac{\ln(2\epochlength)}{K^2\checkperiod\mulower^2}\right),\notag\\
\ell_0 &:= \left\lceil \max\left((NK\epochlength)^{1/\expratio}+1, (2C_2)^{1/(1-\expratio)}, \left(4C_1/(1-\expratio)\right)^{2/(1-\expratio)}\right)\right\rceil, \label{eq:set-of-l0}
\end{align}
and correspondingly $T_0=\ell_0 \epochlength+1$ as the first time slot after which all queues have converged to approximately correct estimates. The performance is then bounded by the following theorem.
\begin{theorem}\label{thm:queue-learning}
If all queues follow $\textsc{DAM.FE}$, then for any $T > 0$ it holds that:
\begin{equation}\label{eq:thmlearning}
\expect{\frac{1}{T}\sum_{t=1}^T \sum_{i=1}^N \lambda_i Q_i(t)} = O\left(\frac{K^2}{\trafficslack^3}\left(\log N + K\right)\checkperiod +  \frac{K\min(T,T_0)^2}{\trafficslack T}\right).
\end{equation}
\end{theorem}
Theorem~\ref{thm:queue-learning} basically shows that, after the $\ell_0$-th epoch, queues behave as if they know service rates accurately, and thus enjoy the same queue length bound in \Cref{thm:queue-nolearning}. The setting of $\ell_0$ stems from two desirable properties. The first value in Eq.~\eqref{eq:set-of-l0} guarantees that the cost of exploration after the $\ell_0$-th epoch is small, i.e., after the $\ell_0$-th epoch, queues run Algorithm~\ref{algo:DAM-queue} to clear jobs with probability at least $1 - \frac{NK}{NK\epochlength}=1-\frac{1}{\epochlength}$. Second, the last two values in Eq.~\eqref{eq:set-of-l0} ensure that, after the $\ell_0$ epoch, with high probability, we collected enough samples to have refined estimates. 

Specifically, the second property relies on the fact that after $T_0$, the estimated service rate $\hat{\mu}_{i,j}$ is close to the ground truth $\mu_{i,j}$ with high probability. In particular, we define \emph{accurate service rate estimation} in the sense that at a starting time slot $t_0$, we have for any $i \in \set{N}, j \in \set{K}$ with $\mu_{i,j} > 0$ that
a) the estimation is within $\Delta_{i,j}(t_0) \coloneqq \sqrt{\frac{3\ln t_0}{n_{i,j}(t_0)}}$, i.e.,~$|\hat{\mu}_{i,j}(t_0) - \mu_{i,j}| \leq \Delta_{i,j}(t_0)$ and b) the confidence bound is small enough such that $\Delta_{i,j}(t_0) \leq \frac{1}{16}\trafficslack\mulower.$
We show that after the $\ell_0$-th epoch, queues have accurate service rate estimations with high probability. To formalize the above two properties, we focus on an epoch starting at time slot $t_0$ and define the following three events, 
\begin{itemize}
    \item event $\set{E}_P$ where there exists at least one queue choosing to explore for this epoch;
    \item event $\set{E}_W$ where some service rate estimations are incorrect, i.e., there exists a pair of $(i,j) \in \set{N} \times \set{K}$ with $\mu_{i,j} > 0$ such that either the estimation is out of the confidence bound or the confidence bound $\Delta_{i,j}(t_0)$ exceeds the threshold $\frac{1}{16}\trafficslack\mulower$;
    \item event $\set{E}_P^c \cap \set{E}_W^c$, i.e., all queues exploit and queues have accurate service rate estimations.
\end{itemize}

A key lemma establishes that, after the $\ell_0$-th epoch, queues stop exploration and accurately estimate service rates, i.e., $\set{E}_P$ and $\set{E}_W$ happen with small probability (proof in Appendix~\ref{app:damfe_convergence_estimates}).
\begin{lemma}\label{lem:estimate-good}
If $t_0 \geq T_0$, queues explore with low probability and estimate accurate service rate estimations with high probability, i.e., $\Pr\{\set{E}_P\} \leq \frac{1}{\epochlength}$ and $\Pr\{\set{E}_W\} \leq \frac{1}{NKt_0^2}.$ 
\end{lemma}
Combining this lemma with the analysis of Lemma~\ref{lem:epoch-drift-converge}, we obtain the following bounds on the contribution of those two events on the expected drift during a later epoch (proof in Appendix~\ref{app:exploration_incorrect_estimates}).

\begin{lemma}\label{cor:bound-drift-ew} \label{lem:bound-drift-ep}
If $t_0 \geq T_0$, it holds that:
\begin{align*} &\expect{V(\bold{Q}(t_0 + \epochlength)) - V(\bold{Q}(t_0)) \mid \set{E}_P}\Pr\{\set{E}_P\}\leq 2\sum_{i=1}^N \lambda_i \expect{Q_i(t_0)} + K\epochlength;\\&\expect{V(\bold{Q}(t_0 + \epochlength)) - V(\bold{Q}(t_0)) \mid \set{E}_P^c \cap \set{E}_W}\Pr\{\set{E}_P^c \cap \set{E}_W\} \leq 3.\end{align*}
\end{lemma}
What is left is to account for estimation errors after epoch $\ell_0$ and update our drift analysis of Lemma~\ref{lem:epoch-drift-bound} to the case where no queue is exploring and all operate with refined-enough but not perfectly accurate estimates. This is shown in the following lemma (proof in Appendix~\ref{app:exploitation_refined_estimates})

\begin{lemma} \label{lem:bound-drift-ei}
If $t_0 \geq T_0$, it holds that: \begin{align*}
\expect{V(\bold{Q}(t_0 + \epochlength)) - V(\bold{Q}(t_0)) \mid \set{E}_P^c \cap \set{E}_W^c}\Pr\{\set{E}_P^c \cap \set{E}_W^c\}\leq 5781 \frac{K}{\trafficslack^2}\convlength^2 - \frac{3}{2}\convlength\sum_{i=1}^N \lambda_i \expect{Q_i(t_0)}.\end{align*}
\end{lemma}

\begin{proof}[Proof sketch of Theorem \ref{thm:queue-learning}]
To capture the effect of erroneous estimation and exploration of the queues, we expand the expected drift within an epoch into terms that depend on conditional expectations conditioned on events $\set{E}_P,\set{E}_W$ and then apply Lemmas~\ref{lem:bound-drift-ep} and \ref{lem:bound-drift-ei}:
\begin{align*}
\expect{V(\bold{Q}(t_0 + \epochlength)) - V(\bold{Q}(t_0))} 
&= \expect{V(\bold{Q}(t_0 + \epochlength)) - V(\bold{Q}(t_0)) \mid \set{E}_P}\Pr\{\set{E}_P\} \\
&\mspace{25mu} + \expect{V(\bold{Q}(t_0 + \epochlength)) - V(\bold{Q}(t_0)) \mid \set{E}_P^c \cap \set{E}_W}\Pr\{\set{E}_P^c \cap \set{E}_W\} \\
&\mspace{25mu} + \expect{V(\bold{Q}(t_0 + \epochlength)) - V(\bold{Q}(t_0)) \mid \set{E}_P^c \cap \set{E}_W^c}\Pr\{\set{E}_P^c \cap \set{E}_W^c\}
\\&\leq 5782\frac{K}{\trafficslack^2}\convlength^2 - \convlength\sum_{i=1}^N \lambda_i \expect{Q_i(t_0)}.
\end{align*}
The final guarantee follows similarly to the proof of Theorem~\ref{thm:queue-nolearning} by summing across epochs and using the fact that the queues are initially empty (which ensures that this sum is non-negative). The full proof is provided in Appendix~\ref{app:damfe_guarantee}.
\end{proof}

%% file: DAM-online-master.tex
\begin{algorithm}[H]
\LinesNumbered
\DontPrintSemicolon
\caption{Decentralized Auction Mechanism with Forced Exploration ($\textsc{DAM.FE}$)}
\label{algo:DAM-online-master}
\SetKwInOut{Input}{input}\SetKwInOut{Output}{output}
\Input{
Traffic slackness $\trafficslack$; lower bound of non-$0$ service rates $\mulower$; exploration parameter~$\expratio$}
Initialize check period $\checkperiod$, converging length $\convlength$, and 
epoch length $\epochlength$ as in
\eqref{eq:parameter-setting}\;
\tcc{initialize a random price perturbation in $(0,10^{-9})$ for tie-breaking}
$\eta_i \gets \text{a uniform random number in }(0,10^{-9})$\; \label{algoline:tiebreaking}
Sample mean $\hat{\mu}_{i,j}(t_0) \longleftarrow 0$, number of samples $n_{i,j}(t_0) \longleftarrow 0$ \;
\For{$\ell = 1\ldots$}{
    $t_0\gets (\ell-1)\epochlength+1$\;
    $E_{\ell} \longleftarrow $ a Bernoulli sample with mean $\min(1, \frac{K}{\ell^{\expratio}})$ \; \label{algoline:explore-rate}
    \If{$E_{\ell} = 1$}{
        \tcc{randomly choose a server to explore and request with high bid}
        $\sigma(i) \longleftarrow $ a uniform sample from $\set{K}$ \;\label{algoline:uniform_server}
        call $\textsc{DAM.commit}(t_0,t_0+\epochlength-1,\sigma(i), (t_0 + \epochlength + 1) (1+\eta_i))$\;\label{algoline:high_bid}
        \tcc{update estimates with all samples after the first success} $\hat{\mu}_{i,\sigma(i)}(t_0 + \epochlength), n_{i,\sigma(i)}(t_0 + \epochlength) \gets \textsc{DAM.update}(t_0,t_0 + \epochlength - 1, \sigma(i), \hat{\mu}_{i,\sigma(i)}(t_0), n_{i,\sigma(i)}(t_0))$ 
    }
    \Else{
    \tcc{optimistically estimate service rates with at least one sample}
    $\bar{\mu}_{i,j}(t_0) = \left\{
    \begin{aligned}
    &\min\left(1, \hat{\mu}_{i,j}(t_0) + \sqrt{\frac{3\ln t_0}{n_{i,j}(t_0)}}\right) & \text{if }n_{i,j}(t_0) \geq 1 \\
    &0 & \text{otherwise.}
    \end{aligned}
    \right.$ \label{algoline:set-ucb-rate}\;
    \tcc{queues converge to matching $\sigma$ and bids $\bold{p}$ in      $\convlength$ time slots}
    $\sigma(i),p_{i,\sigma(i)}\gets \textsc{DAM.converge}(t_0,\convlength, \checkperiod,\trafficslack,\{\bar{\mu}_{i,j}(t_0)\}_{j \in \set{K}},\eta_i)$\;
    \tcc{queues submit jobs to converged server until epoch's end} 
    call $\textsc{DAM.commit}(t_0 + \convlength, t_0 + \epochlength - 1, \sigma(i), p_{i,\sigma(i)})$
    }
}
\end{algorithm}

%% file: DAM-online-estimation.tex
\begin{algorithm}[H]
\LinesNumbered
\DontPrintSemicolon
\caption{Estimation updates with in-epoch samples ($\textsc{DAM.update}$)
}
\label{algo:DAM-online-estimation}
\SetKwInOut{Input}{input}\SetKwInOut{Output}{output}
\Input{Starting time slot $t_s$; Ending time slot $t_e$; Committed server $j$; service rate estimates $\{\hat{\mu}_{i,j}(t_0)\}$; number of samples $\{n_{i,j}(t_0)\}$;}
\tcc{Denote $Y_{i,j}(t)$ by $\mathbbm{1}\{\text{request to  }j \text{ is successful at time slot }t\}, \forall t \in [t_s,t_e]$}
\textbf{if } $\sum_{t=t_s}^{t_e} Y_{i,j}(t) = 0$ \textbf{then return } $\hat{\mu}_{i,j}(t_0),n_{i,j}(t_0)$\;
$\tau \gets \min \{t \in\{t_s,\ldots, t_e\} \colon Y_{i,j}(t) = 1\}$\;
\tcc{Update sample mean $\hat{\mu}_{i,j}$ with new samples $Y_{i,j}(\tau + 1),\ldots,Y_{i,j}(t_e)$}
$\hat{n} \gets n_{i,j}(t_0) + t_e - \tau$ \text{ and }
$\hat{\mu} \gets \frac{n_{i,j}(t_0)\hat{\mu}_{i,j}+\sum_{t=\tau + 1}^{t_e} Y_{i,j}(t)}{\hat{n}}$ \;
\Return{$\hat{\mu}, \hat{n}$}
\end{algorithm}

%% file: algo-ucb.tex
The previous section relied on forced exploration: queues decided to explore with some preselected probability in order to update their service-rate estimates. A natural question that arises is whether we can reduce the amount of exploration via a more adaptive approach similar to how algorithms such as \textsc{UCB} achieve that in multi-armed bandits. 
In this section, we answer this question affirmatively by designing \textsc{DAM.UCB}, a queueing analogue of \textsc{UCB}. In Section~\ref{sec:dynamic}, we then show that the adaptivity of \textsc{DAM.UCB} allows it to seamlessly extend to more dynamic settings.

Before describing the algorithm and the result, we note that, unlike the previous sections, the results in this section require that all service rates are non-zero, i.e., $\mu_{i,j} \geq \mulower > 0$ for every pair of $(i,j) \in \set{N} \times \set{K}$. This occurs because we employ implicit exploration and therefore we cannot distinguish between pairs with $\mu_{i,j}=0$ and queues that are not served due to interference.

\subsection{Our algorithm and main result}
While overall similar to \textsc{DAM.FE}, the key distinction of DAM.UCB (Algorithm~\ref{algo:DAM-ucb}) is that queues explore servers implicitly by optimistically estimating service rates, and gradually improving estimations through samples collected in the commit phase of the DAM algorithm. To collect samples, DAM.UCB calls the estimation update function (Algorithm~\ref{algo:DAM-online-estimation}) and therefore only collects samples after the first success in $[t_0 + \convlength, t_0 + \epochlength - 1]$. Moreover, every queue keeps requesting the same server with the same bid in $[t_0 + \convlength, t_0 + \epochlength - 1]$, which implies that $I(j,t) = I(j,t')$ for every $j \in \set{K}, t,t' \in [t_0 + \convlength,t_0 + \epochlength - 1]$. The two above design choices are similar to the collection process of $\textsc{DAM.FE}$ and again ensure that those samples are unbiased and independent.

\input{DAM-ucb}

\begin{theorem}\label{thm:queue-ucb} Assume $\mu_{i,j} > 0$ and $\lambda_i > 0$ for every $(i,j) \in \set{N} \times \set{K}$, and let $\lambda^{\star} = \sum_{i=1}^N \frac{1}{\lambda_i}.$
If all queues follow DAM.UCB, then for any $T > 0$, it holds that:
\[
\expect{\frac{1}{T}\sum_{t=1}^T\sum_{i=1}^N \lambda_i Q_i(t)} = O\left(\frac{K^2\checkperiod}{\trafficslack^3}(\log N + K) + \lambda^{\star}\frac{(K\epochlength)^2}{\trafficslack^2}\frac{\ln^2 (T+K)}{T}\right).
\]
\end{theorem}
We now contrast the bounds in Theorems~\ref{thm:queue-learning} and \ref{thm:queue-ucb}. In both bounds, the first term is identical to the guarantee of \textsc{DAM.K} (Theorem~\ref{thm:queue-nolearning}) and is due to decentralization even without the learning component. The second term is due to learning and vanishes as $T\rightarrow \infty$. The rate in which the second term of Theorem~\ref{thm:queue-learning} vanishes is ~$O\left(\frac{K(NK\epochlength)^{2/\expratio}\epochlength^2}{\trafficslack T}\right)$ as opposed to $\small{\tilde{O}\left(\lambda^{\star}\left(\frac{K\epochlength}{\trafficslack}\right)^2\frac{1}{T}\right)}$ in Theorem~\ref{thm:queue-ucb}. Even if we set the exploration parameter $\expratio=1$ (the value that minimizes the corresponding term), the second term of the guarantee for \textsc{DAM.UCB} is still superior by a factor of $\tilde{O}\left(K^5N^2\trafficslack^{-3}/\lambda^{\star}\right)$. This improvement is due to the adaptive exploration of \textsc{DAM.UCB}: 1) the frequency that a queue explores on a server depends on the potential service speed (similar to \textsc{UCB} in  bandits); 2) the frequency that a queue explores depends on its difficulty to stabilize (reflected by its queue length) so that queues do not waste epochs because of exploration of queues that are easy to stabilize. The second advantage is also highlighted in the dynamic setting (Section~\ref{sec:dynamic}).

\subsection{Proof sketch of main result for DAM.UCB}
We denote the total number of epochs by $\ell_T = \lceil\frac{T}{\epochlength}\rceil$ and focus on epoch $\tau \leq \ell_T$.  Let $t_0(\tau) = (\tau - 1)\epochlength + 1$. We define $\sigma_{\tau}$ as the output of $\textsc{DAM.converge}$ in Line~\ref{algoline:dam-ucb-converge} of Algorithm~\ref{algo:DAM-ucb}; note that $\sigma_{\tau}$ may not be a matching. We will also omit the dependence on $\tau$ when clear from context.

We now define $\set{G}_{\tau}$ as the event where for all time slots $t \in \{t_0 + \checkperiod - 1,\ldots,t_0 + 2\convlength - 1\}$ and server $j \in \set{K}$, there exists $t' \in [t - \checkperiod + 1,t]$, such that $S_{I(j,t'),j}(t') = 1.$ The difference between $\set{G}_{\tau}$ and the good checking event $\set{E}_{\tau}$ defined in Section~\ref{sec:algorithm-nolearning} is that $\set{G}_\tau$ posits the condition for all $t$ within a period of two exploration lengths $2\convlength$ rather than one. Hence $\set{G}_{\tau} \subseteq \set{E}_{\tau}$, which implies that, conditioning on $\set{G}_{\tau}$,
$\sigma_{\tau}$ must be
a matching by Lemma~\ref{lem:algo-converge-main}. 
When $\set{G}_{\tau}$ holds,
any queue that is matched during the interval $[t_0 + \convlength,t_0 + 2\convlength - 1]$ receives a successful request and updates its estimates.

Let $\sigma^{\star}_{\tau}$ be the max-weight matching with weight $w_{i,j}=\mu_{i,j}Q_i(\tau \epochlength + 1).$ The key proof idea is to show that under DAM.UCB, the drift under the schedule $\sigma_{\tau}$ produced when all queues follow \textsc{DAM.UCB} is close to that under $\sigma^{\star}_{\tau}$ and thus the respective averaged queue lengths are also close. The next lemma  (proof in Appendix~\ref{sec:ucb-decompose}) formalizes this by decomposing the drift of the entire time horizon into a) drift under matching $\sigma^{\star}_{\tau}$ and b) total weight differences between $\sigma^*_{\tau}$ and $\sigma_{\tau}$.

\begin{lemma}\label{lem:ucb-drift-decompose}
The total drift within the time-horizon is upper bounded by
\begin{align*}
\expect{V(\bold{Q}(\ell_T \epochlength + 1)) - V(\bold{Q}(1))} &\leq  \left(4\convlength + (2 + \trafficslack / 8)(\epochlength - 2\convlength)\right)\sum_{\tau=1}^{\ell_T}\sum_{i=1}^N \lambda_i \expect{Q_i(t_0(\tau))} \\
&\hspace{-0.2in}-2(1-\trafficslack/16)(\epochlength - 2\convlength)\sum_{\tau=1}^{\ell_T}\sum_{i=1}^N \expect{\mu_{i,\sigma^{\star}_{\tau}(i)}Q_i(t_0(\tau))} \\
&\hspace{-0.2in}+2(\epochlength-2\convlength)\expect{\sum_{\tau=1}^{\ell_T} \sum_{i=1}^N (\mu_{i,\sigma^{\star}_{\tau}(i)} - \mu_{i,\sigma_{\tau}(i)})Q_i(t_0(\tau))\indic{\set{G}_{\tau}}}\\
&\hspace{-0.2in}+4K\ell_T(\epochlength-2\convlength)(\epochlength+2\convlength).
\end{align*}
\end{lemma}

The first two terms on the right hand side resemble the drift under \textsc{DAM.K}, where queues work with
exact service rates. The third term introduces the additional cost of learning due to unknown service rates. We note that this term is closely related to the concept of regret in multi-armed bandits since we compare the weight of the benchmark policy that operates with known service rates (\textsc{MaxWeight}) to the matching of our decentralized setting where queues operate based on \textsc{DAM.UCB} with the maximum weight. However, different from multi-armed bandits, a suboptimal match selection can correlate with future weights through $Q_i(t_0(\tau))$; bounding this weight difference requires new analytical ideas and is achieved via the next lemma (proof in Section~\ref{ssec:bounding_matching_regret}).

\begin{lemma}\label{lem:ucb-match-difference}
The weight difference compared to \textsc{MaxWeight} at $t_0(\tau)$ is upper bounded by
\begin{align*}
\expect{\sum_{\tau=1}^{\ell_T} \sum_{i=1}^N (\mu_{i,\sigma^{\star}_{\tau}(i)} - \mu_{i,\sigma_{\tau}(i)})Q_i(t_0(\tau))\indic{\set{G}_{\tau}}} 
&\leq\frac{3}{16}\trafficslack \sum_{\tau=1}^{\ell_T}\sum_{i=1}^N \expect{\mu_{i,\sigma_\tau^{\star}(i)}Q_i(t_0(\tau))}
\\&\hspace{-1in}+\frac{1}{8}\trafficslack \sum_{\tau=1}^{\ell_T}\sum_{i=1}^N \lambda_i \expect{Q_i(t_0(\tau))}
+\frac{896K^2\lambda^\star \epochlength }{\trafficslack}\ln^2(T+K+1).
\end{align*}	
\end{lemma}
\begin{proof}[Proof sketch of Theorem~\ref{thm:queue-ucb}]
The proof structure is similar to the one of Theorem~\ref{thm:queue-nolearning} but upper bounds the drift via Lemmas~\ref{lem:ucb-drift-decompose}~and~\ref{lem:ucb-match-difference}. The complete proof is provided in Appendix~\ref{sec:ucb-full-proof}. 
\end{proof}
 
\subsection{Bounding the regret compared to MaxWeight (Lemma~\ref{lem:ucb-match-difference})}\label{ssec:bounding_matching_regret}
We now move to the key technical contribution of this section (Lemma~\ref{lem:ucb-match-difference}) whose proof highlights the additional complexity in analyzing adaptive learning in a queueing setting compared to its multi-armed bandit analogue. Similar to the analysis of \textsc{DAM.FE}, for epoch $\tau$, we define $\Delta_{i,j}(t_0(\tau)) = \sqrt{\frac{3\ln(t_0(\tau)+K)}{n_{i,j}(t_0(\tau))}}$ and $\Delta_{i,\perp} = 0$. In addition, for each $i \in \set{N}$, we define
\begin{align*}
\set{E}_{\tau,i}^1 = \left\{\exists_{j\in\set{K}}, |\hat{\mu}_{i,j}(t_0(\tau))-\mu_{i,j}| > \Delta_{i,j}(t_0(\tau))\right\} \text{ and }
\set{E}_{\tau,i}^2 = \left\{ \Delta_{i,\sigma_{\tau}(i)}(t_0(\tau)) > \frac{1}{16}\trafficslack\mulower\right\}.
\end{align*}
as the event of overestimating any service rate for a particular queue and selecting a server with large confidence interval respectively. We note that we define $\set{E}_{\tau,i}^2$ as the event that a queue $i$ \emph{selects}
a server for which it has a large confidence bound. It is different from event $\set{E}_W$ for \textsc{DAM.FE}, which examines whether any queue-server pair (not only selected ones) has large confidence interval.

When neither  $\set{E}_{\tau,i}^1$ nor $\set{E}_{\tau,i}^2$ holds, the upper confidence bound $\bar{\mu}_{i,j}$ is within a small confidence bound $\Delta_{i,j}(t_0) \leq \frac{\trafficslack\mulower}{16}$ compared with $\mu_{i,j}$. By our analysis of Section~\ref{sec:algorithm-nolearning}, the queues converge to a matching $\sigma$ which is approximate max-weight for weight~$w_{i,j}=\bar{\mu}_{i,j}Q_i(t_0).$ Hence, under event $\bigcap_{i\in\mathcal{N}}\left((\set{E}_{\tau,i}^1)^c \cap (\set{E}_{\tau,i}^2)\right)^c$, the weight of the matching $\sigma$ is approximately the same with the max-weight based on the ``true'' weights generated by accurate service rates. This is formalized in the next lemma. The first term in the upper bound is due to the fact that \textsc{DAM.converge} finds an approximate max-weight matching, while the second measures the impact of errors due to learning. The proof bounds the overestimation caused due to the UCB selection, i.e., $\bar{\mu}_{i,j}-\mu_{i,j}$, is similar to the analogue in multi-armed bandits, and is provided in Appendix~\ref{sec:ucb-diff-conf}.

\begin{lemma}\label{lem:ucb-diff-conf}
The weight difference compared to \textsc{MaxWeight} at $t_0(\tau)$ is upper bounded by
\begin{align*}
\expect{\sum_{\tau=1}^{\ell_T}\sum_{i=1}^N (\mu_{i,\sigma^{\star}_{\tau}(i)} - \mu_{i,\sigma_{\tau}(i)})Q_i(t_0(\tau))\indic{\set{G}_{\tau}}} \notag
&\leq \frac{3\trafficslack}{16}\expect{\sum_{\tau=1}^{\ell_T}\sum_{i=1}^N \mu_{i,\sigma_\tau^\star(i)}Q_i(t_0(\tau))} \\&+ 2\sum_{\tau=1}^{\ell_T}\sum_{i=1}^N \expect{Q_i(t_0(\tau))\cdot \indic{\set{E}^1_{\tau,i}\cup\set{E}^2_{\tau,i}}\indic{\set{G}_\tau}}\label{eq:ucb-diff-conf}.
\end{align*}
\end{lemma}
The crux in our analysis of adaptive learning lies in bounding the second term, which measures the impact of learning. Similar to the analysis of UCB in multi-armed bandits (e.g., see \cite{slivkins2019introduction}), this term charges the suboptimality to individual estimation errors and connects any such error to a decrease in the confidence interval. However, unlike multi-armed bandits, the impact of an estimation error is not restricted to the size of the confidence interval around the service rates but it also needs to account for the queue lengths, which can vary across queues and epochs (and can, in principle, be as large as the current time slot).

Our key analytical idea that allows us to overcome this difficulty of bounding the queue length when an error is made is to spread the impact of an error over previous epochs. In particular, for queue $i \in \set{N}$, we define $e_i$ as its total number of errors, i.e., $e_i = \sum_{\tau=1}^{\ell_T} \indic{\set{G}_\tau}\indic{\set{E}_{\tau,i}^1 \cup \set{E}_{\tau,i}^2}.$
The following lemma provides a uniform upper bound on $Q_i(t_0(\tau))$ for every $\tau$. 
\begin{figure}
    \centering
    \includegraphics[scale=0.2]{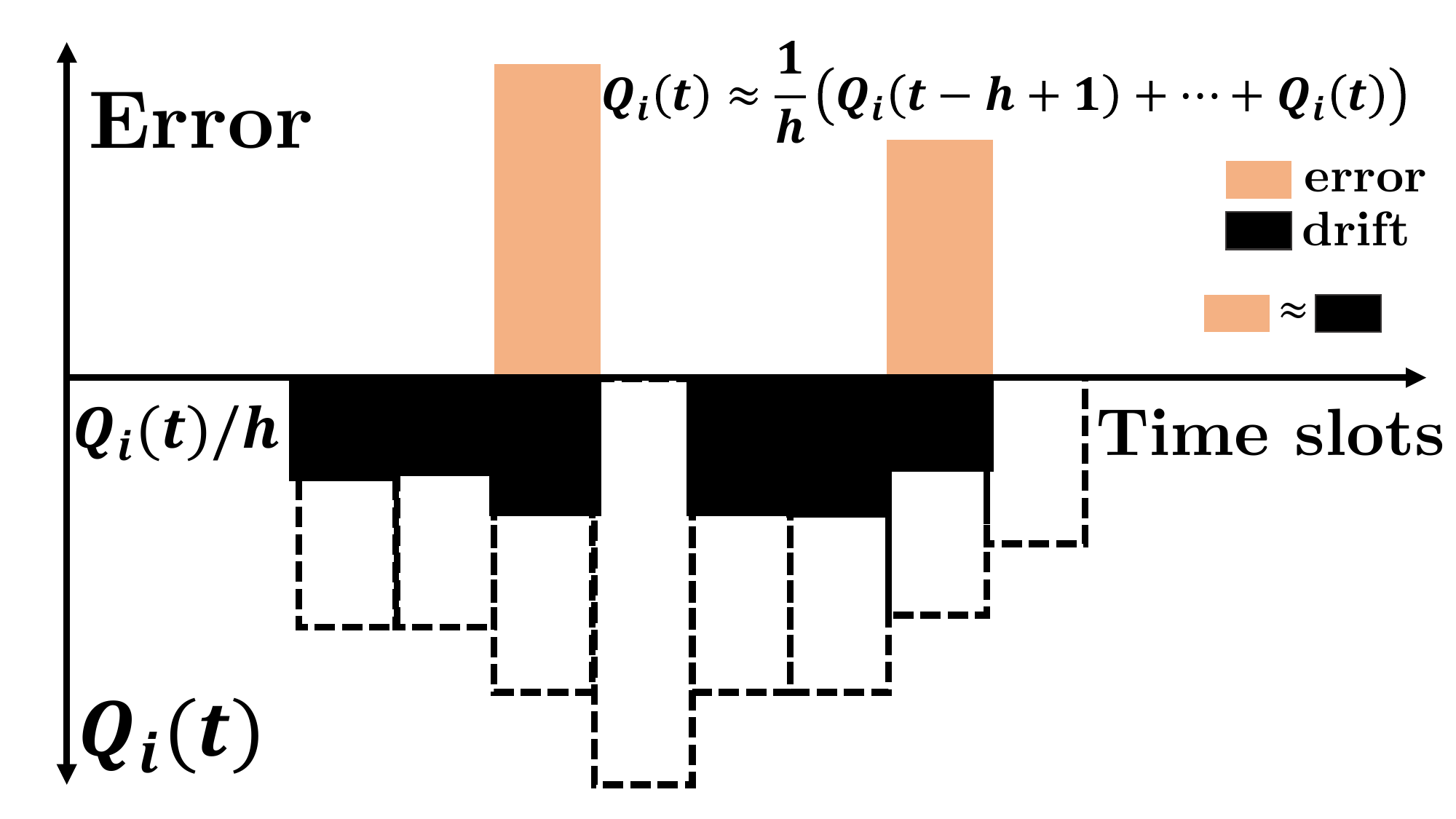}
    \caption{The size of each error (top bars) is the same as the queue length (dashed bar below). We approximate each error by a fraction of the sum of queue lengths in previous epochs (bottom shaded bars). For a fixed period $h$, $Q_i(t_0(\tau)) \approx \frac{1}{h}\sum_{v=0}^{h-1} Q_i(t_0(\tau - v)).$ The difference between $Q_i(t_0(\tau))$ and $Q_i(t_0(\tau - v))$ is at most $v\epochlength$ for every $v$, so the approximation error is at most $h^2\epochlength$.}
    \label{fig:cost-spread}
\end{figure}

\begin{lemma}\label{lem:ucb-spread-cost}
Suppose $e_i \geq 1$. For any fixed epoch $\tau$ and $i \in \set{N}$, we have almost surely
\[
Q_i(t_0(\tau)) \leq \frac{\trafficslack}{16e_i}\sum_{\tau'=1}^{\ell_T}\lambda_i Q_i(t_0(\tau'))+\frac{32e_i\epochlength}{\lambda_i\trafficslack}.
\]
\end{lemma}
\begin{proof}[Proof of Lemma~\ref{lem:ucb-spread-cost}] The proof intuition is illustrated pictorially in Fig.~\ref{fig:cost-spread}. Formally, we spread each individual error over the previous
$h = \lceil\frac{16e_i}{\lambda_i \trafficslack}\rceil$ epochs. If
$\tau \leq h,$ it holds that $Q_i(t_0(\tau)) \leq t_0(\tau) \leq hL + 1 \leq  \frac{32e_i}{\lambda_i\trafficslack}\epochlength.$ Else , since queue lengths change by at most $\epochlength$ within an epoch:
\begin{align*}
Q_i(t_0(\tau)) &\leq \sum_{\tau'=\tau-h+1}^\tau \frac{\lambda_i\trafficslack}{16e_i}\left(Q_i(t_0(\tau'))+(\tau-\tau')\epochlength\right) \\
&\leq \frac{\trafficslack}{16e_i}\sum_{\tau'=\tau-h+1}^{\tau}\lambda_i Q_i(t_0(\tau'))+\frac{\lambda_i\trafficslack}{32e_i}\left\lceil\frac{16e_i}{\lambda_i\trafficslack}\right\rceil^2\epochlength \leq\frac{\trafficslack}{16e_i}\sum_{\tau'=1}^{\ell_T}\lambda_i Q_i(t_0(\tau'))+\frac{32e_i\epochlength}{\lambda_i\trafficslack}.
\end{align*}
\end{proof}
Lemma~\ref{lem:ucb-spread-cost} provides a uniform upper bound on queue lengths with two terms. The first term is only related to drift (and queue lengths), and the second term is only related to the number of errors. Armed with this, we can bound the total drift due to errors via the following lemma.
\begin{lemma}\label{lem:ucb-spread-cost-error}
For a fixed queue $i \in \set{N}$, it holds that
\[
\expect{\sum_{\tau=1}^{\ell_T} Q_i(t_0(\tau))\cdot \indic{\set{G}_\tau}\indic{\set{E}_{\tau,i}^1 \cup \set{E}_{\tau,i}^2}} \leq \frac{\trafficslack}{16}\sum_{\tau=1}^{\ell_T}\lambda_i \expect{Q_i(t_0(\tau))} + \frac{32\epochlength}{\epsilon\lambda_i}\expect{e_i^2}.
\]
\end{lemma}
\begin{proof}
When $e_i \geq 1,$ Lemma~\ref{lem:ucb-spread-cost} implies that
\begin{align*}
\sum_{\tau=1}^{\ell_T} Q_i(t_0(\tau))\indic{\set{G}_{\tau}}\indic{\set{E}_{\tau,i}^1 \cup \set{E}_{\tau,i}^2}
&\leq \sum_{\tau=1}^{\ell_T}\indic{\set{E}_{\tau,i}^1 \cup (\set{G}_{\tau}\cap\set{E}_{\tau,i}^2)}\left(\frac{\trafficslack}{16e_i}\sum_{\tau'=1}^{\ell_T}\lambda_i Q_i(t_0(\tau'))+\frac{32e_i\epochlength}{\lambda_i\trafficslack}\right) \\
&\hspace{-1.3in}=e_i\left(\frac{\trafficslack}{16e_i}\sum_{\tau'=1}^{\ell_T}\lambda_i Q_i(t_0(\tau'))+\frac{32e_i\epochlength}{\lambda_i\trafficslack}\right)
= \frac{\trafficslack}{16}\sum_{\tau=1}^{\ell_T}\lambda_i Q_i(t_0(\tau)) + \frac{32\epochlength}{\trafficslack}\lambda_i^{-1} e_i^2.
\end{align*}
The inequality also holds for $e_i = 0.$ The proof concludes by taking expectation on both sides.
\end{proof}
It remains to bound $\expect{e_i^2}$. This can be done via arguments similar to multi-armed bandits in the following way and is formalized in the following lemma (proof in Appendix~\ref{sec:ucb-spread-cost}).
\begin{lemma}\label{lem:ucb-bound-error}
For a fixed queue $i\in\mathcal{N}$, it holds that $\expect{e_i^2} \leq 14K^2\ln^2(T+K+1).$
\end{lemma}

\begin{proof}[Proof of Lemma~\ref{lem:ucb-match-difference}]The proof follows directly by combining Lemmas~\ref{lem:ucb-diff-conf},~\ref{lem:ucb-spread-cost-error}, and \ref{lem:ucb-bound-error}.
\end{proof}

%% file: DAM-ucb.tex
\begin{algorithm}[H]
\LinesNumbered
\DontPrintSemicolon
\caption{Decentralized Auction Mechanism with UCB exploration~($\textsc{DAM.UCB}$) 
\label{algo:DAM-ucb}
}
\SetKwInOut{Input}{input}\SetKwInOut{Output}{output}
\Input{
Traffic slackness $\trafficslack$; Lower bound of \emph{all} service rates $\mulower$;}
Initialize check period $\checkperiod$, converging length $\convlength$, and epoch length $\epochlength$ as in \eqref{eq:parameter-setting}\;
$\eta_i \gets \text{a uniform random number in }(0,10^{-9})$\;
Sample mean $\hat{\mu}_{i,j}(t_0) \longleftarrow 0$, number of samples $n_{i,j}(t_0) \longleftarrow 0$ \;
\For{$\ell = 1\ldots$}{
    $t_0\gets (\ell-1)\epochlength+1$\;
    $\bar{\mu}_{i,j}(t_0) = 
    \max\left(\mulower,\min\left(1, \hat{\mu}_{i,j}(t_0) + \sqrt{\frac{3\ln (t_0 +K)}{n_{i,j}(t_0)}}\right)\right)$\label{line:ucb-esti}\;
    $\sigma(i),p_{i,\sigma(i)}\gets \textsc{DAM.converge}(t_0,\convlength, \checkperiod,\trafficslack,\{\bar{\mu}_{i,j}(t_0)\}_{j \in \set{K}},\eta_i)$ \label{algoline:dam-ucb-converge}\;
    call $\textsc{DAM.commit}(t_0 + \convlength, t_0 + \epochlength - 1, \sigma(i), p_{i,\sigma(i)})$\;
   \tcc{update estimates with all samples after the first success}
    $\hat{\mu}_{i,j}(t_0 + \epochlength), n_{i,j}(t_0 + \epochlength) \gets \textsc{DAM.update}(t_0 + \convlength, t_0 + \epochlength - 1)$ \label{algoline:dam-commit-sample}
}
\end{algorithm}

%% file: algo-dynamic.tex
\textsc{DAM.UCB} and its analysis also extend to a setting in which queues arrive and depart dynamically. Specifically, consider the following model, inspired by dynamic-population games \cite{LykourisST16}. At the beginning of each time slot $t$, new queues may join the system. A newly arrived queue $i$ has arrival rate $\lambda_i$ and service rate $\mu_{i,j}$ for server $j$. Each new queue originally has queue length zero, i.e., if $i$ joins at time slot $t$, then $Q_i(t) = 0.$ Queues in the current system may also leave. The set of queues at time $t$ (after departures) is defined as $\setQ(t)$. We impose the following assumptions: 
\begin{enumerate}
    \item the set of queues present in each time slot, $(\setQ(t))_{t \geq 1}$, is generated by an oblivious adversary,
    \item there is a known traffic slackness $\trafficslack> 0$ (Definition~\ref{as:stability}) that holds true for all time slots,
    \item  service rates are lower bounded by a known value $\delta$, i.e., $\mu_{i,j} \geq \delta$ for all $i$ and $j$, 
    \item arrival rates are  lower bounded by a, possibly unknown, positive constant $\ubar{\lambda}$,
    \item there is a known maximum number of queues $N,$ such that $|\setQ(t)| \leq N$ for all $t$. 
\end{enumerate}
Observe that the limit of $N$ on $|\setQ(t)|$ plays the same role as the number of queues in previous sections. The parameters $\checkperiod$, $\convlength $, and $\epochlength$ (Eq.~\ref{eq:parameter-setting})
are set with respect to this value of $N$. For a queue $i$, we denote by~$\ts(i)$ the time slot that it joins the system and by~$\te(i)$  the last time slot it stays in the system. If queue $i$ never leaves, we write $\te(i) = \infty.$ We assume $\te(i) \geq \ts(i),$ i.e., each queue stays in the system for at least one time slot.

\vspace{0.1in}
\noindent\textbf{The need for adaptive exploration}
We start with an example to showcase the failure of forced exploration, or specifically \textsc{DAM.FE}, to stabilize the system under this dynamic model of queues. To adapt \textsc{DAM.FE} to the dynamic setting, we assume each new queue  waits until the start of the next epoch and follows \textsc{DAM.FE} (Algorithm~\ref{algo:DAM-online-master}) from that point onwards. For a queue that joins in epoch~$\tau$, the exploration probability in a later epoch~$\tau'$ is~$\min(1,K(\tau'-\tau+1)^{-\expratio})$ (see line \ref{algoline:explore-rate} in Algorithm~\ref{algo:DAM-online-master}). In the following example, (dynamic) DAM.FE fails to stabilize the system.

\begin{example}\label{example:dynamic}
Consider a system with two queues and two servers, in which $\lambda_1 = 0.7, \lambda_2 = 0.4$ and~$\mu_{1,1}=\mu_{2,2}=0.9, \mu_{1,2}=\mu_{2,1}=0.3.$ Queue $1$ is always in the system, whereas queue $2$ is replaced by a new (identical) copy at the start of every epoch of \textsc{DAM.FE} (this queue has the same arrival rate and service rates, but needs to restart its learning). As the new queue explores server $1$ with probability $0.5$, \textsc{DAM.FE} can match server $1$ with queue $1$ only in the remaining periods, i.e., for at most $50\%$ of the time. Therefore, queue $1$ obtains a time-averaged service rate of at most $\frac{1}{2}(0.9+0.3)=0.6<\lambda_1=0.7$, which leads to an average queue length of $\expect{Q_1(T)} \geq 0.1T$. Hence, dynamic DAM.FE fails to stabilize the system.
\end{example}

\noindent\textbf{Dynamic DAM.UCB and its guarantee.}
The failure of forced exploration in Example~\ref{example:dynamic} creates the natural question if the system can be stabilized in such a dynamic environment when agents use adaptive exploration. Indeed, we show that an adaptation of \textsc{DAM.UCB} which we refer to as Dynamic \textsc{DAM.UCB} achieves this guarantee. In particular, at time slot $t \geq \ts(i)$, suppose that queue~$i$ knows the elapsed time since its appearance, $t - \ts(i)$ and whether $t$ is the start of an epoch, i.e., whether there exists an integer $\ell \geq 1$ such that $t = (\ell-1) \epochlength + 1$. Queue $i$ starts to run \textsc{DAM.UCB} (Algorithm~\ref{algo:DAM-ucb}) when the first epoch after its appearance starts, i.e., at time slot $\left(\left\lceil \frac{\ts(i)-1}{\epochlength}\right\rceil\right) \epochlength + 1$. Moreover, the upper confidence estimation is adapted to account for the shorter number of periods queue $i$ has spent in the system, i.e., $t_0$ is replaced by $t_0-\ts(i)+1$ in Line~\ref{line:ucb-esti} of Algorithm~\ref{algo:DAM-ucb}. Finally, a queue terminates its algorithm whenever it leaves. The intuition is that, by using adaptive exploration based on queue lengths, queues which are new (and hence short) do not get assigned to in-demand servers since their weight is small even with optimistic service rates. Unless a queue stays in the system for long, it does not waste server capacity by exploring thus bypassing Example~\ref{example:dynamic}.

The analysis for dynamic \textsc{DAM.UCB} requires some additional notation: for horizon $T$, the survival time of queue $i$, i.e., how many time slots it stays in the system until $T$, is denoted by $s_T(i)=\max\left(0,\min(T,\te(i))-\ts(i)+1\right).$
We also define $\lambda^\star_T = \sum_{i \in \cup_{t \leq T}\set{I}(t)} \lambda_i^{-1}\ln^2(s_T(i)+K+1).$  Theorem~\ref{thm:dynamic-ucb} offers an analogue of Theorem~ \ref{thm:queue-ucb} for dynamic \textsc{DAM.UCB}. We note that the guarantees are similar with the only difference being the dependence on $\lambda^\star_T$ instead of $\lambda^\star\ln^2(T+K)$. This term measures the impact of the frequency of queues joining the system and their survival time. 
\begin{theorem}\label{thm:dynamic-ucb}
If all queues follow dynamic DAM.UCB, then for any $T \geq 1$, it holds that:
\[
\expect{\frac{1}{T}\sum_{t=1}^T\sum_{i\in \setQ(t)} \lambda_i Q_i(t)} = O\left(\frac{K^2\checkperiod}{\trafficslack^3}(\log N + K) + \frac{(K\epochlength)^2}{\trafficslack^2}\frac{\lambda^{\star}_T}{T}\right).
\]
\end{theorem}
\noindent We can naturally extend the notion of strong stability from Section~\ref{sec:prelims} to the dynamic setting by requiring $\lim_{T \to \infty} \expect{\frac{1}{T}\sum_{t=1}^T\sum_{i\in \setQ(t)} Q_i(t)} < \infty$. Extensions of other stability notions to our dynamic settings may not be achievable even in a centralized system with known parameters (see Appendix~\ref{app:stability-dynamic}). To prove strong stability in the dynamic setting, we need the bound in Theorem \ref{thm:dynamic-ucb} to be independent of $T$, i.e.,  $\lambda^\star_T=\sum_{i \in \cup_{t \leq T}\set{I}(t)} \lambda_i^{-1}\ln^2(s_T(i)+K+1)$ can grow at most linearly in $T$. We prove this in corollary \ref{cor:dynamic} using our assumption that arrival rates $\lambda_i$ are bounded from below by $\ubar{\lambda}$: in particular, our proof bounds $\lambda^\star_T$  from above by connecting the gross life time of queues and the total number of queues. Consequently, dynamic \textsc{DAM.UCB} stabilizes the system even with dynamic arrivals and departures of queues.
\begin{corollary}\label{cor:dynamic}
If all queues follow dynamic \textsc{DAM.UCB} and arrival rates are all lower bounded by a positive constant $\ubar{\lambda}$, then for any $T \geq 1$, it holds that:
\[
\expect{\frac{1}{T}\sum_{t=1}^T\sum_{i\in \setQ(t)} \lambda_i Q_i(t)} = O\left(\frac{K^2\checkperiod}{\trafficslack^3}(\log N + K) + \frac{N\ln^2(K+1)(K\epochlength)^2}{\trafficslack^2\ubar{\lambda}}\right).
\]
\end{corollary}

The proof of Theorem~\ref{thm:dynamic-ucb} requires \textsc{DAM.converge} to obtain an approximate maximum weight matching when queues may leave in an epoch. Moreover, it needs an analysis of \textsc{DAM.UCB} adapting to a system with dynamic queues. Fortunately, the main proof ideas remain the same. We showcase the essential proof changes for Theorem~\ref{thm:dynamic-ucb} in Appendix~\ref{app:section-dynamic}. The proof of Theorem~\ref{thm:dynamic-ucb} is in Appendix~\ref{app:proof-dynamic} and the proof of Corollary~\ref{cor:dynamic} is in Appendix~\ref{app:proof-cor-dynamic}.

%% file: algo-simulation.tex
In this section we compare \textsc{DAM.FE} and \textsc{DAM.UCB} numerically with \textsc{ADEQUA} proposed by~\cite{sentenac2021decentralized} and EXP3.P.1 applied by~\cite{gaitonde2023price}. Since queues under \textsc{DAM.FE} and \textsc{DAM.UCB} learn to imitate  DAM.K (queues know service rates but act decentrally), which in turn aims to decentrally implement \textsc{MaxWeight} (queues know service rates and act centrally); we include these as benchmark comparisons. Overall, our results in this section highlight the benefits of our algorithms with respect to both convergence and robustness. Additional details of our implementation, including the exact arrival and service rates, are included in Appendix~\ref{app:simulation}.

\begin{figure}[!h]
    \centering
      \scalebox{0.4}{\input{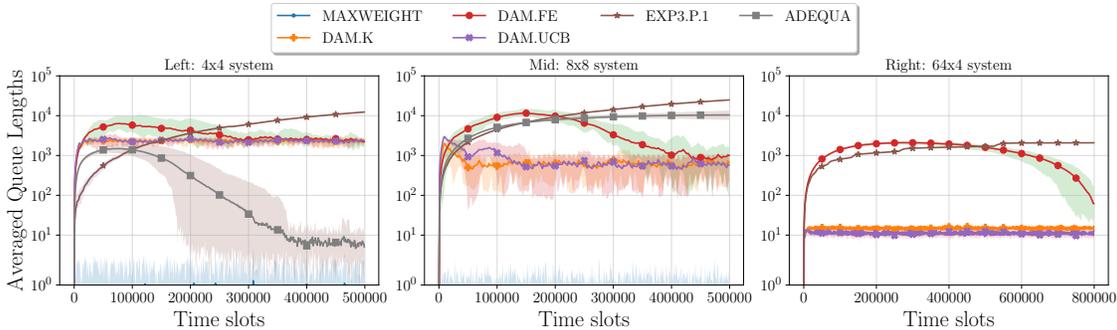}}
      \caption{Convergence of averaged queue lengths under different algorithms in environments of various sizes. DAM.FE and DAM.UCB in general converge substantially faster than ADEQUA and EXP3.P.1 (previous algorithms). DAM.UCB has better convergence than DAM.FE and is nearly indistinguishable from DAM.K for which queues know exact service rates. We note that there is a large gap between decentralized (DAM.K) and centralized algorithms (\textsc{MaxWeight}).}
      \label{fig:packets-n}
\end{figure}

\vspace{0.1in}
\noindent\textbf{Convergence benefits.} The following three instances consider systems of different sizes to compare the scaling behavior of the queue sizes under each of the algorithms. The first instance (left plot in Fig.~\ref{fig:packets-n}) is the hard instance described in \cite{sentenac2021decentralized}. It consists of~4 agents and~4 servers, where both $\trafficslack, \mulower$ are small (0.25 and 0.1875 respectively), leading our algorithms to require a large epoch length for queues to converge to an approximate max-weight matching. Therefore, \textsc{DAM.FE} and \textsc{DAM.UCB} have larger asymptotic 
(as $t \to \infty$) averaged queue lengths compared to \textsc{ADEQUA} , which can asymptotically match a centralized scheduler. In contrast, our second instance, with $N=K=8$ and moderate $\trafficslack$ (central plot in Fig.~\ref{fig:packets-n}), shows that for a system with more queues and servers, \textsc{DAM.FE} and \textsc{DAM.UCB} can learn substantially faster than \textsc{ADEQUA}. Finally, we include a system with $N=64,K=4$ (right plot in Fig.~\ref{fig:packets-n}): for that setting we were not able to simulate \textsc{ADEQUA} due to computational constraints. In contrast, the light computational requirements of our algorithms allow them to scale to larger instance, and we moreover 
find that \textsc{DAM.UCB} quickly converges even there (\textsc{DAM.FE} also converges but
slower). We also see that the time-average queue length of MaxWeight (the centralized counterpart of DAM.K) is much shorter than its decentralized counterpart. Overall, we observe that \textsc{DAM.FE} and \textsc{DAM.UCB} converge much faster than \textsc{ADEQUA} for larger systems, but there is still a big gap in performance between decentralized and centralized algorithms.

\paragraph{Robustness benefits.} 
In Appendix~\ref{app:sim-robust}, we study the robustness of our algorithms. We showcase how \textsc{DAM.FE} and \textsc{DAM.UCB} can stabilize the system under non-stationary arrivals (left plot of Figure~\ref{fig:fluctuate_asym}) and under asymmetric service rates (right plot of Figure~\ref{fig:fluctuate_asym}). We then simulate variations of Example~\ref{example:dynamic} to illustrate how dynamic \textsc{DAM.FE} fails and dynamic \textsc{DAM.UCB} succeeds in stabilizing the system with frequent dynamic departures/arrivals (Figure~\ref{fig:dynamic}). 

%% file: conclusions.tex
We studied a discrete-time queueing system where each queue requests service from at most one server and the goal is to design a decentralized algorithm for the queues that achieves system stability under a simple server selection rule. We devised \textsc{DAM.K} that allows queues to decentrally converge to approximate maximum-weight matchings. We then extended this algorithm to the setting where queues operate without knowledge of the system parameters with \textsc{DAM.FE} and \textsc{DAM.UCB} that employed forced and adaptive exploration respectively. The latter was enabled by a novel analysis of queue lengths under \textsc{DAM.UCB}, showing provable convergence benefits of adaptive exploration in a queueing system. We also showed an additional advantage of adaptive exploration as it can stabilize the system when queues dynamically join or depart from the system (unlike forced exploration). Simulations confirm the advantageous convergence and robustness properties of our algorithms.
Our work opens up several interesting directions for future research.
\begin{itemize}
    \item First, there is a significant performance gap between efficient decentralized algorithms and centralized algorithms. Vanishing queueing delay is possible for a centralized algorithm when the bipartite graph satisfies certain flexibility structure \cite{TsitsiklisX17}. A drift analysis of \textsc{MaxWeight} 
    shows that its time-averaged queue length is $O\left(\nicefrac{K}{\trafficslack}\right)$, which is lower than our
    $\tilde{O}\left(\nicefrac{K^3}{\trafficslack^3}\right)$ guarantees.
    A natural question is thus to either design a decentralized algorithm with an improved queue-length bound (possibly with assumptions on the graph structure) or provide a lower bound
    that shows 
    a gap in performance between centralized and decentralized algorithms.
    \item Second, our algorithms require knowledge of lower bounds on both traffic slackness $\trafficslack$ and nonzero service rates $\mulower$.
    It would be interesting to relax this requirement and design a decentralized multi-agent learning algorithm that does not require this prior knowledge.
    \item Relatedly, the \textsc{DAM.UCB} algorithm with adaptive exploration requires all service rates to be positive. This is necessary for \textsc{DAM.UCB} to deal with the effect of interference due to unobserved collisions, which can be indistinguishable from zero service rates when the exploration is not explicit. Hence, relaxing this assumption requires new algorithm design. 
    \item Moreover, our algorithms are synchronous as queues start at the same time and synchronize on epochs. We partially relax the synchronization by allowing dynamic queue arrivals and departures (that only need to know when an epoch begins). Designing a truly asynchronous algorithm where queues do not even share the epoch starting points seems like a really challenging task for which we do not know if any of our algorithmic ideas would extend. In multi-player multi-armed bandit such an asynchronous algorithm is possible~\cite{boursier2019sic}.
    \item Another intriguing direction involves the selection of tie-breaking rules on the server side. In this work, we assume that servers select the higher-bidding queue when multiple queues request service. It would be interesting to design efficient algorithms with other tie-breaking rules, such as the collision rule \cite{rosenski2016multi}, or the age-based rule \cite{gaitonde2023price}.
    \item Furthermore, it is useful to extend our model and algorithms beyond bipartite queueing settings, e.g., different network structures such as heterogeneous load balancing \cite{WengZhouSrikant20,choudhury2021job} or interference networks~\cite{JiangWalrand10}.
    \item Finally, our model of dynamic queues assumes an oblivious
    queue changing process that is fixed and independent of job arrivals, services and queue lengths. Studying settings where the aforementioned are correlated is an exciting open direction.
\end{itemize}

%% file: app_example.tex
In this section, we summarize the constraints and the assumptions in our decentralized setting and discuss several motivating examples. On the one hand, our model imposes the following constraints:
\begin{itemize}
    \item \emph{Decentralization}: queues possess only local information, i.e., their own queue lengths. There is no shared randomness and no  communication among queues or servers;
    \item \emph{Asymmetry}: service rates may depend on both queues and servers;
    \item \emph{Unique requests}:
     queues request service from at most one server per time slot;
    \item  \emph{Learning}: queues do not initially know the service probabilities of any server;
    \item \emph{Feedback}: a queue receives the same feedback when its request is rejected and when its request is accepted but the service fails.
\end{itemize}
On the other hand, our model assumes that the following are allowed:
\begin{itemize}
    \item \emph{Synchronization}: there exists a mechanism through which queues can agree on the time slots in which each epoch starts; 
    \item \emph{Lower bounds}: queues share a common lower bound on the traffic slackness as well as a lower bound on the minimum positive service probability;
    \item \emph{Collision handling}: when a server receives multiple job requests, these requests do not collide with each other (in contrast to the wireless network literature, see discussion in Section~\ref{ssec:related_work}).  Instead, each queue communicates a scalar signal (bid) to a server and the server picks the queue with the highest bid.
    \item \emph{Collaboration}: queues follow a collaborative protocol and do not act selfishly.
\end{itemize}

\begin{table}[]
\begin{tabular}{|c|c|c|c|c|c|}
\hline
                                                                     & Decentralization & Asymmetry    & Unique requests    & Learning      & Feedback     \\ \hline
Cognitive radio                                                      & $\checkmark$     &              & $\checkmark$       & $\checkmark$  & $\checkmark$ \\ \hline
\begin{tabular}[c]{@{}l@{}}Online service\\ platform\end{tabular}    & $\checkmark$     & $\checkmark$ &                    & $\checkmark$  &              \\ \hline
                                                                     & Synchronization  & Lower bound  & Collision handling & Collaboration &              \\ \hline
Cognitive radio                                                      & $\checkmark$     & $\checkmark$ &                    & $\checkmark$  &              \\ \hline
\begin{tabular}[c]{@{}l@{}}Online service \\  platforms\end{tabular} & $\checkmark$     & $\checkmark$ & $\checkmark$       &               &              \\ \hline
\end{tabular}
\label{tab:example}
\caption{Properties of two motivating examples for the decentralized model}
\end{table}

We next discuss two motivating examples for our model. Table~\ref{tab:example} summarizes how each one reflects our assumptions and constraints. The first example is that of cognitive radio in communication network systems \cite{avner2014concurrent}. In this example there is a set of primary users (PU) who individually possess a communication channel. There is also a set of secondary users (SU) who do not possess channels themselves but can request access to channels from the primary users. In this example, we can model secondary users as agents and primary users as servers. Each SU has a queue of jobs (e.g. TV signals). In each time slot, a SU requests channel access from at most one PU. Each PU can grant access to at most one SU since two SUs accessing the same channel causes interference. In addition, since PUs have their own jobs, the availability of a channel to SUs varies in each time slot and is captured by the service probability of PUs, which must be learned by SUs (as a result, the service rates in this example are likely symmetric). A PU notifies an SU only when the PU is not using its channel and the PU does not accept others' requests. This is a deviation of our model from real cognitive radio settings, in which multiple job requests to the same PUs, cause requests to collide with each other and all requests to fail (see the discussion at the end of Section~\ref{ssec:related_work})\footnote{Our model represents a twist to the status quo: instead of the case that SUs probe channel usability independently from PUs, SUs in our model first send requests to PUs and only use channels after receiving confirmation from the corresponding PUs.}. Synchronization is also difficult to achieve in wireless networks (although not impossible \cite{NiTS12}). When there is no synchronization, our algorithm breaks down. But if agents can sense whether a server received requests in the last time slot, previous CSMA algorithms may stabilize the queues without coordination albeit with exponential queue lengths (see Section~\ref{ssec:related_work} for a further discussion.)  

Another motivation for our model arises from online service platforms. In this setting, customers (queues) observe a stream of jobs and there are a number of freelancers (servers) on the platform. Each customer has a queue of job requests but types of jobs vary across customers (art, writing, website design etc.). Each customer independently sends job requests to freelancers based on local information. On the freelancer side, each of them can take at most one request per time slot. Due to varying job types and skill sets, service probabilities are asymmetric between customers and freelancers. Customers then need to learn freelancers' service speed in an online fashion. However, such a setting may be simpler than ours in that freelancers may signal a customer when a request is accepted (and, consequently, queues receive more feedback). Another distinction to our model is that customers may not want to collaborate in this setting.

%% file: app_stability.tex
When the set of queues is fixed, three notions of stability from the literature are immediately relevant in a queueing system with learning: \emph{moment stability}, \emph{strong stability} and \emph{mean rate stability}. We first review these stability notions in the classical setting where the set of queues is fixed throughout. We then discuss their variants in the dynamic setting.
\subsection{Stability with a static set of queues (Section~\ref{sec:prelims})}\label{app:stability-static}
We first consider the setting where the set of queues is fixed as $\set{N}$. The first notion of stability, moment stability, is used in \cite{gaitonde2023price,sentenac2021decentralized} and requires that every moment of the total queue length is upper bounded throughout the horizon.
\begin{definition}
The queueing process is moment stable if for every positive integer $r$, there exists a constant $C_r$ such that for every $T \geq 1$, $\expect{(\sum_{i \in \set{N}} Q_i(T))^r} \leq C_r$.
\end{definition}
Moment stability is usually hard to establish and may be a problematic definition in certain cases, see the discussion of Section IV.B in \cite{DBLP:journals/corr/abs-1003-3396}. In the networking literature, strong stability is more commonly used, which only requires the time average total queue length to be upper bounded. When the system can be described by a discrete-time Markov chain where queue lengths of agents fully describe the states, strong stability implies positive recurrence of the Markov chain (see Theorem~2.8 in \cite{DBLP:series/synthesis/2010Neely}.)
\begin{definition}
The queueing process is strongly stable if $\lim\limits_{\substack{T \to \infty}}  \frac{\sum_{t=1}^T\expect{\sum_{i\in \set{N}} Q_i(t)}}{T} < \infty$.
\end{definition}
The last notion, mean rate stability, requires that the expected number of serviced jobs is close to that of arrived jobs \cite{DBLP:series/synthesis/2010Neely}. In particular, let us define $\tilde{S}_i(t) = \min(Q_i(t)+A_i(t), S_i(t))$ which is one if there is an actual job in queue $i$ which gets service in time slot $t$. Then mean rate stability requires the following property.
\begin{definition}
The queueing process is mean rate stable if $\lim\limits_{\substack{T \to \infty}} \frac{\sum_{t=1}^T \expect{\sum_{i \in \set{N}} (A_i(t) - \tilde{S}_i(t))}}{T} = 0$.
\end{definition}
If the set of queues is fixed, mean rate stability is equivalent to $\lim_{T \to \infty} \frac{\expect{\sum_{i \in \set{N}} Q_i(T)}}{T} = 0$. As a result, we have the following hierarchy among these three notions of stability. 
\begin{proposition}[Hierarchy of stability notions]
If the set of queues is fixed, moment stability implies strong stability and strong stability implies mean rate stability.
\end{proposition}
The first result is immediate and the second result is by Theorem 4 in \cite{DBLP:journals/corr/abs-1003-3396}. 

\subsection{Stability with dynamic queues (Section~\ref{sec:dynamic})}\label{app:stability-dynamic}
When the set of queues changes over time, a simple way to extend the three notions of stability is to only count the set of agents $\set{I}(t)$ for a time slot $t$. Then, one may naturally wonder whether the hierarchy between these notions is preserved in the dynamic setting. The following example shows that this is not the case even when the time that each agent stays in the system is lower bounded by a constant.
\begin{proposition}\label{prop:mean-rate-stable-dynamic}
For any constant $b \geq 1$, there exists a centralized setting such that: 1) every agent stays in the system for at least $b$ time slots; 2) moment stability and strong stability are trivially satisfied 3) no algorithm can achieve mean rate stability.
\end{proposition}
\begin{proof}
Consider a system with one server ($K = 1$) and originally with two agents. Every $b$ time slots, a new pair of agents replaces the previous pair. Therefore, every agent stays for at least $b$ time slots in the system. Arrival probabilities for all agents are $\frac{1}{3}$ and the service probabilities for the only server are one for all agents, i.e., $\lambda_i = \frac{1}{3},\mu_{i,j} = 1$ for any agent $i$ and $j\in \set{K}$. Moment stability and strong stability are clearly satisfied for any algorithm since the queue length for any agent is at most $b$. We claim that, however, for any $T$ that is a multiple of $b$, we must have $\frac{1}{T}\sum_{t = 1}^T \expect{\sum_{i \in \set{I}(t)} (A_i(t) - \tilde{S}_i(t))} \geq \frac{1}{9b}$. As a result, mean rate stability is impossible.

To prove the claim, fix a $T$ which is a multiple of $b$. The set of agents remains fixed during $[T-b+1,T]$. In addition, we must have $\sum_{t = T-b+1}^{T-1}\sum_{i \in \set{I}(t)} (A_i(t) - \tilde{S}_i(t)) \geq 0$ since $Q_i(T-b+1) = 0$ for any agent $i \in \set{I}(t-b+1)$. In addition, with probability $\frac{1}{9}$, we have $A_i(T) = 1$ for all $i \in \set{I}(T)$. But $\sum_{i \in \set{I}(T)} \tilde{S}_i(T) \leq 1$ since the server accepts at most one request. Therefore, with probability at least $\frac{1}{9}$, we have $\sum_{i \in \set{I}(T)} (A_i(T) - \tilde{S}_i(T)) \geq |\set{I}(T)| - 1 = 1$. Taking expectation gives $\expect{\sum_{t = T-b+1}^{T}\sum_{i \in \set{I}(t)} (A_i(t) - \tilde{S}_i(t))} \geq \frac{1}{9}$. Summing across previous time slots gives $\frac{1}{T}\expect{\sum_{t = 1}^{T}\sum_{i \in \set{I}(t)} (A_i(t) - \tilde{S}_i(t))} \geq \frac{1}{9b}$, which finishes the proof.
\end{proof}
Proposition \ref{prop:mean-rate-stable-dynamic} shows that even in a centralized system, mean rate stability is unachievable for dynamic queues and thus our paper focuses on establishing the strong stability result.

%% file: comparison_to_static.tex
We discuss the extra difficulties brought by decentralized queueing system compared to static matching and more generally the multi-player multi-armed bandit (MMAB) setting. In MMAB, players only need to learn an optimal matching between players and servers (exploration), and fix to it until the game ends (exploitation). However, in our setting, a fixed matching cannot stabilize the system. Queues may alternative the choice of servers to maximize server utilization because service assigned to an empty queue would be wasted. By \Cref{as:stability}, queues need to schedule using a distribution of matchings. In this case, for each time slot, queues must communicate such that all queues are choosing the same matching from the distribution. Otherwise, there will be collision between queues and the schedule would fail. See Example \ref{ex:failure}. 
\begin{example}\label{ex:failure}
Consider a two-queue two-server system, where $\lambda_1 = \lambda_2 = 0.5, \mu_{1,1}=\mu_{2,1}=0.8, \mu_{1,2}=\mu_{2,2}=0.4.$ A feasible distribution over matchings is choosing the permutation $(1,2)$ and the other one $(2,1)$ with equal probability $0.5.$ Fix a time slot $T$ large enough. In the time horizon $[T]$, denote $P_{a,b}$ by the fraction of time queue $1$ selects server $a$ and queue $2$ selects server $b$ with $a,b \in \{1,2\}.$ Since queue $1$ and queue $2$ sample the distribution independently, it holds $P_{a,b} = \frac{1}{4}$ for all $a,b$. Then the total expected service the two queues could obtain is upper bounded by $\frac{0.8}{4}+\frac{1.2}{4}+\frac{0.8}{4}+\frac{0.4}{4}=0.8 < \lambda_1 + \lambda_2 = 1.$ By the transition (\ref{eq:queuedynamic}), for every large enough $T$, we have $\expect{Q_1(T)+Q_2(T)} \geq 0.2T$, which surely would make (\ref{eq:average-queue-length}) dependent to $T$.
\end{example}

In this example, a fixed matching such as either $\sigma_1 = 1,\sigma_2 = 2$ or $\sigma_1 = 2,\sigma_2 = 1$ fails to stabilize the system since the arrival rate of each queue is $0.5$, and the service rate of server $2$ is always $0.4$. Therefore, the structural assumption of \cite{KrishnasamySenJohariShakkottai21} fails in this case even though a centralized scheduler scan stabilize the system. We also note that even if queues share the same distribution over matchings, they require communication between each other to decide which matching to choose for each time slot in order to maintain stability. It is possible when queues are allowed to possess shared randomness \cite{sentenac2021decentralized}. However, in our setting, queues are oblivious to each other, and thus it is unlikely that they can unanimously choose a same matching. In addition, the above solution requires accurate knowledge of $\lambda_i, \mu_{i,j}$ which is absent for queues.

%% file: app_section_no_learning.tex

\subsection{Bounding the probability of the good event (Lemma~\ref{lem:checkperiod-good})}\label{app:good_event}

Before providing the proof, we first have the following lemma which shows that the probability of $\goodevent_{\ell}$ decreases exponentially fast with respect to the length of checking period $\checkperiod$.
\begin{lemma}\label{lem:prob-good-event}
The probability of $\goodevent_{\ell}$ is at least $1 - K\convlength(1-\mulower)^{\checkperiod}.$
\end{lemma}
\begin{proof}
Let $t_0 = (\ell - 1)\epochlength + 1$. Fix $t \in \{t_0+\checkperiod-1,\ldots,t_0+\convlength-1\}$ and a server $j \in \mathcal{K}.$ 
We first upper bound the probability that $\forall t' \in [t - \checkperiod + 1,t], S_{I(j,t'),j}(t') = 0$ and then apply a union bound over all $t$ and $j$ to complete the proof of the lemma. 

Define the events $\mathcal{W}_{d} = \{S_{I(j,t'),j}(t') = 0, \forall t' \in \{t-\checkperiod + 1,\ldots,t - \checkperiod + d\}\}$ for $d \in \{1,\ldots,\checkperiod\}.$ For ease of notation, let $\set{W}_0$ be the full sample space. In addition, we define $\mathcal{I}_j = \{i \in \mathcal{N} \colon \mu_{i,j} > 0\} \cup \{\perp\}$ as the set of possible values of $I(j, t')$ for all time slots $t'$. We now prove $\Pr\{\mathcal{W}_{d}\} \leq (1-\mulower)^{d}$ by induction. Notice that it holds for each $d \in \{1,\ldots,\checkperiod\},$
\begin{align*}
\Pr\{\mathcal{W}_{d}\} &= \Pr\{\{S_{I(j,t - \checkperiod + d),j}(t - \checkperiod + d) = 0\} \cap \mathcal{W}_{d - 1}\} \\
&= \sum_{i \in \mathcal{I}_j} \Pr\left\{S_{i,j}(t - \checkperiod + d) = 0 \mid I(j,t-\checkperiod+d) = i, \mathcal{W}_{d-1}\right\} \notag\\ 
&\mspace{64mu}\cdot\Pr\left\{I(j,t-\checkperiod+1) = i \mid \mathcal{W}_{d - 1}\right\} \Pr\{\mathcal{W}_{d - 1}\} \\
&= \sum_{i \in \mathcal{I}_j} \Pr\left\{S_{i,j}(t - \checkperiod + d) = 0\right\} \Pr\left\{I(j,t-\checkperiod+1) = i\mid \mathcal{W}_{d - 1}\right\}\Pr\{\set{W}_{d-1}\} \\
&\leq (1-\mulower)\Pr\{\set{W}_{d-1}\}\sum_{i \in \mathcal{I}_j} \Pr\left\{I(j,t-\checkperiod+1) = i \mid \set{W}_{d - 1}\right\} \\
&= (1-\mulower)\Pr\{\set{W}_{d - 1}\}, 
\end{align*}
where the second equation is by the Law of Total Probability; the third one is because $S_{i,j}(t - \checkperiod + d)$ is independent of both $I(j,t - \checkperiod + d)$ and $\set{W}_{d - 1}$;  and the inequality is because $\mu_{i,j} \geq \mulower$ when $\mu_{i,j} > 0$ and $\mu_{\perp,j} = 1$ by definition. Since $\Pr\{\set{W}_0\} = 1,$ we then have $\Pr\{\set{W}_{\checkperiod}\} \leq (1-\mulower)^{\checkperiod}.$ With a union bound over $\convlength$ values of $t$ and $K$ servers we obtain
\begin{equation*}
\Pr\{\goodevent_{\ell}\} = 1 - \Pr\{\goodevent_{\ell}^{c}\} \geq 1 - K\convlength(1-\mulower)^{\checkperiod}.
\end{equation*}
\end{proof}
Recall that $\checkperiod = \left\lceil\max\left(2, \left(\frac{2}{\ln(1-\mulower)}\right)^2, \frac{2\ln \xi}{\ln(1-\mulower)}\right)\right\rceil.$ We now show that with~$\checkperiod$ set in this way $\Pr\{\goodevent_{\ell}\}$ is sufficiently large. Recall that $\convlength = \left \lceil\frac{99K\checkperiod}{\trafficslack}(K+\log N)\right\rceil$, and $\epochlength = \left \lceil (\frac{32}{\trafficslack} + 1)\convlength\right \rceil$.
We now complete the proof of \Cref{lem:checkperiod-good}.
\begin{proof}[Proof of Lemma~\ref{lem:checkperiod-good}.]
By \Cref{lem:prob-good-event}, we have
\[
\Pr\{\goodevent_{\ell}\} \geq 1 - K\convlength (1 - \mulower)^{\checkperiod}.
\]
We need to show that $K\convlength(1-\delta)^{\checkperiod} \leq \frac{\trafficslack}{32}$. With $\convlength\leq \frac{100K\checkperiod}{\trafficslack}(\log N +K)$ defined in Eq.~\ref{eq:parameter-setting}, it holds $K\convlength(1-\delta)^{\checkperiod} \leq \frac{100K^2\checkperiod}{\trafficslack}(\log N+K)(1-\delta)^{\checkperiod}$. It thus suffices to show
\[
\checkperiod\left(1 - \mulower\right)^{\checkperiod} \leq \xi \coloneqq \frac{\trafficslack^2}{3200K^2(\log N + K)}.
\]
Define $g(x) = x(1 - \mulower)^x$ for $x > 0$. Then $g'(x) = (1-\mulower)^x + x \ln(1 - \mulower)(1 - \mulower)^x$. When $x \geq x_1 \coloneqq \frac{1}{-\ln(1 - \mulower)},$ we have $g'(x) \leq 0,$ and thus $g(x)$ is a decreasing function after this threshold. Our goal is to find $x$ such that $g(x) \leq \xi$, which is equivalent to finding $\ln x + x\ln(1 - \mulower) \leq \ln \xi.$ Define $h(x) = \ln x + x\ln(1 - \mulower).$

Let $x_2 = \max\left(2,\left(\frac{2}{\ln(1-\mulower)}\right)^2\right).$ Note that for all $x \geq 1$, we have $\frac{\ln x}{x} \leq \frac{1}{\sqrt{x}}.$ Therefore, when $x \geq x_2,$ it holds $\frac{\ln x}{x} \leq \frac{1}{\sqrt{x}} \leq \frac{-\ln(1-\mulower)}{2},$ and thus $h(x) = \ln x + x\ln(1 - \mulower) \leq \frac{x\ln(1-\mulower)}{2}.$ To ensure $\frac{x\ln(1-\mulower)}{2} \leq \ln \xi, $ we only need $x \geq \frac{2\ln \xi}{\ln(1 - \mulower)}.$ Summarizing the above result, if
\[\checkperiod \geq \left\lceil\max\left(2, \left(\frac{2}{\ln(1-\mulower)}\right)^2, \frac{2\ln \xi}{\ln(1-\mulower)}\right)\right\rceil,\] we always have $\checkperiod(1-\mulower)^{\checkperiod} \leq \xi = \frac{\trafficslack^2}{3200K^2(\log N + K)}$, which implies  $\Pr\{\goodevent_{\ell}\} \geq 1 - \frac{1}{32}\trafficslack.$
\end{proof}

\subsection{Final guarantee for $\textsc{DAM.K}$ (full proof of \Cref{thm:queue-nolearning})}\label{app:proof_thm_queue_nolearning}

\begin{proof}[Proof of \Cref{thm:queue-nolearning}]
Fix a time slot $T$. Let $\ell = \lfloor \frac{T - 1}{\epochlength} \rfloor$ so that $T$ is in the $\ell+1$-th epoch. Now because the number of jobs of a queue can increase by at most one per time slot, we have:
\begin{align}
\expect{\sum_{i=1}^N \lambda_i\sum_{t=1}^T Q_i(t)} &\leq \expect{\sum_{i=1}^N \lambda_i \sum_{\tau=0}^{\ell} (\epochlength Q_i(\tau \epochlength + 1) + \epochlength^2)}\notag \\
&= \epochlength\expect{\sum_{i=1}^N \lambda_i \sum_{\tau=0}^{\ell} Q_i(\tau \epochlength + 1)} + K\epochlength^2(\ell+1).\label{eq:total_queue_decomp}
\end{align}
It is now sufficient to bound $\expect{\sum_{i=1}^N \lambda_i \sum_{\tau=0}^{\ell} Q_i(\tau \epochlength + 1)}$, which is the sum of weighted queue lengths in each starting time slot of previous epochs. We use \Cref{lem:epoch-drift-bound} and obtain:
\begin{align*}
&\mspace{25mu}\sum_{\tau=0}^{\ell} \expect{V(\bold{Q}((\tau+1)\epochlength + 1)) - V(\bold{Q}(\tau \epochlength + 1))} \\
&\leq \sum_{\tau=0}^{\ell} \left(\frac{5780K\convlength^2}{\trafficslack^2} - 2\convlength\sum_{i=1}^N \lambda_i \expect{Q_i(\tau \epochlength + 1)}\right) \\
&= \frac{5780(\ell+1)K\convlength^2}{\trafficslack^2} - 2\convlength\expect{\sum_{i=1}^N \lambda_i \sum_{\tau=0}^{\ell} Q_i(\tau \epochlength + 1)}.
\end{align*}
However, we also know that
\begin{align*}
\sum_{\tau=0}^{\ell} \expect{V(\bold{Q}((\tau+1)\epochlength + 1)) - V(\bold{Q}(\tau \epochlength + 1))} &= \expect{V(\bold{Q}((\ell + 1)\epochlength + 1)) - V(\bold{Q}(1))} \\
&= \expect{V(\bold{Q}((\ell + 1)\epochlength + 1))} \\
&\geq 0
\end{align*}
since all queues are assumed to be empty at the beginning. As a result we can write
\[
2\convlength\expect{\sum_{i=1}^N \lambda_i \sum_{\tau=0}^{\ell} Q_i(\tau \epochlength + 1)}\leq \frac{5780(\ell+1)K\convlength^2}{\trafficslack^2},
\]
which simplifies to
\[
\expect{\sum_{i=1}^N \lambda_i \sum_{\tau=0}^{\ell} Q_i(\tau \epochlength + 1)} \leq \frac{2890(\ell+1)K\convlength}{\trafficslack^2}.
\]
Consider $T \geq \epochlength + 1.$ Then we can substitute this expression in \eqref{eq:total_queue_decomp}, with $(\ell + 1)\epochlength \leq 2T$, to bound
\[
\expect{\sum_{i=1}^N \lambda_i\sum_{t=1}^T Q_i(t)} \leq \epochlength\frac{2890(\ell+1)K\convlength }{\trafficslack^2} +K\epochlength^2(\ell + 1) \leq 2T\left( \frac{2890K\convlength}{\trafficslack^2}+K\epochlength\right).
\]
Next, dividing both sides by $T$, and recalling that $\convlength = O\left(\frac{K\checkperiod}{\trafficslack}(K+\log N)\right), \epochlength = O\left(\frac{\convlength}{\trafficslack}\right)$, we have
\[
\expect{\frac{1}{T}\sum_{t=1}^T\sum_{i=1}^N \lambda_i Q_i(t)} = O\left(\frac{K\convlength}{\trafficslack} + \frac{K^2\checkperiod}{\trafficslack^3}\left(\log N + K\right)\right) = O\left(\frac{K^2\checkperiod}{\trafficslack^3}\left(\log N + K\right)\right).
\]
On the other hand, when $T < \epochlength + 1$, we know that 
\[
\expect{\frac{1}{T}\sum_{t=1}^T\sum_{i=1}^N \lambda_i Q_i(t)} \leq \epochlength\sum_{i=1}^N \lambda_i \leq K\epochlength = O\left(\frac{K^2\checkperiod}{\trafficslack^2}(K + \log N)\right),
\]
which completes the proof of \Cref{thm:queue-nolearning}. 
\end{proof}

\subsection{Matched servers do not become unmatched (Lemma~~\ref{lem:def-converge})}\label{app:not_unmatched}

\begin{proof}[Proof of Lemma~\ref{lem:def-converge}.]
We prove the result by induction. It suffices to show that if at time slot $t$, for each server, there is at most one request, then at time slot $t + 1$ with $R_j(t+1)=R_j(t)$. 
If $t \geq t_0 + \convlength - 1$, this
holds vacuously since queues commit to their last requested servers in the commit phase. Let us suppose $t < t_0 + \convlength - 1.$ consider a queue $i \in \mathcal{N}$ at time~$t + 1$ under two different cases: 
First, suppose that queue $i$ has negative payoff for all servers at the current prices and therefore it stopped requesting by time slot~$t$. Then $i$ does not request in time slot $t + 1$ either. Second, suppose $i$ requests server $j$ in time slot~$t$. The inductive assumption implies that $R_j(t) = \{i\}$, and the inductive step requires us to show that $i$ does not request a new server in time slot $t + 1$. Suppose that instead queue $i$ did request a new server which requires $ \tau_i(t) < t + 1 - \checkperiod$; we derive a contradiction to $\goodevent_{\ell}$. In this case it must be the case that $i$ has requested service in all periods from $t+1-\checkperiod$ to $t$. Within these periods, queue $i$ has not received service in any period as otherwise we would have $\tau_i(t) \geq t+1-\checkperiod$. However, the good checking event requires that server $j$ must fulfill service to some queue $i'$ with $\mu_{i',j} > 0$ in one of these periods. If $i' \neq i,$ then $i'$ requests service from server $j$ in period $t$ (as it would not update between receiving service and~$t$) and~$i' \in R_j(t)$ is a contradiction. Thus, we must have $i'=i$; but then we have $\tau_i(t) \geq t + 1 - \checkperiod$  which is also a contradiction. 
Therefore, queue $i$ does not update the server it requests from at $t + 1$.  
As this applies to all queues we have $R_j(t + 1) = R_j(t), \forall j \in \mathcal{K}$.
\end{proof}

\subsection{Unmatched queues increase their prices (Lemma~\ref{lem:interval-update})}\label{app:unmatched_queues_increase_prices}

\begin{proof}[Proof of Lemma~\ref{lem:interval-update}]
Let $i^\star$ be the queue in $R_j(t)$ referenced by the lemma, i.e., the one that does not update its price during $[t + 1,t + \checkperiod + 1].$ If no such $i^\star$ exists, the result trivially holds. Now define \begin{equation}
\begin{aligned}
\overline{R} &= \left\{i \in R_j(t) \mid j \text{ favors }i\text{ over }i^\star \text{ at }t\right\} \\
\ubar{R} &= \left\{i \in R_j(t) \mid j \text{ favors }i^\star\text{ over }i \text{ at }t\right\}.
\end{aligned}
\end{equation}
Since $i^\star$ does not update, it must successfully receive service from $j$ in some time slot $t' \in [t + 1,t + \checkperiod + 1].$ Now, each $i \in \overline{R}$ must do an update before $t'.$ If not, then at $t'$, server $j$ would still favor~$i$ over $i^\star,$ and thus not serve $i^\star$ which yields a contradiction. In addition, for each $i \in \ubar{R}$, we know that $i$ must do an update during $[t + 1,t + \checkperiod + 1]$. Indeed, if $i \in \ubar{R}$ never updates during the interval, then it must get a service according to Algorithm \ref{algo:DAM-queue}. However, since $i^{\star}$ is requesting service for all time slots in $[t+1,t+\checkperiod+1]$ and $i^\star$ is favored by server $j$ over all $i\in\ubar{R}$, this implies that $i\in\ubar{R}$ cannot receive service during this interval and will therefore have their price updated at the end of the interval.
\end{proof}

\subsection{Bound on convergence length within an epoch (Lemma~\ref{lem:algo-converge})}\label{app:algo_converge}
We prove the convergence by bounding the number of updates in the prices $p_{i,j}$ within an epoch. To do so, we let $\mathcal{A}(t)$ denote the set of queues that have at least one positive-payoff server at time slot $t$ and define the potential function $\Psi(t)=\sum_{i\in\mathcal{A}(t)}C_i(t)$ where $C_i(t)=\sum_{j\in\mathcal{K}}(1+\lceil\frac{w_{i,j}-p_{i,j}}{\beta_{i,j}}\rceil)$ denotes how many times a price of queue $i$ can be updated if all updates increment the price by $\beta_{i,j}=\frac{1}{16}\varepsilon w_{i,j}$. Our first lemma bounds the potential function of a system that has not converged.
\begin{lemma}\label{lem:potential_function}
Condition on $\goodevent_{\ell}$ and let $T_u = t_0 + (u - 1)(\checkperiod + 1)$ and $\mathcal{A}(T_u)=n$ for some $u \geq 1$ and $n\leq N$. Then, for any $z\geq u$, at time slot $T_{z}$, it holds that a) the system has converged or b) the potential function is upper bounded by $\Psi(T_{z})\leq n\frac{17K}{\trafficslack} - (z - u)$.
\end{lemma}

\begin{proof}
Consider an interval $[T_{z}, T_{z + 1} - 1], u \leq z < v$. 
By Lemma~\ref{lem:interval-update}, the potential function will decrease by at least~$\sum_{j \in \mathcal{K}: |R_j(T_{z})| > 1} |R_j(T_z)| - 1 \geq 1.$ If the algorithm does not converge at $T_z$, the potential function is upper bounded by
\begin{equation*}
\Psi(T_{z})=\sum_{i \in \mathcal{A}(T_u)} C_i(T_{z}) \leq \sum_{i \in \mathcal{A}(T_u)} C_i(T_u) - (z - u) \leq n\frac{17K}{\trafficslack} - (z - u).
\end{equation*}
\end{proof}
\begin{lemma}\label{lem:algo-conv-finite}
Condition on $\goodevent_{\ell}$ and let $T_u = t_0 + (u - 1)(\checkperiod + 1)$ and $\mathcal{A}(T_u)=n$ for some $u \geq 1$ and $n\leq N$. Then, if \textsc{DAM.converge} was run uninterrupted, the system would converge no later than at time slot  $T_u + \frac{24nK}{\trafficslack}\checkperiod$. Moreover, the system converges in a finite number of time slots.
\end{lemma}
\begin{proof}
Let $v = u + \lceil \frac{17nK}{\trafficslack}\rceil + 1$. Recall that the algorithm converges at time $t$ if for every server~$j$, it holds that $|R_j(t)| \leq 1.$ In addition, by \Cref{lem:def-converge}, as long as the algorithm converges at a time~$t$, the set $R_j$ will not change any more. Therefore, it suffices to show that $|R_j(T_v)| \leq 1$ for all $j$. 

Applying Lemma~\ref{lem:potential_function} with $z=v$, unless the algorithm converges at $T_v$, we have~$\sum_{i \in \mathcal{A}(T_v)} C_i(T_v) < 0$; this is a contradiction as $C_i(t)\geq0$. Hence, the decentralized queueing system must converge at $T_v$, and thus at~$T_u + \frac{18nK}{\trafficslack}(\checkperiod+1) \leq T_u +  \frac{24nK}{\trafficslack}\checkperiod$ because $\checkperiod \geq 3$ by assumption. Since $|\mathcal{A}(t_0)| = N$, the system converges before~$t_0 + \frac{24NK}{\trafficslack} \checkperiod$ (after finite time slots).
\end{proof}

\begin{proof}[Proof of Lemma~\ref{lem:algo-converge}] 
By Lemma~\ref{lem:algo-conv-finite}, the decentralized queueing system converges in $\frac{24 NK}{\trafficslack}\checkperiod$ time slots. When $N \leq 4K$, this already implies \Cref{lem:algo-converge}. However, when $N$ is much larger than $K$, this is much weaker than \Cref{lem:algo-converge} which depends only logarithmically on $N$. Let us now assume that $N \geq 4K$ and prove \Cref{lem:algo-converge}. The key observation that enables the proof is that $|\mathcal{A}(t)|$ decreases quickly from $N$ to $4K$ since at each time slot, at most $K$ queues can be selected. After $|\mathcal{A}(t)|$ drops below $4K$, \Cref{lem:algo-conv-finite} indicates that the system takes at most another  $\frac{96K^2}{\trafficslack}\checkperiod$ time slots to converge, and thus proves \Cref{lem:algo-converge}. We next present a formal proof of this result.

Fix a sample-path. To argue about how many queues are in $\mathcal{A}(t)$, we define a sequence $n_s = \lceil \frac{N}{2^s}\rceil$ for any $s\geq 0$ and denote by $p=\max\{s: n_{s}\geq 2K\}$ the final point in this sequence that we will consider. In addition, let $\hat{u}(s) = \min \left\{u \geq 0 \colon |\mathcal{A}(T_u)| \leq n_s\right\}$. Then, $\hat{u}(0) = 0.$ Note that~$\hat{u}(s)$ is well-defined and finite for every $s$ due to \Cref{lem:algo-conv-finite}. 

Now let us bound $\hat{u}_s - \hat{u}_{s-1}$ for $1 \leq s \leq p.$ By definition of $n_{s - 1},$ we have $\mathcal{A}(T_{\hat{u}(s-1)}) \leq n_{s - 1},$ and thus $\sum_{i \in \mathcal{A}(T_{\hat{u}(s-1)})} C_i(T_{\hat{u}(s-1)}) \leq \frac{17n_{s - 1}K}{\trafficslack}$ by Lemma~\ref{lem:potential_function}. Using the relation $\sum_{j = 1}^K |R_j(t)| = |\mathcal{A}(t)|$ for every time slot $t$ we know that for $t\in[T_{\hat{u}(s - 1)},T_{\hat{u}(s)})$. As a result, the number of queues that are updating their price vector is at least
\begin{equation*}
    \sum_{j \in \mathcal{K}, |R_j(t)| \geq 1} \big(|R_j(t)| - 1 \big)\geq |\mathcal{A}(t)| - K \geq n_s - K.
\end{equation*}
Furthermore, by \Cref{lem:interval-update}, for intervals $[T_{\hat{u}(s - 1)},T_{\hat{u}(s - 1) + 1}),\ldots,[T_{\hat{u}(s) - 1},T_{\hat{u}(s)})$, it follows that the number of updates per interval is lower bounded by $n_s-K$. Since the total potential at $T_{\hat{u}(s-1)}$ is bounded above by $\frac{17n_{s - 1}K}{\trafficslack}$, it then holds that the number of intervals is bounded by that quantity divided by $n_s-K$, i.e.,
\begin{equation*}
\hat{u}(s) - \hat{u}(s - 1) \leq \frac{1}{n_s - K}\frac{17n_{s - 1}K}{\trafficslack} \leq \frac{68K}{\trafficslack},
\end{equation*}
where the last inequality is due to $n_s \geq 2K$ and $n_{s - 1} \leq 2n_s.$ (as we take a geometric sequence starting from $2K$). As a result, 
\begin{equation*}
T_{\hat{u}(p)} - T_{\hat{u}(0)} \leq \left(\frac{68pK}{\trafficslack}\right)(\checkperiod + 1) \leq \frac{91\log N K}{\trafficslack}\checkperiod.
\end{equation*}
In addition, we know at $T_{\hat{u}(p)}$, $|\mathcal{A}(T_{\hat{u}(p)})| \leq 4K$ by the definition of $p$. Then, by \Cref{lem:algo-conv-finite}, it takes another $\frac{96K^2\checkperiod}{\trafficslack}$ time slots for the systems to converge. Summing the two terms, the number of times slots for the system to converge is at most
\begin{equation*}
\frac{91\log N K}{\trafficslack} \checkperiod+ \frac{96K^2\checkperiod}{\trafficslack} \leq \frac{99K\checkperiod}{\trafficslack}(K + \log N).
\end{equation*}
\end{proof}

\subsection{From complementary slackness to max-weight matching (Lemma~\ref{lem:lowslack-goodweight})
}\label{app:complementary_slackness_to_approx_matching}
\begin{proof}[Proof of Lemma~\ref{lem:lowslack-goodweight}]
Let $f_{i,j} = 1$ if $\sigma(i) = j.$ For notational convenience, we define 
\begin{equation}\label{eq:def-slack}
\begin{aligned}
u_{i,j} &= (\hat{\pi}_i + \hat{p}_j - w_{i,j})f_{i,j}, \qquad
v_i &= (1 - \sum_{j=1}^K f_{i,j})\hat{\pi}_i, \qquad and \quad
v'_j &= (1 - \sum_{i=1}^N f_{i,j})\hat{p}_j.
\end{aligned}
\end{equation}
Notice that $f$ is a feasible solution to (\ref{eq:primal}) and $(\hat{\boldsymbol{\pi}}, \hat{\bold{p}})$ is a feasible solution to (\ref{eq:dual}). Since the feasible set of (\ref{eq:primal}) is exactly $\Phi$, by weak duality, 
\begin{equation*}
\max_{\phi \in \Phi} \sum_{i=1}^N \sum_{j=1}^K \phi_{i,j}w_{i,j} \leq \sum_{i=1}^N \hat{\pi}_i + \sum_{j=1}^N \hat{p}_j.
\end{equation*}
In addition, rearranging terms using Eq.~\ref{eq:def-slack} implies that
\begin{equation*}
\left(\sum_{i=1}^N \hat{\pi}_i + \sum_{j=1}^N \hat{p}_j\right) - \sum_{i=1}^N w_{i, \sigma(i)}  = \sum_{i=1}^N \sum_{j=1}^K u_{i,j} + \sum_{i = 1}^N v_i + \sum_{j=1}^K v'_j.
\end{equation*}
By the complementary slackness assumption (Definition~\ref{def:comp-slack}),
if queue $i$ is not matched, then $\hat{\pi}_i = 0$, and similarly if server $j$ is not matched, then~$\hat{p}_j = 0$. Then, for every queue $i$ and every server $j$, we have $v_i = v'_j = 0$ by their definitions in (\ref{eq:def-slack}). In addition, $u_{i,j}$ = 0 when $\sigma(i) \neq j$ and $u_{i,\sigma(i)} \leq \alpha w_{i,\sigma(i)}$ by the third property in $\alpha-$complementary slackness. Thus, we obtain
\begin{align*}
\max_{\phi \in \Phi} \sum_{i=1}^N \sum_{j=1}^K \phi_{i,j}w_{i,j} - \sum_{i=1}^N w_{i, \sigma(i)} & \leq \sum_{i=1}^N \sum_{j=1}^K u_{i,j} + \underbrace{\sum_{i = 1}^N v_i + \sum_{j=1}^K v'_j}_{=0} \\
&\leq \sum_{i \in \mathcal{N}, \sigma(i) \not = \perp} \alpha w_{i,\sigma(i)} 
= \alpha \sum_{i=1}^N w_{i,\sigma(i)}
\leq \alpha \max_{\phi \in \Phi} \sum_{i=1}^N \sum_{j=1}^K \phi_{i,j}w_{i,j},
\end{align*}
which completes the proof.
\end{proof}

\subsection{Approximate complementary slackness of converged matching  (Lemma~\ref{lem:algo-efficient})}\label{app:satisfies_slackness}
We first present several preliminary results that pave the way to \Cref{lem:algo-efficient}. The next lemma shows that once a server receives a request, it will receive at least one request every time slot from then on for the whole epoch.
\begin{lemma}\label{lem:server-occupy}
Condition on $\goodevent_{\ell}$. If, for a server $j$ and a time slot $t \in [t_0,t_0 + \epochlength - 1]$, we have~$R_j(t) \neq \emptyset$, then for all $t' \in [t, t_0 + \epochlength - 1],$ it holds that $R_j(t') \neq \emptyset$.
\end{lemma}
\begin{proof}
Fix a server $j$. Suppose $|R_j(t)| \geq 1$ at some time slot $t$ and further that $|R_j(t')| = 0$ for all~$t'$ such that $t_0 \leq t' < t.$ If $t \geq t_0 + \convlength$, then the result trivially holds since it is in the commit phase. Suppose $t < t_0 + \convlength.$ Then the first request to server $j$ arrives at time $t$. Define an \emph{effective} switch as
the time that server $j$ successfully offers service to a queue different from the last successful one. Denote the time slots of effective switches at server $j$ before $t_0 + \convlength$ by $t_0 \leq t_1,t_2,\ldots,t_{b} \leq t_0 + \convlength - 1$ where $t_1$ is defined as the first time 
of a successful request and $b$ is the number of effective switches.
We note that $t_1$ must exist since the queue in $R_j(t)$ will not leave until $t + \checkperiod$. But by definition of event $\goodevent_{\ell}$, server $j$ offers service at least once during the interval $[t,t + \checkperiod - 1]$. As a result, it also holds that $t_1 \leq t + \checkperiod - 1.$ Further, the queue, that made the request in time slot $t$, remains at server $j$ throughout the interval $[t,t_1]$. 

Next, for each $1 \leq u < b$, we show there is at least one request to server $j$ in $[t_u, t_{u + 1} - 1].$ Denote the queue which gets served at $t_u$ by queue $i$. Queue $i$ continues to request server $j$ during~$[t_u,t_u + \checkperiod]$, which guarantees that $j$ receives at least one request in every time slot of the interval $[t_u, t_{u + 1} - 1]$ if $t_{u + 1} \leq t_u + \checkperiod$. If instead~$t_{u + 1} > t_u + \checkperiod$, by definition of $\goodevent_{\ell},$ server $j$ successfully offers service at least once in $[t_u + 1,t_u + \checkperiod].$ Since the next time server $j$ serves another queue successfully is $t_{u + 1},$ we know the request from $i$ is successful at least one time slot during $[t_u + 1,t_u + \checkperiod]$, and thus queue $i$ continues to request service from $j$ throughout $[t_u + 1,t_u + \checkperiod + 1].$ Then, by induction, it keeps requesting server $j$ until at least~$t_{u + 1} - 1,$ and thus $R_j(t') \neq \emptyset$ for $t' \in [t_u,t_{u + 1} - 1].$

Finally, let us check time slots in $[t_{b},t_0 + \epochlength - 1]$. Denote by $i$ the queue that successfully gets served at $t_{b}$. Following the same argument as above, it holds that queue $i$ will keep requesting server $j$ from $t_{b}$, which thus finishes the proof. 
\end{proof}

The following lemma shows that the price $\hat{p}_j(t)$ defined at Eq.~\ref{eq:def-hat-price} in the proof sketch is monotone increasing in $t$, and the price $p_{i,j}(t)$ maintained by each queue is an underestimate of $\hat{p}_j(t).$

\begin{lemma}\label{lem:monotone-price}
Condition on $\goodevent_{\ell}$. Fix a time slot $t \geq t_0$. For each server $j$, price $\hat{p}_j(t)$ satisfies that
\begin{enumerate}
\item $\hat{p}_j(t) \leq \hat{p}_j(t + 1);$
\item for a queue $i \in \mathcal{N}$, if $p_{i,j}(t + 1) > \hat{p}_j(t),$ then $i \in R_j(t + 1);$
\item for every queue $i \in \mathcal{N}$, $p_{i,j}(t) \leq p_{i,j}(t + 1) \leq \hat{p}_j(t + 1).$
\end{enumerate}
\end{lemma}
\begin{proof}
For a queue $i$ and a server $j$, \textsc{DAM.converge} never decreases its price estimate $p_{i,j}(t)$; hence  $p_{i,j}(t + 1) \geq p_{i,j}(t).$ We now prove the remaining results by induction.  

\textbf{Induction basis.} We first show that the lemma holds for $t=t_0$. 1) We know that $\hat{p}_j(t_0 + 1) \geq 0 = \hat{p}_j(t_0).$ 2) In addition, if $p_{i,j}(t_0 + 1) > 0,$ then queue $i$ must update its price at $t_0 + 1$, and request server $j$, so $i \in R_j(t_0 + 1).$ 3) Moreover, $\hat{p}_j(t_0 + 1) = \max_{i' \in R_j(t_0+1)} p_{i',j}(t_0 + 1) \geq \max_{i'} p_{i',j}(t_0 + 1)$. 

\textbf{Induction from $t-1$ to $t$.} Now suppose the results hold for all time slots before $t$, and $t \geq t_0 + 1.$
1) First, we show $\hat{p}_j(t) \leq \hat{p}_j(t + 1).$ If $\hat{p}_j(t) = 0,$ the inequality holds vacuously true. Otherwise, there exists $i' \in R_j(t),$ such that $\hat{p}_j(t) = p_{i',j}(t)$ by Eq.~\eqref{eq:def-hat-price}. If $i'$ is also in $R_j(t + 1),$ then $\hat{p}_j(t+1) \geq p_{i',j}(t+1) \geq p_{i',j}(t) = \hat{p}_j(t).$ If instead~$i' \not \in R_j(t+1), $ then $i'$ must do a price update, which means that $t + 1 - \tau_{i'}(t) > \checkperiod, $ and thus $i'$ requests $j$ throughout $[t - \checkperiod + 1, t]$ without receiving service. By definition of $\goodevent_{\ell}$, server $j$ must successfully provide service at least once in $[t - \checkperiod + 1,t].$ Denote the queue that receives the service by $x$, and the time of the service by $t'$. We know $x$ must be requesting $j$ at $t + 1$ by line~\ref{algoline:no-update-price} in Algorithm \ref{algo:DAM-queue}. As a result,
\begin{equation*}
\hat{p}_j(t + 1) \geq p_{x,j}(t + 1) \geq p_{x,j}(t') \geq p_{i',j}(t') = p_{i',j}(t) = \hat{p}_j(t),
\end{equation*}
where the first relation is because $x \in R_j(t+1)$; the second one is because $t'<t+1$; the third one is because $j$ selects $x$ instead of $i'$ at time $t'$; the forth one is because~$i'$ does not update in~$[t - \checkperiod + 1, t]$ and $t'$ is in this interval.

2) Next for a queue $i$, we know by definition of $\hat{p}_j(t)$ that $p_{i,j}(t) \leq \hat{p}_j(t).$ Since $p_{i,j}(t+1) > \hat{p}_j(t) \geq p_{i,j}(t),$ queue $i$ updates its price of server $j$ at time slot $t+1$, and we must have $i \in R_j(t+1).$

3) Finally, to show $p_{i,j}(t + 1) \leq \hat{p}_j(t + 1),$ notice that if $p_{i,j}(t + 1) > \hat{p}_j(t),$ then $i \in R_j(t + 1)$, and thus $p_{i,j}(t + 1) \leq \hat{p}_j(t + 1).$ On the other hand, if~$p_{i,j}(t + 1) \leq \hat{p}_j(t),$ then $p_{i,j}(t + 1) \leq \hat{p}_j(t) \leq \hat{p}_j(t + 1)$ where the last inequality was shown above.
\end{proof}

Recall (proof sketch of Lemma~\ref{lem:algo-efficient}) that the payoff of agent $i$ based on its own price is
$\pi_i(t) = \max\left(\max_{j \in \mathcal{K}} w_{i,j} - p_{i,j}(t),0\right)$
Based on \Cref{lem:monotone-price}, the next lemma shows that the defined payoff $\pi_i(t)$ and prices $\hat{\bold{p}}$ always satisfy the first and third conditions of $\frac{1}{16}\trafficslack-$complementary slackness.
\begin{lemma}\label{lem:feasible-slack}
Condition on $\goodevent_{\ell}$ and fix a queue $i$. For all $t \geq t_0,$  it holds that
\begin{itemize}
\item for each server $j \in \mathcal{K}$, 
\begin{equation}\label{eq:dual-feasible}
    \pi_i(t) + \hat{p}_j(t) \geq w_{i,j};
\end{equation}
\item if $J(i,t) \neq \perp$, then 
\begin{equation}\label{eq:close-slack}
    \pi_i(t) + p_{i,J(i,t)}(t) \leq w_{i,J(i,t)} + \frac{1}{16}\trafficslack w_{i,J(i,t)}.
\end{equation}
\end{itemize}
\end{lemma}
\begin{proof}
Fix a queue $i$, we first prove the result for $t=t_0$. Then it holds that $\pi_i(t_0) = \max_{j \in \mathcal{K}} w_{i,j}$.

If $\pi_i(t_0) > 0,$ then $J(i,t_0) \neq \perp$. In addition, since $p_{i,j}(t_0) = 0$ for all $j$, (\ref{eq:dual-feasible}) and (\ref{eq:close-slack}) satisfy naturally. Now we prove the lemma by induction; assuming it is true in every time slot $t_0,\ldots, t-1$ we prove it for $t$. 

There are two cases. First, if $i$ does not update its price at $t$, then $\pi_i(t) = \pi_i(t - 1), J(i,t) = J(i,t - 1), p_{i,j}(t) = p_{i,j}(t-1)$ for all $j$. Therefore, since (\ref{eq:close-slack}) holds for $t - 1$, it also holds for $t$. For (\ref{eq:dual-feasible}), since $\hat{p}_j(t) \geq \hat{p}_j(t - 1)$ by \Cref{lem:monotone-price}, we also have $\pi_i(t) + \hat{p}_j(t) \geq w_{i,j}.$ 

For the second case suppose queue $i$ updates it price at $t$. We first prove that (\ref{eq:dual-feasible}) still holds. Recall that $\pi_i(t) = \max(\max_j(w_{i,j} - p_{i,j}(t)), 0).$ Then for each $j \in \mathcal{K}$, we have
\begin{equation*}
\pi_i(t) + \hat{p}_j(t) \geq w_{i,j}-p_{i,j}(t) + \hat{p}_j(t) \geq w_{i,j},
\end{equation*}
where the second inequality is because $\hat{p}_j(t) \geq p_{i,j}(t)$ by \Cref{lem:monotone-price}. Next we prove (\ref{eq:close-slack}). Let $j^{\star}$ denote  $J(i,t) \not = \perp.$ Note that $j^{\star}$ is chosen to maximize $w_{i,j^{\star}} - p_{i,j^{\star}}(t - 1).$ Therefore,
\begin{align*}
\pi_i(t) &= \max\left(0,\max_j w_{i,j} - p_{i,j}(t)\right) \\
&= \max\left(0, \max_j w_{i,j} - p_{i,j}(t-1) - (p_{i,j}(t) - p_{i,j}(t-1))\right) \\
&\overset{(a)}{\leq} \max\left(0, \max_j w_{i,j} - p_{i,j}(t-1)\right) \\
&\overset{(b)}{=} \max(0, w_{i,j^{\star}} - p_{i,j^{\star}}(t-1)) \\
&\overset{(c)}{=} w_{i,j^{\star}} - p_{i,j^{\star}}(t-1),
\end{align*}
where (a) holds because $p_{i,j}(t) \geq p_{i,j}(t-1)$; (b) holds because $j^{\star}$ is chosen to maximize $w_{i,j^{\star}} - p_{i,j^{\star}}(t - 1)$; (c) holds because if we had $w_{i,j^{\star}} - p_{i,j^{\star}}(t-1) < 0,$ queue $i$ would not be requesting. Since the update rule in Algorithm \ref{algo:DAM-queue} implies $p_{i,j^{\star}}(t) - p_{i,j^{\star}}(t-1) = \frac{1}{16}\trafficslack w_{i,j^{\star}},$ we have
\begin{equation*}
\pi_i(t) + p_{i,j^{\star}}(t) = \pi_i(t) + p_{i,j^{\star}}(t-1) + \frac{1}{8}\trafficslack\mu_{i,j^{\star}} Q_i(t_0) \leq w_{i,j^{\star}} + \frac{1}{16}\trafficslack w_{i,j^{\star}},
\end{equation*}
which completes the proof.
\end{proof}
The next lemma provides additional connections between $\pi_i$ and the induced matching when the algorithm converges.
In the next lemma we derive additional properties for queues that are unmatched by the algorithm.
\begin{lemma}\label{lem:unmatched-value}
Condition on $\goodevent_{\ell}$. Suppose the system converges at time $t$. Then if queue $i$ is not matched, i.e., $i \not \in \cup_{j\in\mathcal{K}}R_j(t),$ then: 1) $\pi_i(t) = 0;$ 2) every server $j$ with $w_{i,j} > 0$ has $|R_j(t)| = 1.$
\end{lemma}
\begin{proof}
Suppose the system
converges in time slot $t$. Convergence of queues in Algorithm \ref{algo:DAM-queue} requires for each server $j$ that~$|R_j(t)| \leq 1.$ In addition, since queue $i$ is not matched, we know $J(i,t) = \perp,$ and thus $w_{i,j} \leq p_{i,j}(t)$ for all $j$ according to line~\ref{algoline:when-to-request} in Algorithm \ref{algo:DAM-queue}. By the definition of $\pi_i(t)$ that $\pi_i(t) = \max\left(\max_{j \in \mathcal{K}} w_{i,j} - p_{i,j}(t),0\right)$, it holds $\pi_i(t) = 0.$ On the other hand, consider a server $j$ with $w_{i,j} > 0.$ We know $p_{i,j}(t) \geq w_{i,j} > 0$. For~$p_{i,j}(t)>0$ to hold, $i$ must have requested $j$ at some time $t' < t$ before increasing $p_{i,j}(t)$ from 0; then, by Lemma~\ref{lem:server-occupy}, $R_j(t') \neq \emptyset, $ guarantees that we have $|R_j(t)| > 0,$ which completes the proof.
\end{proof}

\begin{proof}[Proof of \Cref{lem:algo-efficient}]
Suppose the algorithm converges in time slot $t$. Define the matching $\sigma$ by setting $\sigma(I(j,t)) = j$ for server $j$ with $I(j,t) \neq \perp$. In addition, define $\hat{\pi}_i = \max(0, \max_{j \in \mathcal{K}} w_{i,j} - \hat{p}_j(t)).$ Note that $\hat{\pi}_i$ need not be the same as $\pi_i(t)$. We proceed by verifying the conditions of $\frac{\trafficslack}{16}-$complementary slackness. The first property is satisfied since $\hat{\pi}$ is defined as required. For a queue $i$, if $\sigma(i) = \perp,$ by \Cref{lem:unmatched-value}, we have $\pi_i(t) = \max(0, \max_j w_{i,j} - p_{i,j}(t)) = 0.$ Since \Cref{lem:monotone-price} shows $\hat{p}_j(t) \geq p_{i,j}(t),$ we know $\hat{\pi}_i \leq \pi_i(t) = 0,$ so $\hat{\pi}_i = 0.$ 

Finally, if $\sigma(i) \neq \perp$, we still have $\hat{\pi}_i \leq \pi_i(t)$. Since we also know $J(i,t) = \sigma(i) \neq \perp,$ by Lemma~\ref{lem:feasible-slack}, we have
\begin{equation*}
\hat{\pi}_i + \hat{p}_{\sigma(i)}(t) \leq \pi_i(t) + \hat{p}_{\sigma(i)}(t) = \pi_i(t) + p_{i,\sigma(i)}(t) \leq w_{i,\sigma(i)} + \frac{1}{16}\trafficslack w_{i,\sigma(i)}.
\end{equation*}
\end{proof}

\subsection{Bounding the drift during $\textsc{DAM.converge}$ (Lemma~\ref{lem:epoch-drift-converge})}\label{app:epoch-drift-converge}
\begin{lemma}\label{lem:bound-general-drift}
Fix a time slot $t_1 \geq 1$. Consider a future interval $[t_2,t_3 + 1], t_1 \leq t_2 \leq t_3,$ and an event $\set{W}$ that is independent of all arrivals in $[t_1,t_3].$ Then it holds that 
\begin{equation*}
\expect{V(\bold{Q}(t_3 + 1)) - V(\bold{Q}(t_2)) \mid \bold{Q}(t_1), \set{W}}  \leq  (t_3 - t_2 + 1)\left(2\sum_{i=1}^N \lambda_{i}Q_{i}(t_1) + K\left(1 +  t_3+t_2-2t_1\right)\right).
\end{equation*}
\end{lemma}
\begin{proof}
For $t' \in [t_2,t_3]$, by the dynamics of queue $i$ (Section~\ref{sec:prelims}),
it holds that $Q_i(t') \leq Q_i(t_1) + A_i(t_1) + \ldots + A_i(t' - 1)$.
As a result,
\begin{equation}\label{eq:bound-simple-queue}
\begin{aligned}
\expect{Q_i(t') \mid Q_i(t_1),\set{W}} &\leq Q_i(t_1) + \expect{A_i(t_1)+\ldots+A_i(t' - 1) \mid Q_i(t_1),\set{W}} \\
&= Q_i(t_1) + \lambda_i(t'-t_1),
\end{aligned}
\end{equation}
where the equality is due to independence between $Q_i(t_1), \set{W}$ and arrivals in $[t_1,t_2]$.

Moreover, notice that $((Q_i(t') + A_i(t') - S_i(t'))^+)^2 \leq (Q_i(t')+A_i(t'))^2$. Thus, we have
\begin{align}
\expect{V(\bold{Q}(t' + 1)) - V(\bold{Q}(t')) \mid \bold{Q}(t_1), \set{W}} 
&= \expect{\sum_{i=1}^N ((Q_{i}(t')+A_{i}(t')-S_{i}(t'))^+)^2 - Q^2_{i}(t') \mid \bold{Q}(t_1),\set{W}}\nonumber \\
&\leq \expect{\sum_{i=1}^N (Q_{i}(t')+A_{i}(t'))^2 - Q^2_{i}(t') \mid \bold{Q}(t_1),\set{W}}\nonumber \\
&= \expect{2\sum_{i=1}^N A_i(t')Q_i(t')+\sum_{i=1}^N A^2_{i}(t') \mid \bold{Q}(t_1), \set{W}}\nonumber \\
&= 2\expect{\sum_{i=1}^N \lambda_{i}Q_{i}(t') \mid \bold{Q}(t_1), \set{W}} + \sum_{i=1}^N \lambda_i \nonumber\\
&\leq 2\expect{\sum_{i=1}^N \lambda_{i}Q_{i}(t') \mid \bold{Q}(t_1), \set{W}}+K, \label{eq:bound-one-step}
\end{align}
where the last equality follows from $A_i(t')$ being a Bernoulli random variable with mean $\lambda_i$ that is independent of $\bold{Q}(t_1)$ and $\set{W}$. The last inequality is due to \Cref{as:stability}.
Finally, we can see that
\begin{align*}
\expect{V(\bold{Q}(t_3 + 1)) - V(\bold{Q}(t_2)) \mid \bold{Q}(t_1), \set{W}}
&= \sum_{t'=t_2}^{t_3} \expect{V(\bold{Q}(t' + 1)) - V(\bold{Q}(t')) \mid \bold{Q}(t_1), \set{W}} \\
&\overset{(\ref{eq:bound-one-step})}{\leq} \sum_{t'=t_2}^{t_3} \left( 2\expect{\sum_{i=1}^N \lambda_i Q_i(t') \mid \bold{Q}(t_1), \set{W}}+K\right) \\
&\overset{(\ref{eq:bound-simple-queue})}{\leq} 2\sum_{t'=t_2}^{t_3} \sum_{i=1}^N \lambda_i\Big(Q_i(t_1) + \lambda_i(t'-t_1)\Big) + K(t_3-t_2+1) \\
&\leq (t_3-t_2+1)\left(2\sum_{i=1}^N \Big(\lambda_iQ_i(t_1)\Big) + K\right) + \sum_{t'=t_2}^{t_3}2(t'-t_1)K \\
&=(t_3 - t_2 + 1)\left(2\sum_{i=1}^N \lambda_{i}Q_{i}(t_1) + K\left(1 +  t_3+t_2-2t_1\right)\right),
\end{align*}
where the last inequality is because $\sum_{i=1}^N \lambda^2_i \leq \sum_{i=1}^N \lambda_i \leq K$ under \Cref{as:stability}. 
\end{proof}

\begin{proof}[Proof of Lemma~\ref{lem:epoch-drift-converge}]
\Cref{lem:bound-general-drift} immediately implies Lemma~\ref{lem:epoch-drift-converge} by showing an upper bound of~$\expect{V(\bold{Q}(t_0+\convlength)) - V(\bold{Q}(t_0)) \mid \bold{Q}(t_0)}$. Specifically, take $t_1 = t_2 = t_0, t_3 = t_0 + \convlength - 1,\set{W}=\emptyset,$ Lemma~\ref{lem:bound-general-drift} implies
\begin{align*}
&\mspace{32mu}\expect{V(\bold{Q}(t_0+\convlength)) - V(\bold{Q}(t_0)) \mid \bold{Q}(t_0)} \\
&\leq (t_0+\convlength-1-t_0+1)\left(2\sum_{i=1}^N \lambda_i Q_i(t_0) + K(1+t_0+\convlength-1+t_0-2t_0)\right) \\
&= 2\convlength\sum_{i=1}^N \lambda_i Q_i(t_0) + K\convlength^2.
\end{align*}
\end{proof}

\subsection{Bounding the drift during $\textsc{DAM.commit}$ (Lemma~\ref{lem:epoch-drift-matching})}\label{app:epoch-drift-matching}

\paragraph{Drift when good checking event does not hold.} We first upper bound the drift conditioning on $\goodevent_{\ell}^c$ (second term in Eq.~\eqref{eq:drift-decomposition}). Notice that $\goodevent_{\ell}^c$ is independent from all arrivals in $[t_0, t_0 + \epochlength]$ since arrivals are exogenous. Therefore, applying \Cref{lem:bound-general-drift} again gives us the following bound. 
\begin{lemma}\label{lem:drift-bad-event}
It holds that 
\begin{align*}
&\mspace{32mu}
\expect{V(\bold{Q}(t_0+\epochlength)) - V(\bold{Q}(t_0+\convlength)) \mid \bold{Q}(t_0), \goodevent_{\ell}^c} \\
&\leq (\epochlength-\convlength)\left(2\sum_{i=1}^N \lambda_i Q_i(t_0) + K(\epochlength + \convlength)\right).
\end{align*}
\end{lemma}

\paragraph{Drift when good checking event holds}
It remains to upper bound the drift conditioning on $\goodevent_{\ell}$. Due to the ``converge-commit'' framework of Algorithm~\ref{algo:DAM-master}, we have that $\goodevent_{\ell}$ is independent of the services happening in $[t_0 + \convlength, t_0 + \epochlength - 1]$. 
In addition, recall from \Cref{lem:algo-converge} in Section \ref{sec:property}  that, conditioning on $\goodevent_{\ell}$, with Algorithm \ref{algo:DAM-queue} the queues are guaranteed to converge to a fixed matching $\sigma$ by time slot $t_0 + \convlength$.
Moreover, by \Cref{lem:algo-converge-main}, we know that
\begin{equation}\label{eq:esti-approx-opt}
\sum_{i=1}^N \mu_{i,\sigma(i)}Q_i(t_0) \geq (1 - \frac{1}{16}\trafficslack)\max_{\phi \in \Phi} \sum_{i=1}^N \sum_{j=1}^K \phi_{i,j}Q_i(t_0),
\end{equation}
where $\set{S}$ is the set of feasible solutions of (\ref{eq:primal}).
Next, we show that the weight of the converged matching, $\sum_{i=1}^N \mu_{i,\sigma(i)}Q_i(t_0)$, can be approximately bounded from below by $\sum_{i=1}^N\lambda_i Q_i(t_0)$. 

\begin{lemma}\label{lem:drift-matching-weight}
Condition on $\goodevent_{\ell}$, the converged matching $\sigma$ fulfills
\[
\sum_{i=1}^N \mu_{i,\sigma(i)}Q_i(t_0) \geq (1+\frac{1}{2}\trafficslack)\sum_{i=1}^N \lambda_i Q_i(t_0).
\]
\end{lemma}
\begin{proof}
By the stability assumption (\Cref{as:stability}), there exists a feasible solution $\phi$ to the matching problem (\ref{eq:primal}), such that for all $i \in \set{N}$, $(1+\trafficslack)\lambda_i \leq \sum_{j=1}^K \mu_{i,j}\phi_{i,j}.$ As a result, we have
\[
(1+\trafficslack)\sum_{i=1}^N \lambda_i Q_i(t_0) \leq \sum_{i=1}^N \sum_{j=1}^K \phi_{i,j}\mu_{i,j}Q_i(t_0).
\]
Observe that  (\ref{eq:esti-approx-opt}) implies that
\[
\sum_{i=1}^N \tilde{\mu}_{i,\sigma(i)}Q_i(t_0) \geq (1-\frac{1}{8}\trafficslack)\sum_{i=1}^N \sum_{j=1}^K \phi_{i,j} \tilde{\mu}_{i,j} Q_i(t_0).
\]
In addition, since $\tilde{\mu}_{i,j} = \mu_{i,j}$ we can combine these inequalities as 
\[
\sum_{i=1}^N \tilde{\mu}_{i,\sigma(i)}Q_i(t_0) \geq (1-\frac{1}{8}\trafficslack)\sum_{i=1}^N \sum_{j=1}^K \phi_{i,j} \tilde{\mu}_{i,j}
\geq (1-\frac{1}{8}\trafficslack)(1+\trafficslack)\sum_{i=1}^N \lambda_i Q_i(t_0).
\]
We further bound this expression from below by $(1 + \frac{\trafficslack}{2})\sum_{i=1}^N \lambda_i Q_i(t_0)$, which completes the proof of the lemma.
\end{proof}
The above lemma shows that conditioning on $\goodevent_{\ell}$, the queues maintain a matching with provably large weight throughout time slots $t_0 + \convlength, \ldots,t_0 + \epochlength - 1.$ 
The following lemma translates such approximate optimality into an upper bound on the single-step drift.
\begin{lemma}\label{lem:single-step-drift-good-event}
Fix $d \in [\convlength, \epochlength - 1].$ It holds
\begin{equation}\label{eq:single-step-drift-good-event}
\expect{V(\bold{Q}(t_0 + d + 1)) - V(\bold{Q}(t_0 + d)) \mid \bold{Q}(t_0), \goodevent_{\ell}} \leq 6d K - \frac{\trafficslack}{2}\sum_{i=1}^N \lambda_i Q_i(t_0).
\end{equation}
\end{lemma}
\begin{proof}
Condition on $\goodevent_{\ell}$. \Cref{lem:algo-converge} shows that queues converge to the matching $\sigma$ in time slot $t_0 + \convlength$, and stay with it during $[t_0 + \convlength, t_0 + \epochlength - 1].$ By invoking the definition of drift:
\begin{equation}
\begin{aligned}
&\mspace{25mu}\expect{V(d + 1)) - V(\bold{Q}(t_0 + d)) \mid \bold{Q}(t_0), \goodevent_{\ell}} \\
&= \expect{\sum_{i=1}^N \left(Q_i(t_0+d)+A_i(t_0+d)-S_{i,\sigma(i)}(t_0+d)\right)^2 - \sum_{i=1}^N Q_i^2(t_0+d) \mid \bold{Q}(t_0),\goodevent_{\ell}} \\
&\leq \expect{2\sum_{i=1}^N Q_i(t_0+d)A_i(t_0+d)-2\sum_{i=1}^N Q_i(t_0+d)S_{i,\sigma(i)}(t_0+d)\mid \bold{Q}(t_0), \goodevent_{\ell}} \\
&\mspace{32mu}+\underbrace{\expect{\sum_{i=1}^N A_i^2(t_0+d)\mid \bold{Q}(t_0), \goodevent_{\ell}}}_{\leq K} +\underbrace{\expect{\sum_{i=1}^N S^2_{i,\sigma(i)}(t_0+d)\mid \bold{Q}(t_0), \goodevent_{\ell}}}_{\leq K} \label{eq:single-step-drift}.
\end{aligned}
\end{equation}
Now we know that $A_i(t_0+d)$ is independent of $\bold{Q}(t_0), \goodevent_{\ell}$ and $\expect{A_i(t_0+d)} = \expect{A^2_i(t_0+d)} = \lambda_i$. In addition, since $\sigma$ is a matching, at most $K$ of $S_{i,\sigma(i)}, i \in \mathcal{N}$ can be one, and the others are all zero. Furthermore, by the queueing dynamic (\ref{eq:queuedynamic}), each queue $i$ fulfills 
\begin{equation}
    Q_i(t_0) - d \leq Q_i(t_0 + d) \leq Q_i(t_0) + d.
\end{equation}
Therefore, we have
\begin{align}
\eqref{eq:single-step-drift} &\leq 2\expect{\sum_{i=1}^N \underbrace{Q_i(t_0+d)}_{\leq Q_i(t_0)+d}\lambda_i - 2\sum_{i=1}^N \underbrace{Q_i(t_0+d)}_{\geq Q_i(t_0)-d}S_{i,\sigma(i)}(t_0+d) \mid \bold{Q}(t_0), \goodevent_{\ell}} + 2K \notag\\
&\leq 2 \expect{\sum_{i=1}^N (Q_i(t_0)\lambda_i + d \lambda_i) \mid \bold{Q}(t_0), \goodevent_{\ell}} - 2 \expect{\sum_{i=1}^N  (Q_i(t_0) -  d)S_{i,\sigma(i)}(t_0+d) \mid \bold{Q}(t_0), \goodevent_{\ell}}+2K \notag\\
&\leq 6d K + 2\sum_{i=1}^N \lambda_i Q_i(t_0) - 2\sum_{i=1}^N Q_i(t_0)\expect{S_{i,\sigma(i)}(t_0+d) \mid \bold{Q}(t_0), \goodevent_{\ell}}.\label{eq:single-step-drift-final}
\end{align}
The last inequality is based  on $K$ upper bounding $\sum_{i=1}^N\lambda_i$ and $\sum_{i=1}^N \expect{S_{i,\sigma(i)}(t_0+d) \mid \bold{Q}(t_0), \goodevent_{\ell}}$. We now lower bound $\sum_{i=1}^N Q_i(t_0)\expect{S_{i,\sigma(i)}(t_0+d) \mid \bold{Q}(t_0), \goodevent_{\ell}}$. By the tower property of conditional expectation, 
\begin{align*}
\sum_{i=1}^N Q_i(t_0)\expect{S_{i,\sigma(i)}(t_0+d) \mid \bold{Q}(t_0), \goodevent_{\ell}} &= \expect{\expect{\sum_{i=1}^N Q_i(t_0)S_{i,\sigma(i)}(t_0+d) \mid \sigma,\bold{Q}(t_0), \goodevent_{\ell}} \mid \bold{Q}(t_0), \goodevent_{\ell}} \\
&= \expect{\expect{\sum_{i=1}^N Q_i(t_0)\mu_{i,\sigma(i)} \mid \sigma,\bold{Q}(t_0), \goodevent_{\ell}} \mid \bold{Q}(t_0), \goodevent_{\ell}} \\
&= \expect{\sum_{i=1}^N Q_i(t_0)\mu_{i,\sigma(i)} \mid \bold{Q}(t_0), \goodevent_{\ell}} \\
&\geq (1 + \frac{1}{2}\trafficslack)\sum_{i=1}^N \lambda_i Q_i(t_0),
\end{align*}
where the second equation is because $\bold{Q}(t_0),\goodevent_{\ell}$ are independent of $S_{i,j}(t_0 + d)$ for any fixed $i \in \set{N}, j \in \set{K}, d \geq \convlength$, and the last inequality is due to Lemma~\ref{lem:drift-matching-weight}. Substituting the right-handside into (\ref{eq:single-step-drift-final}), we can conclude that 
\[
\expect{V(\bold{Q}(t_0 + d + 1)) - V(\bold{Q}(t_0 + d)) \mid \bold{Q}(t_0), \goodevent_{\ell}} \leq 6dK - \frac{1}{2}\trafficslack \sum_{i=1}^N \lambda_i Q_i(t_0).
\]
\end{proof}

We are now ready to bound $\expect{V(\bold{Q}(t_0+\epochlength)) - V(\bold{Q}(t_0 + \convlength)) \mid \bold{Q}(t_0)}$ by conditioning on either~$\goodevent_{\ell}$ or~$\goodevent_{\ell}^c$, which finishes the proof of Lemma~\ref{lem:epoch-drift-matching}.
\begin{proof}[Proof of Lemma~\ref{lem:epoch-drift-matching}]
We have 
\[
\begin{aligned}
&\mspace{25mu}
\expect{V(\bold{Q}(t_0+\epochlength)) - V(\bold{Q}(t_0 + \convlength)) \mid \bold{Q}(t_0)} \\
&= \expect{V(\bold{Q}(t_0+\epochlength)) - V(\bold{Q}(t_0 + \convlength)) \mid \bold{Q}(t_0), \goodevent_{\ell}}\Pr\{\goodevent_{\ell}\} \\
&\mspace{25mu}+ \expect{V(\bold{Q}(t_0+\epochlength)) - V(\bold{Q}(t_0 + \convlength)) \mid \bold{Q}(t_0),\goodevent_{\ell}^c}\Pr\{\goodevent_{\ell}^c\}.
\end{aligned}
\]
By \Cref{lem:drift-bad-event}, 
\begin{align*}
&\mspace{32mu}\expect{V(\bold{Q}(t_0+\epochlength)) - V(\bold{Q}(t_0 + \convlength)) \mid \bold{Q}(t_0),\goodevent_{\ell}^c} \\
&\leq (\epochlength-\convlength)\left(2\sum_{i=1}^N \lambda_i Q_i(t_0) + K(\epochlength + \convlength)\right).
\end{align*}
In addition, \Cref{lem:single-step-drift-good-event} shows that 
\[
\expect{V(\bold{Q}(t_0+\epochlength)) - V(\bold{Q}(t_0 + \convlength)) \mid \bold{Q}(t_0),\goodevent_{\ell}} \leq \sum_{d=\convlength}^{\epochlength - 1} (6d K - \frac{1}{2}\trafficslack \sum_{i=1}^N \lambda_i Q_i(t_0)).
\]
Therefore,
\[
\begin{aligned}
&\mspace{25mu}\expect{V(\bold{Q}(t_0+\epochlength)) - V(\bold{Q}(t_0 + \convlength)) \mid \bold{Q}(t_0)} \\
&\leq \Pr\{\goodevent_{\ell}^c\} (\epochlength-\convlength)\left(2\sum_{i=1}^N \lambda_i Q_i(t_0) + K(\epochlength + \convlength)\right) \\
&\mspace{25mu} + \Pr\{\goodevent_{\ell}\}\left(3K(\epochlength-1+\convlength)(\epochlength-\convlength)-\frac{1}{2}\trafficslack(\epochlength-\convlength)\sum_{i=1}^N \lambda_i Q_i(t_0)\right).
\\
&\leq (\epochlength-\convlength)\left[5K(\epochlength-1+\convlength) - \sum_{i=1}^N \lambda_i Q_i(t_0)\left(\frac{1}{2}\Pr\{\goodevent_{\ell}\}\trafficslack - 2(1 - \Pr\{\goodevent_{\ell}\})\right)\right].
\end{aligned} 
\]
By assumption, $\Pr\{\goodevent_{\ell}\} \geq 1 - \frac{1}{8}\trafficslack$. It holds
\[
\frac{1}{2}\Pr\{\goodevent_{\ell}\}\trafficslack - 2(1 - \Pr\{\goodevent_{\ell}\}) \geq \frac{1}{2}\trafficslack - \frac{1}{8}\trafficslack^2-\frac{1}{4}\trafficslack \geq \frac{1}{8}\trafficslack.
\]
Therefore, we have
\begin{align*}
&\mspace{32mu}\expect{V(\bold{Q}(t_0+\epochlength)) - V(\bold{Q}(t_0 + \convlength)) \mid \bold{Q}(t_0)} \\
&\leq (\epochlength - \convlength)\left(5K(\epochlength-1+\convlength)-\frac{1}{8}\trafficslack\sum_{i=1}^N \lambda_i Q_i(t_0)\right).
\end{align*}
\end{proof}

%% file: app_dam_fe.tex

\subsection{Unbiased estimation of service rates~(Lemma~\ref{lem:update-unbiased})}\label{app:dam_update}
\begin{proof}[Proof of Lemma~\ref{lem:update-unbiased}]
Fix $i,j,t_0$, and suppose $t_0 = \ell \epochlength + 1$. Note that the event that queue $i$ will explore server $j$ is independent from samples of previous epochs in Algorithm \ref{algo:DAM-online-estimation}. In addition, whether a sample at time $t \in [t_0, t_0 + \epochlength - 1]$ is collected only depends on whether there is a success in $[t_0,t - 1]$, and is independent from previous samples. Therefore, we know samples collected in epoch $\ell$ are independent from samples from previous epochs. 

Now it remains to show samples collected in epoch $\ell$ are almost surely independent Bernoulli random variables of mean $\mu_{i,j}$. First almost surely, there is no tie between bids of queues. Condition on the event $\set{E}_{\ell,i,j}$ that queue $i$ explores server $j$, and $I(j,t) = i, \forall t \in [t_0,t_0+\epochlength - 1]$. If the event $\set{E}_{\ell,i,j}$ does not hold, there is no collected sample as the bid of queue $i$ is higher than the bid of any non-exploring queue and the bids of all exploring queues are constant within the epoch thanks to our consistent tie-breaking. 
Given $\set{E}_{\ell,i,j}$, we have that
$Y_{i,j}(t) = S_{i,j}(t), \forall t \in [t_0,t_0 + \epochlength - 1].$ Finally, we note that success outcomes $S_{i,j}(t)$ are independent. For these $\epochlength$ independent Bernoulli random variables, if we only keep those after the first success, they are still independent with the same mean. Therefore, the samples Algorithm \ref{algo:DAM-online-estimation} collects after the first success are independent and Bernoulli distributed with mean $\mu_{i,j}$, which completes the proof.
\end{proof}

\subsection{Bounding the probability of exploration or incorrect estimation (Lemma~\ref{lem:estimate-good})}\label{app:damfe_convergence_estimates}
In this section we establish that when $t_0 \geq T_0$, we have high probability that all queues learn accurate service rates. In particular, fix the $\tau-$th epoch starting in time slot $t_0 = (\tau-1)\epochlength+1$. Recall that $\Delta_{i,j}(t_0) = \sqrt{\frac{3\ln t_0}{n_{i,j}(t_0)}}$ and define events
\begin{align}
\set{E}^1_\tau &= \left\{\exists_{i \in \mathcal{N}, j \in \mathcal{K}}, |\hat{\mu}_{i,j}(t_0) - \mu_{i,j}| > \Delta_{i,j}(t_0)\right\} \\ \set{E}_\tau^2 &= \left\{\exists_{i \in \mathcal{N}, j \in \mathcal{K} \colon \mu_{i,j} > 0}, \Delta_{i,j}(t_0) > \frac{1}{16}\trafficslack\mulower\right\}\label{eq:damfe-a2}.
\end{align}
Then the event that some service rate estimations are inaccurate is exactly $\set{E}_\tau^1 \cup \set{E}_\tau^2$. We bound the probability of the two events individually, and then use union bound to obtain the probability of accurate service rates. Specifically, we have the following two lemmas.
\begin{lemma}\label{lem:prob-a1}
We have $\Pr\{\set{E}_\tau^1\} \leq \frac{2NK}{t_0^4}$.
\end{lemma}
\begin{lemma}\label{lem:prob-a2}
When $t_0 \geq T_0$, we have $\Pr\{\set{E}_\tau^2\} \leq \frac{1}{2NKt_0^2}.$
\end{lemma}
\begin{proof}[Proof of \Cref{lem:estimate-good}]
Note that the event of inaccurate service rates is exactly $\set{E}_W = \set{E}_\tau^1\cup \set{E}_\tau^2$ for the $\tau-$th epoch. Using union bound over \Cref{lem:prob-a1} and \Cref{lem:prob-a2}, we have
\[
\Pr\{\set{E}_W\} \leq \frac{2NK}{t_0^4} + \frac{1}{2NKt_0^2} \leq \frac{1}{NKt_0^2},
\]
where the last inequality is because $t_0 \geq T_0 \geq \frac{1}{2}NK\epochlength^2 \geq 2NK$ by \eqref{eq:set-of-l0} and \eqref{eq:parameter-setting}, so~$t_0^2 \geq 4N^2K^2.$ 

Moreover, let $\ell = \frac{t_0 + \epochlength - 1}{\epochlength}.$ By union bound and the fact that each queue explores with probability at most $\frac{K}{\ell^{\expratio}}$ in \Cref{algoline:explore-rate} of Algorithm~\ref{algo:DAM-online-master}, we know $\Pr\{\set{E}_P\} \leq \frac{NK}{\ell^{\expratio}}.$ By assumption, $t_0 \geq T_0$, and thus $\ell \geq \ell_0$, which implies $\ell \geq \ell_0 \geq (NK\epochlength)^{1/\expratio} + 1$ by \eqref{eq:set-of-l0}. As a result, we have $\ell^{\expratio} \geq NK\epochlength$, and $\Pr\{\set{E}_P\} \leq \frac{1}{\epochlength}$.
\end{proof}
We next provide the proofs of Lemma~\ref{lem:prob-a1} and Lemma~\ref{lem:prob-a2}. Lemma~\ref{lem:prob-a1} follows from standard concentration arguments. For completeness, we provide its proof here.
\begin{proof}[Proof of \Cref{lem:prob-a1}]
Fix $i \in \mathcal{N}, j \in \mathcal{K}.$ If $n_{i,j} = 0,$ we have $\Delta_{i,j} = \infty$ and trivially $|\hat{\mu}_{i,j}(t_0)-\mu_{i,j}| \leq \Delta_{i,j}(t_0).$ Assume $n_{i,j} > 0$. Recall that $\set{Y}_{i,j}(t_0)$ is the set of samples collected by queue $i$ for server $j$ until time slot $t_0$. By Algorithm \ref{algo:DAM-online-estimation}, the estimation is given by the average of samples in $\set{Y}_{i,j}(t_0).$ By Lemma~\ref{lem:update-unbiased}, $\set{Y}_{i,j}(t_0) = \{X_1,\ldots,X_{n_{i,j}(t_0)}\}$ contains independent Bernoulli random variables  with mean $\mu_{i,j}$ $X_u, 1 \leq u \leq n_{i,j}(t_0)$. We then have
\[
\Pr\left\{|\hat{\mu}_{i,j}(t_0) - \mu_{i,j}| > \Delta_{i,j}(t_0)\right\} = \Pr\left\{\left|\sum_{u=1}^{n_{i,j}(t_0)}X_u - n_{i,j}(t_0)\mu_{i,j} \right| > n_{i,j}(t_0)\Delta_{i,j}(t_0)\right\}.
\]
Using Chernoff-Hoeffding's Inequality (provided in Fact~\ref{fact:chernoff}) and a union bound over $n_{i,j}(t_0) \leq t_0$ shows that 
\begin{align*}
\Pr\left\{\left|\sum_{u=1}^{n_{i,j}(t_0)}X_u - n_{i,j}(t_0)\mu_{i,j} \right| > n_{i,j}(t_0)\Delta_{i,j}(t_0)\right\} &\leq 2t_0\exp\left(-2(\Delta_{i,j}(t_0)n_{i,j}(t_0))^2 / n_{i,j}(t_0)\right) \\
&= 2t_0\exp(-6\ln(t_0)) = \frac{2}{t_0^4}.
\end{align*}
Using union bound over the $NK$ pairs of $i,j$ gives
the desired result.
\end{proof}
We now prove \Cref{lem:prob-a2}. The crux of establishing this lemma is to show that after $T_0$, each queue $i$ has sufficient samples for every server $j$ such that $\mu_{i,j} > 0.$ However, different from the classical multi-armed bandit setting, in Algorithm \ref{algo:DAM-online-estimation}, samples are collected in batches instead of in time slots. We call an epoch successful if there is at least one fulfilled request in $[t_0,t_0 + T_c - 1]$. In a successful epoch, queue $i$ collects at least $L - T_c$ samples. It then remains to show that the number of successful epoch is sufficiently large for $t_0 \geq T_0.$

Fix a pair of queue $i$ and server $j$, and a starting time slot of an epoch $t_0$ where $t_0 \geq T_0$. Let $\ell = \frac{t_0 + \epochlength - 1}{\epochlength},$ and define $Z_{i,j}(1),\ldots,Z_{i,j}(\ell - 1)$ such that $Z_{i,j}(r) = 1$ if queue $i$ successfully collects samples from server $j$ in the epoch $[(r - 1)\epochlength + 1, r\epochlength]$ for $1 \leq r \leq \ell - 1.$ We know there are exactly $\ell - 1$ epochs ahead of time slot $t_0$, and we have
\[
n_{i,j}(t_0) = (\epochlength - \checkperiod)\sum_{r=1}^{\ell - 1} Z_{i,j}(r).
\]
The next lemma shows that the number of successful epochs is large with high probability if $\mu_{i,j} > 0$.
\begin{lemma}\label{lem:bound-success-epoch}
Let $\ell = \frac{t_0 + \epochlength - 1}{\epochlength}$. If $t_0 \geq T_0$, and $\mu_{i,j} > 0$, then we have 
\begin{equation*}
\Pr\left\{\sum_{r=1}^{\ell-1}Z_{i,j}(r) \geq \frac{1}{12}\ell^{1-\expratio}\right\} \geq 1 - \exp\left(-\frac{1}{40}\ell^{1-\expratio}\right),
\end{equation*}
\end{lemma}
\begin{proof}
Fix $i,j$ such that $\mu_{i,j} > 0$. For our analysis, it is useful to consider the epoch $r_0 = \lceil (3N)^{1/\expratio}\rceil$; in any epoch after $r_0$, with high probability, at most one queue is exploring any particular server. Recall by  \eqref{eq:set-of-l0} the definition of $\ell_0$ and note that $\ell \geq \ell_0 \geq r_0 + 1 \geq 3$. We have $\sum_{r=1}^{\ell - 1} Z_{i,j}(r) \geq \sum_{r = r_0}^{\ell - 1} Z_{i,j}(r)$ since $Z_{i,j}(r) \geq 0$ for all $1 \leq r \leq \ell - 1.$ Next, we lower bound the probability that $Z_{i,j}(r) = 1$ for $r \geq r_0$. Note that when only queue $i$ explores server $j$, server $j$ must keep favoring queue $i$ in the whole epoch $[t_0,t_0 + \epochlength - 1].$ The probability that only queue $i$ explores server $j$ is at least $\frac{1}{\ell^{\expratio}}\left(1 - \frac{N}{\ell^{\expratio}}\right)$, where the first term is the probability that queue $i$ explores server $j$, and the second term is a lower bound on the probability that all other queues do not explore $j$. In addition, when $I(j,t) = i$, the probability that there is at least one successful request in $[t_0,t_0 + \checkperiod - 1]$ is at least $(1 - (1 - \mulower)^{\checkperiod})$. Following the same proof as in \Cref{lem:checkperiod-good}, we have $(1 - \mulower)^{\checkperiod} \leq \frac{\trafficslack}{16K\convlength} \leq \frac{1}{16K\convlength}.$ Therefore, for $r_0 \leq r \leq \ell - 1,$ 
\begin{equation}\label{eq:bound-zij}
\begin{aligned}
\Pr\{Z_{i,j}(r) = 1\} &\geq \frac{1}{r^{\expratio}}\left(1 - \frac{N}{r^{\expratio}}\right)(1 - (1 - \mulower)^{\checkperiod}) 
\geq \frac{1}{r^{\expratio}}\cdot\frac{2}{3}\left(1 - \frac{1}{16K\convlength}\right) 
\geq \frac{1}{2r^{\expratio}},
\end{aligned}
\end{equation}
where the second inequality is because $r^{\expratio} \geq r_0^{\expratio} \geq 3N,$ and the last inequality is because of the fact that $\frac{1}{16K\convlength} \leq \frac{1}{4}.$

Note that $X_{i,j}(r)$'s are independent, and take value $[0,1]$. Therefore, we can apply Bernstein's Inequality (Fact \ref{fact:bernstein}) to show $Z_s \coloneqq Z_{i,j}(r_0) + \cdots + Z_{i,j}(\ell - 1)$ is large with high probability. Since the function $f(x)=x^{-\expratio}$ is a decreasing function for $x > 0$, it holds for any $n \geq 2$,
\begin{equation}\label{eq:bound-sum-by-integral}
\int_2^{n+1} f(x)\mathrm{d}x \leq \sum_{r=2}^n f(r) \leq \int_1^n f(x)\mathrm{d}x.
\end{equation}

By \eqref{eq:bound-zij} and \eqref{eq:bound-sum-by-integral} and the fact that $r_0 \geq 3$, we have
\begin{equation}\label{eq:lower-bound-zs}
\begin{aligned}
\expect{Z_s} &\geq \sum_{r = r_0}^{\ell - 1} \frac{1}{2r^{\expratio}} \geq \frac{1}{2}\left(\sum_{r=2}^{\ell - 1} \frac{1}{r^{\expratio}} - \sum_{r = 2}^{r_0} \frac{1}{r^{\expratio}}\right) \\
&\geq \frac{1}{2(1-\expratio)}\left(\ell^{1-\expratio} - 2^{1-\expratio} - r_0^{1-\expratio}+1\right) \\
&\geq \frac{1}{2(1-\expratio)}\ell^{1-\expratio}\left(1 - \left(\frac{2r_0}{\ell}\right)^{1-\expratio}\right) \geq \frac{1}{4}\ell^{1-\expratio},
\end{aligned}
\end{equation}
where the last inequality is because $t_0 \geq T_0$, and by \eqref{eq:set-of-l0} it holds \[\ell \geq (N\epochlength)^{1/\expratio} \geq (\epochlength/6)^{1/\expratio}r_0 \geq 500^{1/\expratio}r_0,\] and thus $\frac{1}{2(1-\expratio)}\left(1-\left(\frac{2r_0}{\ell}\right)^{1-\expratio}\right) \geq \frac{1}{2(1-\expratio)}\left(1-\left(\frac{1}{250}\right)^{1-\expratio}\right) \geq \frac{1}{4}.$

Since $Z_{i,j}(r)$'s are independent, and $Z_{i,j}(r) \in \{0,1\}$, we also have
\begin{equation}\label{eq:var-less-mean}
\begin{aligned}
\var(Z_s) &= \sum_{r = r_0}^{\ell - 1} \var(Z_{i,j)(r)} = \sum_{r = r_0}^{\ell - 1} \Pr\{Z_{i,j}(r) = 1\}\Pr\{Z_{i,j}(r) = 0\} \\
&\leq \sum_{r = r_0}^{\ell - 1} \Pr\{Z_{i,j}(r) = 1\} = \expect{Z_s}.
\end{aligned}
\end{equation}
Applying Bernstein's Inequality (Fact \ref{fact:bernstein} in Appendix~\ref{app:standard_concentration}) gives
\begin{align*}
\Pr\left\{Z_s < \frac{1}{3}\expect{Z_s}\right\} \leq \exp\left(-\frac{\frac{4}{9}\expect{Z_s}^2}{2(\var(Z_s) + \frac{1}{9}\expect{Z_s})}\right) \overset{\eqref{eq:var-less-mean}}{\leq} \exp\left(\frac{-1}{10}\expect{Z_s}\right).
\end{align*}
Using \eqref{eq:lower-bound-zs}, we have
\[
\Pr\left\{\sum_{r = 1}^{\ell - 1}Z_{i,j}(r) \geq \frac{1}{12}\ell^{1-\expratio}\right\} \geq \Pr\{Z_s \geq \frac{1}{12}\ell^{1-\expratio}\} \geq 1 - \exp\left(-\frac{1}{40}\ell^{1-\expratio}\right).
\]
\end{proof}
Before proving \Cref{lem:prob-a2}, we provide the following lemma on properties of $\ell_0$.
\begin{lemma}
Every $\ell \geq \ell_0$ satisfies
\begin{align}
\ell^{1-\expratio} &\geq 40\left(2\ln(2NK\epochlength) + 2\ln \ell\right) \label{eq:t0-enough-probability}\\
\ell^{1-\expratio} &\geq \frac{\ln \ell + \ln(2\epochlength)}{K^2 \checkperiod \mulower^2} \label{eq:t0-enough-sample}.
\end{align}
\end{lemma}
\begin{proof}
Recall that $C_1 = \max\left(80, \frac{1}{K^2\checkperiod \mulower^2}\right), C_2 = \max\left(80\ln(2NK\epochlength), \frac{\ln(2\epochlength)}{K^2\checkperiod\mulower^2}\right)$ and by Eq.~\eqref{eq:set-of-l0},
\[
\ell_0 = \left\lceil \max\left((NKL)^{1/\expratio}+1, (2C_2)^{1/(1-\expratio)}, \left(4C_1/(1-\expratio)\right)^{2/(1-\expratio)}\right)\right\rceil.
\]
It is sufficient to prove that for all $\ell \geq \ell_0$, we have $\ell^{1-\expratio} \geq C_1 \ln \ell + C_2.$ It holds
\[
\frac{1}{2}\ell^{1-\expratio} \geq \frac{1}{2}\ell_0^{1-\expratio} \geq C_2
\]
by the definition of $\ell_0.$ It remains to show $\frac{1}{2}\ell^{1-\expratio} \geq C_1 \ln \ell$ for $\ell \geq \ell_0.$ Note that for every $x,y > 0$, $\ln x \leq \frac{1}{y}x^y$ since $y\ln x = \ln(x^y) \leq x^y.$ Take $y = \frac{1-\expratio}{2}$. It is sufficient to prove $\ell^{(1-\expratio)/2} \geq \left(\frac{4C_1}{1-\expratio}\right)^{2/(1-\expratio)}$, which holds vacuously by the definition of $\ell_0.$
\end{proof}

\begin{proof}[Proof of \Cref{lem:prob-a2}]
Recall that we want to bound $\Pr\{\set{E}_\tau^2\}$. Fix a pair of $i \in \set{N}, j \in \set{K}$ with $\mu_{i,j} > 0$. We next show that the probability that $\Delta_{i,j}(t_0) > \frac{1}{16}\trafficslack\mulower$ is less than $\frac{1}{2N^2K^2t_0^2}$. Using a union bound over at most $NK$ pairs of $i,j$ gives the desired result.

We know $\Delta_{i,j}(t_0) = \sqrt{\frac{3\ln t_0}{n_{i,j}(t_0)}}$, where $n_{i,j}(t_0) = (\epochlength-\checkperiod)\sum_{r = 1}^{\ell - 1}Z_{i,j}(r)$ with $\ell = \frac{t_0 + \epochlength - 1}{\epochlength}.$
To make $\Delta_{i,j}(t_0) \leq \frac{1}{16}\trafficslack\mulower$, it is sufficient to require that $n_{i,j}(t_0) \geq \frac{768\ln t_0}{(\trafficslack\mulower)^2}.$

Using \Cref{lem:bound-success-epoch} gives
\[
\Pr\left\{n_{i,j}(t_0) \geq \frac{\epochlength-\checkperiod}{12}\ell^{1-\expratio}\right\} \geq 1 - \exp\left(\frac{-1}{40}\ell^{1-\expratio}\right).
\]
Since $t_0 \geq T_0$, we have $\ell \geq \ell_0$ and thus $\ell^{1-\expratio} \geq \frac{\ln(\ell) + \ln(2\epochlength)}{K^2 \checkperiod \mulower^2}$ by \eqref{eq:t0-enough-sample}. It then implies that
\[
\frac{\epochlength-\checkperiod}{12}\ell^{1-\expratio} \geq \frac{32\cdot 98 K^2\checkperiod}{12\trafficslack^2}\ell^{1-\expratio} \geq \frac{768\ln(2\epochlength\ell)}{(\trafficslack\mulower)^2} \geq \frac{768\ln t_0}{(\trafficslack\mulower)^2},
\]
where the last inequality is because $t_0 \leq 2\ell \epochlength.$
Now, by \eqref{eq:t0-enough-probability}, we have $\frac{1}{40}\ell^{1-\expratio} \geq 2\ln(2NK\epochlength\ell_0)$,
and thus $\exp(-\frac{1}{40}\ell^{1-\expratio}) \leq \frac{1}{2N^2K^2t_0^2}.$ Therefore, we have $\Pr\{n_{i,j}(t_0) \geq \frac{768\ln t_0}{(\trafficslack\mulower)^2}\} \geq 1 - \frac{1}{2N^2K^2t_0^2},$ which completes the proof.
\end{proof} 

\subsection{Expected drift due to exploration or incorrect estimation (Lemma~\ref{lem:bound-drift-ep})}\label{app:exploration_incorrect_estimates}
\begin{proof}[Proof of Lemma \ref{lem:bound-drift-ep}]
Note that the event that at least one queue explores, $\set{E}_P$, is independent of arrivals in time interval $[t_0,t_0 + \epochlength - 1].$ Therefore, \Cref{lem:bound-general-drift} shows that 
\[
\expect{V(\bold{Q}(t_0 + \epochlength)) - V(\bold{Q}(t_0)) \mid \bold{Q}(t_0), \set{E}_P} \leq \epochlength\left(2\sum_{i=1}^N \lambda_i Q_i(t_0) + K\epochlength\right).
\]
Taking expectation on both sides with respect to $\bold{Q}(t_0)$. Since $\set{E}_P$ is independent of $\bold{Q}(t_0)$ by the exploration policy (line~\ref{algoline:explore-rate} in Algorithm \ref{algo:DAM-online-master}), we have
\[
\expect{V(\bold{Q}(t_0 + \epochlength)) - V(\bold{Q}(t_0)) \mid \set{E}_P} \leq \epochlength\left(2\sum_{i=1}^N \lambda_i \expect{Q_i(t_0)} + K\epochlength\right).
\]

Since, $t_0 \geq T_0$ (by the assumption of the lemma), Lemma~\ref{lem:estimate-good} implies $\Pr\{\set{E}_P\} \leq \frac{1}{\epochlength}$. Therefore,
\[
\expect{V(\bold{Q}(t_0 + \epochlength)) - V(\bold{Q}(t_0)) \mid \set{E}_P}\Pr\{\set{E}_P\} \leq 2\sum_{i=1}^N \lambda_i \expect{Q_i(t_0)} + K\epochlength.
\]

Next, we show $\expect{V(\bold{Q}(t_0 + \epochlength)) - V(\bold{Q}(t_0)) \mid \set{E}_P^c \cap \set{E}_W}\Pr\{\set{E}_P^c \cap \set{E}_W\}\leq 3$. Since arrivals in $[t_0,t_0 + \epochlength - 1]$ are independent of $\set{E}_P, \set{E}_W$, they are independence of $\set{E}_P^c \cap \set{E}_W.$ By Lemma \ref{lem:bound-general-drift},
\begin{align*}
\expect{V(\bold{Q}(t_0 + \epochlength)) - V(\bold{Q}(t_0)) \mid \set{E}_P^c \cap \set{E}_W} &= \expect{\expect{V(\bold{Q}(t_0 + \epochlength)) - V(\bold{Q}(t_0)) \mid \bold{Q}(t_0),\set{E}_P^c \cap \set{E}_W}} \\
&\leq \epochlength\left(2\sum_{i=1}^N \lambda_i \expect{Q_i(t_0) \mid \set{E}_P^c \cap \set{E}_W} + K\epochlength\right) \\
&\leq 2\epochlength Kt_0 + K\epochlength^2,
\end{align*}
where the last inequality is because $Q_i(t_0) \leq t_0$ and $\sum_{i=1}^N \lambda_i \leq K.$ Since $t_0 \geq T_0 \geq \frac{1}{2}NK\epochlength^2$, we then have
\begin{align*}
\expect{V(\bold{Q}(t_0 + \epochlength)) - V(\bold{Q}(t_0)) \mid \set{E}_P^c \cap \set{E}_W}\Pr\{\set{E}_P^c \cap \set{E}_W\} &\leq \frac{1}{NKt_0^2}\left(2\epochlength Kt_0 + K\epochlength^2\right) \\
&\leq \frac{2\epochlength}{Nt_0} + \frac{\epochlength^2}{Nt_0^2} \leq 1 + 2 = 3.
\end{align*}
\end{proof}

\subsection{Expected drift with exploitation and refined estimates (Lemma~\ref{lem:bound-drift-ei})}\label{app:exploitation_refined_estimates}
For ease of notation, let $\set{E}_I = \set{E}_P^c \cap \set{E}_W^c$. We first derive a bound on the drift conditioning on $\set{E}_I.$
\begin{lemma}\label{lem:bound-conddrift-ei}
We have
\begin{equation}
\expect{V(\bold{Q}(t_0 + \epochlength)) - V(\bold{Q}(t_0))) \mid \set{E}_I} \leq 5780 \frac{K}{\trafficslack^2}\convlength^2 - 2\convlength\sum_{i=1}^N \lambda_i \expect{Q_i(t_0) \mid \set{E}_I},
\end{equation}
\end{lemma}
We are ready to establish \Cref{lem:bound-drift-ei} by \Cref{lem:bound-conddrift-ei}.
\begin{proof}[Proof of \Cref{lem:bound-drift-ei}]
We first remove the conditioning on $\set{E}_I$ in the drift bound of \Cref{lem:bound-conddrift-ei}. Note that $\set{E}_P$ is independent of $\bold{Q}(t_0)$ and $\set{E}_W$. Therefore, for a fixed queue $i$,
\[
\expect{Q_i(t_0)} = \expect{Q_i(t_0) \mid \set{E}_P^c} = \expect{Q_i(t_0) \mid \set{E}_P^c, \set{E}_W^c}\Pr\{\set{E}_W^c\} + \expect{Q_i(t_0) \mid \set{E}_P^c, \set{E}_W}\Pr\{\set{E}_W\}.
\]
We know $\set{E}_P^c \cap \set{E}_W^c = \set{E}_I,$ and $Q_i(t_0) \leq t_0$.
Note that $\set{E}_P$ is independent of the $\sigma-$field generated by $\bold{Q}(t_0)$ and $\set{E}_W$. It implies $\indic{E_P^c}$ and $Q_i(t_0)\indic{E_W^c}$ are two independent random variables. As a result,
\begin{align*}
\expect{Q_i(t_0) \mid \set{E}_I} = \frac{\expect{Q_i(t_0) \indic{\set{E}_W^c}\indic{\set{E}_P^c}}}{\Pr\{\set{E}_W^c \cap \set{E}_P^c\}} = \frac{\expect{\indic{\set{E}_P^c}}\expect{Q_i(t_0)\indic{\set{E}_W^c}}}{\Pr\{\set{E}_P^c\}\Pr\{\set{E}_W^c\}} &= \frac{\expect{Q_i(t_0)\indic{\set{E}_W^c}}}{\Pr\{\set{E}_W^c\}} \\
&= \expect{Q_i(t_0) \mid \set{E}_W^c}.
\end{align*}
By \Cref{lem:estimate-good}, we have $\Pr\{\set{E}_W\} \leq \frac{1}{NKt_0^2}$, and thus
\begin{align*}
\expect{Q_i(t_0) \mid \set{E}_I} = \expect{Q_i(t_0) \mid \set{E}_W^c}\geq \expect{Q_i(t_0)} - \Pr\{\set{E}_W\}t_0 
\geq \expect{Q_i(t_0)} - \frac{t_0}{NKt_0^2} = \expect{Q_i(t_0)} - \frac{1}{NKt_0}.
\end{align*}
Combining with \Cref{lem:bound-conddrift-ei} gives
\begin{align*}
\expect{V(\bold{Q}(t_0 + \epochlength)) - V(\bold{Q}(t_0))) \mid \set{E}_I} &\leq 5780 \frac{K}{\trafficslack^2}\convlength^2 - 2\convlength\sum_{i=1}^N \lambda_i \expect{Q_i(t_0)} + 2\convlength\sum_{i=1}^N \frac{\lambda_i}{NKt_0} \\
&\leq 5781 \frac{K}{\trafficslack^2}\convlength^2 - 2\convlength\sum_{i=1}^N \lambda_i \expect{Q_i(t_0)}.
\end{align*}
where the last inequality holds as $2\convlength\sum_{i=1}^N\frac{\lambda_i}{NKt_0}\leq 2$. Finally, observe that
\[
\Pr\{\set{E}_I\} \geq 1 - \Pr\{\set{E}_P\} - \Pr\{\set{E}_W\} \geq 1 - \frac{NK}{\ell^{\expratio}} - \frac{1}{NKt_0^2} \geq 1 - \frac{2}{\epochlength}
\]
where $\ell = \frac{t_0 + \epochlength - 1}{\epochlength}$, and the last inequality is because $\ell \geq \ell_0 \geq (NK\epochlength)^{1/\expratio}$ by \eqref{eq:set-of-l0}.
Therefore,
\begin{align*}
&\expect{V(\bold{Q}(t_0 + \epochlength)) - V(\bold{Q}(t_0)) \mid \set{E}_P^c \cap \set{E}_W^c}\Pr\{\set{E}_P^c \cap \set{E}_W^c\} = \expect{V(\bold{Q}(t_0 + \epochlength)) - V(\bold{Q}(t_0))) \mid \set{E}_I} \Pr\{\set{E}_I\} \\
&\hspace{2em}\leq 5781 \frac{K}{\trafficslack^2}\convlength^2 - 2\convlength\left(1 - \frac{2}{\epochlength}\right)\sum_{i=1}^N \lambda_i \expect{Q_i(t_0)} \leq 5781 \frac{K}{\trafficslack^2}\convlength^2 - 1.5\convlength\sum_{i=1}^N \lambda_i \expect{Q_i(t_0)},
\end{align*}
where the last inequality is because $\epochlength \geq 32$ by \eqref{eq:parameter-setting}.
\end{proof}

What is left is to prove \Cref{lem:bound-conddrift-ei} which we do via
a similar argument as Lemma~\ref{lem:epoch-drift-bound}. Condition on $\set{E}_I$. The event $\set{E}_I$ is independent of arrivals and service events that happen in time interval $[t_0,t_0 + \epochlength - 1]$. In addition, we notice that results established in \Cref{sec:property} are valid for general setting of $\tilde{\mu}_{i,j}.$ Furthermore, one can verify that the proof of Lemma \ref{lem:epoch-drift-bound} remains valid conditioning on $\set{E}_I$ as long as we establish an analog of \Cref{lem:drift-matching-weight} when $\tilde{\mu}_{i,j}$ is close to $\mu_{i,j}$. We show that when service rates are accurate, this is indeed the case.
\begin{lemma}\label{lem:close-mu-close-weight}
Assume that events $\set{E}_W^c$ and $\goodevent_{\ell}$ hold. Then the converged matching $\sigma$ fulfills
\[
\sum_{i=1}^N \mu_{i,\sigma(i)}Q_i(t_0) \geq (1+\frac{1}{2}\trafficslack)\sum_{i=1}^N \lambda_i Q_i(t_0).
\]
\end{lemma}
\begin{proof}
Recall that $\bar{\mu}_{i,j} = \min(1,\hat{\mu}_{i,j}(t_0) + \Delta_{i,j}(t_0))$ (Line~\ref{algoline:set-ucb-rate} in Algorithm~\ref{algo:DAM-online-master}.) In addition, if $\mu_{i,j} = 0$, let $\bar{\mu}_{i,j} = 0$. Since service rates are accurate,  when $\mu_{i,j} > 0$, we have either $\bar{\mu}_{i,j}=1\geq \mu_{i,j}$, or\[
\bar{\mu}_{i,j} - \mu_{i,j} = \hat{\mu}_{i,j} - \mu_{i,j} + \Delta_{i,j}(t_0) \geq -\Delta_{i,j}(t_0) + \Delta_{i,j}(t_0) = 0.
\]
In addition,
\[
\bar{\mu}_{i,j} - \mu_{i,j} \leq \hat{\mu}_{i,j} - \mu_{i,j} + \Delta_{i,j}(t_0) \leq \Delta_{i,j}(t_0) + \Delta_{i,j}(t_0) \leq \frac{1}{8}\trafficslack\mulower.
\]
Therefore, for any $\phi \in \Phi$, 
\begin{equation}\label{eq:upper-matching-weight}
\sum_{i=1}^N \sum_{j=1}^K \phi_{i,j}w_{i,j} = \sum_{i=1}^N \sum_{j=1}^K \phi_{i,j}\bar{\mu}_{i,j}Q_i(t_0) \geq \sum_{i=1}^N \sum_{j=1}^K \phi_{i,j}\mu_{i,j}Q_i(t_0).
\end{equation}
Notice that when $\mu_{i,j} = 0$, we almost surely have $n_{i,j}(t_0) = 0$ and thus $w_{i,j} = \bar{\mu}_{i,j} = 0$ by Algorithm \ref{algo:DAM-online-master}. Therefore, if $\sigma(i) \neq \perp,$ we would have $w_{i,j} > 0$ by Algorithm \ref{algo:DAM-queue} and $\mu_{i,\sigma(i)} > 0$ by the previous argument. As a result,
\begin{align}
\sum_{i=1}^N w_{i,\sigma(i)} = \sum_{i=1}^N \bar{\mu}_{i,\sigma(i)}Q_i(t_0) \notag
&\leq \sum_{i \in \mathcal{N} \colon \sigma(i) \neq \perp} (\mu_{i,\sigma(i)}+\frac{1}{8}\trafficslack\mulower) Q_i(t_0) \notag\\
&\leq \sum_{i \in \mathcal{N} \colon \sigma(i) \neq \perp} (1 + \frac{1}{8}\trafficslack)\mu_{i,\sigma(i)}Q_i(t_0) \notag\\
&= (1 + \frac{1}{8}\trafficslack)\sum_{i = 1}^N \mu_{i,\sigma(i)}Q_i(t_0), \label{eq:lower-matching-weight}
\end{align}
where the last inequality is because $\mu_{i,\sigma(i)} > 0$ and by Assumption that $\mu_{i,\sigma(i)} \in \{0\} \cup [\mulower,1]$ we have $\mu_{i,\sigma(i)} \geq \delta.$
By \Cref{lem:algo-converge-main}, the converged matching $\sigma$ satisfies
\[
\sum_{i=1}^N w_{i,\sigma(i)} \geq (1 - \frac{1}{16}\trafficslack)\max_{\phi \in \Phi} \sum_{i=1}^N \sum_{j=1}^K \phi_{i,j}w_{i,j}.
\]
Therefore, using \Cref{lem:algo-converge-main}, \eqref{eq:upper-matching-weight}, \eqref{eq:lower-matching-weight}, we have
\begin{align}
(1+\frac{1}{8}\trafficslack)\sum_{i = 1}^N \mu_{i,\sigma(i)}Q_i(t_0) &\geq (1 - \frac{1}{16}\trafficslack)\max_{\phi \in \Phi} \sum_{i=1}^N \sum_{j=1}^K \phi_{i,j}w_{i,j} \notag\\
&\geq (1 - \frac{1}{16}\trafficslack)\max_{\phi \in \Phi} \sum_{i=1}^N \sum_{j=1}^K \phi_{i,j}\mu_{i,j}Q_i(t_0).
\end{align}
By \Cref{as:stability}, we further know that
\[
(1+\frac{1}{8}\trafficslack)\sum_{i = 1}^N \mu_{i,\sigma(i)}Q_i(t_0) \geq (1 - \frac{1}{16}\trafficslack)(1+\trafficslack)\sum_{i=1}^N \lambda_i Q_i(t_0).
\]
Note that $\left(1 + \frac{1}{8}\trafficslack\right)^{-1} \geq 1 - \frac{1}{8}\trafficslack$. Therefore,
\begin{align*}
\sum_{i = 1}^N \mu_{i,\sigma(i)}Q_i(t_0) &\geq (1-\frac{1}{8}\trafficslack)(1-\frac{1}{16}\trafficslack)(1+\trafficslack)\sum_{i=1}^N \lambda_i Q_i(t_0) \\
&\geq (1 + \frac{1}{2}\trafficslack)\sum_{i=1}^N \lambda_i Q_i(t_0),
\end{align*}
since $(1 - a)(1-b) \geq 1 - (a + b)$ for $a,b \geq 0,$ and 
$(1 - \frac{3}{16}\trafficslack)(1+\trafficslack) \geq 1 + \trafficslack - \frac{3}{8}\trafficslack \geq 1 + \frac{1}{2}\trafficslack$.
\end{proof}
\begin{proof}[Proof of \Cref{lem:bound-conddrift-ei}]
The only part of the proof of Lemma~\ref{lem:epoch-drift-bound} that does not directly transfer here is Lemma~\ref{lem:drift-matching-weight} which relies on $\sigma$ being an approximate max-weight matching. Lemma~\ref{lem:close-mu-close-weight} shows that indeed conditioning on $\set{E}_I$, $\sigma$ is an approximate maximum-weight matching and therefore can replace  Lemma~\ref{lem:drift-matching-weight} in the proof of \Cref{lem:epoch-drift-bound}. As a result, Lemma~\ref{lem:epoch-drift-bound} is still valid, which implies:
\[
\expect{V(\bold{Q}(t_0 + \epochlength)) - V(\bold{Q}(t_0))) \mid \bold{Q}(t_0),\set{E}_I} \leq 5780 \frac{K}{\trafficslack^2}\convlength^2 - 2\convlength\sum_{i=1}^N \lambda_i Q_i(t_0).
\]
Taking expectation on both side with respect to $\bold{Q}(t_0)$ gives 
\[
\expect{V(\bold{Q}(t_0 + \epochlength)) - V(\bold{Q}(t_0))) \mid \set{E}_I} \leq 5780 \frac{K}{\trafficslack^2}\convlength^2 - 2\convlength\sum_{i=1}^N \lambda_i \expect{Q_i(t_0) \mid \set{E}_I}.
\]
\end{proof}

\subsection{Final guarantee for \textsc{DAM.FE} (full proof of Theorem~\ref{thm:queue-learning})}\label{app:damfe_guarantee}
\begin{proof}[Proof of Theorem \ref{thm:queue-learning}]
Using \Cref{cor:bound-drift-ew}, \Cref{lem:bound-drift-ep} and \Cref{lem:bound-drift-ei}, we have for an epoch starting time slot $t_0 \geq T_0$,
\begin{equation} \label{eq:learning-drift-bound}
\begin{aligned}
\expect{V(\bold{Q}(t_0 + \epochlength)) - V(\bold{Q}(t_0))}&= \expect{V(\bold{Q}(t_0 + \epochlength)) - V(\bold{Q}(t_0)) \mid \set{E}_P}\Pr\{\set{E}_P\} \\
&\hspace{2em}+ \expect{V(\bold{Q}(t_0 + \epochlength)) - V(\bold{Q}(t_0)) \mid \set{E}_P^c \cap \set{E}_W}\Pr\{\set{E}_P^c \cap \set{E}_W\} \\
&\hspace{2em} + \expect{V(\bold{Q}(t_0 + \epochlength)) - V(\bold{Q}(t_0)) \mid \set{E}_P^c \cap \set{E}_W^c}\Pr\{\set{E}_P^c \cap \set{E}_W^c\}\\
&\hspace{-5em}\leq 2\sum_{i=1}^N \lambda_i \expect{Q_i(t_0)} + K\epochlength + 5781 \frac{K}{\trafficslack^2}\convlength^2 - 1.5\convlength\sum_{i=1}^N \lambda_i \expect{Q_i(t_0)} - 3 \\
&\hspace{-5em}\leq 5782\frac{K}{\trafficslack^2}\convlength^2 - \convlength\sum_{i=1}^N \lambda_i \expect{Q_i(t_0)},
\end{aligned}
\end{equation}
where the last inequality is because $\convlength \geq 4, K\epochlength + 3 \leq \frac{K\convlength^2}{\trafficslack^2}$ since $\convlength \geq 99, \epochlength + 3 \leq 35\convlength \leq \convlength^2.$

Following the same proof technique of \Cref{thm:queue-learning}, recall that $T_0 = \ell_0 \epochlength + 1$. Fix a time slot $T > 0$, and with abuse of notation, let $\ell = \lceil\frac{T - 1}{\epochlength}\rceil$, so $T$ is in the $\ell + 1-$th epoch. Note that if $T \leq T_0$, \Cref{thm:queue-learning} holds vacuously since \[
\expect{\frac{1}{T}\sum_{t=1}^T \sum_{i=1}^N \lambda_i Q_i(t)} \leq \frac{1}{T} \sum_{t=1}^T T\left(\sum_{i=1}^N \lambda_i\right) \leq KT.
\]
Now suppose $T \geq T_0$, and thus $\ell \geq \ell_0.$ We have
\begin{equation}
\label{eq:learning-sum-of-drift}
\begin{aligned}
\expect{V(T_0)} + \sum_{\tau = \ell_0}^{\ell} \expect{V(\bold{Q}((\tau + 1)\epochlength + 1)) - V(\bold{Q}(\tau \epochlength + 1))} &= \expect{V(\bold{Q}((\ell + 1)\epochlength + 1))}\\
&\geq 0.
\end{aligned}
\end{equation}
In addition, taking $t_1 = t_2 = 1, t_3 = T_0 - 1$ and let $\set{W}$ be the full sample space in \Cref{lem:bound-general-drift} and by the fact that $Q_i(1) = 0$ for $i \in \set{N}$, it holds $\expect{V(T_0)} \leq KT_0^2$. Combining \eqref{eq:learning-sum-of-drift} with \eqref{eq:learning-drift-bound}, we obtain
\[
(\ell - \ell_0 + 1)5782\frac{K}{\trafficslack^2}\convlength^2 - \convlength\sum_{\tau=\ell_0}^{\ell} \sum_{i=1}^N\lambda_i \expect{Q_i(\tau \epochlength + 1)} + KT_0^2 \geq 0,
\]
which gives
\[
\expect{\sum_{\tau = \ell_0}^\ell \sum_{i=1}^N \lambda_i Q_i(\tau \epochlength + 1)} \leq 5782\frac{(\ell - \ell_0 + 1)K}{\trafficslack^2}\convlength + \frac{KT_0^2}{\convlength},
\]
and
\begin{align*}
\expect{\sum_{\tau = 0}^\ell \sum_{i=1}^N \lambda_i Q_i(\tau \epochlength + 1)} &\leq 5782\frac{(\ell - \ell_0 + 1)K}{\trafficslack^2}\convlength + \frac{KT_0^2}{\convlength} + \ell_0 K T_0 \\
&\leq 5782\frac{(\ell + 1)K}{\trafficslack^2}\convlength + \frac{2KT_0^2}{\convlength}.
\end{align*}
since $Q_i(\tau \epochlength + 1) \leq T_0$ for $0 \leq \tau < \ell_0,$ and $\ell_0 T_0 \leq \frac{T_0^2}{\epochlength} \leq \frac{T_0^2}{\convlength}.$
Using \eqref{eq:total_queue_decomp} and the fact that~$(\ell + 1)\epochlength \leq 2T$, we obtain
\begin{align*}
\expect{\sum_{i=1}^N \lambda_i\sum_{t=1}^T Q_i(t)} &\leq \epochlength\frac{5782(\ell+1)K\convlength }{\trafficslack^2} + \frac{2\epochlength KT_0^2}{\convlength} +K\epochlength^2(\ell + 1) \\
&\overset{\epochlength \leq \frac{34 \convlength}{\trafficslack}}{\leq} 2T\left( \frac{5782K\convlength}{\trafficslack^2}+\frac{34 KT_0^2}{\trafficslack T}+K\epochlength\right).
\end{align*}
Finally, divide both sides by $T$, and by \eqref{eq:parameter-setting} $\convlength = O\left(\frac{K\checkperiod}{\trafficslack}(K+\log N)\right), \epochlength = O\left(\frac{\convlength}{\trafficslack}\right)$, we have
\begin{align*}
\expect{\frac{1}{T}\sum_{t=1}^T\sum_{i=1}^N \lambda_i Q_i(t)} &= O\left(\frac{K\convlength}{\trafficslack} + \frac{KT_0^2}{\trafficslack T} +  \frac{K^2\checkperiod}{\trafficslack^3}\left(\log N + K\right)\right) \\
&= O\left(\frac{K^2\checkperiod}{\trafficslack^3}\left(\log N + K\right) + \frac{KT_0^2}{\trafficslack T}\right).
\end{align*}
\end{proof}

\subsection{Standard concentration inequalities}
\label{app:standard_concentration}
We first state two useful facts that we will use later: the Chernoff-Hoeffding's inequality and the Bernstein's Inequality, which we adopt and simplify from \cite{boucheron2013concentration}.
\begin{fact}[Chernoff-Hoeffding's Inequality]\label{fact:chernoff}
Given $n$ independent random variables $X_u$ taking value in $[0,1]$ almost surely. Let $X = \sum_{u=1}^n X_u$. Then for every $a > 0$,
\begin{equation}\label{eq:chernoff-bound}
    \Pr\{|X-\expect{X}| > a\} \leq 2e^{-2a^2/n}.
\end{equation}
\end{fact}
\begin{fact}[Bernstein's Inequality]\label{fact:bernstein}
Given $n$ independent random variables $X_u$ taking value in $[0,1]$ almost surely. Let $X = \sum_{u=1}^n X_u$. Then for every $a > 0$,
\begin{equation}\label{eq:bernstein-bound}
    \Pr\{|X-\expect{X}| > a\} \leq \exp\left(-\frac{a^2}{2(\var(X) + \frac{1}{3}a)}\right).
\end{equation}
\end{fact}

%% file: app_dam_ucb.tex
\subsection{Bounding drift via  max-weight decomposition (Lemma~\ref{lem:ucb-drift-decompose})}\label{sec:ucb-decompose}
The proof follows a similar epoch-based drift analysis but we use a stronger good checking event $\set{G}_{\tau}$ instead than the previous $\set{E}_{\tau}$ (used in the proofs of Theorem~\ref{thm:queue-nolearning} and Theorem~\ref{thm:queue-learning}). The former 
requires successful requests every $\checkperiod$ time slots during $[t_0,t_0 + 2\convlength - 1],$ while the latter 
only requires this property over $[t_0,t_0 + \convlength - 1].$ We start by providing some analogous lemmas.
\begin{lemma}\label{lem:prob-ucb-good-event}
It holds that $\Pr\{\set{G}_\tau\} \geq 1 - \frac{1}{16}\trafficslack.$
\end{lemma}
\begin{proof}
Similar to the proof of Lemma~\ref{lem:checkperiod-good}:
$\Pr\{\set{G}_\tau\} \geq 1 - 2K\convlength(1-\mulower)^{\checkperiod} \geq 1 - \frac{1}{16}\trafficslack.$
\end{proof}

\begin{lemma}
Let $t_0$ be the start of an
epoch. We can bound the drift before convergence by
\[
\expect{V(\bold{Q}(t_0 + 2\convlength)) - V(\bold{Q}(t_0))} \leq 4\convlength\sum_{i=1}^N \lambda_i\expect{Q_i(t_0)}+4K\convlength^2 
\]
and we can bound the drift when the stronger good checking event does not hold by
\begin{align*}
&\expect{V(\bold{Q}(t_0 + \epochlength)) - V(\bold{Q}(t_0+2\convlength)) \mid \set{G}_\tau^c} \\&\mspace{32mu}\leq (\epochlength-2\convlength)\left(2\sum_{i=1}^N \lambda_i\expect{Q_i(t_0)}+K(\epochlength+2\convlength)\right).
\end{align*}
\end{lemma}
\begin{proof}
The lemma follows directly from Lemma~\ref{lem:bound-general-drift}.
\end{proof}

\begin{lemma}\label{lem:ucb-drift-good}
Let $t_0(\tau)$ be the start of the $\tau$-th
epoch. We can bound the drift after convergence by the drift under the max-weight matching $\sigma_\tau^\star$ and the weight gap between $\sigma_{\tau}$ and $\sigma_\tau^\star$:
\begin{align*}
&\expect{V(\bold{Q}(t_0 + \epochlength)) - V(\bold{Q}(t_0 + 2\convlength)) \mid \set{G}_\tau}\Pr\{\set{G}_\tau\}\\
&\mspace{32mu}\leq (\epochlength-2\convlength)\left(2\sum_{i=1}^N \lambda_i \expect{Q_i(t_0)}-2(1-\trafficslack/16)\sum_{i=1}^N \expect{\mu_{i,\sigma_\tau^\star(i)}Q_i(t_0)}\right. \\
&\mspace{48mu}+\left.2\expect{\sum_{i=1}^N (\mu_{i,\sigma_\tau^\star(i)} - \mu_{i,\sigma_\tau(i)})Q_i(t_0)\indic{\set{G}_\tau}}+3K(\epochlength+2\convlength)+6K\right).
\end{align*}
\end{lemma}
\begin{proof}[Proof sketch.]
The proof follows similar arguments with the one of Lemma~\ref{lem:epoch-drift-matching}, but has the additional component of matching weight difference. The proof is provided in Appendix~\ref{app:ucb-drift-good}.
\end{proof}

\begin{proof}[Proof of Lemma~\ref{lem:ucb-drift-decompose}]
Fix an epoch $\tau$ and let $t_0 = t_0(\tau)$ be the start of the $\tau$-th epoch. The key idea is to decompose the drift within one epoch $[t_0,t_0 + \epochlength - 1]$ into the sum of drifts in two intervals, $[t_0,t_0+2\convlength-1],[t_0+2\convlength,t_0 + \epochlength - 1].$ We bound the drift within epoch $\tau$ by
\begin{align*}
\expect{V(\bold{Q}(t_0 + \epochlength)) - V(\bold{Q}(t_0))}
&= \expect{V(\bold{Q}(t_0 + 2\convlength)) - V(\bold{Q}(t_0))} \\
&+ \expect{V(\bold{Q}(t_0 + \epochlength)) - V(\bold{Q}(t_0 + 2\convlength)) \mid \set{G}_{\tau}^c}\Pr\{\set{G}_{\tau}^c\}\\
&+ \expect{V(\bold{Q}(t_0 + \epochlength)) - V(\bold{Q}(t_0 + 2\convlength)) \mid \set{G}_\tau}\Pr\{\set{G}_\tau\}\\
&\mspace{-264mu}\leq 4\convlength\sum_{i=1}^N \lambda_i\expect{Q_i(t_0)}+4K\convlength^2 + \frac{1}{16}\trafficslack (\epochlength-2\convlength)\left(2\sum_{i=1}^N \lambda_i\expect{Q_i(t_0)}+K(\epochlength+2\convlength)\right) \\
&\mspace{-256mu}+(\epochlength-2\convlength)\left(2\sum_{i=1}^N \lambda_i \expect{Q_i(t_0)}-2(1-\trafficslack/16)\sum_{i=1}^N \expect{\mu_{i,\sigma_\tau^\star(i)}Q_i(t_0)}\right. \\
&\mspace{-256mu}+\left.2\expect{\sum_{i=1}^N (\mu_{i,\sigma_\tau^\star(i)} - \mu_{i,\sigma_\tau(i)})Q_i(t_0)\indic{\set{G}_\tau}}+3K(\epochlength+2\convlength)+6K\right) \\
\end{align*}
where the inequalities hold by Lemma~\ref{lem:prob-ucb-good-event}, Lemma~\ref{lem:checkperiod-good}, and Lemma~\ref{lem:ucb-drift-good}. As a result,
\begin{align*}
\expect{V(\bold{Q}(t_0 + \epochlength)) - V(\bold{Q}(t_0))}
&\leq \left(4\convlength+(2+\trafficslack/8)(\epochlength+2\convlength)\right)\sum_{i=1}^N \lambda_i \expect{Q_i(t_0)} \\
&-2(1-\trafficslack/16)(\epochlength-2\convlength)\sum_{i=1}^N \expect{\mu_{i,\sigma_\tau^\star(i)}Q_i(t_0)} \\
&+2(\epochlength-2\convlength)2\expect{\sum_{i=1}^N (\mu_{i,\sigma_\tau^\star(i)} - \mu_{i,\sigma_\tau(i)})Q_i(t_0)\indic{\set{G}_\tau}}\\
&+\underbrace{4K\convlength^2+K(\epochlength-2\convlength)\left((3+\trafficslack/16)(\epochlength+2\convlength)+6\right)}_{\leq 4K(\epochlength-2\convlength)(\epochlength+2\convlength)}.
\end{align*}
The lemma then follows by summing across all epochs.
\end{proof}

\subsection{Bounding drift after convergence
under good checking event (Lemma~\ref{lem:ucb-drift-good})}\label{app:ucb-drift-good}

\begin{proof}[Proof of Lemma~\ref{lem:ucb-drift-good}]
Condition on the event $\set{G}_\tau$. By Lemma~\ref{lem:algo-converge-main},
$\sigma_\tau$ is a matching and hence:
\begin{align*}
&V(\bold{Q}(t_0 + \epochlength)) - V(\bold{Q}(t_0 + 2\convlength)) = \sum_{t = t_0 + 2\convlength}^{t_0 + \epochlength - 1} \sum_{i=1}^N \left(\left(\left(Q_i(t)+A_i(t) - S_{i,\sigma_\tau(i)}(t)\right)^+\right)^2 - Q_i^2(t)\right) \\
&\mspace{32mu}= \sum_{t = t_0 + 2\convlength}^{t_0 + \epochlength - 1} \sum_{i=1}^N \left(\left(\left(Q_i(t)+A_i(t) - S_{i,\sigma^{\star}_\tau(i)}(t)\right)^+\right)^2 - Q_i^2(t)\right) \\
&\mspace{48mu} + \sum_{t = t_0 + 2\convlength}^{t_0 + \epochlength - 1} \sum_{i=1}^N \left(\left(\left(Q_i(t)+A_i(t) - S_{i,\sigma_\tau(i)}(t)\right)^+\right)^2 - \left(\left(Q_i(t)+A_i(t) - S_{i,\sigma^{\star}_\tau(i)}(t)\right)^+\right)^2\right).
\end{align*}
Note that $Q_i(t) + A_i(t) - S_{i,\sigma^*_\tau(i)}(t) \geq -S_{i,\sigma^*_\tau(i)}(t)$. As a result, for $t \in [t_0 + 2\convlength,t_0 + \epochlength - 1],$
\begin{align*}
&\sum_{i=1}^N \left(\left(\left(Q_i(t)+A_i(t) - S_{i,\sigma_\tau(i)}(t)\right)^+\right)^2 - \left(\left(Q_i(t)+A_i(t) - S_{i,\sigma^*_\tau(i)}(t)\right)^+\right)^2\right) \\
&\mspace{32mu}\leq \sum_{i=1}^N \left(\left(Q_i(t)+A_i(t) - S_{i,\sigma_\tau(i)}(t)\right)^2 - \left(Q_i(t)+A_i(t) - S_{i,\sigma^{\star}_\tau(i)}(t)\right)^2 + S^2_{i,\sigma^{\star}_\tau(i)}(t)\right) \\
&\mspace{32mu}\leq K + \sum_{i=1}^N (S_{i,\sigma_\tau^{\star}(i)}(t) - S_{i,\sigma_\tau(i)}(t))(2Q_i(t) + 2A_i(t) - S_{i,\sigma_\tau(i)}(t) - S_{i,\sigma^{\star}_\tau(i)}(t)) \\
&\mspace{32mu}\leq K + 2\sum_{i=1}^N A_i(t) + \sum_{i=1}^N S^2_{i,\sigma_\tau(i)}(t) + 2 \sum_{i=1}^N (S_{i,\sigma_\tau^{\star}(i)}(t) - S_{i,\sigma_\tau(i)}(t))Q_i(t).
\end{align*}
where the first inequality holds because for all real value $a$ it holds $a^2-(\min(a,0))^2\leq ((a)^+)^2 \leq a^2$ and in the last inequality we drop a $-S_{i,\sigma_\tau^\star(i)}(t)^2$ term.
In addition, again using $(a^+)^2 \leq a^2$:
\[\left(\left(Q_i(t)+A_i(t) - S_{i,\sigma^\star_\tau(i)}(t)\right)^+\right)^2 - Q_i^2(t) \leq A_i^2(t) + S_{i,\sigma^\star_\tau(i)}^2(t) + 2(A_i(t) - S_{i,\sigma^\star_\tau(i)}(t))Q_i(t).
\]
Since
$A_i^2(t) = A_i(t),S_{i,j}^2(t)=S_{i,j}(t),\sum_{i=1}^N S_{i,\sigma(i)}(t) \leq K$ for any matching $\sigma$:
\begin{align*}
&\mspace{32mu}V(\bold{Q}(t_0+\epochlength)) - V(\bold{Q}(t_0+2\convlength)) \\
&\leq \sum_{t=t_0+2\convlength}^{t_0+\epochlength-1}\sum_{i=1}^N 2(A_i(t) - S_{i,\sigma^\star_\tau(i)}(t))Q_i(t) + \sum_{t=t_0+2\convlength}^{t_0+\epochlength-1}\sum_{i=1}^N A_i(t) + (\epochlength-2\convlength)K \\
&\mspace{32mu}+2\sum_{t=t_0+2\convlength}^{t_0+\epochlength-1}\sum_{i=1}^N A_i(t)+2(\epochlength-2\convlength)K+2\sum_{t=t_0+2\convlength}^{t_0+\epochlength-1}\sum_{i=1}^N (S_{i,\sigma^\star_\tau(i)}(t) - S_{i,\sigma_\tau(i)}(t))Q_i(t) \\
&\leq \sum_{t=t_0+2\convlength}^{t_0+\epochlength-1}\sum_{i=1}^N 2(A_i(t) - S_{i,\sigma^{\star}_\tau(i)}(t))Q_i(t) +2\sum_{t=t_0+2\convlength}^{t_0+\epochlength-1}\sum_{i=1}^N (S_{i,\sigma^\star_\tau(i)}(t) - S_{i,\sigma_\tau(i)}(t))Q_i(t) \\
&\mspace{32mu}+ 3\sum_{t=t_0+2\convlength}^{t_0+\epochlength-1}\sum_{i=1}^N A_i(t) + 3(\epochlength-2\convlength)K.
\end{align*}
Since every time slot, there is at most one arriving job for each queue,
$|Q_i(t) - Q_i(t_0)| \leq |t - t_0|$ for any $t$ and hence
\begin{equation}\label{eq:ucb-decomp-preexpectation}
\begin{aligned}
&\mspace{32mu}V(\bold{Q}(t_0+\epochlength)) - V(\bold{Q}(t_0+2\convlength)) \\
&\leq \sum_{t=t_0+2\convlength}^{t_0+\epochlength-1}\sum_{i=1}^N 2(A_i(t) - S_{i,\sigma^{\star}_\tau(i)}(t))Q_i(t_0) +2\sum_{t=t_0+2\convlength}^{t_0+\epochlength-1}\sum_{i=1}^N (S_{i,\sigma^\star_\tau(i)}(t) - S_{i,\sigma_\tau(i)}(t))Q_i(t_0) \\
&\mspace{32mu}+2\sum_{t=t_0+2\convlength}^{t_0+\epochlength-1}\sum_{i=1}^N A_i(t)(t-t_0) + 6\sum_{t=t_0+2\convlength}^{t_0+\epochlength-1} K(t-t_0)+ 3\sum_{t=t_0+2\convlength}^{t_0+\epochlength-1}\sum_{i=1}^N A_i(t)\\ &\mspace{32mu}+3(\epochlength-2\convlength)K.
\end{aligned}
\end{equation}
We next consider $\expect{V(\bold{Q}(t_0+\epochlength)) - V(\bold{Q}(t_0+2\convlength)) \mid \set{G}_\tau}.$ For any time slot $t$, define the history filtration $\set{H}_t$ be the $\sigma$-field generated by $(\{A_i(t'),S_{i,j}(t'),Q_i(t')\} \colon i\in \set{N},j\in \set{K},t' \leq t).$ For $t \in [t_0 + 2\convlength, t_0 + \epochlength - 1],$ we have $A_i(t), S_{i,j}(t)$ independent of $\set{H}_{t_0 + 2\convlength - 1},$ for any fixed $i \in \set{N},j\in \set{K}.$ On the other hand, we have $\bold{Q}(t_0), \set{G}_\tau, \sigma_\tau,\sigma^\star_\tau \in \set{H}_{t_0 + 2\convlength - 1}.$ Therefore, it holds for $t \in [t_0 + 2\convlength,t_0 + \epochlength - 1], i \in \set{N},$
\[
\expect{A_i(t)Q_i(t_0) \mid \set{G}_\tau} = \expect{Q_i(t_0)\expect{A_i(t) \mid \bold{Q}(t_0),\set{G}_\tau} \mid \set{G}_\tau} = \expect{\lambda_i Q_i(t_0) \mid \set{G}_\tau} = \lambda_i\expect{Q_i(t_0)},
\]
where the last equation is because $Q_i(t_0)$ is independent of $\set{G}_\tau.$ Similarly,
$\expect{A_i(t) \mid \set{G}_\tau} = \lambda_i$, $\expect{S_{i,\sigma_\tau(i)}(t)Q_i(t_0) \mid \set{G}_\tau} = \expect{\mu_{i,\sigma_\tau(i)}(t)Q_i(t_0) \mid \set{G}_\tau}$, and $\expect{S_{i,\sigma_\tau^\star(i)}(t)Q_i(t_0) \mid \set{G}_\tau}= \expect{\mu_{i,\sigma_\tau^\star(i)}(t)Q_i(t_0)}$,
where the third equation is because $\sigma_\tau^\star$ is a function of $\bold{Q}(t_0)$ which is independent of $\set{G}_\tau.$
Putting the above back to Eq.~(\ref{eq:ucb-decomp-preexpectation}) and taking expectation we obtain
\begin{align*}
&\mspace{32mu}\expect{V(\bold{Q}(t_0+\epochlength)) - V(\bold{Q}(t_0+2\convlength)) \mid \set{G}_\tau} \\
&\leq (\epochlength - 2\convlength)\left(\sum_{i=1}^N 2\expect{(\lambda_i - \mu_{i,\sigma_\tau^\star(i)})Q_i(t_0)} +2\sum_{i=1}^N \expect{(S_{i,\sigma^\star_\tau(i)}(t) - S_{i,\sigma_\tau(i)}(t))Q_i(t_0) \mid \set{G}_\tau}\right) \\
&\mspace{32mu}+ (\epochlength-2\convlength)\left(3K(\epochlength + 2\convlength) + 6K\right).
\end{align*}
We then finish the proof by noting that Lemma~\ref{lem:prob-ucb-good-event} shows $\Pr\{\set{G}_\tau\} \geq 1 - \trafficslack / 16,$ and that
\[
\sum_{i=1}^N \expect{(S_{i,\sigma^\star_\tau(i)}(t) - S_{i,\sigma_\tau(i)}(t))Q_i(t_0) \mid \set{G}_\tau}\Pr\{\set{G}_\tau\} =\sum_{i=1}^N \expect{(S_{i,\sigma^\star_\tau(i)}(t) - S_{i,\sigma_\tau(i)}(t))Q_i(t_0)\indic{\set{G}_\tau}}.
\]
\end{proof}

\subsection{Final guarantee for \textsc{DAM.UCB} (full proof of Theorem~\ref{thm:queue-ucb})}\label{sec:ucb-full-proof}
\begin{proof}
Fix a time slot $T > 0$. Recall $\ell_T = \lceil \frac{T}{\epochlength} \rceil.$ We know by merging terms in Lemma~\ref{lem:ucb-drift-decompose} and Lemma~\ref{lem:ucb-match-difference} that
\begin{align*}
\expect{V(\bold{Q}(\ell_T \epochlength + 1)) - V(\bold{Q}(1))}
&\leq \sum_{\tau=1}^{\ell_T} (4\convlength + (2 + 3\trafficslack / 8)(\epochlength-2\convlength))\sum_{i=1}^N \lambda_i\expect{Q_i(t_0(\tau))} \\
&\mspace{32mu}-(\epochlength-2\convlength)(2-\trafficslack / 2)\sum_{\tau=1}^{\ell_T}\sum_{i=1}^N \expect{\mu_{i,\sigma_\tau^\star(i)}Q_i(t_0(\tau))} \\
&\mspace{32mu}+2(\epochlength-2\convlength)\frac{896K^2\lambda^\star \epochlength}{\trafficslack }\ln^2 (T+K+1) \\ &\mspace{32mu}+4K\ell_T(\epochlength-2\convlength)(\epochlength+2\convlength).
\end{align*}
Recall that by the stability assumption (Definition~\ref{as:stability}), we have for every $\tau$,
\[
\sum_{i=1}^N \expect{\mu_{i,\sigma_\tau^\star(i)}Q_i(t_0(\tau))} \geq (1+\trafficslack)\sum_{i=1}^N \lambda_iQ_i(t_0(\tau))
\]
since $\sigma_\tau^{\star}$ is the max-weight matching for weight $w_{i,j}=\mu_{i,j}Q_i(t_0(\tau)).$ In addition, we have $(2-\trafficslack / 2)(1+\trafficslack) \geq 2 + \trafficslack.$ Therefore,
\begin{equation}\label{eq:drift-bound-ucb}
\begin{aligned}\expect{V(\bold{Q}(\ell_T \epochlength + 1)) - V(\bold{Q}(1))}
&\leq \sum_{\tau=1}^{\ell_T} (4\convlength - 0.625\trafficslack(\epochlength-2\convlength))\sum_{i=1}^N \lambda_i\expect{Q_i(t_0(\tau))} \\
&\mspace{32mu}+2(\epochlength-2\convlength)\frac{896K^2\lambda^\star \epochlength}{\trafficslack }\ln^2 (T+K+1)\\ &\mspace{32mu}+4K\ell_T(\epochlength-2\convlength)(\epochlength+2\convlength).
\end{aligned}
\end{equation}
Recall (Eq.~\ref{eq:parameter-setting}) that $\epochlength \geq \frac{32}{\trafficslack}\convlength$, and thus $4\convlength - 0.625\trafficslack(\epochlength-2\convlength) \leq -14\convlength.$ Since \[\expect{V(\bold{Q}(\ell_T \epochlength + 1)) - V(\bold{Q}(1))} = \expect{V(\bold{Q}(\ell_T \epochlength + 1))} \geq 0,\]
we know the right hand side is positive. This implies that 
\begin{align*}
&\mspace{32mu}14\convlength\sum_{\tau=1}^{\ell_T} \sum_{i=1}^N \lambda_i \expect{Q_i(t_0(\tau))} \\
&\leq (\epochlength - 2\convlength)\left(\frac{896K^2\lambda^\star \epochlength}{\trafficslack}\ln^2 (\ell_T\epochlength+K+2) + 4K\ell_T(\epochlength+2\convlength)\right),
\end{align*}
and thus 
\begin{equation}\label{eq:ucb-epoch-queue}
\frac{1}{\ell_T}\sum_{\tau=1}^{\ell_T}\sum_{i=1}^N \lambda_i \expect{Q_i(t_0(\tau))} = O\left(\lambda^\star \frac{\epochlength^2 K^2}{\trafficslack^2}\frac{\ln^2(T+K+1)}{\ell_T \epochlength} +\frac{K\epochlength}{\trafficslack} \right).
\end{equation}
Notice that if $T < \epochlength$, then Theorem~\ref{thm:queue-ucb} holds vacuously since the time-averaged queue length does not exceed $\epochlength$. On the other hand, if $T \geq \epochlength$, we have $\frac{\ln^2(T+K+1)}{\ell_T \epochlength} = O\left(\frac{\ln^2 (T+K)}{T}\right).$ Finally, using Eq.~\eqref{eq:total_queue_decomp}, we obtain
\begin{align*}
\frac{1}{T}\sum_{t = 1}^T \sum_{i=1}^N \lambda_i \expect{Q_i(t)} &= \frac{1}{\ell_T}\sum_{\tau=1}^{\ell_T}\sum_{i=1}^N \lambda_i \expect{Q_i(t_0(\tau))} + O(K\epochlength)\\
&= O\left(\lambda^\star \frac{\epochlength^2 K^2}{\trafficslack^2}\frac{\ln^2(T+K)}{T} +\frac{KL}{\trafficslack} \right),
\end{align*}
which finishes the proof of Theorem~\ref{thm:queue-ucb} by using the fact that $\epochlength = O\left(\frac{K\checkperiod(\log N + K)}{\trafficslack^2}\right).$
\end{proof}

\subsection{Bounding weight difference by error impact (Lemma~\ref{lem:ucb-diff-conf})} \label{sec:ucb-diff-conf}
\begin{proof}[Proof of Lemma~\ref{lem:ucb-diff-conf}]
Fix an epoch $\tau$ and let $t_0 = t_0(\tau)$. Note that for every pair  $(i,j)\in\set{N}\times\set{K},$ if $\bar{\mu}_{i,j}(t_0) = 1$ then $\bar{\mu}_{i,j}(t_0) \geq \mu_{i,j} - \indic{\set{E}_{\tau,i}^1}$ as $\mu_{i,j} \leq 1$; otherwise,
it holds that $\bar{\mu}_{i,j}(t_0) = \hat{\mu}_{i,j}(t_0) + \Delta_{i,j}(t_0) \geq \indic{(\set{E}_{\tau,i}^1)^c}\left(\mu_{i,j}-\Delta_{i,j}(t_0) + \Delta_{i,j}(t_0)\right) \geq \mu_{i,j} - \indic{\set{E}_{\tau,i}^1}$.  In addition, condition on $\set{G}_\tau,$ we know by Lemma~\ref{lem:algo-converge-main} that $\sigma_\tau$ is a matching and 
\begin{align*}
\sum_{i=1}^N \bar{\mu}_{i,\sigma_\tau(i)}(t_0)Q_i(t_0) &\geq \left(1 - \frac{\trafficslack}{16}\right)\max_{\phi \in \Phi} \sum_{i=1}^N \sum_{j=1}^K \phi_{i,j}\bar{\mu}_{i,j}(t_0)Q_i(t_0) \\
&\geq \left(1 - \frac{\trafficslack}{16}\right)\sum_{i=1}^N \bar{\mu}_{i,\sigma_\tau^\star(i)}(t_0)Q_i(t_0) \\
&\geq \left(1 - \frac{\trafficslack}{16}\right)\sum_{i=1}^N \left(\mu_{i,\sigma_\tau^\star(i)}-\indic{\set{E}_{\tau,i}^1}\right)Q_i(t_0) \\
&\geq \left(1 - \frac{\trafficslack}{16}\right)\sum_{i=1}^N \mu_{i,\sigma_\tau^\star(i)}Q_i(t_0) -\sum_{i=1}^N\indic{\set{E}_{\tau,i}^1}Q_i(t_0).
\end{align*}
On the other hand, for each $i \in \set{N}$ with $\indic{\set{E}_{\tau,i}^2} = 0$, we have $\Delta_{i,\sigma_\tau(i)} \leq \frac{1}{16}\trafficslack\mulower$ by definition of $\set{E}_{\tau,i}^2$. Since $\bar{\mu}_{i,\sigma_\tau(i)} \leq 1$, this gives \[
\bar{\mu}_{i,\sigma_\tau(i)}(t_0)-\mu_{i,\sigma_\tau(i)} \leq \indic{(\set{E}_{\tau,i}^1)^c,(\set{E}_{\tau,i}^2)^c}\frac{1}{8}\trafficslack\mulower + \indic{\set{E}_{\tau,i}^1 \cup \set{E}_{\tau,i}^2} \leq \frac{1}{8}\trafficslack\mulower + \indic{\set{E}_{\tau,i}^1 \cup \set{E}_{\tau,i}^2}.
\]
The latter implies that
\begin{align*}
\sum_{i=1}^N \mu_{i,\sigma_\tau(i)}Q_i(t_0)
&\geq \sum_{i\in\set{N}\colon \sigma_\tau(i) \neq \perp} \left(\bar{\mu}_{i,\sigma_\tau(i)}-\frac{1}{8}\trafficslack\mulower-\indic{\set{E}_{\tau,i}^1 \cup \set{E}_{\tau,i}^2}\right)Q_i(t_0) \\
&\geq \left(1-\frac{1}{8}\trafficslack\right)\sum_{i\in\set{N}\colon \sigma_\tau(i) \neq \perp}\bar{\mu}_{i,\sigma_\tau(i)}Q_i(t_0) - \sum_{i=1}^N \indic{\set{E}_{\tau,i}^1\cup\set{E}_{\tau,i}^2}Q_i(t_0),
\end{align*}
where the second inequality is because $\bar{\mu}_{i,j} \geq \delta$. Combining the above, the first term is lower bounded by
\begin{align*}
\left(1-\frac{1}{8}\trafficslack\right)\sum_{i\in\set{N}\colon \sigma_\tau(i) \neq \perp}\bar{\mu}_{i,\sigma_\tau(i)}Q_i(t_0)&\geq\left(1-\frac{\trafficslack}{16}\right)\left(1-\frac{\trafficslack}{8}\right)\sum_{i=1}^N \mu_{i,\sigma_\tau^\star(i)}Q_i(t_0) - \sum_{i=1}^N \indic{\set{E}_{\tau,i}^1}Q_i(t_0) \\
&\geq \left(1-\frac{3\trafficslack}{16}\right)\sum_{i=1}^N \mu_{i,\sigma_\tau^\star(i)}Q_i(t_0) - \sum_{i=1}^N \indic{\set{E}_{\tau,i}^1}Q_i(t_0).
\end{align*}
As a result, taking expectations and conditioning on $\set{G}_\tau$, it holds that
\begin{align*}
\expect{\sum_{i=1}^N (\mu_{i,\sigma^{\star}_{\tau}(i)} - \mu_{i,\sigma_{\tau}(i)})Q_i(t_0(\tau))\mid \set{G}_{\tau}} 
&\leq \frac{3\trafficslack}{16}\expect{\sum_{i=1}^N (\mu_{i,\sigma^{\star}_{\tau}(i)}Q_i(t_0(\tau))\mid \set{G}_{\tau}}\\&+2\expect{\sum_{i=1}^N \indic{\set{E}_{\tau,i}^1 \cup \set{E}_{\tau,i}^2}Q_i(t_0) \mid \set{G}_{\tau}}.
\end{align*}
This then implies that
\begin{align*}
\expect{\sum_{i=1}^N (\mu_{i,\sigma^{\star}_{\tau}(i)} - \mu_{i,\sigma_{\tau}(i)})Q_i(t_0(\tau))\indic{\set{G}_\tau}}
&= \expect{\sum_{i=1}^N (\mu_{i,\sigma^{\star}_{\tau}(i)} - \mu_{i,\sigma_{\tau}(i)})Q_i(t_0(\tau))\mid \set{G}_{\tau}}\Pr\{\set{G}_\tau\} \\
& \hspace{-1in}\leq \frac{3\trafficslack}{16}\expect{\sum_{i=1}^N (\mu_{i,\sigma^{\star}_{\tau}(i)}Q_i(t_0(\tau))}+2\expect{\sum_{i=1}^N \indic{\set{E}_{\tau,i}^1 \cup \set{E}_{\tau,i}^2}\indic{\set{G}_\tau}Q_i(t_0)}.
\end{align*}
Summing over all epochs $\tau \leq \ell_T$ completes the proof.
\end{proof}

\subsection{Bounding the number of errors (Lemma~\ref{lem:ucb-bound-error})}
\label{sec:ucb-spread-cost}
\begin{lemma}\label{lem:ucb-bound-e1}
For a fixed queue $i\in\mathcal{N}$, let $\hat{e}_{i,1} = \sum_{\tau=1}^{\ell_T} \indic{\set{E}_{\tau,i}^1}$. It holds that $\expect{\hat{e}_{i,1}^2} \leq 5.$
\end{lemma}
\begin{proof}[Proof of Lemma~\ref{lem:ucb-bound-e1}]
Expanding the definition of $\hat{e}_{i,1}$, we obtain:
\begin{align*}
\expect{(\hat{e}_{i,1})^2} = \expect{\left(\sum_{\tau=1}^{\ell_T} \indic{\set{E}_{\tau,i}^1}\right)^2} &= \expect{\sum_{\tau=1}^{\ell_T} \indic{\set{E}_{\tau,i}^1}} + 2\expect{\sum_{\tau=1}^{\ell_T}\sum_{\tau'=\tau+1}^{\ell_T}\indic{\set{E}_{\tau,i}^1}\indic{\set{E}_{\tau',i}^1}} \\
&\leq \expect{\sum_{\tau=1}^{\ell_T} \indic{\set{E}_{\tau,i}^1}} + 2\expect{\sum_{\tau'=1}^{\ell_T}(\tau'-1)\indic{\set{E}_{\tau',i}^1}} \leq 2\sum_{\tau=1}^{\ell_T}\tau\Pr\{\set{E}_{\tau,i}^1\}.
\end{align*}
Similar to \textsc{DAM.FE}, \textsc{DAM.UCB} collects unbiased samples (Lemma~\ref{lem:update-unbiased} also holds for \textsc{DAM.UCB}). Then following the same proof of Lemma~\ref{lem:prob-a1}, we know for a fixed epoch $\tau$, the probability of $\set{E}_{\tau}^1$ can be upper bounded by $\Pr\{\set{E}_{\tau,i}^1\} \leq \frac{2K}{(K+t_0(\tau))^4}$; indeed, the probability is smaller since we use a larger confidence bound in DAM.UCB. As a result,
\[
\expect{(\hat{e}_{i,1})^2} \leq \sum_{\tau=1}^{\ell_T} \frac{4K\tau}{(K+(\tau-1)\epochlength+1)^4} \leq 4 +  \sum_{\tau=2}^{\ell_T}\frac{4\tau}{(\tau-1)^4\epochlength^4} \leq 5.
\]
The last inequality holds since, for $\tau \geq 2$, we have $\frac{4\tau}{(\tau-1)\epochlength^4} \leq \frac{8}{\epochlength^4} \leq 0.5$, and $\sum_{i=1}^{\infty} \frac{1}{i^3} \leq 2.$
\end{proof}

\begin{lemma}\label{lem:ucb-prop-tc}
It holds that $\mulower^2\checkperiod \geq 1.$
\end{lemma}
\begin{proof}
Recall that by the definition of $\checkperiod$ (Eq.~\ref{eq:parameter-setting}), we set $\checkperiod \geq 2$ and $\checkperiod \geq \left(\frac{2}{\ln(1-\mulower)}\right)^2.$ If $\mulower \geq \frac{1}{\sqrt{2}},$ the claim follows directly. Otherwise, we have $\mulower^2\checkperiod \geq 4\left(\frac{\mulower}{\ln(1-\mulower)}\right)^2.$ We claim that $\frac{\ln(1-\mulower)}{\mulower}$ is a decreasing function in $(0,1).$ Since it is also negative, we have
\[
\mulower^2\checkperiod \geq 4\left(\frac{\mulower}{\ln(1-\mulower)}\right)^2 \geq 4\left(\frac{1/\sqrt{2}}{\ln(1-1/\sqrt{2})}\right)^2 \geq 4 \cdot \frac{1}{2^2} = 1.
\]
To prove the claim, we define $f(x) = \frac{\ln(1-x)}{x}$. We have $f'(x)=\frac{1}{x^2}\left(\frac{-2x}{1-x}-\ln(1-x)\right).$ Let $g(x) = \frac{-2x}{1-x}-\ln(1-x).$ It holds $g(x) = 0,$ and $g'(x) = \frac{-2(1-x)+(-2x)}{(1-x)^2}+\frac{1}{1-x}=\frac{-(1+x)}{(1-x)^2} < 0.$ Hence, $f'(x) < 0$ for $x \in [0,1),$ which implies that $f(x)$ is a decreasing function and completes the proof.
\end{proof}

\begin{lemma}\label{lem:ucb-bound-e2}
For $i\in\mathcal{N}$, let $\hat{e}_{i,2} = \sum_{\tau=1}^{\ell_T} \indic{\set{G}_\tau,\set{E}_{\tau,i}^2}
$. Then, almost surely, $\hat{e}_{i,2} \leq K+\frac{\ln(T+K)}{2K}$.
\end{lemma}

\begin{proof}[Proof of Lemma~\ref{lem:ucb-bound-e2}]
By definition,
\begin{align*}
\hat{e}_{i,2} &= \sum_{\tau=1}^{\ell_T} \indic{\set{G}_\tau}\indic{\set{E}_{\tau,i}^2} \leq \sum_{\tau=1}^{\ell_T} \indic{\Delta_{i,\sigma_\tau(i)}(t_0(\tau)) > \frac{\trafficslack\mulower}{16}, \set{G}_\tau} \\
&= \sum_{\tau=1}^{\ell_T}\sum_{j=1}^K \indic{\Delta_{i,\sigma_\tau(i)}(t_0(\tau)) > \frac{\trafficslack\mulower}{16}, \set{G}_\tau,\sigma_\tau(i) = j} \\
&= \sum_{j=1}^K \sum_{\tau=1}^{\ell_T} \indic{\Delta_{i,\sigma_\tau(i)}(t_0(\tau)) > \frac{\trafficslack\mulower}{16}, \set{G}_\tau,\sigma_\tau(i) = j}.
\end{align*}
Fix $j \in \set{K}$. Define event $\set{B}_{\tau}^{i,j} = \{\Delta_{i,\sigma_\tau(i)}(t_0(\tau)) > \frac{\trafficslack\mulower}{16}\}\cap \set{G}_\tau \cap \{\sigma_\tau(i) = j\}$. We want to upper bound $\sum_{\tau=1}^{\ell_T} \indic{\set{B}^{i,j}_\tau}.$
Let $\tau'$ be the last epoch with $\indic{\set{B}^{i,j}_{\tau'}} = 1.$ We have $\Delta_{i,j}(t_0(\tau')) > \frac{\trafficslack\mulower}{16}$, which implies
\begin{equation}\label{eq:error2-sample-bound}
\frac{3\ln(T+K)}{n_{i,j}(t_0(\tau'))} > \left(\frac{\trafficslack\mulower}{16}\right)^2= 2^{-8}\trafficslack^2\mulower^2,
\end{equation}
and thus $n_{i,j}(t_0(\tau')) \leq \frac{768\ln(T+K)}{\trafficslack^2\mulower^2}.$ In addition, we know that by $\textsc{DAM.update}$ (Algorithm~\ref{algo:DAM-online-estimation}), if queue $i$ sees a successful request of server $\sigma_\tau(i)$ in $[t_0(\tau)+\convlength,t_0(\tau)+2\convlength-1]$, it collects all samples from $[t_0(\tau)+2\convlength,t_0(\tau)+\epochlength-1].$ Therefore, every $\tau$ for which $\set{B}^{i,j}_\tau$ happens ensure that queue $i$ collects at least $\epochlength - 2\convlength$ samples for queue $j$. As a result, we also have
\[
n_{i,j}(t_0(\tau')) \geq (\epochlength-2\convlength)\sum_{\tau=1}^{\tau'-1}\indic{\set{B}^{i,j}_\tau}.
\]
Therefore, it holds that 
\[
\sum_{\tau=1}^{\ell_T}\indic{\set{B}^{i,j}_\tau} \leq \sum_{\tau=1}^{\tau'-1}\indic{\set{B}^{i,j}_\tau} + 1 \leq \frac{n_{i,j}(t_0(\tau'))}{\epochlength-2\convlength} + 1 \leq \frac{768\ln(T+K)}{\trafficslack^2\mulower^2(\epochlength-2\convlength)}+1.
\]
Furthermore, by the setting of $\epochlength,\convlength$ in Eq.~\ref{eq:parameter-setting} and Lemma~\ref{lem:ucb-prop-tc}, we have \[\sum_{\tau=1}^{\ell_T}\indic{\set{B}^{i,j}_\tau} \leq 1+\frac{\ln(T+K)}{2K^2\mulower^2\checkperiod}\leq 1+\frac{\ln(T+K)}{2K^2}.\]
Therefore, $\hat{e}_{i,2} = \sum_{j=1}^K \sum_{\tau=1}^{\ell_T}\indic{\set{B}^{i,j}_\tau} \leq K + \frac{\ln(T+K)}{2K}.$
\end{proof}

\begin{proof}[Proof of Lemma~\ref{lem:ucb-bound-error}]
By Lemma~\ref{lem:ucb-bound-e2}, we have
$2\expect{(\hat{e}_{i,2})^2} \leq 4K^2\ln^2(T+K)$. 
Combined with Lemma~\ref{lem:ucb-bound-e1}:
\begin{align*}
\expect{(e_i)^2} \leq 2\expect{(\hat{e}_{i,1})^2} + 2\expect{(\hat{e}_{i,2})^2} \leq 10 + 4K^2\ln^2(T+K) \leq 14K^2\ln^2(T+K+1).
\end{align*}
\end{proof}

%% file: app_dam_dynamic.tex

In this section we provide the proof of Theorem~\ref{thm:dynamic-ucb}. Our proof relies on two new lemmas that reflect results proven for \textsc{DAM.converge} and \textsc{DAM.UCB} adapted to the system of dynamic queues. We discuss these two lemmas and present the proof of Theorem~\ref{thm:dynamic-ucb}. When results resemble previous ones, we highlight extra terms or key differences brought by the dynamic setting in \markcol{red}.

The proof of Theorem~\ref{thm:dynamic-ucb} first requires a new version of Lemma~\ref{lem:algo-converge-main} for \textsc{DAM.converge} when queues depart dynamically in an epoch. Specifically, fix an epoch $\ell$ and define $$\set{X}(\ell) = \set{I}(t_0(\ell)) \cap \set{I}(t_0(\ell+1))$$ as the set of queues that stay in the system for the whole $\ell$th epoch.
Let $\hat{\sigma}^\star_\ell$ be the maximum-weight matching between $\set{X}(\ell)$ and the set of servers with the optimistic weight $w_{i,j}=\bar{\mu}_{i,j}(t_0(\ell))Q_i(t_0(\ell))\in [0,Q_i(t_0(\ell))]$ for $i \in \set{I}(t_0(\ell)), j \in [K].$ Recall $\sigma_\ell$ is the output of \textsc{DAM.converge} in epoch $\ell$. If a queue $i$ leaves the system, we write $\sigma_\ell(i) = \perp.$ Then, similar to Lemma~\ref{lem:algo-converge-main} we obtain the following result.
\begin{lemma}\label{lem:dynamic-converge-main}
Fix an epoch $\ell$. Assume that all queues follow \textsc{DAM.converge} in time slots $\{t_0(\ell),\ldots,t_0(\ell)+\convlength - 1\}$. Under the good checking event $\set{E}_{\ell}$, it holds that $\sigma_\ell$ is a matching where for $i \neq i'$, either $\sigma_\ell(i) = \perp$ or $\sigma_\ell(i) \neq \sigma_\ell(i')$. Moreover, there is a set $\set{Q}(\ell) \subseteq \set{I}(t_0(\ell)) \setminus \set{X}(\ell)$ of size at most $K$, such that
\[
\sum_{i \in \set{X}(\ell)} w_{i,\sigma_\ell(i)} \geq (1-\frac{1}{16}\trafficslack) \sum_{i \in \set{X}(\ell)} w_{i,\hat{\sigma}_\ell^\star(i)} - \markcol{2\sum_{i \in \set{Q}(\ell)} Q_i(t_0(\ell))}.
\]
\end{lemma}
Comparing this result with Lemma~\ref{lem:algo-converge-main}, we find that even when queues leave within an epoch, following \textsc{DAM.converge} still leads to a matching  that is approximately max-weight up to an additional loss that relates only to the length of queues that depart during the epoch. The proof is similar to that of Lemma~\ref{lem:algo-converge-main}, and we highlight the changes in Appendix~\ref{app:dynamic-algo-converge}. 
Based on this lemma, we are able to conduct a drift analysis similar to the proof of Theorem~\ref{thm:queue-ucb} for \textsc{DAM.UCB}. In particular, define the set-based Lyapunov function $V_{\set{I}}(\bold{Q}) = \sum_{i \in \set{I}} Q_i^2.$ Note that if a  queue $i$ is not in the system, i.e., $i \not \in \set{I}(t)$, we set $Q_i(t) = 0.$ We note that $\bold{Q}$ may be an infinite dimensional vector in the dynamic case. However, in each time slot $t$, only queues in $\set{I}(t)$ have nonzero entries in $\bold{Q}(t)$. Accounting for this fact, although $V(\bold{Q}(t)) = \sum_{i} Q_i^2(t)$ is an infinite sum, only terms corresponding to indices in $\set{I}(t)$ are non-zero, and we have $V(\bold{Q}(t)) = V_{\set{I}(t)}(\bold{Q}(t))$. Recall that $\lambda_T^\star = \sum_{i \in \cup_{t \leq T} \set{I}(t)} \lambda_i^{-1}\ln^2(s_T(i)+K+1)$ and $s_T(i)$ is the survival time of queue $i$. We have the following drift bound similar to Eq. \eqref{eq:drift-bound-ucb} in the proof of Theorem~\ref{thm:queue-ucb}.
\begin{lemma}\label{lem:dynamic-drift-bound}
Let $\ell_T = \lceil \frac{T}{\epochlength}\rceil$. It holds that 
\begin{align*}
&\mspace{32mu}\expect{V(\bold{Q}(\ell_T \epochlength + 1)) - V(\bold{Q}(1))} \\
&\leq \sum_{\tau=1}^{\ell_T} (4\convlength - 0.625\trafficslack(\epochlength-2\convlength))\sum_{i \in \set{I}(t_0(\tau))} \lambda_i\expect{Q_i(t_0(\tau))} \\
&\mspace{32mu}+2(\epochlength-2\convlength)\frac{896K^2\markcol{\lambda^\star_{T}} \epochlength K^2}{\trafficslack}\\ &\mspace{32mu}+4K\ell_T(\epochlength-2\convlength)(\epochlength+2\convlength) +\markcol{19K\ell_T \epochlength^2}.
\end{align*}
\end{lemma}
The only additional term is the last term, $19K\ell_T \epochlength^2$, and it is of the same order as the second-to-last term. We establish Lemma~\ref{lem:dynamic-drift-bound} in Appendix~\ref{app:dynamic-drift-bound}. 
\subsection{Proof of Theorem~\ref{thm:dynamic-ucb}}\label{app:proof-dynamic}

\begin{proof}[Proof of Theorem~\ref{thm:dynamic-ucb}]
Replace \eqref{eq:drift-bound-ucb} in the proof of Theorem~\ref{thm:queue-ucb} by Lemma~\ref{lem:dynamic-drift-bound}. We follow the same argument after \eqref{eq:drift-bound-ucb} in the proof of Theorem~\ref{thm:queue-ucb} (Appendix~\ref{sec:ucb-full-proof}). Then, similar to \eqref{eq:ucb-epoch-queue}, it holds that 
\[
\frac{1}{\ell_T}\sum_{\tau=1}^{\ell_T}\sum_{i \in \set{I}(t_0(\tau))} \lambda_i \expect{Q_i(t_0(\tau))} = O\left(\frac{\epochlength^2 K^2}{\trafficslack^2}\frac{\lambda^\star_T}{\ell_T \epochlength} +\frac{K\epochlength}{\trafficslack} \right).
\]
Notice that for a fixed epoch $\tau$ and every $t \in \{t_0(\tau),\ldots,t_0(\tau+1)-1\}$, we have
\begin{align*}
\sum_{i \in \set{I}(t)} \lambda_i \expect{Q_i(t)} &\leq \sum_{i \in \set{I}(t_0(\tau))} \lambda_i (\expect{Q_i(t_0)} + \epochlength) + \sum_{i \in \set{I}(t) \cap \set{I}^c(t_0(\tau))} \lambda_i(t-t_0) \\
&\leq \sum_{i \in \set{I}(t_0(\tau))} \lambda_i \expect{Q_i(t_0)} + K\epochlength+\sum_{i \in \set{I}(t) \cap \set{I}^c(t_0(\tau))} \lambda_i \epochlength \\
&\leq \sum_{i \in \set{I}(t_0(\tau))} \lambda_i \expect{Q_i(t_0)} + 2K\epochlength.
\end{align*}
We can conclude that
\begin{align*}
\frac{1}{T}\sum_{t = 1}^T \sum_{i \in \set{I}(t)} \lambda_i \expect{Q_i(t)} &= \frac{1}{\ell_T}\sum_{\tau=1}^{\ell_T}\left(O(K\epochlength)+\sum_{i \in \set{I}(t_0(\tau))} \lambda_i \expect{Q_i(t_0(\tau))}\right)\\
&= O\left( \frac{\epochlength^2 K^2}{\trafficslack^2}\frac{\lambda^\star_T}{T} +\frac{K\epochlength}{\trafficslack} \right),
\end{align*}
which completes the proof by replacing $\epochlength$ in the non-vanishing term by Eq.~\ref{eq:parameter-setting}.
\end{proof}
\subsection{Proof of Corollary~\ref{cor:dynamic}}\label{app:proof-cor-dynamic}
\begin{proof}[Proof of Corollary~\ref{cor:dynamic}]
It suffices to upper bound $\lambda_T^\star.$ For ease of notation, let $\hat{\set{I}} = \cup_{t \leq T} \set{I}(t).$ Assume $|\hat{\set{I}}| \geq 1$; otherwise $\lambda_T^\star = 0$. Since $\lambda_i \geq \ubar{\lambda}$ for every queue $i$ by assumption, we have $\lambda_T^\star \leq \frac{1}{\ubar{\lambda}}\sum_{i \in \hat{\set{I}}} \ln^2(s_T(i)+K+1)$. Recall that for $i \in \hat{\set{I}}$, $s_T(i) = \min(T,\te(i))-\ts(i)+1$ is the number of time slots that queue $i$ stays in the system. Therefore, $s_T(i) = \sum_{t = 1}^T \indic{i \in \set{I}(t)}$ and 
\[
\sum_{i \in \hat{\set{I}}} s_T(i) = \sum_{i \in \hat{\set{I}}}\sum_{t=1}^T \indic{i \in \set{I}(t)} = \sum_{t = 1}^T \sum_{i \in \hat{\set{I}}} \indic{i \in \set{I}(t)} = \sum_{t=1}^T |\set{I}(t)| \leq NT,
\]
where the inequality is by assumption that $\set{I}(t) \leq N$ for every time slot $t$. Note that $\ln^2(x)$ is a concave function for $x > e$ because its second derivative $\frac{2(1-\ln(x))}{x^2}$ is strictly negative when $x > e$. We know $s_T(i)+K+1 \geq 3$ for all $i \in \hat{\set{I}}$. Then by Jensen's inequality, we have
\begin{equation}\label{eq:cor-jensen}
\frac{1}{|\hat{\set{I}}|}\sum_{i \in \hat{\set{I}}} \ln^2(s_T(i)+K+1) \leq \ln^2\left(\frac{1}{|\hat{\set{I}}|}\sum_{i\in \hat{\set{I}}}\left(s_T(i)+K+1\right)\right) \leq \ln^2\left(\frac{NT}{|\hat{\set{I}}|} + (K+1)\right),
\end{equation}
where the second inequality is because $\ln^2(x)$ is an increasing function for $x > 1.$ 

Let us now upper bound $|\hat{\set{I}}|\ln^2\left(\frac{NT}{|\hat{\set{I}}|} + (K+1)\right).$ Note that we must have $|\hat{\set{I}}| \in \{1,\ldots,NT\}$. If $|\hat{\set{I}}|=NT$, then $|\hat{\set{I}}|\ln^2\left(\frac{NT}{|\hat{\set{I}}|} + (K+1)\right) = NT\ln^2(K+1)$. Let us consider the case $|\hat{\set{I}}| \leq NT-1.$ We first have 
\[
\ln\left(\frac{NT}{|\hat{\set{I}}|} + (K+1)\right) \leq \ln\left(\frac{NT}{|\hat{\set{I}}|}\right) + \ln(K+1),
\]
because $\ln(a+b) \leq \ln(a)+\ln(b)$ for $a,b > 1.$ It implies 
\[
\ln^2\left(\frac{NT}{|\hat{\set{I}}|} + (K+1)\right) \leq \left(\ln\left(\frac{NT}{|\hat{\set{I}}|}\right) + \ln(K+1)\right)^2 \leq  2\ln^2\left(\frac{NT}{|\hat{\set{I}}|}\right)+2\ln^2(K+1),
\]
and thus $|\hat{\set{I}}|\ln^2\left(\frac{NT}{|\hat{\set{I}}|} + (K+1)\right) \leq 2|\hat{\set{I}}|\ln^2\left(\frac{NT}{|\hat{\set{I}}|}\right)+2NT\ln^2(K+1).$ We next upper bound the first term. For ease of notation, we replace $|\hat{\set{I}}|$ by $x$ and consider a function $f(x) = 2x\ln^2\left(\frac{NT}{x}\right)$ with $x$ in $[1,NT]$. We know $f(1) = 2\ln^2(NT), f(NT) = 2\ln^2(1) = 0.$ In addition, $f'(x) = 2\ln^2\left(\frac{NT}{x}\right)-4\ln\left(\frac{NT}{x}\right).$ The two zero points of $f'(x)$ are $x = e^{-2}NT$ and $x = NT.$ Therefore, $\max_{x \in [1,NT]}f(x) = \max(2\ln^2(NT),f(e^{-2}NT))=f(e^{-2}NT)=8NT.$ We then have \[|\hat{\set{I}}|\ln^2\left(\frac{NT}{|\hat{\set{I}}|} + (K+1)\right) \leq 8NT+2NT\ln^2(K+1),\]
which by \eqref{eq:cor-jensen} implies
\[
\lambda_T^\star \leq \frac{1}{\ubar{\lambda}}\sum_{i \in \hat{\set{I}}} \ln^2(s_T(i)+K+1) \leq \frac{1}{\ubar{\lambda}}|\hat{\set{I}}|\ln^2\left(\frac{NT}{|\hat{\set{I}}|} + (K+1)\right) \leq \frac{8NT+2NT\ln^2(K+1)}{\ubar{\lambda}}.
\]
Combining with Theorem~\ref{thm:dynamic-ucb}, we conclude that
\begin{align*}
\expect{\frac{1}{T}\sum_{t=1}^T\sum_{i\in \setQ(t)} \lambda_i Q_i(t)} &= O\left(\frac{K^2\checkperiod}{\trafficslack^3}(\log N + K) + \frac{(K\epochlength)^2}{\trafficslack^2}\frac{\lambda^{\star}_T}{T}\right) \\
&= O\left(\frac{K^2\checkperiod}{\trafficslack^3}(\log N + K) + \frac{N\ln^2(K+1)(K\epochlength)^2}{\trafficslack^2\ubar{\lambda}}\right).
\end{align*}
\end{proof}
\subsection{Convergence of \textsc{DAM.converge} with dynamic queues (Lemma~\ref{lem:dynamic-converge-main})}\label{app:dynamic-algo-converge}
To establish Lemma~\ref{lem:dynamic-converge-main} in epoch $\ell$, it suffices to
consider queues in $\set{I}(t_0(\ell))$ (other queues are either not present or not requesting service throughout the epoch).  Some of the queues in $\set{I}(t_0(\ell))$ may leave during $\{t_0(\ell),\ldots,t_0(\ell) + \convlength - 1\}$. For simplicity we  consider an auxiliary system in which only queues in $\set{I}(t_0(\ell))$ exist; these queues run \textsc{DAM.converge} from $t_0(\ell)$ unless they depart. If a queue $i$ leaves during $\{t_0(\ell),\ldots,t_0(\ell) + \convlength - 1\}$ in the original system, it also leaves in the same time slot in the auxiliary system. Otherwise, it stays in the auxiliary system. Moreover, we consider a fixed sample path for which $\set{E}_\ell$ holds for the original system. The auxiliary system has the same job arrivals and services as the original system. Establishing Lemma~\ref{lem:dynamic-converge-main} in the auxiliary system immediately implies the result for the original system. We thus focus on this auxiliary system for this section. We first establish that $\sigma_\ell$ is a matching.
\begin{lemma}
Fix a sample path for which $\set{E}_{\ell}$ holds. The output of \textsc{DAM.converge} in time slot $t_0(\ell)+\convlength-1$, $\sigma_{\ell}$, must be a matching.
\end{lemma}
\begin{proof}[Proof sketch.]
The proof is similar to that of Lemma~\ref{lem:algo-converge}, and we provide a proof sketch here. 
The main difference is that we need to change the definition of ``converge'': in the auxiliary system we say that the system converges at time $t$ if for every $i\in \set{I}(t_0(\ell))$ it is the case that either (i) $i$ departs in $\{t_0(\ell),\ldots,t\}$, or (ii) $i$ requests a server that no other queue requests and either $\tau_i(t) + \checkperiod - 1 > t_0(\ell)+\convlength - 1$ or $i$ obtains service from that server during $[t,\tau_i(t)+\checkperiod-1]$ (recall that $\tau_i(t)$ is the event log defined in \textsc{DAM.converge}); see initiations in Algorithm~\ref{algo:DAM-queue}). The same argument of Lemma~\ref{lem:def-converge} implies that a queue continues to request from the same server until $t_0(\ell)+\convlength - 1$ unless it leaves. However, such a departure does not lead to any two queues requesting from the same server, and thus the queues form a matching with the servers they request service from. 

We next show that the system converges by time slot $t_0(\ell)+\convlength - 1$ following the same argument as Lemma~\ref{lem:algo-converge}. Note that Lemma~\ref{lem:interval-update} is still valid, i.e.,  for a time slot $t$, if $|R_j(t)| > 1$ for a server $j$, then all but one queue in $R_j(t)$ either updates one of their prices at least once in $[t + 1,t+\checkperiod+1]$ or leaves the system in those periods. Similarly, if a queue $i$ is the sole queue requesting service of a server~$j$ and it does not obtain service until $\tau_i(t)+\checkperiod$, then it either updates its price or leaves the system. This also implies that, if the system has not converged in time slot $t$, at least one queue updates its price or leaves the system by time slot $t+\checkperiod+1$. We can then follow the argument in the proof of Lemma~\ref{lem:algo-converge}. The first step is to consider the time it takes for the number of active queues to reduce to $2K$. The second step is to use Lemma~\ref{lem:algo-conv-finite} to upper bound the remaining time for the system to converge. Combining both steps as in the proof of Lemma~\ref{lem:algo-converge} finishes the proof of the convergence result.
\end{proof}
It remains to establish that $\sigma_{\ell}$ is an approximate maximum weight matching for $\set{X}(\ell).$ 
To do so we again employ the dual program \eqref{eq:dual} but only account for the queues in $\set{X}(\ell)$ which will incur an additional loss from queues that depart in the $\ell$th epoch.
Let us define $\bar{p}_j(t)$ as  the maximum bid server $j$  receives from the start $t_0(\ell)$ to time slot $t$. Let $\eta_j(t)$ be the queue that offers this highest bid. We set $\bar{p}_j(t) = 0$ and $\eta_j(t) = \perp$ if no queue ever proposes a bid to server $j$. Note that $\eta_j(t)$ may already leave the system before time slot $t$. Nevertheless, the next lemma shows that if $\eta_j(t)$ is present at time slot $t$, then it must be requesting server $j$.
\begin{lemma}\label{lem:max-bid-request}
Condition on the good checking event $\set{E}_{\ell}$.
For a time slot $t$ in epoch $\ell$ and a server~$j$, if $\eta_j(t) \in \set{I}(t),$ then $\eta_j(t) \in R(j,t).$ Moreover, if $\eta_j(t) \neq \perp,$ then $\eta_j(t) \neq \eta_{j'}(t)$ for any $j' \neq j$. In other words, if $\eta_j(t)$ is present at time $t$ then it must be requesting service from $j$ and, regardless of whether $\eta_j(t)$ is present at $t$, no server other than $j$ can also have received its highest bid from queue $\eta_j(t)$.
\end{lemma}
\begin{proof}
Fix a time slot $t$ and a server $j$ with $\eta_j(t) \neq \perp$ and $\eta_j(t) \in \set{I}(t).$ We prove the result by contradiction. Suppose $\eta_j(t) \not \in R(j,t)$. Let $t_1 < t$ be the first time that $\eta_j(t)$ proposed the highest bid to server $j$. Let $t_2 > t_1$ be the time slot when $\eta_j(t)$ first requests another server; for $\eta_j(t) \not \in R(j,t)$ to hold we must have $t\geq t_2>t_1$. As $\eta_j(t)$ continues to bid at $j$ for at least $\checkperiod$ time slots after $t_1$ we must have $t_2 \geq t_1 + \checkperiod$. Moreover, for $\eta_j(t)$ to depart server $j$ in time slot $t_2$, $\eta_j(t)$ must  not receive service from $j$ in any of the time slots $[t_2 - \checkperiod+1,t_2]$. However, by $\set{E}_{\ell}$, this implies that $\eta_j(t)$ is not the queue with the highest bid for at least one time slot in $[t_2 - \checkperiod+1,t_2]$; this contradicts the assumption that $\eta_j(t)$ proposes the highest bid to server $j$ up to time slot $t$.  

To prove the second result, suppose there is $j' \neq j$ such that $\eta_{j'}(t) = \eta_j(t)$. Suppose $\eta_j(t)$ proposes the highest bid to $j'$ in time slot $t_3 \leq t$. Clearly, we cannot have $t_1 = t_3$ as a queue makes only one bid per time slot. In addition, without loss of generality we can relabel $j$ and $j'$ and thus may assume $t_3 > t_1$. Then, since $t_1 < t_3 \leq t,$ we have $\bar{p}_j(t_1) = \bar{p}_j(t_3) = \bar{p}_j(t)$ and thus $\eta_j(t_3) = \eta_j(t).$ However, since $\eta_j(t) \in \set{I}(t_3)$, i.e., queue $\eta_j(t)$ is present in time slot $t_3$, we must have $\eta_j(t) \in R(j,t_3)$ by the first part of the lemma. This contradicts that $\eta_j(t)$ requests service from $j'$ at time $t_3$ and thus gives a contradiction. 
 \end{proof}
 We can now finish the proof of Lemma~\ref{lem:dynamic-converge-main}.
 \begin{proof}[Proof of Lemma~\ref{lem:dynamic-converge-main}]
Let us define $\pi_i(t) = \max\left(0,\max_{j \in [K]} \left(w_{i,j} - p_{i,j}(t)\right)\right)$. To simplify notations, we write $t_0 = t_0(\ell)$. Consider the matching $\sigma_{\ell}$ obtained by \textsc{DAM.converge} in time slot $t_0(\ell)+\convlength - 1.$ Moreover, recall that $\hat{\sigma}^\star_{\ell}$ is the maximum weight matching between $\set{I}(t_0)$ and the set of servers with weight $w_{i,j}$. We next pick suitable dual variables to bound the difference between $\sum_{i \in \set{X}(\ell)} w_{i,\sigma_{\ell}(i)}$ and $\sum_{i \in \set{X}(\ell)} w_{i,\hat{\sigma}^\star_{\ell}(i)}.$

With abuse of notations, let $\bar{p}_j = \bar{p}_j(t_0+\convlength - 1)$, which is the highest bid server $j$ received during the run of \textsc{DAM.converge}. For $i \in \set{X}(\ell)$, let us define $\hat{\pi}_i = \max\left(0, \max_{j \in [K]} (w_{i,j} - \bar{p}_j)\right)$. Consider the dual program of maximum weight matching between $\set{X}(\ell)$ and the set of servers; this reflects \eqref{eq:dual} except for that rather than indexing over $i\in[N]$) we index over $i\in\set{X}(\ell)$. The definition of $\hat{\boldsymbol{\pi}}$ immediately implies that $\hat{\boldsymbol{\pi}},\bar{\boldsymbol{p}}$ are feasible solutions for this dual. Following the proof of Lemma~\ref{lem:lowslack-goodweight}, define $f_{i,j} = 1$ if $\sigma_{\ell}(i) = j$ and zero otherwise. Let 
\begin{align*}
u_{i,j} = (\hat{\pi}_i + \bar{p}_j - w_{i,j})f_{i,j}, \qquad
v_i = (1 - \sum_{j=1}^K f_{i,j})\hat{\pi}_i, \qquad and \quad
v'_j = \left(1 - \sum_{i \in \set{X}(\ell)} f_{i,j}\right)\bar{p}_j.
\end{align*}
As in the proof of Lemma \ref{lem:lowslack-goodweight}, weak duality implies that
\[
\sum_{i \in \set{X}(\ell)} w_{i,\hat{\sigma}^\star_\ell(i)} - \sum_{i \in \set{X}(\ell)} w_{i,\sigma_\ell(i)} \leq \sum_{i\in \set{X}(\ell),j\in[K]} u_{i,j}+\sum_{i \in \set{X}(\ell)} v_i + \sum_{j \in [K]} v'_j.
\]
We next bound the three terms on the right hand side. First, for the second term, the same argument as in Lemma~\ref{lem:unmatched-value} implies that an unmatched queue must have $\hat{\pi}_i = 0.$ Therefore, the second term is zero. To bound the third term, let $t' = t_0(\ell) + \convlength - 1$ and note that if $\bar{p}_j > 0$ but server $j$ is not matched with a queue in $\set{X}(\ell)$, then by Lemma~\ref{lem:max-bid-request} it must be the case that $\eta_j(t') \in \set{I}(t_0)\setminus \set{X}(\ell)$. For ease of notations, we write $\sigma^{-1}_{\ell}(j) = \perp$ for an unmatched server, and $\sigma^{-1}_{\ell}(j) = i$ if $\sigma_{\ell}(i) = j$. Therefore, 
\[
\sum_{j \in [K]} v'_j \leq \sum_{j\colon \sigma^{-1}(j)=\perp} \bar{p}_j = \sum_{j\colon \sigma^{-1}(j)=\perp} \bar{p}_j\indic{\eta_j(t') \in \set{I}(t_0) \setminus \set{X}(\ell)}.
\]

For the first term, it is equal to $\sum_{i \in \set{X}(\ell),\sigma_\ell(i) \neq \perp} \left(\hat{\pi}_i + \bar{p}_{\sigma_\ell(i)} - w_{i,\sigma_\ell(i)}\right)$. Note that  $\hat{\pi}_i \leq \pi_i(t')$ because $p_{i,j}(t') \leq \bar{p}_j$ for every server $j$. Therefore, 
\begin{align*}
\sum_{i \in \set{X}(\ell),\sigma_\ell(i) \neq \perp} \left(\hat{\pi}_i + \bar{p}_{\sigma_\ell(i)} - w_{i,\sigma_\ell(i)}\right) &\leq \sum_{i \in \set{X}(\ell),\sigma_\ell(i) \neq \perp} \left(\pi_i(t') + \bar{p}_{\sigma_\ell(i)} - w_{i,\sigma_\ell(i)}\right) \\
&\leq \sum_{i \in \set{X}(\ell),\sigma_\ell(i) \neq \perp} \frac{1}{16}\trafficslack w_{i,\sigma_\ell(i)} + \sum_{i \in \set{X}(\ell),\sigma_\ell(i) \neq \perp} \left(\bar{p}_{\sigma_\ell(i)}-p_{i,\sigma_\ell(i)}(t')\right),
\end{align*}
where the second inequality uses the fact that $\pi_i(t') \leq w_{i,J(i,t')}+\frac{1}{16}\trafficslack w_{i,J(i,t')}-p_{i,J(i,t')}(t')$, which holds by the as same argument as in Lemma~\ref{lem:feasible-slack}. Recall that by Lemma~\ref{lem:max-bid-request}, for a server $j$, if $\eta_j(t') \in \set{I}(t'),$ we must have $\eta_j(t') \in R(j,t')$ and thus $\sigma_\ell(\eta_j(t')) = j.$ Therefore, we have 
\[
\sum_{i \in \set{X}(\ell),\sigma_\ell(i) \neq \perp} \left(\bar{p}_{\sigma_\ell(i)}-p_{i,\sigma_\ell(i)}(t')\right) \leq \sum_{j \in [K]\colon \sigma^{-1}_\ell(j) \neq \perp} \bar{p}_j\indic{\eta_j(t') \not \in \set{I}(t')}.
\]
Moreover, we know
\[
\sum_{i \in \set{X}(\ell),\sigma_\ell(i) \neq \perp} \frac{1}{16}\trafficslack w_{i,\sigma_\ell(i)} \leq \frac{\trafficslack}{16}\sum_{i \in \set{X}(\ell)} w_{i,\hat{\sigma}^\star_\ell(i)}
\]
because $\hat{\sigma}^\star_\ell$ is the maximum weight matching between $\set{X}(\ell)$ and the set of servers. Therefore, we have
\begin{align*}
\sum_{i\in \set{X}(\ell),j\in[K]} u_{i,j} &\leq \frac{\trafficslack}{16}\sum_{i \in \set{X}(\ell)} w_{i,\hat{\sigma}^\star_\ell(i)} + \sum_{j \in [K]\colon \sigma^{-1}_\ell(j) \neq \perp} \bar{p}_j\indic{\eta_j(t') \not \in \set{I}(t')} \\
&\leq \frac{\trafficslack}{16}\sum_{i \in \set{X}(\ell)} w_{i,\hat{\sigma}^\star_\ell(i)} + \sum_{j \in [K]\colon \sigma^{-1}_\ell(j) \neq \perp} \bar{p}_j\indic{\eta_j(t') \in \set{I}(t_0) \setminus \set{X}(\ell)},
\end{align*}
where the last inequality holds because $\eta_j(t') \not \in \set{I}(t')$ implies $\eta_j(t') \in \set{I}(t_0) \setminus \set{X}(\ell)$. Therefore, we have
\begin{align*}
&\mspace{32mu}\sum_{i \in \set{X}(\ell)} w_{i,\hat{\sigma}^\star_\ell(i)} - \sum_{i \in \set{X}(\ell)} w_{i,\sigma_\ell(i)} \\
&\leq \sum_{i\in \set{X}(\ell),j\in[K]} u_{i,j}+\sum_{i \in \set{X}(\ell)} v_i + \sum_{j \in [K]} v'_j \\
&\leq \frac{\trafficslack}{16}\sum_{i \in \set{X}(\ell)} w_{i,\hat{\sigma}^\star_\ell(i)} + \sum_{j \in [K]\colon \sigma^{-1}_\ell(j) \neq \perp} \bar{p}_j\indic{\eta_j(t') \in \set{I}(t_0) \setminus \set{X}(\ell)} \\
&\mspace{32mu}+ \sum_{j\colon \sigma^{-1}(j)=\perp} \bar{p}_j\indic{\eta_j(t') \in \set{I}(t_0) \setminus \set{X}(\ell)} \\
&\leq \frac{\trafficslack}{16}\sum_{i \in \set{X}(\ell)} w_{i,\hat{\sigma}^\star_\ell(i)}+ \sum_{j \in [K]} \bar{p}_j \indic{\eta_j(t') \in \set{I}(t_0) \setminus \set{X}(\ell)}
\end{align*}
Lemma~\ref{lem:max-bid-request} shows that $\eta_j(t') \neq \eta_{j'}(t')$ for $j \neq j'$. In addition, the weight and the bid of a queue $i$ is at most $(1+\nicefrac{\trafficslack}{16})Q_i(t_0) \leq 2Q_i(t_0)$. Therefore, 
\begin{align*}
&\mspace{32mu}\sum_{i \in \set{X}(\ell)} w_{i,\hat{\sigma}^\star_\ell(i)} - \sum_{i \in \set{X}(\ell)} w_{i,\sigma_\ell(i)} \\
&\leq \frac{\trafficslack}{16}\sum_{i \in \set{X}(\ell)} w_{i,\hat{\sigma}^\star_\ell(i)} + 2\sum_{i \in \set{I}(t_0) \setminus \set{X}(\ell)} Q_i(t_0) \indic{\exists j \in [K], \eta_j(t') = i}.
\end{align*}
Let $\set{Q}$ be the set of queues that are present at $t_0$, leave during the epoch, and propose the highest bid to one of the servers during the epoch, i.e., $\{i\in \set{I}(t_0) \setminus \set{X}(\ell)\colon \exists j \in [K], \eta_j(t') = i\}$. Then we have $\set{Q} \subseteq \set{I}(t_0) \setminus \set{X}(\ell)$, and $|\set{Q}| \leq K$. Moreover, 
\[
\sum_{i \in \set{X}(\ell)} w_{i,\hat{\sigma}^\star_\ell(i)} - \sum_{i \in \set{X}(\ell)} w_{i,\sigma_\ell(i)} \leq \frac{\trafficslack}{16}\sum_{i \in \set{X}(\ell)} w_{i,\hat{\sigma}^\star_\ell(i)} + 2\sum_{i \in \set{Q}} Q_i(t_0),
\]
which completes the proof.
\end{proof}

\subsection{Drift analysis of Dynamic DAM.UCB (Lemma~\ref{lem:dynamic-drift-bound})}\label{app:dynamic-drift-bound}
Recall that $V(\bold{Q}(t)) = \sum_{i} Q_i^2(t) = V_{\set{I}(t)}(\bold{Q}(t))$ and $V_{\set{I}}(\bold{Q}) = \sum_{i \in \set{I}} Q_i^2.$ Let us define $\sigma_\tau^\star$ by the maximum-weight matching between $\set{I}(t_0(\tau))$ and the set of servers with weight $\mu_{i,j}Q_i(t_0(\tau))$. Note that it is different from the matching $\hat{\sigma}_\tau^\star$ that we defined above which is based on the weight $w_{i,j} = \bar{\mu}_{i,j}(t_0(\tau))Q_i(t_0(\tau))$. For an epoch $\tau$, we define the event $\set{G}(\tau)$ by the same event in the analysis of \textsc{DAM.UCB} such that under this event, for all time slots $t \in \{t_0(\tau) + \checkperiod - 1,\ldots,t_0(\tau) + 2\convlength - 1\}$ and server $j \in \set{K}$, there exists $t' \in [t - \checkperiod + 1,t]$ with $S_{I(j,t'),j}(t') = 1.$ To establish Lemma~\ref{lem:dynamic-drift-bound}, we first need a similar drift decomposition as Lemma~\ref{lem:ucb-drift-decompose}, and we prove this lemma in Appendix~\ref{app:dynamic-drift-decompose}
\begin{lemma}\label{lem:dynamic-drift-decompose}
It holds that
\begin{align*}
&\mspace{32mu}\expect{V(\bold{Q}(\ell_T \epochlength + 1)) - V(\bold{Q}(1))} \\
&\leq  \left(4\convlength + (2 + \trafficslack / 8)(\epochlength - 2\convlength)\right)\sum_{\tau=1}^{\ell_T}\sum_{i \in \set{I}(t_0(\tau))} \lambda_i \expect{Q_i(t_0(\tau))} \\
&\mspace{32mu}-2(1-\trafficslack/16)(\epochlength - 2\convlength)\sum_{\tau=1}^{\ell_T}\sum_{i \in \set{I}(t_0(\tau))} \expect{\mu_{i,\sigma^{\star}_{\tau}(i)}Q_i(t_0(\tau))} \\
&\mspace{32mu}+4K\ell_T(\epochlength-2\convlength)(\epochlength+2\convlength)+\markcol{\ell_T K\epochlength^2} \\
&\mspace{32mu}+2(\epochlength-2\convlength)\expect{\sum_{\tau=1}^{\ell_T} \sum_{i \in \set{X}(\tau)} (\mu_{i,\sigma^{\star}_{\tau}(i)} - \mu_{i,\sigma_{\tau}(i)})Q_i(t_0(\tau))\indic{\set{G}_{\tau}}}\\
&\mspace{32mu}+\markcol{2(\epochlength-2\convlength)\sum_{\tau=1}^{\ell_T}\sum_{i \in \set{I}(t_0(\tau)) \setminus \set{X}(\tau)}\expect{\mu_{i,\sigma_\tau^\star(i)}Q_i(t_0(\tau))}-\sum_{\tau=1}^{\ell_T}\expect{V_{\set{I}(t_0(\tau)) \setminus \set{X}(\tau)}(\bold{Q}(t_0(\tau)))}}.
\end{align*}
\end{lemma}
In our next lemma we bound the weight difference term $$\expect{\sum_{\tau=1}^{\ell_T} \sum_{i \in \set{X}(\tau)} (\mu_{i,\sigma^{\star}_{\tau}(i)} - \mu_{i,\sigma_{\tau}(i)})Q_i(t_0(\tau))\indic{\set{G}_{\tau}}}$$ following the proof of Lemma~\ref{lem:ucb-match-difference}. A proof of Lemma~\ref{lem:dynamic-match-difference} is provided in Appendix~\ref{app:dynamic-matching-difference}.
\begin{lemma}\label{lem:dynamic-match-difference}
We have
\begin{align*}
&\mspace{32mu}\expect{\sum_{\tau=1}^{\ell_T} \sum_{i \in \set{X}(\tau)} (\mu_{i,\sigma^{\star}_{\tau}(i)} - \mu_{i,\sigma_{\tau}(i)})Q_i(t_0(\tau))\indic{\set{G}_{\tau}}} \\
&\leq\frac{3}{16}\trafficslack \sum_{\tau=1}^{\ell_T}\sum_{i \in \set{X}(\tau)} \expect{\mu_{i,\sigma_\tau^{\star}(i)}Q_i(t_0(\tau))}
+\frac{1}{8}\trafficslack \sum_{\tau=1}^{\ell_T}\sum_{i \in \set{X}(\tau)} \lambda_i \expect{Q_i(t_0(\tau))} \\
&\mspace{32mu}+\frac{896K^2\markcol{\lambda_T^\star} \epochlength}{\trafficslack} + \markcol{2\sum_{\tau=1}^{\ell_T} \expect{\sum_{i \in \set{Q}(\tau)} Q_i(t_0(\tau))}}, 
\end{align*}	
where $\set{Q}(\tau)$ is the random set defined in Lemma~\ref{lem:dynamic-converge-main} if $\set{G}_\tau$ holds; and is the empty-set otherwise.
\end{lemma}
\begin{proof}[Proof of Lemma~\ref{lem:dynamic-drift-bound}]
Similar to the proof of Theorem~\ref{thm:queue-ucb} (Appendix~\ref{sec:ucb-full-proof}), by the stability assumption, we have for every $\tau$,
\[
\sum_{i \in \set{I}(t_0(\tau))} \expect{\mu_{i,\sigma_\tau^\star(i)}Q_i(t_0(\tau))} \geq (1+\trafficslack)\sum_{i \in \set{I}(t_0(\tau))} \lambda_iQ_i(t_0(\tau))
\]
since $\sigma_\tau^{\star}$ is the max-weight matching for weight $w_{i,j}=\mu_{i,j}Q_i(t_0(\tau))$ between queues $\set{I}(t_0(\tau))$ and the set of servers. Combining the bounds in Lemma~\ref{lem:dynamic-drift-decompose} and Lemma~\ref{lem:dynamic-match-difference}, we obtain
\begin{align*}
&\mspace{32mu}\expect{V(\bold{Q}(\ell_T \epochlength + 1)) - V(\bold{Q}(1))} \\
&\leq \sum_{\tau=1}^{\ell_T} (4\convlength - 0.625\trafficslack(\epochlength-2\convlength))\sum_{i \in \set{I}(t_0(\tau))} \lambda_i\expect{Q_i(t_0(\tau))} \\
&\mspace{32mu}+2(\epochlength-2\convlength)\frac{896K^2\lambda^\star_T \epochlength K^2}{\trafficslack }\\ &\mspace{32mu}+4K\ell_T(\epochlength-2\convlength)(\epochlength+2\convlength) + \ell_T K\epochlength^2\\
&\mspace{32mu}+4(\epochlength-2\convlength)\sum_{\tau=1}^{\ell_T} \expect{\sum_{i \in \set{Q}(\tau)} Q_i(t_0(\tau))} \\
&\mspace{32mu}+2(\epochlength-2\convlength)\sum_{\tau=1}^{\ell_T}\sum_{i \in \set{I}(t_0(\tau)) \setminus \set{X}(\tau)}\expect{\mu_{i,\sigma_\tau^\star(i)}Q_i(t_0(\tau))}-\sum_{\tau=1}^{\ell_T}\expect{V_{\set{I}(t_0(\tau)) \setminus \set{X}(\tau)}(\bold{Q}(t_0(\tau)))}.
\end{align*}
The first three terms are the same as that in the desired bound of Lemma~\ref{lem:dynamic-drift-bound}. Thus, it suffices to show that the remaining terms are bounded by $19K\epochlength^2$ for each $\tau \leq \ell_T$, i.e., 
\begin{equation}\label{eq:dynamic-drift-departure}
\begin{aligned}
19K\epochlength^2 &\geq K\epochlength^2 +4(\epochlength-2\convlength)\expect{\sum_{i \in \set{Q}(\tau)} Q_i(t_0(\tau))} \\
&\mspace{32mu}+2(\epochlength-2\convlength)\sum_{i \in \set{I}(t_0(\tau)) \setminus \set{X}(\tau)}\expect{\mu_{i,\sigma_\tau^\star(i)}Q_i(t_0(\tau))}\\
&\mspace{32mu}-\expect{V_{\set{I}(t_0(\tau)) \setminus \set{X}(\tau)}(\bold{Q}(t_0(\tau)))}.
\end{aligned}
\end{equation}
Recall that by Lemma~\ref{lem:dynamic-converge-main} and Lemma~\ref{lem:dynamic-match-difference}, the set $\set{Q}(\tau)$ is a subset of $\set{I}(t_0(\tau)) \setminus \set{X}(\tau)$ of size at most~$K$. We define $\set{Q}'(\tau) = \set{Q}(\tau) \cup \{i \in \set{I}(t_0(\tau)) \setminus \set{X}(\tau) \colon \sigma_\tau^\star(i) \neq \perp\}$ as  the set of queues that leave in epoch $\tau$ in which they either (i) are in the optimal matching or (ii) have proposed the highest bid to some server. Then $|\set{Q}'(\tau)| \leq |\set{Q}(\tau)| + K = 2K.$ In addition, we have
\begin{align*}
&\mspace{32mu}4(\epochlength-2\convlength)\expect{\sum_{i \in \set{Q}(\tau)} Q_i(t_0(\tau))} +2(\epochlength-2\convlength)\sum_{i \in \set{I}(t_0(\tau)) \setminus \set{X}(\tau)}\expect{\mu_{i,\sigma_\tau^\star(i)}Q_i(t_0(\tau))} \\
&\leq 6\epochlength \expect{\sum_{i \in \set{Q}'(\tau)} Q_i(t_0(\tau))},
\end{align*}
since all indices with positive summands in both sums are contained in $\set{Q}'(\tau)$.
Moreover, we know $V_{\set{I}(t_0(\tau)) \setminus \set{X}(\tau)}(\bold{Q}(t_0(\tau))) = \sum_{i \in \set{I}(t_0(\tau)) \setminus \set{X}(\tau)} Q_i(t_0(\tau))^2 \geq \sum_{i \in \set{Q}'(\tau)} Q_i(t_0(\tau))^2.$ Therefore,
\begin{align*}
6\epochlength \expect{\sum_{i \in \set{Q}'(\tau)} Q_i(t_0(\tau))} -\expect{V_{\set{I}(t_0(\tau)) \setminus \set{X}(\tau)}(\bold{Q}(t_0(\tau)))}&\leq \expect{\sum_{i \in \set{Q}'(\tau)} Q_i(t_0(\tau))(6\epochlength - Q_i(t_0(\tau))} \\
&\leq \expect{9|\set{Q}'(\tau)|\epochlength^2} \leq 18K\epochlength^2.
\end{align*}
We then finish the proof by establishing \eqref{eq:dynamic-drift-departure}.

\end{proof}
\subsection{Proof of drift decomposition (Lemma~\ref{lem:dynamic-drift-decompose})}\label{app:dynamic-drift-decompose}
Recall that $Q_i(t) = 0$ if $i \not \in \set{I}(t)$ (either because it has not arrived or it has left.) The same proof of Lemma~\ref{lem:bound-general-drift} gives the following bound adapted to our new definition of set-based Lyapunov function.
\begin{lemma}\label{lem:set-general-drift}
Fix a time slot $t_1 \geq 1$ and a set of queues $\set{I}$. Consider a future interval $[t_2,t_3 + 1], t_1 \leq t_2 \leq t_3,$ and an event $\set{W}$ that is independent of all arrivals in $[t_1,t_3].$ Then it holds that 
\begin{equation*}
\expect{V_{\set{I}}(\bold{Q}(t_3 + 1)) - V_{\set{I}}(\bold{Q}(t_2)) \mid \bold{Q}(t_1), \set{W}}  \leq  (t_3 - t_2 + 1)\left(2\sum_{i \in \set{I}} \lambda_{i}Q_{i}(t_1) + K\left(1 +  t_3+t_2-2t_1\right)\right).
\end{equation*}
\end{lemma}
We now provide the proof of Lemma~\ref{lem:dynamic-drift-decompose}.
\begin{proof}[Proof of Lemma~\ref{lem:dynamic-drift-decompose}]
Fix an epoch $\tau \leq \ell_T$. It suffices to consider the drift within epoch $\tau$, i.e.,~$\expect{ V(\bold{Q}(t_0(\tau+1))) - V(\bold{Q}(t_0(\tau)))}$, since Lemma~\ref{lem:dynamic-drift-decompose} bounds the sum of these drifts. For ease of notations, we write $t_0$ as a shorthand of $t_0(\tau).$ Recall that $\set{X}(\tau) = \set{I}(t_0) \cap \set{I}(t_0+\epochlength)$. By definition, 
\begin{align*}
&\mspace{32mu}\expect{V(\bold{Q}(t_0(\tau+1))) - V(\bold{Q}(t_0(\tau)))}  \\
&= \expect{V_{\set{I}(t_0(\tau+1))}(\bold{Q}(t_0(\tau+1))) - V_{\set{I}(t_0(\tau))}(\bold{Q}(t_0(\tau))} \\
&= \expect{V_{\set{X}(\tau)}(\bold{Q}(t_0(\tau+1)))-V_{\set{X}(\tau)}(\bold{Q}(t_0(\tau)))} - \expect{V_{\set{I}(t_0(\tau)) \setminus \set{X}(\tau)}(\bold{Q}(t_0(\tau)))}  \\
&\mspace{32mu}+\expect{V_{\set{I}(t_0(\tau+1)) \cap \set{I}^c(t_0(\tau))}(\bold{Q}(t_0(\tau+1))) - V_{\set{I}(t_0(\tau+1)) \cap \set{I}^c(t_0(\tau))}(\bold{Q}(t_0(\tau)))}.
\end{align*}
The second expectation also appears in the bound of Lemma~\ref{lem:dynamic-drift-decompose}. Therefore, it suffices to bound the first and the third expectation. Notice Lemma~\ref{lem:set-general-drift} allows us to bound the third expectation as
\begin{equation}\label{eq:bound-e5-step1}
\begin{aligned}
&\mspace{32mu}\expect{V_{\set{I}(t_0+\epochlength) \cap \set{I}^c(t_0)}(\bold{Q}(t_0+\epochlength)-V_{\set{I}(t_0+\epochlength) \cap \set{I}^c(t_0)}(\bold{Q}(t_0)} \\
&\leq \epochlength\left(2\expect{\sum_{i \in \set{I}(t_0+\epochlength) \cap \set{I}^c(t_0)} \lambda_i Q_i(t_0)} + K\epochlength\right) \\
&= K\epochlength^2,
\end{aligned}
\end{equation}
where the equality is due to $Q_i(t_0) = 0$ for $i \in\set{I}^c(t_0)$ and $t_0$ is used as shorthand for $t_0(\tau)$. 

For the term $\expect{V_{\set{X}(\tau)}(\bold{Q}(t_0+\epochlength)) - V_{\set{X}(\tau)}(\bold{Q}(t_0)}$, we use a bound that is analogous to Lemma~\ref{lem:ucb-drift-decompose}. Recall that $\sigma^\star_\tau$ is the maximum-weight matching between $\set{I}(t_0)$ and the set of servers with weight $\mu_{i,j}Q_i(t_0)$. Then the same analysis of Lemma~\ref{lem:ucb-drift-decompose}, which holds since $\sigma^\star_\tau$ is the maximum-weight matching in time slot $t_0$ and consequently independent of any events (arrivals of jobs or queues, service of jobs, etc) that occur within epoch $\tau$, gives the following bound
\begin{equation}\label{eq:bound-e5-step2}
\begin{aligned}
&\mspace{32mu}\expect{V_{\set{X}(\tau)}(t_0 + \epochlength) - V_{\set{X}(\tau)}(t_0)} \\
&\leq  \left(4\convlength + (2 + \trafficslack / 8)(\epochlength - 2\convlength)\right)\sum_{i \in \set{X}(\tau)} \lambda_i \expect{Q_i(t_0(\tau))} \\
&\mspace{32mu}-2(1-\trafficslack/16)(\epochlength - 2\convlength)\sum_{i \in \set{X}(\tau)} \expect{\mu_{i,\sigma^{\star}_{\tau}(i)}Q_i(t_0(\tau))} \\
&\mspace{32mu}+4K(\epochlength-2\convlength)(\epochlength+2\convlength) \\
&\mspace{32mu}+2(\epochlength-2\convlength)\expect{\sum_{i \in \set{X}(\tau)} (\mu_{i,\sigma^{\star}_{\tau}(i)} - \mu_{i,\sigma_{\tau}(i)})Q_i(t_0(\tau))\indic{\set{G}_{\tau}}}.
\end{aligned}
\end{equation}
We note that
\begin{equation}\label{eq:bound-e5-step3}
\begin{aligned}
&\mspace{32mu}-2(1-\trafficslack/16)(\epochlength - 2\convlength)\sum_{i \in \set{X}(\tau)} \expect{\mu_{i,\sigma^{\star}_{\tau}(i)}Q_i(t_0(\tau))} \\
&\leq -2(1-\trafficslack/16)(\epochlength - 2\convlength)\sum_{i \in \set{I}(t_0(\tau))} \expect{\mu_{i,\sigma^{\star}_{\tau}(i)}Q_i(t_0(\tau))}\\
&\mspace{32mu}+ 2(\epochlength-2\convlength)\sum_{i \in \set{I}(t_0(\tau)) \setminus \set{X}(\tau)} \expect{\mu_{i,\sigma^{\star}_{\tau}(i)}Q_i(t_0(\tau))}.
\end{aligned}
\end{equation}
To finish the proof of Lemma~\ref{lem:dynamic-drift-decompose}, we have
\begin{align*}
&\mspace{32mu}\expect{V(\bold{Q}(t_0(\tau+1))) - V(\bold{Q}(t_0(\tau)))}  \\
&= \expect{V_{\set{X}(\tau)}(\bold{Q}(t_0(\tau+1)))-V_{\set{X}(\tau)}(\bold{Q}(t_0(\tau)))} - \expect{V_{\set{I}(t_0(\tau)) \setminus \set{X}(\tau)}(\bold{Q}(t_0(\tau)))}  \\
&\mspace{32mu}+\expect{V_{\set{I}(t_0(\tau+1)) \cap \set{I}^c(t_0(\tau))}(\bold{Q}(t_0(\tau+1))) - V_{\set{I}(t_0(\tau+1)) \cap \set{I}^c(t_0(\tau))}(\bold{Q}(t_0(\tau)))} \\
&\overset{\eqref{eq:bound-e5-step1}}{\leq}\expect{V_{\set{X}(\tau)}(\bold{Q}(t_0(\tau+1)))-V_{\set{X}(\tau)}(\bold{Q}(t_0(\tau)))}  + K\epochlength^2 - \expect{V_{\set{I}(t_0(\tau)) \setminus \set{X}(\tau)}(\bold{Q}(t_0(\tau)}.
\end{align*}
Using Eq.~\ref{eq:bound-e5-step2} to replace the first expectation, it holds
\begin{align*}
&\mspace{32mu}\expect{V(\bold{Q}(t_0(\tau+1))) - V(\bold{Q}(t_0(\tau)))}  \\
&\leq\left(4\convlength + (2 + \trafficslack / 8)(\epochlength - 2\convlength)\right)\sum_{i \in \set{X}(\tau)} \lambda_i \expect{Q_i(t_0(\tau))} \\
&\mspace{32mu}-2(1-\trafficslack/16)(\epochlength - 2\convlength)\sum_{i \in \set{X}(\tau)} \expect{\mu_{i,\sigma^{\star}_{\tau}(i)}Q_i(t_0(\tau))} \\
&\mspace{32mu}+4K(\epochlength-2\convlength)(\epochlength+2\convlength) \\
&\mspace{32mu}+2(\epochlength-2\convlength)\expect{\sum_{i \in \set{X}(\tau)} (\mu_{i,\sigma^{\star}_{\tau}(i)} - \mu_{i,\sigma_{\tau}(i)})Q_i(t_0(\tau))\indic{\set{G}_{\tau}}} \\
&\mspace{32mu}+ K\epochlength^2 - \expect{V_{\set{I}(t_0(\tau)) \setminus \set{X}(\tau)}(\bold{Q}(t_0(\tau)}.
\end{align*}
Since queue lengths are non-negative and $\set{X}(\tau) \subseteq \set{I}(t_0)$, we can replace the sum in the first term by $\sum_{i \in \set{I}(t_0)} \lambda_i \expect{Q_i(t_0(\tau))}$. Moreover, replacing the second expression by Eq.~\ref{eq:bound-e5-step3}, we obtain
\begin{align*}
&\mspace{32mu}\expect{V(\bold{Q}(t_0(\tau+1))) - V(\bold{Q}(t_0(\tau)))}  \\
&\leq\left(4\convlength + (2 + \trafficslack / 8)(\epochlength - 2\convlength)\right)\sum_{i \in \set{I}(t_0(\tau))} \lambda_i \expect{Q_i(t_0(\tau))} \\
&\mspace{32mu}-2(1-\trafficslack/16)(\epochlength - 2\convlength)\sum_{i \in \set{I}(t_0(\tau))} \expect{\mu_{i,\sigma^{\star}_{\tau}(i)}Q_i(t_0(\tau))} \\
&\mspace{32mu}+\markcol{2(\epochlength-2\convlength)\sum_{i \in \set{I}(t_0(\tau)) \setminus \set{X}(\tau)} \expect{\mu_{i,\sigma^{\star}_{\tau}(i)}Q_i(t_0(\tau))}}\\
&\mspace{32mu}+4K(\epochlength-2\convlength)(\epochlength+2\convlength) \\
&\mspace{32mu}+2(\epochlength-2\convlength)\expect{\sum_{i \in \set{X}(\tau)} (\mu_{i,\sigma^{\star}_{\tau}(i)} - \mu_{i,\sigma_{\tau}(i)})Q_i(t_0(\tau))\indic{\set{G}_{\tau}}} \\
&\mspace{32mu}+\markcol{K\epochlength^2 - \expect{V_{\set{I}(t_0(\tau)) \setminus \set{X}(\tau)}(\bold{Q}(t_0(\tau)}}.
\end{align*}
Summing over epoch $\tau \leq \ell_T$ and arranging terms finish the proof of Lemma~\ref{lem:dynamic-drift-decompose}.
\end{proof}
\subsection{Proof of weight difference bound (Lemma~\ref{lem:dynamic-match-difference})}\label{app:dynamic-matching-difference}
The proof of Lemma~\ref{lem:dynamic-match-difference} naturally follows from the proof of Lemma~\ref{lem:ucb-match-difference}. We provide a proof here for completeness. For an epoch $\tau$, let us set $\Delta_{i,j}(t_0(\tau)) = \sqrt{\frac{3\ln(t_0(\tau)-\ts(i)+1+K)}{n_{i,j}(t_0(\tau)}}.$ Recall the definition of events $\set{E}_{\tau,i}^1,\set{E}_{\tau,i}^2$ where
\begin{align*}
\set{E}_{\tau,i}^1 &= \left\{\exists_{j\in\set{K}}, |\hat{\mu}_{i,j}(t_0(\tau))-\mu_{i,j}| > \Delta_{i,j}(t_0(\tau))\right\} \\
\set{E}_{\tau,i}^2 &= \left\{ \Delta_{i,\sigma_{\tau}(i)}(t_0(\tau)) > \frac{1}{16}\trafficslack\mulower\right\}.
\end{align*}
We first show the following result similar to Lemma~\ref{lem:ucb-diff-conf}.
\begin{lemma}\label{lem:dynamic-diff-conf}
We have
\begin{align*}
&\mspace{32mu}\expect{\sum_{\tau=1}^{\ell_T}\sum_{i \in \set{X}(\tau)} (\mu_{i,\sigma^{\star}_{\tau}(i)} - \mu_{i,\sigma_{\tau}(i)})Q_i(t_0(\tau))\indic{\set{G}_{\tau}}} \notag\\
&\leq \frac{3}{16}\trafficslack\expect{\sum_{\tau=1}^{\ell_T}\sum_{i \in \set{X}(\tau)} \mu_{i,\sigma_\tau^\star(i)}Q_i(t_0(\tau))} + 2\sum_{\tau=1}^{\ell_T}\sum_{i \in \set{X}(\tau)} \expect{\indic{\set{E}^1_{\tau,i}\cup\set{E}^2_{\tau,i}}\indic{\set{G}_\tau}Q_i(t_0(\tau))} \\
&\mspace{32mu}\markcol{+ 2\sum_{\tau=1}^{\ell_T} \expect{\sum_{i \in \set{Q}(\tau)} Q_i(t_0(\tau))}},
\end{align*}
where $\set{Q}(\tau)$ is the random set defined in Lemma~\ref{lem:dynamic-converge-main} if $\set{G}_\tau$ holds; and is the empty-set otherwise.
\end{lemma}
\begin{proof}
It suffices to consider each epoch $\tau \leq \ell_T$ and then sum over epochs to get Lemma~\ref{lem:dynamic-diff-conf}. Fix an epoch $\tau$. Let $t_0 = t_0(\tau).$ For each $i \in \set{X}(\tau)$ and $j \in [K]$, we have 
\begin{equation}\label{eq:bound-ucb-weight}
\mu_{i,j} - \indic{\set{E}_{\tau,i}^1} \leq \bar{\mu}_{i,j} \leq \mu_{i,j}+\frac{\trafficslack\mulower}{8}+\indic{\set{E}_{\tau,i}^1 \cup \set{E}_{\tau,i}^2}.
\end{equation}
Conditioning on $\set{G}_\tau,$ which is a subset of $\set{E}_\tau$, we know that $\sigma_\tau$ is a matching. By Lemma~\ref{lem:dynamic-converge-main}, there is a set $\set{Q}(\tau) \subseteq \set{I}(t_0(\tau)) \setminus \set{X}(\tau)$ of size at most $K$, such that
\[
\sum_{i \in \set{X}(\tau)} \bar{\mu}_{i,\sigma_\tau(i)}Q_i(t_0) \geq (1-\frac{1}{16}\trafficslack)\sum_{i \in \set{X}(\tau)} \bar{\mu}_{i,\hat{\sigma}^\star_\tau(i)}Q_i(t_0)-2\sum_{i \in \set{Q}(\tau)} Q_i(t_0),
\]
where $\hat{\sigma}^\star_\tau$ is the maximum-weight matching between $\set{X}(\tau)$ and the set of servers with weight $\bar{\mu}_{i,j}Q_i(t_0)$. Therefore, we have
\begin{align*}
 &\mspace{32mu}\sum_{i \in \set{X}(\tau)} \mu_{i,\sigma_\tau(i)}Q_i(t_0) \\ &\geq \sum_{i \in \set{X}(\tau)}\bar{\mu}_{i,\sigma_\tau(i)}Q_i(t_0) - \frac{\trafficslack \mulower}{8}\sum_{i \in \set{X}(\tau), \sigma_\tau(i) \neq \perp} Q_i(t_0) - \sum_{i \in \set{X}(\tau)}\indic{\set{E}_{\tau,i}^1 \cup \set{E}_{\tau,i}^2} Q_i(t_0)\\
  &\geq (1-\frac{\trafficslack}{16})\sum_{i \in \set{X}(\tau)} \bar{\mu}_{i,\hat{\sigma}^\star_\tau(i)}Q_i(t_0)-2\sum_{i \in \set{Q}(\tau)} Q_i(t_0)- \frac{\trafficslack}{8}\sum_{i \in \set{X}(\tau), \sigma_\tau(i) \neq \perp} \mulower Q_i(t_0)\\&\mspace{32mu}- \sum_{i \in \set{X}(\tau)}\indic{\set{E}_{\tau,i}^1 \cup \set{E}_{\tau,i}^2} Q_i(t_0) 
 \end{align*}
where the first inequality uses the second bound in \eqref{eq:bound-ucb-weight} for each pair of $i,\sigma_\tau(i)$ and the second inequality uses that by Lemma~\ref{lem:dynamic-converge-main} $\sigma_\tau$ is an approximate maximum-weight matching with weight $w_{i,j}=\bar{\mu}_{i,j}Q_i(t_0)$. We will further bound this expression by observing that  $\bar{\mu}_{i,j} \geq \mulower$ for every pair of $i,j$ (line~\ref{line:ucb-esti}) and $\sum_{i \in \set{X}(\tau)} \bar{\mu}_{i,\hat{\sigma}^\star_\tau(i)}Q_i(t_0) \geq \sum_{i \in \set{X}(\tau)} \bar{\mu}_{i,\sigma_\tau(i)}Q_i(t_0)$ because $\hat{\sigma}^\star_\tau$ is the maximum-weight matching with weight $w_{i,j}=\bar{\mu}_{i,j}Q_i(t_0)$. This gives us
\[\frac{\trafficslack}{8}\sum_{i \in \set{X}(\tau), \sigma_\tau(i) \neq \perp} \mulower Q_i(t_0) \leq \frac{\trafficslack}{8}\sum_{i \in \set{X}(\tau), \sigma_\tau(i) \neq \perp} \bar{\mu}_{i,\sigma_\tau(i)} Q_i(t_0) \leq \frac{\trafficslack}{8}\sum_{i \in \set{X}(\tau)} \bar{\mu}_{i,\hat{\sigma}^\star_\tau(i)} Q_i(t_0),\]
which allows us to obtain the bound
\begin{align*}
  \sum_{i \in \set{X}(\tau)}\mu_{i,\sigma_\tau(i)}Q_i(t_0)&\geq (1-\frac{3\trafficslack}{16})\sum_{i \in \set{X}(\tau)} \bar{\mu}_{i,\hat{\sigma}^\star_\tau(i)}Q_i(t_0)-2\sum_{i \in \set{Q}(\tau)} Q_i(t_0) - \sum_{i \in \set{X}(\tau)}\indic{\set{E}_{\tau,i}^1 \cup \set{E}_{\tau,i}^2} Q_i(t_0).
 \end{align*}
Using again that $\hat{\sigma}^\star_\tau$ is the maximum-weight matching with weight $w_{i,j}=\bar{\mu}_{i,j}Q_i(t_0)$ we then obtain
 \begin{align*}
 &\mspace{32mu}\sum_{i \in \set{X}(\tau)} \mu_{i,\sigma_\tau(i)}Q_i(t_0) \\
 &\geq (1-\frac{3\trafficslack}{16})\sum_{i \in \set{X}(\tau)} \bar{\mu}_{i,\sigma^\star_\tau(i)}Q_i(t_0)-2\sum_{i \in \set{Q}(\tau)} Q_i(t_0) - \sum_{i \in \set{X}(\tau)}\indic{\set{E}_{\tau,i}^1 \cup \set{E}_{\tau,i}^2} Q_i(t_0) \\
 &\overset{\eqref{eq:bound-ucb-weight}}{\geq} (1-\frac{3\trafficslack}{16})\sum_{i \in \set{X}(\tau)} \mu_{i,\sigma^\star_\tau(i)}Q_i(t_0) -\sum_{i \in \set{X}(\tau)} \indic{\set{E}_{\tau,i}^1}Q_i(t_0)- 2\sum_{i \in \set{Q}(\tau)} Q_i(t_0) \\
 &\mspace{32mu}- \sum_{i \in \set{X}(\tau)}\indic{\set{E}_{\tau,i}^1 \cup \set{E}_{\tau,i}^2} Q_i(t_0) \\
 &\geq (1-\frac{3\trafficslack}{16})\sum_{i \in \set{X}(\tau)} \mu_{i,\sigma^\star_\tau(i)}Q_i(t_0) - 2\sum_{i \in \set{Q}(\tau)} Q_i(t_0) - 2\sum_{i \in \set{X}(\tau)}\indic{\set{E}_{\tau,i}^1 \cup \set{E}_{\tau,i}^2} Q_i(t_0) 
\end{align*}
We finish the proof by taking expectation and summing over all epoch $\tau \leq \ell_T.$
\end{proof}
Recall that $\ell_T = \lceil \frac{T}{\epochlength}\rceil$. We can define
$\hat{e}_{i,1} = \sum_{\tau=1}^{\ell_T} \indic{\set{E}_{\tau,i}^1},
\hat{e}_{i,2} = \sum_{\tau=1}^{\ell_T} \indic{\set{G}_\tau,\set{E}_{\tau,i}^2}$
and let $e_i = \hat{e}_{i,1} + \hat{e}_{i,2}$. We next upper bound $\expect{e_i^2}$ following similar argument as the proof of Lemma~\ref{lem:ucb-bound-error}.
\begin{lemma}\label{lem:dynamic-square-error}
It holds that for any queue $i$, $\expect{e_i^2} \leq 14K^2\ln^2(\markcol{s_T(i)}+K+1).$
\end{lemma}
\begin{proof}
We first show $\expect{\hat{e}_{i,1}^2} \leq 5$. Fix a queue $i$. Let $\tau_s(i) = \lfloor \frac{\ts(i)-1}{\epochlength}\rfloor+1$ be the epoch this queue joins into the system. We have
\begin{align*}
\expect{\hat{e}_{i,1}^2} = \expect{\left(\sum_{\tau=\tau_s(i)}^{\ell_T} \indic{\set{E}_{\tau,i}^1}\right)^2} &= \expect{\sum_{\tau = \tau_s(i)}^{\ell_T} \indic{\set{E}_{\tau,i}^1}} + 2\expect{\sum_{\tau=\tau_s(i)}^{\ell_T}\sum_{\tau'=\tau+1}^{\ell_T}\indic{\set{E}_{\tau,i}^1}\indic{\set{E}_{\tau',i}^1}} \\
&\leq \expect{\sum_{\tau=\tau_s(i)}^{\ell_T} \indic{\set{E}_{\tau,i}^1}} + 2\expect{\sum_{\tau=\tau_s(i)}^{\ell_T}(\tau-\tau_s(i))\indic{\set{E}_{\tau,i}^1}} \\
&\leq 2\sum_{\tau=\tau_s(i)}^{\ell_T}(\tau-\tau_s(i)+1)\Pr\{\set{E}_{\tau,i}^1\}.
\end{align*}
Since samples are unbiased, the same proof of Lemma~\ref{lem:prob-a1} shows that for a fixed epoch $\tau$, we have $\Pr\{\set{E}_{\tau,i}^1\} \leq \frac{2K}{(K+t_0(\tau)-\ts(i)+1)^4}$. Therefore,
\begin{equation}\label{eq:dynamic-bound-error1}
\expect{\hat{e}_{i,1}^2} \leq \sum_{\tau=\tau_s(i)}^{\ell_T}
\frac{4K(\tau-\tau_s(i)+1)}{(K+(\tau-\tau_s(i))\epochlength+1)^4} \leq 4 +  \sum_{\tau=\tau_s(i)+1}^{\ell_T}\frac{4(\tau-\tau_s(i)+1)}{(\tau-\tau_s(i))^4\epochlength^4} \leq 5.
\end{equation}
Now we show almost surely that $\hat{e}_{i,2} \leq K + \frac{\ln(s_T(i)+K)}{2K}$ using arguments in the proof of Lemma~\ref{lem:ucb-bound-e2}. The first inequality in the proof of Lemma~\ref{lem:ucb-bound-e2} gives
\[
\hat{e}_{i,2} \leq \sum_{j=1}^K \sum_{\tau=1}^{\ell_T} \indic{\Delta_{i,\sigma_\tau(i)}(t_0(\tau)) > \trafficslack\mulower/16, \set{G}_\tau,\sigma_\tau(i) = j}.
\]
Fix $j \in \set{K}$. We define event $\set{B}_{\tau}^{i,j} = \{\Delta_{i,\sigma_\tau(i)}(t_0(\tau)) > \trafficslack\mulower/16\}\cap \set{G}_\tau \cap \{\sigma_\tau(i) = j\}$ such that the above can be rewritten as $\hat{e}_{i,2} \leq \sum_{j=1}^K \sum_{\tau=1}^{\ell_T}\indic{\set{B}_{\tau}^{i,j}}$. Let $\tau'$ be the last epoch with $\indic{\set{B}^{i,j}_{\tau'}} = 1.$ For that event to hold true we must have $\Delta_{i,j}(t_0(\tau')) > \trafficslack\mulower/16$. Moreover, by the definition of $\Delta_{i,j}(t_0(\tau'))$, we must have 
\[
\trafficslack\mulower/16<\Delta_{i,j}(t_0(\tau'))=\sqrt{\frac{3\ln(t_0(\tau')-\ts(i)+1+K)}{n_{i,j}(t_0(\tau'))}} \leq \sqrt{\frac{3\ln(s_T(i)+K)}{n_{i,j}(t_0(\tau'))}},
\]
where the last inequality is because $i$ is in the system in time slot $t_0(\tau')$ and thus $t_0(\tau') - \ts(i) \leq \min(\te(i),T) - \ts(i).$
It implies 
\[
2^{-8}\trafficslack^2\mulower^2<
\frac{3\ln(s_T(i)+K)}{n_{i,j}(t_0(\tau'))}.
\]
Following the same argument after \eqref{eq:error2-sample-bound}, with $\frac{3\ln(s_T(i)+K)}{n_{i,j}(t_0(\tau'))}$ replacing $\frac{3\ln(T+K)}{n_{i,j}(t_0(\tau'))}$ in the proof of Lemma~\ref{lem:ucb-bound-e2}, gives $\hat{e}_{i,2} \leq K + \frac{\ln(s_T(i)+K)}{2K}$. Thus,
$\expect{\hat{e}_{i,2}^2} \leq 2K^2+\frac{\ln^2(s_T(i)+K)}{2K^2} \leq 4K^2\ln^2(s_T(i)+K)$.

Finally, we have
\[
\expect{e_i^2} = \expect{(\hat{e}_{i,1}+\hat{e}_{i,2})^2} \leq 2\expect{\hat{e}_{i,1}^2}+2\expect{\hat{e}_{i,2}^2} \leq 10+4K^2\ln^2(s_T(i)+K) \leq 14K^2\ln^2(s_T(i)+K+1),
\]
which completes the proof.
\end{proof}

It remains to bound the term $\sum_{\tau=1}^{\ell_T}\sum_{i \in \set{X}(\tau)} \expect{\indic{\set{E}^1_{\tau,i}\cup\set{E}^2_{\tau,i}}\indic{\set{G}_\tau}Q_i(t_0(\tau))}$. Following the proof of Lemma~\ref{lem:ucb-spread-cost-error} gives an upper bound as follows.
\begin{lemma}\label{lem:dynamic-matching-regret}
It holds that
\begin{align*}
&\mspace{32mu}\expect{\sum_{\tau=1}^{\ell_T}\sum_{i \in \set{X}(\tau)} \expect{\indic{\set{E}^1_{\tau,i}\cup\set{E}^2_{\tau,i}}\indic{\set{G}_\tau}Q_i(t_0(\tau))}} \\
&\leq \frac{1}{16}\trafficslack \sum_{\tau=1}^{\ell_T}\sum_{i \in \set{X}(\tau)} \lambda_i \expect{Q_i(t_0(\tau))}+ \frac{448\epochlength K^2}{\trafficslack}\markcol{\lambda^\star_T}.
\end{align*}
\end{lemma}
\begin{proof}
Note that Lemma~\ref{lem:ucb-spread-cost} still applies. Following the proof sketch of Lemma~\ref{lem:ucb-spread-cost-error}, we have
\begin{align*}
&\mspace{32mu}\expect{\sum_{\tau=1}^{\ell_T}\sum_{i \in \set{X}(\tau)} \expect{\indic{\set{E}^1_{\tau,i}\cup\set{E}^2_{\tau,i}}\indic{\set{G}_\tau}Q_i(t_0(\tau))}} \\
&\leq \frac{1}{16}\trafficslack \sum_{\tau=1}^{\ell_T}\sum_{i \in \set{X}(\tau)} \lambda_i \expect{Q_i(t_0(\tau))}+ \frac{32\epochlength}{\trafficslack}\sum_{i \in \cup_{t \in T}\set{I}(t)} \lambda_i^{-1}\expect{e_i^2}.
\end{align*}
By Lemma~\ref{lem:dynamic-square-error}, we have $\expect{e_i^2} \leq 14K^2\ln^2(s_T(i)+K+1)$ for every queue $i$. Therefore,
\[
\frac{32\epochlength}{\trafficslack}\sum_{i \in \cup_{t \in T}\set{I}(t)} \lambda_i^{-1}\expect{e_i^2} \leq \frac{448\epochlength K^2}{\trafficslack}\sum_{i \in \cup_{t \in T}\set{I}(t)} \lambda_i^{-1}\ln^2(s_T(i)+K+1) = \frac{448\epochlength K^2}{\trafficslack}\lambda_T^\star,
\]
which completes the proof.
\end{proof}
We can now prove Lemma~\ref{lem:dynamic-match-difference} using Lemma~\ref{lem:dynamic-diff-conf} and Lemma~\ref{lem:dynamic-matching-regret}.
\begin{proof}[Proof of Lemma~\ref{lem:dynamic-match-difference}]
With Lemma~\ref{lem:dynamic-diff-conf}, we have
\begin{align*}
&\mspace{32mu}\expect{\sum_{\tau=1}^{\ell_T}\sum_{i \in \set{X}(\tau)} (\mu_{i,\sigma^{\star}_{\tau}(i)} - \mu_{i,\sigma_{\tau}(i)})Q_i(t_0(\tau))\indic{\set{G}_{\tau}}} \notag\\
&\leq \frac{3}{16}\trafficslack\expect{\sum_{\tau=1}^{\ell_T}\sum_{i \in \set{X}(\tau)} \mu_{i,\sigma_\tau^\star(i)}Q_i(t_0(\tau))} + 2\sum_{\tau=1}^{\ell_T}\sum_{i \in \set{X}(\tau)} \expect{\indic{\set{E}^1_{\tau,i}\cup\set{E}^2_{\tau,i}}\indic{\set{G}_\tau}Q_i(t_0(\tau))} \\
&\mspace{32mu}+ 2\sum_{\tau=1}^{\ell_T} \expect{\sum_{i \in \set{Q}(\tau)} Q_i(t_0(\tau))} 
\end{align*}
We finish the proof by using Lemma~\ref{lem:dynamic-matching-regret} to replace the second term.
\end{proof}

%% file: app_simulation.tex
\subsection{Implementation details}\label{app:sim-implement}
Parameters in DAM.K, DAM.FE and DAM.UCB are set according to Eq.~(\ref{eq:parameter-setting}) but with tuned constants so that $\convlength = \lceil\frac{K\checkperiod}{4\trafficslack}(\log N + K)\rceil, \epochlength = \lceil \frac{2}{\trafficslack}\convlength\rceil$. The price update step size (Line~\ref{algoline:price-update} in Algorithm~\ref{algo:DAM-queue}) is adjusted to $0.5\trafficslack w_{i,j^\star}.$ For DAM.FE, the exploration probability is $\frac{K}{\ell^{0.8}}$. The codes of ADEQUA and EXP3.P.1 are taken from \cite{sentenac2021decentralized}. The exploration probability of ADEQUA is set as $(N+K)t^{1/4}$ following \cite{sentenac2021decentralized}. Note that for DAM.FE, it can also use samples during \textsc{DAM.commit} phase like (Line~\ref{algoline:dam-commit-sample} in Algorithm~\ref{algo:DAM-ucb}). We include this sample collection into our implementation of DAM.FE. Each simulation is repeated for 15 runs.

\subsection{Instance settings}\label{app:sim-setting}
The first instance (Left of Fig.~\ref{fig:packets-n}) consists of $N=K=4$, $\lambda_i = \frac{N+1}{N^2},\forall i$ and $\mu_{i,1} = 1, \mu_{i,j} = \frac{N-1}{N^2}$ for $j \geq 2.$ Both $\trafficslack, \mulower$ are small, equal to 0.25 and 0.1875 respectively.

The second instance (Middle of Fig.~\ref{fig:packets-n}) is a system with $N=K=8$ and for all $i \in \set{N}$, $\lambda_i = 0.4, \mu_{i,1}=\mu_{i,2} = 0.9, \mu_{i,j}=0.4$ for $j > 2$. This system enjoys $\trafficslack = 0.3125,\mulower=0.4$. 

The third instance (Right of Fig.~\ref{fig:packets-n}) has $N=64, K = 4$. Service rates are given by $\mu_{i,1}=1,\mu_{i,j}=0.4$ for $j \in \{2,3,4\}.$ For arrival rates, $\lambda_1 = \cdots = \lambda_4 = 0.3,$ and $\lambda_i = \frac{1}{600}.$ We have $\trafficslack\approx 0.7,\mulower=0.4$ for this instance.

For the forth instance (Left of Fig.~\ref{fig:fluctuate_asym}), the system is small with $N=K=3$. Service rates are fixed with $\mu_{i,1} = 1, \mu_{i,2}=0.5,\mu_{i,3}=0.3$ for all $i \in \set{N}$. Arrival rates periodically change between ~$[0.7,0.5,0.3],[0.5,0.5,0.5],$ and $[0.4,0.8,0.2]$ every $10^4$ time slots. 

In the fifth instance (Right of Fig.~\ref{fig:fluctuate_asym}), we consider a $N=4,K=4$ system with asymmetric service rates where $\mu_{1,j}=1,\forall j \in \set{K}$ and $\mu_{i,1}=1,\mu_{i,2}=0.5,\mu_{i,3}=0.4,\mu_{i,4}=0.2$ for $i > 1$. Arrival rates are $(5/6,0.7,0.5,0.4)$.

The sixth instance (Fig.~\ref{fig:dynamic}) follows Example~\ref{example:dynamic} with two queues and two servers. Arrival rates are $[0.7,0.4]$ and service rates are $\mu_{1,1}=\mu_{2,2}=0.9,\mu_{1,2}=\mu_{2,1}=0.3.$ Both queues join the system in time slot $1$. At the beginning of each epoch, queue $2$ is replaced by a new but identical queue with probability $p$. The left plot of Fig.~\ref{fig:dynamic} sets $p = 1$. The right plot of Fig.~\ref{fig:dynamic} sets $p \in \{2^{-19},2^{-18},\ldots,2^{0}\}$.

\subsection{Robustness benefits}\label{app:sim-robust}
\begin{figure}[!h]
    \centering
     \scalebox{0.5}{\input{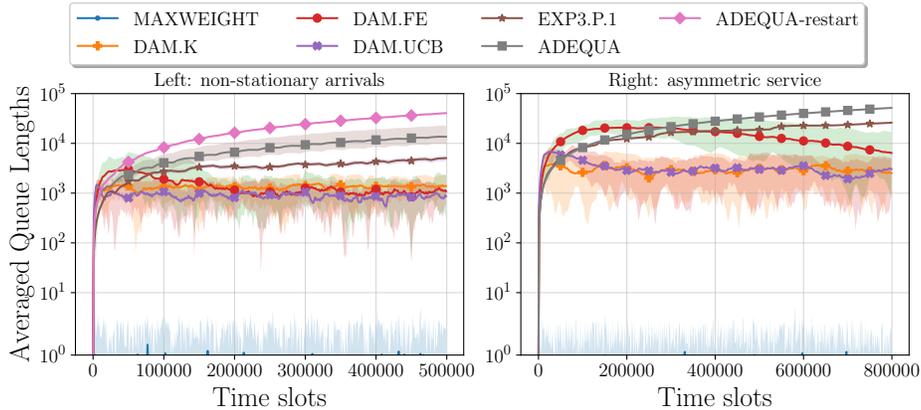}}
      \caption{Convergence of averaged queue lengths under different algorithms when either arrival rates can change (left) or service rates are asymmetric (right). Previous algorithms, ADEQUA and EXP3.P.1., fail to stabilize the system. DAM.FE and DAM.UCB remain robust to these changes and stabilize the system.}
      \label{fig:fluctuate_asym}
\end{figure}

In the fourth instance (left plot in Fig. \ref{fig:fluctuate_asym}), we consider a small system with $N=K=3$. Service rates are fixed, but arrival rates periodically change every $10^4$ time slots. For this we implement another version of \textsc{ADEQUA}, referred to as \textsc{ADEQUA}-restart, where each queue restarts the \textsc{ADEQUA} algorithm each time the arrival rates change. However, neither \textsc{ADEQUA} nor \textsc{ADEQUA}-restart can stabilize the system. In contrast, despite lacking knowledge of the arrival rates changes, \textsc{DAM.FE} and \textsc{DAM.UCB} are unaffected (as they require no such knowledge).  

In the fifth instance we consider a system ($N=K=4$) with asymmetric service rates and fixed arrival rates (right plot in Fig.~\ref{fig:fluctuate_asym}) under which \textsc{ADEQUA} fails to stabilize. Admittedly, this may be an unfair comparison as each queue in \textsc{ADEQUA} assumes the other queues' service rates to be the same as its service rates. Therefore, queues fail to coordinate with other queues that have have different service rates. In contrast, \textsc{DAM.FE} and \textsc{DAM.UCB} manage to stabilize the system because each queue only needs its own service rate information to cooperatively find an approximate max-weight matching.

\begin{figure}[h]
    \centering
     \scalebox{0.5}{\input{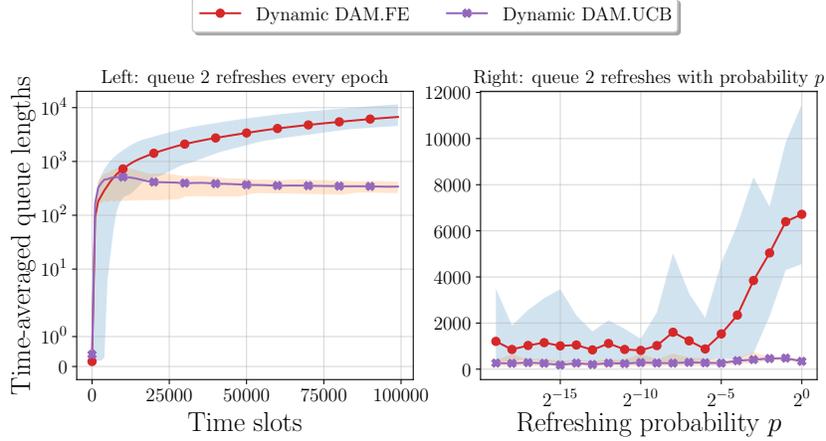}}
      \caption{Comparison of DAM.FE and DAM.UCB in a two-queue two-server system where queue $2$ is dynamically refreshed by an identical queue with no learning memory with certain refreshing probability. DAM.UCB is robust in maintaining low time-averaged queue lengths under different refreshing probability (right plot). DAM.FE has similar performance when the refreshing probability is low (right plot), but fails to stabilize the system when the probability is high (left plot).}
      \label{fig:dynamic}
\end{figure}

In our sixth and seventh instance we consider an extension to Example~\ref{example:dynamic} that highlights the robustness of dynamic \textsc{DAM.UCB}. Here, we focus only on the comparison between dynamic \textsc{DAM.UCB} and dynamic \textsc{DAM.FE} as the comparison with the existing algorithms would be futile. There are two queues and two servers. Queue $1$ always stays in the system and queue $2$ is refreshed by a new but identical queue every start of the epoch with a refreshing probability $p$. We compare dynamic \textsc{DAM.FE} and dynamic \textsc{DAM.UCB} in the left plot of Fig.~\ref{fig:dynamic} with $p = 1$ (as in Example~\ref{example:dynamic}). In this plot we display the value of $\frac{1}{T}\sum_{t=1}^T \sum_{i\in \set{I}(t)} Q_i(t)$ on the $y$-axis.  We observe (left plot of Fig.~\ref{fig:dynamic}) that dynamic \textsc{DAM.FE} fails to stabilize the system when $p=1$ as suggested by Example~\ref{example:dynamic}.
In the right plot of Fig.~\ref{fig:dynamic}, we vary the value of $p$ with $p \in \{2^{-19},\ldots,2^{-0}\}$, and report the time-averaged queue length $\frac{1}{10^5}\sum_{t=1}^{10^5}\sum_{i \in \set{I}(t)} Q_i(t)$ under dynamic \textsc{DAM.FE} and dynamic \textsc{DAM.UCB}.
We find that for \textsc{DAM.FE} the time-average queue length blows up as $p$ increases to $1$ (right plot of Fig.~\ref{fig:dynamic}). However, thanks to adaptive exploration, dynamic \textsc{DAM.UCB} always stabilizes the system, being robust to different choices of $p$.

%% file: app_notation.tex
\begin{table}[h]
\begin{tabular}{|l|l|}
\hline
Notation          & Description      \\ \hline
$\set{K}$ and $K$               & set and number of servers  \\ \hline
$\set{N}$  and $N$               & set and number of agents  \\ \hline
$\delta$          & lower bound on non-zero service probabilities                                                    \\ \hline
$J(i,t)$          & requested server for agent $i$ at time $t$                                                       \\ \hline
$R(j,t)$          & selected agent for server $j$ at time $t$                                                        \\ \hline
$\lambda_i$       & arrival rate of agent $i$                                                                        \\ \hline
$\lambda_T^\star$ & the sum of $\lambda_i^{-1}\ln^2(s_T(i)+K+1)$ for agents that arrived in the first $T$ time slots \\ \hline
$\lambda^\star$   & the sum of inverse of arrival rates                                                              \\ \hline
$\ubar{\lambda}$  & a lower bound of arrival rates                                                                   \\ \hline
$\mu_{i,j}$       & service rate between agent $i$ and server $j$                                                    \\ \hline
$\set{M}$         & the set of all vectors of queue-processing rates                                                 \\ \hline
$\Phi$            & the feasible set of fractions matching agents with servers                                       \\ \hline
$\varepsilon$     & the traffic slackness                                                                            \\ \hline
$\set{I}(t)$      & the set of queues present in time slot $t$                                                       \\ \hline
$A_i(t)$  and $S_i(t)$         & random variables for arriving and served jobs for agent $i$ at time $t$                       \\ \hline
$Q_i(t)$          & queue length of agent $i$ at the beginning of time slot $t$                                      \\ \hline
$t_s(i)$          & the time slot that agent $i$ joins the system                                                         \\ \hline
$t_e(i)$          & the last time slot that agent $i$ is the system                                                 \\ \hline
$s_T(i)$          & how many time slots agent $i$ stay in the system in the first $T$ time slots                           \\ \hline
\end{tabular}
\caption{Notations used in the system model}
\end{table}

\begin{table}[]
\begin{tabular}{|l|l|}
\hline
Notation              & Description                                                                                 \\ \hline
$\epochlength$, $\checkperiod$, $\convlength$       & lengths of an epoch, checking period, and convergence phase                                          \\ \hline
$\eta_i$              & a uniform random number of agent $i$ for perturbation                                       \\ \hline
$\pi_i$               & payoff of queue $i$                                                                         \\ \hline
$\set{E}_{\ell}$      & good checking event for epoch $\ell$                                                        \\ \hline
$\boldsymbol{\sigma}$ and $\sigma_i$  & a bipartite matching between agents and servers  and the committed server of agent $i$               \\ \hline
$\tau_i(t)$           & the last time slot before $t$ that a price changes or a request is successful for agent $i$ \\ \hline
$\ell$   and $\ell_T$               & an epoch $\ell$    and the number of epochs for the first $T$ time slots                \\ \hline
$t_0(\tau)$ and    $t_0$  & the starting time slot of an epoch $\tau$   and of a fixed epoch         \\ \hline
$f_{i,j}$             & an indicator of whether queue $i$ is matched to server $j$                                  \\ \hline
$p_{i,j}$   and    $p_j$         & agent $i$'s price of a server $j$  and price of a server $j$                                         \\ \hline
$\beta_{i,j}$  and $w_{i,j}$        & agent $i$'s increment of server $j$'s price and weight of matching agent $i$ with server $j$              \\ \hline
$\tilde{\mu}_{i,j}$   & estimated service rate between agent $i$ and server $j$                                     \\ \hline
$V(\bold{Q}(t))$      & the sum of squares of queue lengths at time $t$                                             \\ \hline
$\bold{D}_t$          & the drift $V(\bold{Q}(t+1))-V(\bold{Q}(t))$                                                 \\ \hline
$\set{E}_{\ell}$      & good checking event for epoch $\ell$                                                        \\ \hline
$\Delta_{i,j}(t)$     & the size of confidence bound of service rates between agent $i$ and server $j$              \\ \hline
$\gamma$              & exploration parameter for exploration probability in $\textsc{DAM.FE}$                      \\ \hline
$\hat{\mu}_{i,j}(t)$  & empirical average of service rates between agent $i$ and server $j$                         \\ \hline
$\bar{\mu}_{i,j}(t)$  & optimistic esimate of service rates between agent $i$ and server $j$                        \\ \hline
$E_{\ell}$            & a Bernoulli sample on whether to explore in epoch $\ell$                                    \\ \hline
$\set{E}_P$           & event where there exists at least one exploring queue for $\textsc{DAM.FE}$                 \\ \hline
$\set{E}_W$           & event where some service rate estimations are incorrect for $\textsc{DAM.FE}$               \\ \hline
$\set{E}_{\tau,i}^1$  & event of overestimating any service rate for agent $i$                                      \\ \hline
$\set{E}_{\tau,i}^2$  & event of an agent selecting a server with large confidence interval                         \\ \hline
$\set{G}_{\tau}$      & an extension of good checking events                                                        \\ \hline
$e_i$                 & total number of errors of agent $i$ in $\textsc{DAM.UCB}$                                   \\ \hline
\end{tabular}
\caption{Notations used in algorithms and analysis}
\end{table}

%% file: main.bbl
\newcommand{\etalchar}[1]{$^{#1}$}
\begin{thebibliography}{GMBB23}

\bibitem[ACBF02]{auer2002finite}
Peter Auer, Nicolo Cesa-Bianchi, and Paul Fischer.
\newblock Finite-time analysis of the multiarmed bandit problem.
\newblock {\em Machine learning}, 47(2):235--256, 2002.

\bibitem[AIM{\etalchar{+}}15]{armony2015patient}
Mor Armony, Shlomo Israelit, Avishai Mandelbaum, Yariv~N Marmor, Yulia
  Tseytlin, and Galit~B Yom-Tov.
\newblock On patient flow in hospitals: A data-based queueing-science
  perspective.
\newblock {\em Stochastic systems}, 5(1):146--194, 2015.

\bibitem[AM14]{avner2014concurrent}
Orly Avner and Shie Mannor.
\newblock Concurrent bandits and cognitive radio networks.
\newblock In {\em Joint European Conference on Machine Learning and Knowledge
  Discovery in Databases}, pages 66--81. Springer, 2014.

\bibitem[BBS08]{DBLP:journals/tsmc/BusoniuBS08}
Lucian Busoniu, Robert Babuska, and Bart~De Schutter.
\newblock A comprehensive survey of multiagent reinforcement learning.
\newblock {\em {IEEE} Trans. Syst. Man Cybern. Part {C}}, 38(2):156--172, 2008.

\bibitem[BBS21]{bubeck2021cooperative}
S{\'e}bastien Bubeck, Thomas Budzinski, and Mark Sellke.
\newblock Cooperative and stochastic multi-player multi-armed bandit: Optimal
  regret with neither communication nor collisions.
\newblock In {\em Conference on Learning Theory}, pages 821--822. PMLR, 2021.

\bibitem[BBvL11]{BoumanBL11}
Niek Bouman, Sem~C. Borst, and Johan van Leeuwaarden.
\newblock Achievable delay performance in {CSMA} networks.
\newblock In {\em 49th Annual Allerton Conference on Communication, Control,
  and Computing, Allerton 2011, Allerton Park {\&} Retreat Center, Monticello,
  IL, USA, 28-30 September, 2011}, pages 384--391. {IEEE}, 2011.

\bibitem[BCR{\etalchar{+}}12]{BirandCRSZZ12}
Berk Birand, Maria Chudnovsky, Bernard Ries, Paul~D. Seymour, Gil Zussman, and
  Yori Zwols.
\newblock Analyzing the performance of greedy maximal scheduling via local
  pooling and graph theory.
\newblock {\em {IEEE/ACM} Trans. Netw.}, 20(1):163--176, 2012.

\bibitem[Ber88]{bertsekas1988auction}
Dimitri~P Bertsekas.
\newblock The auction algorithm: A distributed relaxation method for the
  assignment problem.
\newblock {\em Annals of operations research}, 14(1):105--123, 1988.

\bibitem[BL21]{bistritz2021game}
Ilai Bistritz and Amir Leshem.
\newblock Game of thrones: Fully distributed learning for multiplayer bandits.
\newblock {\em Mathematics of Operations Research}, 46(1):159--178, 2021.

\bibitem[BLM13]{boucheron2013concentration}
St{\'e}phane Boucheron, G{\'a}bor Lugosi, and Pascal Massart.
\newblock {\em Concentration inequalities: A nonasymptotic theory of
  independence}.
\newblock Oxford university press, 2013.

\bibitem[BP19]{boursier2019sic}
Etienne Boursier and Vianney Perchet.
\newblock Sic-mmab: synchronisation involves communication in multiplayer
  multi-armed bandits.
\newblock {\em Advances in Neural Information Processing Systems}, 32, 2019.

\bibitem[BPSS07]{bayati2007iterative}
Mohsen Bayati, Balaji Prabhakar, Devavrat Shah, and Mayank Sharma.
\newblock Iterative scheduling algorithms.
\newblock In {\em IEEE INFOCOM 2007-26th IEEE International Conference on
  Computer Communications}, pages 445--453. IEEE, 2007.

\bibitem[BSS05]{BayatiSS05}
Mohsen Bayati, Devavrat Shah, and Mayank Sharma.
\newblock Maximum weight matching via max-product belief propagation.
\newblock In {\em Proceedings of the 2005 {IEEE} International Symposium on
  Information Theory, {ISIT} 2005, Adelaide, South Australia, Australia, 4-9
  September 2005}, pages 1763--1767. {IEEE}, 2005.

\bibitem[BSS09]{BuiSS09}
Loc Bui, Sujay Sanghavi, and R.~Srikant.
\newblock Distributed link scheduling with constant overhead.
\newblock {\em {IEEE/ACM} Trans. Netw.}, 17(5):1467--1480, 2009.

\bibitem[CDS20]{chen2020survey}
Jinsheng Chen, Jing Dong, and Pengyi Shi.
\newblock A survey on skill-based routing with applications to service
  operations management.
\newblock {\em Queueing Systems}, 96(1):53--82, 2020.

\bibitem[CJWS21]{choudhury2021job}
Tuhinangshu Choudhury, Gauri Joshi, Weina Wang, and Sanjay Shakkottai.
\newblock Job dispatching policies for queueing systems with unknown service
  rates.
\newblock In {\em Proceedings of the Twenty-second International Symposium on
  Theory, Algorithmic Foundations, and Protocol Design for Mobile Networks and
  Mobile Computing}, pages 181--190, 2021.

\bibitem[CLCD06]{ChenLCD06}
Lijun Chen, Steven~H. Low, Mung Chiang, and John~C. Doyle.
\newblock Cross-layer congestion control, routing and scheduling design in ad
  hoc wireless networks.
\newblock In {\em {INFOCOM} 2006. 25th {IEEE} International Conference on
  Computer Communications, Joint Conference of the {IEEE} Computer and
  Communications Societies, 23-29 April 2006, Barcelona, Catalunya, Spain}.
  {IEEE}, 2006.

\bibitem[DGS86]{demange1986multi}
Gabrielle Demange, David Gale, and Marilda Sotomayor.
\newblock Multi-item auctions.
\newblock {\em Journal of political economy}, 94(4):863--872, 1986.

\bibitem[DW06]{dimakis_walrand_2006}
Antonis Dimakis and Jean Walrand.
\newblock Sufficient conditions for stability of longest-queue-first
  scheduling: second-order properties using fluid limits.
\newblock {\em Advances in Applied Probability}, 38(2):505–521, 2006.

\bibitem[GLS09]{GuptaLS09}
Abhinav Gupta, Xiaojun Lin, and R.~Srikant.
\newblock Low-complexity distributed scheduling algorithms for wireless
  networks.
\newblock {\em {IEEE/ACM} Trans. Netw.}, 17(6):1846--1859, 2009.

\bibitem[GMBB23]{gao2021finite}
Zuguang Gao, Qianqian Ma, Tamer Ba{\c{s}}ar, and John~R Birge.
\newblock Sample complexity of decentralized tabular q-learning for stochastic
  games.
\newblock In {\em 2023 American Control Conference (ACC)}, pages 1098--1103.
  IEEE, 2023.

\bibitem[GS10]{GhaderiS10}
Javad Ghaderi and R.~Srikant.
\newblock On the design of efficient {CSMA} algorithms for wireless networks.
\newblock In {\em Proceedings of the 49th {IEEE} Conference on Decision and
  Control, {CDC} 2010, December 15-17, 2010, Atlanta, Georgia, {USA}}, pages
  954--959. {IEEE}, 2010.

\bibitem[GT23]{gaitonde2023price}
Jason Gaitonde and {\'E}va Tardos.
\newblock The price of anarchy of strategic queuing systems.
\newblock {\em Journal of the ACM}, 2023.

\bibitem[GW10]{DBLP:journals/ior/GurvichW10}
Itay Gurvich and Ward Whitt.
\newblock Service-level differentiation in many-server service systems via
  queue-ratio routing.
\newblock {\em Oper. Res.}, 58(2):316--328, 2010.

\bibitem[HXLB22]{HsuXLB22}
Wei{-}Kang Hsu, Jiaming Xu, Xiaojun Lin, and Mark~R. Bell.
\newblock Integrated online learning and adaptive control in queueing systems
  with uncertain payoffs.
\newblock {\em Oper. Res.}, 70(2):1166--1181, 2022.

\bibitem[JLS09]{JooLS09}
Changhee Joo, Xiaojun Lin, and Ness~B. Shroff.
\newblock Greedy maximal matching: Performance limits for arbitrary network
  graphs under the node-exclusive interference model.
\newblock {\em {IEEE} Trans. Autom. Control.}, 54(12):2734--2744, 2009.

\bibitem[JLWY21]{jin2021v}
Chi Jin, Qinghua Liu, Yuanhao Wang, and Tiancheng Yu.
\newblock V-learning--a simple, efficient, decentralized algorithm for
  multiagent rl.
\newblock {\em arXiv preprint arXiv:2110.14555}, 2021.

\bibitem[JSSW10]{JiangSSW10}
Libin Jiang, Devavrat Shah, Jinwoo Shin, and Jean~C. Walrand.
\newblock Distributed random access algorithm: Scheduling and congestion
  control.
\newblock {\em {IEEE} Trans. Inf. Theory}, 56(12):6182--6207, 2010.

\bibitem[JW10]{JiangWalrand10}
Libin Jiang and Jean~C. Walrand.
\newblock A distributed {CSMA} algorithm for throughput and utility
  maximization in wireless networks.
\newblock {\em {IEEE/ACM} Trans. Netw.}, 18(3):960--972, 2010.

\bibitem[JW11]{JiangW11}
Libin Jiang and Jean~C. Walrand.
\newblock Approaching throughput-optimality in distributed {CSMA} scheduling
  algorithms with collisions.
\newblock {\em {IEEE/ACM} Trans. Netw.}, 19(3):816--829, 2011.

\bibitem[KAA{\etalchar{+}}18]{krishnasamy2018augmenting}
Subhashini Krishnasamy, PT~Akhil, Ari Arapostathis, Rajesh Sundaresan, and
  Sanjay Shakkottai.
\newblock Augmenting max-weight with explicit learning for wireless scheduling
  with switching costs.
\newblock {\em IEEE/ACM Transactions on Networking}, 26(6):2501--2514, 2018.

\bibitem[KAJS18]{KrishnasamyAJS18}
Subhashini Krishnasamy, Ari Arapostathis, Ramesh Johari, and Sanjay Shakkottai.
\newblock On learning the c{\(\mu\)} rule in single and parallel server
  networks.
\newblock In {\em 56th Annual Allerton Conference on Communication, Control,
  and Computing, Allerton 2018, Monticello, IL, USA, October 2-5, 2018}, pages
  153--154. {IEEE}, 2018.

\bibitem[KNJ14]{kalathil2014decentralized}
Dileep Kalathil, Naumaan Nayyar, and Rahul Jain.
\newblock Decentralized learning for multiplayer multiarmed bandits.
\newblock {\em IEEE Transactions on Information Theory}, 60(4):2331--2345,
  2014.

\bibitem[KSJS16]{KrishnasamySenJohariShakkottai16}
Subhashini Krishnasamy, Rajat Sen, Ramesh Johari, and Sanjay Shakkottai.
\newblock Regret of queueing bandits.
\newblock In {\em Advances in Neural Information Processing Systems 29: Annual
  Conference on Neural Information Processing Systems 2016, December 5-10,
  2016, Barcelona, Spain}, pages 1669--1677, 2016.

\bibitem[KSJS21]{KrishnasamySenJohariShakkottai21}
Subhashini Krishnasamy, Rajat Sen, Ramesh Johari, and Sanjay Shakkottai.
\newblock Learning unknown service rates in queues: {A} multiarmed bandit
  approach.
\newblock {\em Oper. Res.}, 69(1):315--330, 2021.

\bibitem[KT75]{KleinrockT75}
Leonard Kleinrock and Fouad~A. Tobagi.
\newblock Packet switching in radio channels: Part i-carrier sense
  multiple-access modes and their throughput-delay characteristics.
\newblock {\em {IEEE} Trans. Commun.}, 23(12):1400--1416, 1975.

\bibitem[LM10]{abs-1009-5944}
Mahdi Lotfinezhad and Peter Marbach.
\newblock Throughput-optimal random access with order-optimal delay.
\newblock {\em CoRR}, abs/1009.5944, 2010.

\bibitem[LM11]{LotfinezhadM11}
Mahdi Lotfinezhad and Peter Marbach.
\newblock Throughput-optimal random access with order-optimal delay.
\newblock In {\em {INFOCOM} 2011. 30th {IEEE} International Conference on
  Computer Communications, Joint Conference of the {IEEE} Computer and
  Communications Societies, 10-15 April 2011, Shanghai, China}, pages
  2867--2875. {IEEE}, 2011.

\bibitem[LM18]{liang2018minimizing}
Qingkai Liang and Eytan Modiano.
\newblock Minimizing queue length regret under adversarial network models.
\newblock {\em Proceedings of the ACM on Measurement and Analysis of Computing
  Systems}, 2(1):1--32, 2018.

\bibitem[LM21]{lugosi2021multiplayer}
G{\'a}bor Lugosi and Abbas Mehrabian.
\newblock Multiplayer bandits without observing collision information.
\newblock {\em Mathematics of Operations Research}, 2021.

\bibitem[LRMJ21]{liu2021bandit}
Lydia~T Liu, Feng Ruan, Horia Mania, and Michael~I Jordan.
\newblock Bandit learning in decentralized matching markets.
\newblock {\em Journal of Machine Learning Research}, 22(211):1--34, 2021.

\bibitem[LS05]{LinS05}
Xiaojun Lin and Ness~B. Shroff.
\newblock The impact of imperfect scheduling on cross-layer rate control in
  wireless networks.
\newblock In {\em {INFOCOM} 2005. 24th Annual Joint Conference of the {IEEE}
  Computer and Communications Societies, 13-17 March 2005, Miami, FL, {USA}},
  pages 1804--1814. {IEEE}, 2005.

\bibitem[LST16]{LykourisST16}
Thodoris Lykouris, Vasilis Syrgkanis, and {\'{E}}va Tardos.
\newblock Learning and efficiency in games with dynamic population.
\newblock In {\em Proceedings of the Twenty-Seventh Annual Symposium on
  Discrete Algorithms (SODA)}, pages 120--129. {SIAM}, 2016.

\bibitem[MBKP20]{mehrabian2020practical}
Abbas Mehrabian, Etienne Boursier, Emilie Kaufmann, and Vianney Perchet.
\newblock A practical algorithm for multiplayer bandits when arm means vary
  among players.
\newblock In {\em International Conference on Artificial Intelligence and
  Statistics}, pages 1211--1221. PMLR, 2020.

\bibitem[MS04]{Mandelbaum_Stolyar_2004}
Avishai Mandelbaum and Alexander~L. Stolyar.
\newblock Scheduling flexible servers with convex delay costs: Heavy-traffic
  optimality of the generalized c$\mu$-rule.
\newblock {\em Operations Research}, 52(6):836–855, Dec 2004.

\bibitem[MS16]{maguluri2016heavy}
Siva~Theja Maguluri and R~Srikant.
\newblock Heavy traffic queue length behavior in a switch under the maxweight
  algorithm.
\newblock {\em Stochastic Systems}, 6(1):211--250, 2016.

\bibitem[MSZ06]{ModianoSZ06}
Eytan~H. Modiano, Devavrat Shah, and Gil Zussman.
\newblock Maximizing throughput in wireless networks via gossiping.
\newblock In Raymond~A. Marie, Peter~B. Key, and Evgenia Smirni, editors, {\em
  Proceedings of the Joint International Conference on Measurement and Modeling
  of Computer Systems, SIGMETRICS/Performance 2006, Saint Malo, France, June
  26-30, 2006}, pages 27--38. {ACM}, 2006.

\bibitem[Nee10a]{DBLP:journals/corr/abs-1003-3396}
Michael~J. Neely.
\newblock Stability and capacity regions or discrete time queueing networks.
\newblock {\em CoRR}, abs/1003.3396, 2010.

\bibitem[Nee10b]{DBLP:series/synthesis/2010Neely}
Michael~J. Neely.
\newblock {\em Stochastic Network Optimization with Application to
  Communication and Queueing Systems}.
\newblock Synthesis Lectures on Communication Networks. Morgan {\&} Claypool
  Publishers, 2010.

\bibitem[NL11]{NaparstekL11}
Oshri Naparstek and Amir Leshem.
\newblock Fully distributed auction algorithm for spectrum sharing in
  unlicensed bands.
\newblock In {\em 4th {IEEE} International Workshop on Computational Advances
  in Multi-Sensor Adaptive Processing, {CAMSAP} 2011, San Juan, PR, USA,
  December 13-16, 2011}, pages 233--236. {IEEE}, 2011.

\bibitem[NTS12]{NiTS12}
Jian Ni, Bo~(Rambo) Tan, and R.~Srikant.
\newblock {Q-CSMA:} queue-length-based {CSMA/CA} algorithms for achieving
  maximum throughput and low delay in wireless networks.
\newblock {\em {IEEE/ACM} Trans. Netw.}, 20(3):825--836, 2012.

\bibitem[PT99]{DBLP:journals/mor/PapadimitriouT99}
Christos~H. Papadimitriou and John~N. Tsitsiklis.
\newblock The complexity of optimal queuing network control.
\newblock {\em Math. Oper. Res.}, 24(2):293--305, 1999.

\bibitem[QWL22]{qu2022scalable}
Guannan Qu, Adam Wierman, and Na~Li.
\newblock Scalable reinforcement learning for multiagent networked systems.
\newblock {\em Operations Research}, 70(6):3601--3628, 2022.

\bibitem[RSS09]{RajagopalanSS09}
Shreevatsa Rajagopalan, Devavrat Shah, and Jinwoo Shin.
\newblock Network adiabatic theorem: an efficient randomized protocol for
  contention resolution.
\newblock In John~R. Douceur, Albert~G. Greenberg, Thomas Bonald, and Jason
  Nieh, editors, {\em Proceedings of the Eleventh International Joint
  Conference on Measurement and Modeling of Computer Systems,
  SIGMETRICS/Performance 2009, Seattle, WA, USA, June 15-19, 2009}, pages
  133--144. {ACM}, 2009.

\bibitem[RSS16]{rosenski2016multi}
Jonathan Rosenski, Ohad Shamir, and Liran Szlak.
\newblock Multi-player bandits--a musical chairs approach.
\newblock In {\em International Conference on Machine Learning}, pages
  155--163. PMLR, 2016.

\bibitem[S{\etalchar{+}}19]{slivkins2019introduction}
Aleksandrs Slivkins et~al.
\newblock Introduction to multi-armed bandits.
\newblock {\em Foundations and Trends{\textregistered} in Machine Learning},
  12(1-2):1--286, 2019.

\bibitem[SBP21]{sentenac2021decentralized}
Flore Sentenac, Etienne Boursier, and Vianney Perchet.
\newblock Decentralized learning in online queuing systems.
\newblock In Marc'Aurelio Ranzato, Alina Beygelzimer, Yann~N. Dauphin, Percy
  Liang, and Jennifer~Wortman Vaughan, editors, {\em Advances in Neural
  Information Processing Systems 34: Annual Conference on Neural Information
  Processing Systems 2021, NeurIPS 2021, December 6-14, 2021, virtual}, pages
  18501--18512, 2021.

\bibitem[SS12]{shah2012randomized}
Devavrat Shah and Jinwoo Shin.
\newblock Randomized scheduling algorithm for queueing networks.
\newblock {\em The Annals of Applied Probability}, 22(1):128--171, 2012.

\bibitem[SSM19]{StahlbuhkSM19}
Thomas Stahlbuhk, Brooke Shrader, and Eytan~H. Modiano.
\newblock Learning algorithms for scheduling in wireless networks with unknown
  channel statistics.
\newblock {\em Ad Hoc Networks}, 85:131--144, 2019.

\bibitem[SSM21]{stahlbuhk2021learning}
Thomas Stahlbuhk, Brooke Shrader, and Eytan Modiano.
\newblock Learning algorithms for minimizing queue length regret.
\newblock {\em IEEE Transactions on Information Theory}, 67(3):1759--1781,
  2021.

\bibitem[SST11]{ShahST11}
Devavrat Shah, Jinwoo Shin, and Prasad Tetali.
\newblock Medium access using queues.
\newblock In Rafail Ostrovsky, editor, {\em {IEEE} 52nd Annual Symposium on
  Foundations of Computer Science, {FOCS} 2011, Palm Springs, CA, USA, October
  22-25, 2011}, pages 698--707. {IEEE} Computer Society, 2011.

\bibitem[Sto04]{Stolyar_2004}
Alexander~L. Stolyar.
\newblock Maxweight scheduling in a generalized switch: State space collapse
  and workload minimization in heavy traffic.
\newblock {\em The Annals of Applied Probability}, 14(1), Feb 2004.

\bibitem[STT11]{ShahTT11}
Devavrat Shah, David N.~C. Tse, and John~N. Tsitsiklis.
\newblock Hardness of low delay network scheduling.
\newblock {\em {IEEE} Trans. Inf. Theory}, 57(12):7810--7817, 2011.

\bibitem[SW12]{shah2012switched}
Devavrat Shah and Damon Wischik.
\newblock Switched networks with maximum weight policies: Fluid approximation
  and multiplicative state space collapse.
\newblock {\em The Annals of Applied Probability}, 22(1):70--127, 2012.

\bibitem[SY14]{Srikant_Ying_2014}
R.~Srikant and Lei Ying.
\newblock {\em Communication networks: an optimization, control, and stochastic
  networks perspective}.
\newblock Cambridge University Press, 2014.

\bibitem[SZL{\etalchar{+}}21]{sayin2021decentralized}
Muhammed Sayin, Kaiqing Zhang, David Leslie, Tamer Basar, and Asuman Ozdaglar.
\newblock Decentralized q-learning in zero-sum markov games.
\newblock In {\em Advances in Neural Information Processing Systems},
  volume~34, pages 18320--18334, 2021.

\bibitem[Tas98]{tassiulas1998linear}
Leandros Tassiulas.
\newblock Linear complexity algorithms for maximum throughput in radio networks
  and input queued switches.
\newblock In {\em Proceedings. IEEE INFOCOM'98, the Conference on Computer
  Communications. Seventeenth Annual Joint Conference of the IEEE Computer and
  Communications Societies. Gateway to the 21st Century (Cat. No. 98},
  volume~2, pages 533--539. IEEE, 1998.

\bibitem[TE92]{Tassiulas_Ephremides_1992}
Leandros Tassiulas and Anthony Ephremides.
\newblock Stability properties of constrained queueing systems and scheduling
  policies for maximum throughput in multihop radio networks.
\newblock {\em IEEE Transactions on Automatic Control}, 37(12):1936–1948, Dec
  1992.

\bibitem[TE93]{DBLP:journals/tit/TassiulasE93}
Leandros Tassiulas and Anthony Ephremides.
\newblock Dynamic server allocation to parallel queues with randomly varying
  connectivity.
\newblock {\em {IEEE} Trans. Inf. Theory}, 39(2):466--478, 1993.

\bibitem[TX17]{TsitsiklisX17}
John~N. Tsitsiklis and Kuang Xu.
\newblock Flexible queueing architectures.
\newblock {\em Oper. Res.}, 65(5):1398--1413, 2017.

\bibitem[Wal14]{walton2014two}
Neil~S Walton.
\newblock Two queues with non-stochastic arrivals.
\newblock {\em Operations Research Letters}, 42(1):53--57, 2014.

\bibitem[WX21]{walton2021learning}
Neil Walton and Kuang Xu.
\newblock Learning and information in stochastic networks and queues.
\newblock In {\em Tutorials in Operations Research: Emerging Optimization
  Methods and Modeling Techniques with Applications}, pages 161--198. INFORMS,
  2021.

\bibitem[WZS20]{WengZhouSrikant20}
Wentao Weng, Xingyu Zhou, and R.~Srikant.
\newblock Optimal load balancing with locality constraints.
\newblock {\em Proc. {ACM} Meas. Anal. Comput. Syst.}, 4(3):45:1--45:37, 2020.

\bibitem[YSY23]{YangSrikantYing22}
Zixian Yang, R.~Srikant, and Lei Ying.
\newblock Learning while scheduling in multi-server systems with unknown
  statistics: Maxweight with discounted ucb.
\newblock In {\em Proceedings of The 26th International Conference on
  Artificial Intelligence and Statistics}, volume 206 of {\em Proceedings of
  Machine Learning Research}, pages 4275--4312. PMLR, 25--27 Apr 2023.

\bibitem[YYSE12]{YunYSE12}
Se{-}Young Yun, Yung Yi, Jinwoo Shin, and Do~Young Eun.
\newblock Optimal {CSMA:} {A} survey.
\newblock In {\em {IEEE} International Conference on Communication Systems,
  {ICCS} 2012, Singapore, November 21-23, 2012}, pages 199--204. {IEEE}, 2012.

\bibitem[ZBW22]{zhong2022learning}
Yueyang Zhong, John~R Birge, and Amy Ward.
\newblock Learning the scheduling policy in time-varying multiclass many server
  queues with abandonment.
\newblock {\em Available at SSRN}, 2022.

\bibitem[ZSP08]{zavlanos2008distributed}
Michael~M Zavlanos, Leonid Spesivtsev, and George~J Pappas.
\newblock A distributed auction algorithm for the assignment problem.
\newblock In {\em 2008 47th IEEE Conference on Decision and Control}, pages
  1212--1217. IEEE, 2008.

\end{thebibliography}
